\documentclass[10pt,twocolumn,letterpaper]{article}
\usepackage{thunlp_cvpr}
\usepackage{booktabs, tabularx, multirow, multicol, makecell, colortbl}

\usepackage{amsmath, booktabs, multirow, array, tikz, nccmath, enumitem}
\usepackage[table]{xcolor}

\usepackage{xspace}
\newcommand{\methodname}{HieroSA\xspace}

\title{Enabling Stroke-Level Structural Analysis of Hieroglyphic Scripts without Language-Specific Priors}

\setthunlparticlemeta{RESEARCH ARTICLE} 
\thunlplogooff   

\author{
  Fuwen Luo\textsuperscript{1}\quad
  Zihao Wan\textsuperscript{1}\quad
  Ziyue Wang\textsuperscript{1}\quad
  Yaluo Liu\textsuperscript{3}\quad
  Pau Tong Lin Xu\textsuperscript{1}\quad
  Xuanjia Qiao\textsuperscript{4}\quad
  Xiaolong Wang\textsuperscript{1}\quad
  Peng Li\textsuperscript{2,$\dagger$}\quad
  Yang Liu\textsuperscript{1,2,$\dagger$} \\
  \thunlpauthoraffildivider
}
\thunlpaffiliation{
  \textsuperscript{1}Dept. of Comp. Sci. \& Tech., Institute for AI, Tsinghua University \\
  \textsuperscript{2}Institute for AI Industry Research (AIR), Tsinghua University \\
  \textsuperscript{3}Rixin College, Tsinghua University \\
  \textsuperscript{4}Department of Foreign Languages and Literatures, Tsinghua University \\
  {\small \texttt{lfw23@mails.tsinghua.edu.cn}, \texttt{lipeng@air.tsinghua.edu.cn}, \texttt{liuyang2011@tsinghua.edu.cn}.}
}
\date{April 2026}

\begin{document}


\makefrontmatter
{Hieroglyphs, as logographic writing systems, encode rich semantic and cultural information within their internal structural composition. Yet, current advanced Large Language Models (LLMs) and Multimodal LLMs (MLLMs) usually remain structurally blind to this information. LLMs process characters as textual tokens, while MLLMs additionally view them as raw pixel grids. Both fall short to model the underlying logic of character strokes. Furthermore, existing structural analysis methods are often script-specific and labor-intensive. In this paper, we propose \textbf{Hiero}glyphic \textbf{S}troke \textbf{A}nalyzer (\methodname), a novel and generalizable framework that enables MLLMs to automatically derive stroke-level structures from character bitmaps without handcrafted data. It transforms modern logographic and ancient hieroglyphs character images into explicit, interpretable line-segment representations in a normalized coordinate space, allowing for cross-lingual generalization. Extensive experiments demonstrate that \methodname effectively captures character-internal structures and semantics, bypassing the need for language-specific priors. Experimental results highlight the potential of our work as a graphematics analysis tool for a deeper understanding of hieroglyphic scripts.\footnote{Code: \url{https://github.com/THUNLP-MT/HieroSA}.}}

\begingroup
\renewcommand{\thefootnote}{\fnsymbol{footnote}}
\footnotetext[2]{Corresponding author.}
\endgroup

\section{Introduction}\label{sec:introduction}

Hieroglyphs are among the earliest known writing systems in human history, originating as visually motivated symbols that depict objects, actions, or abstract concepts through recognizable forms~\cite{boltz1986early,davies1990egyptian,henderson1997world}. As a foundational medium of written communication, hieroglyphic scripts preserve rich information about early human cognition, cultural practices, and systems of knowledge representation~\cite{allen2000middle,woods2011ancient,zhang2017overview,wong2018ancient,qi2025ancientglyphnet}. Moreover, some modern writing systems continue to exhibit hieroglyphic properties, such as Chinese characters~\cite{zhao2005chinese,liu2019ancient} and Japanese Kanji~\cite{heisig2008remembering}, whose logographic characters possess intrinsic linguistic and cultural significance~\cite{joyce2011significance,fan2014study}.

Unlike phonetic scripts, hieroglyphs integrate imagery and language, in which logographic forms are carriers of meaning rather than phonetic placeholders. Semantic information is encoded in the internal structure of characters, particularly in the configuration and combination of strokes. Systematic variations in these patterns reflect semantic relationships between characters. Structural organization of characters plays a central role in understanding hieroglyphic writing systems, motivating computational approaches that aim to model character-internal structure~\cite{munoz-sanchez-2024-hieroglyphs}.

\begin{figure}[t]
    \centering
    \includegraphics[width=0.45\textwidth]{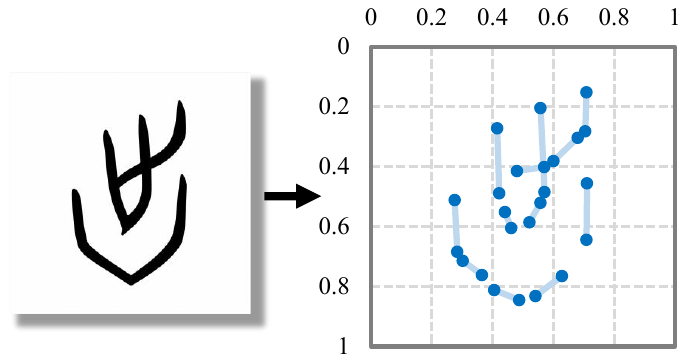}
    \vspace{-10pt}
    \caption{Given a bitmap image of a hieroglyphic character (for example, the Oracle Bone Script character on the left), our proposed \methodname infers stroke-level structure and converts the image into explicit line-segment representations in a normalized coordinate space (right).}
    \label{fig:introduction}
    \vspace{-12pt}
\end{figure}

However, capturing stroke-level structure remains challenging, despite the rapid development of large language models (LLMs) and multimodal large language models (MLLMs). LLMs usually encode text into unified symbolic representations~\cite{bouma2009normalized,sennrich2016neural,kudo2018sentencepiece,land2024fishing}, where characters or tokens are mapped to discrete indices and processed as abstract sequences, discarding character-internal structural information. As a complement to textual tokens, MLLMs additionally operate on visual inputs and process characters as images composed of pixels~\cite{dosovitskiy2020image,fei2024vitron,tschannen2025siglip,wu2025impact}. While such pixel-level representations allow models to incorporate visual appearance, they do not effectively model the structural relationships among strokes or the underlying compositional principles of character formation, limiting the ability of current models to achieve generalizable understanding of hieroglyphic characters.

Other existing methods for structural analysis are largely confined to specific writing systems and often rely on language-dependent assumptions~\cite{cao2018cw2vec,xiong2021learning,chen-etal-2025-multi}. Moreover, existing approaches typically rely on substantial amounts of labor-intensive annotation or external linguistic knowledge, such as predefined stroke inventories or decomposition rules~\cite{assael2022restoring, sommerschield2023machine}. As a result, their applicability is limited to well-studied scripts, and extending them to other hieroglyphic or pictographic writing systems requires significant additional effort, which constrains their scalability and generalization. Computer-vision-based methods inspired by sketch learning provide a more general alternative by modeling characters as line drawings or skeletal patterns~\cite{suarez2022elsed,pautrat2023deeplsd}. However, they are less effective at capturing compositional roles and symbolic functions of strokes within characters. These limitations motivate the need for a stroke-centric framework that can infer character structure directly from visual patterns, without relying on script-specific stroke definitions or manual annotations.

In this work, we propose \textbf{Hiero}glyphic \textbf{S}troke \textbf{A}nalyzer (\textbf{\methodname}), a novel and generalizable approach for automatically deriving stroke-level structure from bitmap images of hieroglyphic characters without annotated training data. Specifically, our method enables MLLMs to transform character images into explicit line-segment representations, yielding a compact and interpretable representation of stroke structure, as shown in Figure~\ref{fig:introduction}. It does not rely on manual annotations or prior knowledge of language-specific stroke systems, allowing the method to generalize naturally across diverse hieroglyphic and logographic writing systems. Extensive experiments demonstrate that \methodname effectively models stroke representations, and presents the potential as a scalable tool for computational graphematics analysis and further exploration of hieroglyphic scripts. Our contributions are threefold:\vspace{-5pt}
\begin{itemize}[left=0.3cm, itemsep=2pt, parsep=0pt]
    \item We introduce a novel training framework that enables MLLMs to effectively capture and model stroke-level structures in hieroglyphic characters, providing explicit access to character-internal structural information.
    \item Extensive experimental results verify that our proposed \textbf{\methodname} consistently generalizes across multiple pictographic writing systems, without relying on external linguistic expertise or annotations.
    \item We show that incorporating stroke representations generated by \methodname benefits downstream task of glyph recognition, and demonstrates potential as a scalable tool for broader graphematics analysis and further exploration of hieroglyphic scripts.
\end{itemize}

\section{Method}\label{sec:method}

We design an approach based on reinforcement learning (RL) that enables a flexible and expressive representation of strokes as line segments. The supervision signals are derived solely from bitmap images, which frees the training process from reliance on predefined stroke definitions, external linguistic expertise, or additional annotations. We introduce the stroke representation, the corresponding reward design, and the overall training paradigm in Sections~\ref{sec:stroke_representation}, \ref{sec:reward_design_for_stroke_representation}, and \ref{sec:training_paradigm}, respectively.

\begin{figure*}[t]
    \centering
    \includegraphics[width=1.\textwidth]{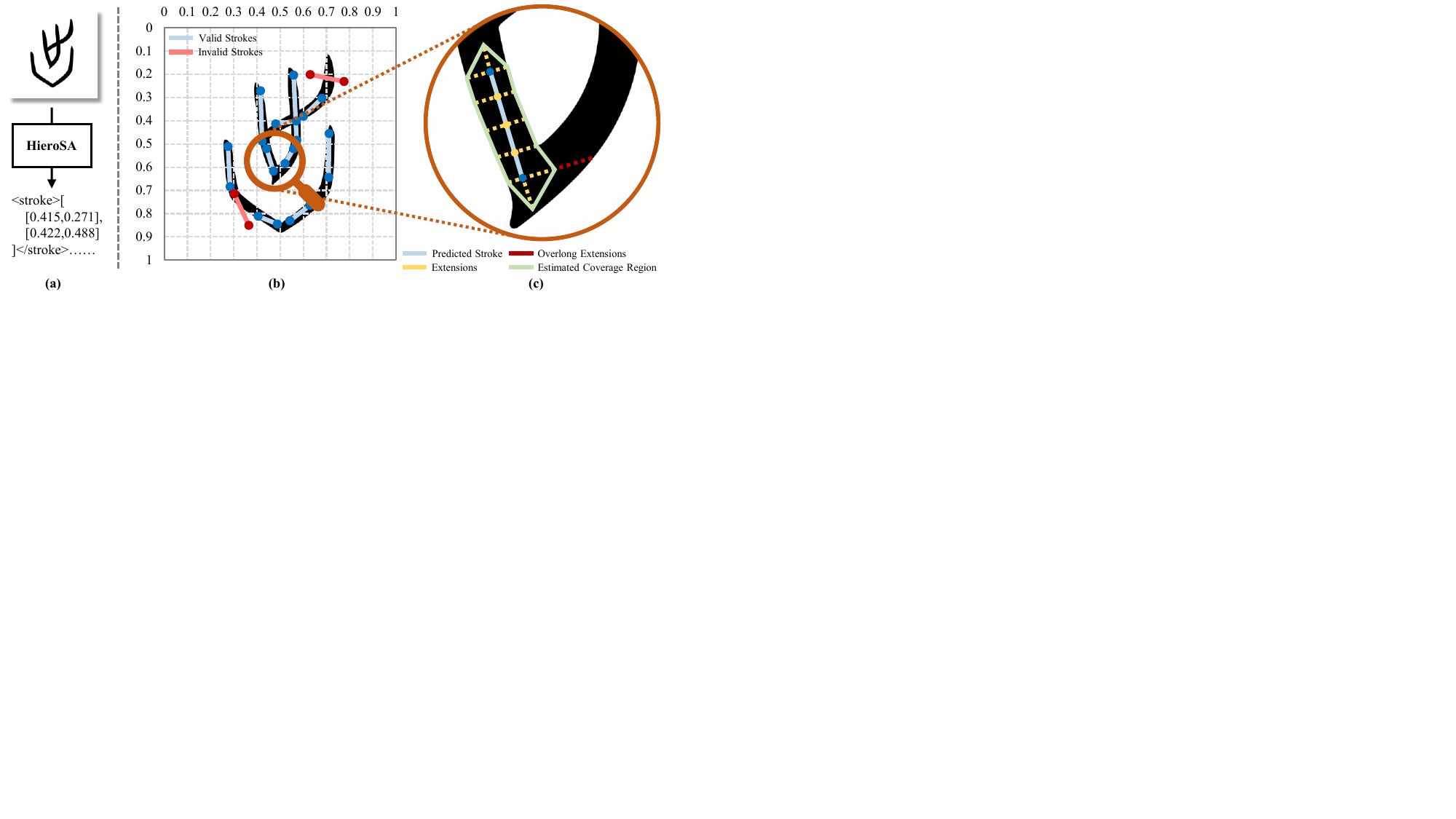}
    \vspace{-21pt}
    \caption{Overview of \textbf{\methodname}. (a) \methodname takes a binarized character image as input and outputs a structured stroke representation, where strokes are represented as a set of line segments in a normalized coordinate space. (b) Illustration of the training objective: the predicted stroke segments are optimized to maximize their overlap with the black pixels in the binarized character image. (c) Geometric estimation of black-pixel coverage for a single stroke.}
    \label{fig:method}
    \vspace{-12pt}
\end{figure*}

\subsection{Stroke Representation}\label{sec:stroke_representation}

As illustrated in Figure~\ref{fig:method}(a), our method maps a binarized character image, with black strokes as the foreground against a white background, to an geometric representation in a normalized coordinate space. A character is modeled as a set of line segments, each specified by coordinates of its two endpoints. Formally, the stroke structure is represented as\vspace{-3pt}
\begin{equation}
    \mathcal{S}=\left\{\left(\mathbf{p}_s^k,\mathbf{p}_e^k\right)\right\}_{k=1}^n,\;\mathbf{p}_s^k,\mathbf{p}_e^k\in\mathbb{R}^2,\vspace{-3pt}
\end{equation}
where $n$ denotes the number of strokes.

This coordinate-based representation allows the model to learn stroke structures directly from bitmap images, without relying on language-specific or rule-based stroke definitions. In particular, curved or complex stroke shapes are not explicitly parameterized; instead, the model learns to approximate such structures through flexible decompositions into multiple line segments that best explain the observed visual evidence.

We note that the automatically inferred stroke decompositions may not always align with human-intuitive or conventionally defined stroke segmentations. However, this flexibility enables the model to learn more adaptive and generalizable representations, particularly for scripts where standard stroke definitions are ambiguous or absent. For example, for the oracle bone script character shown in Figure~\ref{fig:introduction}, to the best of our knowledge, no standard stroke decomposition is available; nevertheless, our model produces a set of line segments that effectively capture its structural characteristics.

\subsection{Reward Design for Stroke Representation}\label{sec:reward_design_for_stroke_representation}

We use bitmap images as the sole source of supervision. Our training objectives involve maximizing the overlap between generated strokes and black pixels in the bitmap image, increasing coverage over black-pixel regions, and decreasing the number of strokes outside black pixels. Our reward computation consists of three steps: identifying valid strokes, estimating stroke coverage, and aggregating the reward.

\subsubsection{Valid Stroke Identification}\label{sec:valid_stroke_identification}\vspace{3pt}

Our first step is to identify valid strokes by examining whether each stroke lies within the black-pixel regions. For each predicted stroke defined by its endpoints $\left(\mathbf{p}_s,\mathbf{p}_e\right)$, we uniformly sample points along the line segment:\vspace{-6pt}
\begin{fleqn}
\begin{equation}
    \mathbf{p}_i=\frac{i\,\mathbf{p}_s+\left(m+1-i\right)\,\mathbf{p}_e}{m+1},\;i=1,\cdots,m,\vspace{-6pt}
\end{equation}
\end{fleqn}
where $m$ is the minimum value such that the distance between neighboring sampled points is smaller than a threshold $D$. We then check whether all sampled points $\left\{\mathbf{p}_i\right\}_{i=1}^m$ and endpoints $\mathbf{p}_s$ and $\mathbf{p}_e$ lie within the black-pixel region of the binarized image. If any point falls outside the black-pixel area, the corresponding stroke is marked as invalid. For example, as illustrated in Figure~\ref{fig:method}(b), the two red strokes are identified as invalid under this criterion. By filtering out such strokes, the model is encouraged to generate stroke segments that closely follow the foreground structure of the character.

\subsubsection{Single-Stroke Coverage Estimation}\label{sec:single_stroke_coverage_estimation}

Given a stroke represented by a line segment, we estimate its coverage region by extending the segment along both tangential and normal directions within the black-pixel region, as shown in Figure~\ref{fig:method} (c). To avoid over-extension that may intrude into neighboring strokes, excessively long extensions are truncated, yielding a polygonal region that approximates the spatial coverage of the stroke.

Specifically, given endpoints $\left(\mathbf{p}_s,\mathbf{p}_e\right)$ of a stroke and sampled points $\{\mathbf{p}_i\}_{i=1}^m$ obtained in the previous step, we define\vspace{-5pt}
\begin{equation}
    \mathbf{t}=\frac{\mathbf{p}_e-\mathbf{p}_s}{\|\mathbf{p}_e-\mathbf{p}_s\|}\vspace{-3pt}
\end{equation}
to denote the unit tangent direction of the stroke, and let $\mathbf{n}$ be the corresponding counterclockwise unit normal direction, defined as\vspace{-5pt}
\begin{equation}
    \mathbf{n}=(-\mathbf{t}_y,\mathbf{t}_x),\vspace{-5pt}
\end{equation}
where $(\mathbf{t}_x,\mathbf{t}_y)$ are the coordinates of $\mathbf{t}$. For each point $\mathbf{p}_j\in\{\mathbf{p}_s,\mathbf{p}_1,\cdots,\mathbf{p}_m,\mathbf{p}_e\}$, we extend rays along both normal directions until reaching the black-white boundary:\vspace{-5pt}
\begin{equation}
    \begin{cases}
        d_j^{+}=\max\left\{d>0\;\middle|\;\left[\mathbf{p}_j,\mathbf{p}_j+d\,\mathbf{n}\right]\subset\Omega_B\right\}, \\
        d_j^{-}=\max\left\{d>0\;\middle|\;\left[\mathbf{p}_j,\mathbf{p}_j-d\,\mathbf{n}\right]\subset\Omega_B\right\},\vspace{-3pt}
    \end{cases}
\end{equation}
where $\Omega_B$ denotes the set of black pixels in the binarized image.

We first compute the mean normal extension length:\vspace{-6pt}
\begin{equation}
    \bar d=\frac{1}{2(m+2)}\sum_{j}\left(d_j^{+}+d_j^{-}\right).\vspace{-6pt}
\end{equation}
Then, excessively large extension lengths are regarded as abnormal and excluded:\vspace{-5pt}
\begin{equation}
    \mathcal{A}=\left\{j\;\middle|\;\max(d_j^{+},d_j^{-})>\lambda\bar d\right\},\vspace{-5pt}
\end{equation}
where $\lambda$ is the threshold for identifying abnormal extensions. Using the remaining samples, we compute a refined mean:\vspace{-6pt}
\begin{equation}
    \tilde d=\frac{1}{2|\mathcal{I}|}\sum_{j\in\mathcal{I}}\left(d_j^{+}+d_j^{-}\right),\vspace{-6pt}
\end{equation}
where $\mathcal{I}=\{s,1,\dots,m,e\}\setminus\mathcal{A}$. Extension lengths are then truncated as\vspace{-5pt}
\begin{equation}
    \hat d_j^{\pm}=\min\left(d_j^{\pm},\lambda\tilde d\right),\vspace{-5pt}
\end{equation}

For the endpoints $\mathbf{p}_s$ and $\mathbf{p}_e$, we further extend along the tangent direction. The tangential extension length is\vspace{-5pt}
\begin{equation}
    \begin{cases}
        \ell_s=\displaystyle\frac{\hat d_s^{+}+\hat d_s^{-}}{2}, \\[4pt]
        \ell_e=\displaystyle\frac{\hat d_e^{+}+\hat d_e^{-}}{2}.\vspace{-5pt}
    \end{cases}
\end{equation}
The endpoints are thus extended to\vspace{-5pt}
\begin{equation}
    \begin{cases}
        \mathbf{p}_s'=\mathbf{p}_s-\ell_s\,\mathbf{t}, \\
        \mathbf{p}_e'=\mathbf{p}_e+\ell_e\,\mathbf{t}.\vspace{-5pt}
    \end{cases}
\end{equation}

Finally, the set of offset vertices $\mathcal{Q}$ computes as\vspace{-5pt}
\begin{fleqn}
    \begin{equation}
        \mathcal{Q}=\{q_j^\pm\mid q_j^\pm=p_j\pm d_j^\pm\mathbf{n},\;j=1,\cdots,m\},\vspace{-5pt}
    \end{equation}
\end{fleqn}
together with the tangentially extended endpoints $\mathbf{p}_s'$ and $\mathbf{p}_e'$ define a polygonal coverage region $\mathcal{C}$ for the stroke. Examples of the polygonal coverage regions estimated by our method are provided in Appendix~\ref{app:examples_of_stroke_coverage_estimation}. The region is subsequently used for reward computation in the Section~\ref{sec:reward_aggregation}.

\subsubsection{Reward Aggregation}\label{sec:reward_aggregation}

Given the coverage regions $\{\mathcal{C}_k\}_{k=1}^n$ estimated for individual strokes, we aggregate them sequentially to compute the final reward, while penalizing invalid strokes. Let\vspace{-6pt}
\begin{equation}
    \mathcal{U}_k=\bigcup_{i\in\mathcal{V}_k}\mathcal{C}_i\vspace{-6pt}
\end{equation}
where $\mathcal{V}_k$ denotes the set of previously accepted valid strokes. For the $k$-th stroke, we examine whether its coverage region $\mathcal{C}_k$ provides sufficient novel contribution. Specifically, the stroke is marked as invalid if\vspace{-3pt}
\begin{equation}
    \frac{|\left(\mathcal{C}_k\setminus\mathcal{U}_k\right)\cap\Omega_B|}{|\Omega_B|}<\tau,\vspace{-3pt}
\end{equation}
where $\tau$ is the overlap ratio threshold. This encourages the model to cover new foreground regions with each stroke while penalizing redundant strokes, thereby preventing solutions that repeatedly exploit the same local areas.

All invalid strokes, including those identified in Section~\ref{sec:valid_stroke_identification} as well as those identified during aggregation, are discarded. Let $\mathcal{V}$ denote the index set of remaining valid strokes, and let\vspace{-6pt}
\begin{equation}
    \mathcal{C}_{\mathrm{final}}=\bigcup_{k\in\mathcal{V}}\mathcal{C}_k\vspace{-6pt}
\end{equation}
be the final aggregated coverage region. The overall reward is then defined as\vspace{-5pt}
\begin{equation}
    r_s=\frac{|\mathcal{C}_{\mathrm{final}}\cap\Omega_B|}{|\Omega_B|}\cdot\left(1-\alpha\,N_{\mathrm{invalid}}\right),\vspace{-5pt}
\end{equation}
where $N_{\mathrm{invalid}}$ denotes the number of invalid strokes and $\alpha$ is a penalty coefficient. This formulation provides a compact training signal that balances foreground coverage promotion with penalties on invalid strokes.

\subsection{Training Paradigm}\label{sec:training_paradigm}\vspace{3pt}

We adopt Group Relative Policy Optimization (GRPO)~\cite{shao2024deepseekmath}, a reinforcement learning (RL) algorithm that has been widely used to improve the reasoning abilities of LLMs, as our training method. Prior studies commonly apply GRPO together with a rule-based answer reward and a format reward, leading to strong and reliable model performance. Following this established setting, we integrate our stroke-representation reward $r_s$ with a format reward $r_f$, encouraging the model to generate outputs that conform to the structured format as exemplified in Figure~\ref{sec:method}(a), thereby facilitating reliable parsing. The final reward is
\vspace{-5pt}
\begin{equation}
    r=r_s+\beta\,r_f,\vspace{-5pt}
\end{equation}
where $\beta$ is a hyperparameter that balances the contribution of the format reward.

\section{Experiments}\label{sec:experiments}

\begin{table*}[t]
\centering
\small
\resizebox{1.\textwidth}{!}{
\begin{tabular}{l|ccc|ccc|ccc|ccc}
\toprule
\multirow{2}{*}[-3pt]{\textbf{Model}} & \multicolumn{3}{c|}{\textbf{Chinese (ZH)}} & \multicolumn{3}{c|}{\textbf{Japanese (JA)}} & \multicolumn{3}{c|}{\textbf{Oracle Bone Script (OBS)}} & \multicolumn{3}{c}{\textbf{AVG}} \\
\cmidrule(rl){2-4}\cmidrule(rl){5-7}\cmidrule(rl){8-10}\cmidrule(rl){11-13}
& \textbf{RE$\uparrow$} & \textbf{CO (\%)$\uparrow$} & \textbf{IS (\%)$\downarrow$} & \textbf{RE$\uparrow$} & \textbf{CO (\%)$\uparrow$} & \textbf{IS (\%)$\downarrow$} & \textbf{RE$\uparrow$} & \textbf{CO (\%)$\uparrow$} & \textbf{IS (\%)$\downarrow$} & \textbf{RE$\uparrow$} & \textbf{CO (\%)$\uparrow$} & \textbf{IS (\%)$\downarrow$} \\
\midrule
GPT-5 & 0.133 & \hphantom{0}3.6 & 88.2 & 0.129 & \hphantom{0}1.5 & 92.8 & 0.139 & \hphantom{0}3.8 & 92.0 & 0.134 & \hphantom{0}3.0 & 91.0 \\
Claude Sonnet 4 & 0.137 & 11.9 & 86.9 & 0.131 & 10.1 & 89.9 & 0.129 & \hphantom{0}6.3 & 93.4 & 0.132 & \hphantom{0}9.4 & 90.1 \\
Qwen3-VL-4B & 0.032 & \hphantom{0}0.5 & 97.9 & 0.028 & \hphantom{0}0.5 & 98.0 & 0.063 & \hphantom{0}0.7 & 98.5 & 0.041 & \hphantom{0}0.6 & 98.1 \\
ELSED & 0.026 & \hphantom{0}0.4 & 83.8 & 0.014 & \hphantom{0}0.3 & 80.3 & 0.100 & \hphantom{0}0.8 & 73.9 & 0.047 & \hphantom{0}0.5 & 79.3 \\
DeepLSD & 0.092 & \hphantom{0}1.1 & 85.2 & 0.086 & \hphantom{0}2.7 & 71.1 & 0.064 & \hphantom{0}0.3 & 72.4 & 0.081 & \hphantom{0}1.4 & 76.2 \\
\midrule
\methodname~(ZH) & \textbf{0.837} & \textbf{78.5} & \hphantom{0}\textbf{6.1} & \textbf{0.591} & \textbf{60.9} & \underline{19.7} & \textbf{0.522} & \underline{50.5} & \textbf{17.6} & \textbf{0.650} & \textbf{63.3} & \textbf{14.5} \\
\methodname~(JA) & \underline{0.756} & \underline{72.2} & \underline{10.2} & \underline{0.584} & \underline{59.4} & \textbf{19.1} & \underline{0.455} & 45.0 & \underline{23.9} & \underline{0.598} & \underline{58.9} & \underline{17.7}  \\
\methodname~(OBS) & 0.446 & 64.6 & 23.1 & 0.295 & 52.2 & 33.9 & 0.344 & \textbf{53.3} & 25.2 & 0.362 & 56.7 & 27.4 \\
\bottomrule
\end{tabular}}
\vspace{-9pt}
\caption{Results on the stroke structure parsing task.}
\label{tab:results}
\vspace{-6pt}
\end{table*}

\begin{table}[t]
\centering
\small
\resizebox{1.\linewidth}{!}{
\begin{tabular}{l|cccccc}
\toprule
\textbf{Model} & \textbf{RE$\uparrow$} & \textbf{CO$\uparrow$} & \textbf{IS$\downarrow$} & \textbf{CD$\downarrow$} & \textbf{MD$\downarrow$} & \textbf{LD$\downarrow$} \\
\midrule
GPT-5 & 0.118 & 1 & 94.9 & 9.5 & 0.56 & 0.244 \\
Claude Sonnet 4 & 0.012 & 0.3 & 96.8 & 11.4 & 0.705 & 0.303 \\
Qwen3-VL-4B & 0.023 & 0.2 & 98.8 & 11.4 & 0.656 & 0.285 \\
ELSED & 0.003 & 0 & 92.2 & 12.4 & 0.759 & 0.323 \\
DeepLSD & 0.063 & 0.6 & 91 & 21.5 & 0.439 & 0.288 \\
\midrule
\methodname~(ZH) & 0.719 & 68.2 & 7.6 & 5.5 & 0.069 & 0.175 \\
\methodname~(JA) & 0.586 & 57.5 & 13.1 & 4.3 & 0.07 & 0.166 \\
\methodname~(OBS) & 0.386 & 55.1 & 25.2 & 10.7 & 0.077 & 0.205 \\
\bottomrule
\end{tabular}}
\vspace{-9pt}
\caption{Results on Make Me A Hanzi dataset.}
\label{tab:results_make_me_a_hanzi}
\vspace{-12pt}
\end{table}

\begin{table}[t]
\centering
\small
{\setlength{\tabcolsep}{4pt}
\renewcommand{\arraystretch}{1.2}
\begin{tabular}{>{\raggedright\arraybackslash}m{0.03\linewidth}|>{\raggedright\arraybackslash}m{0.16\linewidth}|>{\centering\arraybackslash}m{0.20\linewidth}>{\centering\arraybackslash}m{0.20\linewidth}>{\centering\arraybackslash}m{0.20\linewidth}}
\toprule
& & a & b & c \\\midrule
& \textbf{Model} & \textbf{Chinese (ZH)} & \textbf{Japanese (JA)} & \textbf{OBS} \\\midrule\vspace{-6pt}
& Input Bitmap Image &
\includegraphics[width=0.75\linewidth]{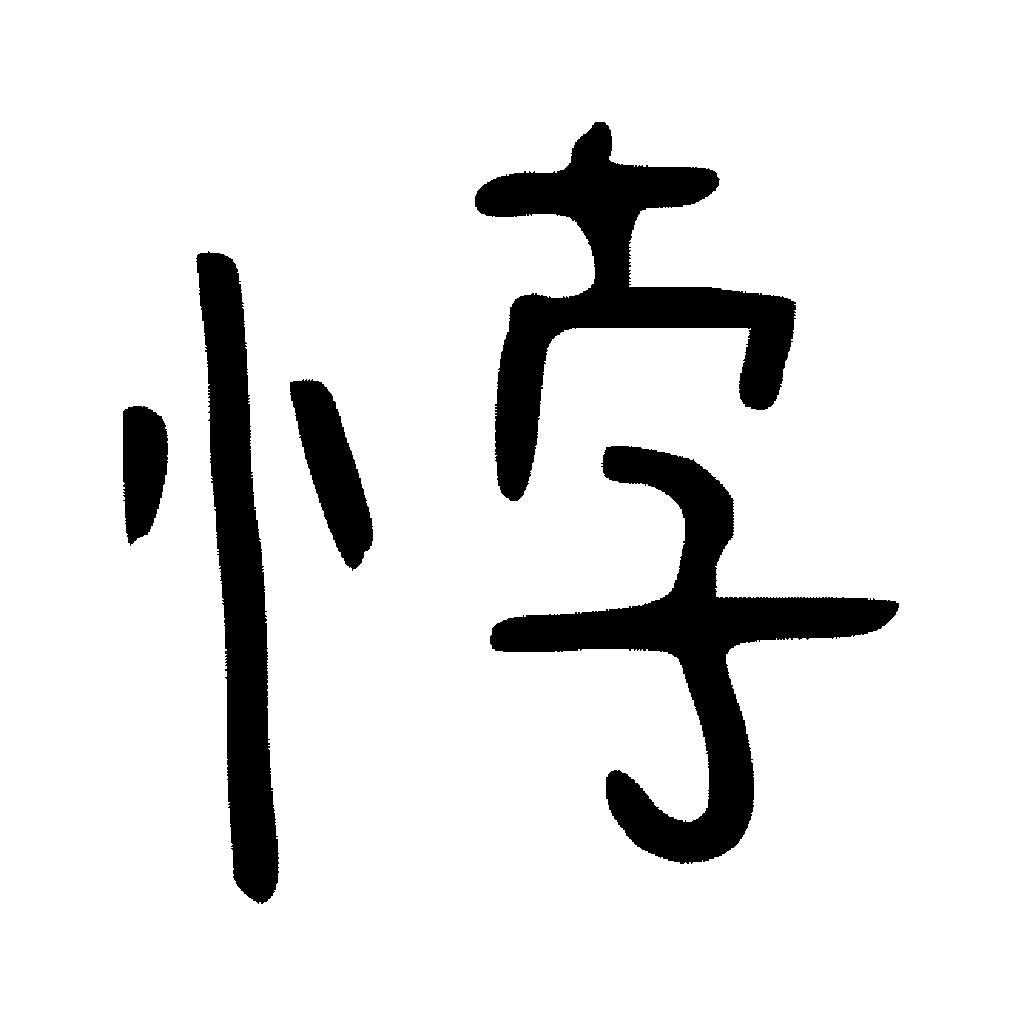} \vspace{-6pt}&
\includegraphics[width=0.75\linewidth]{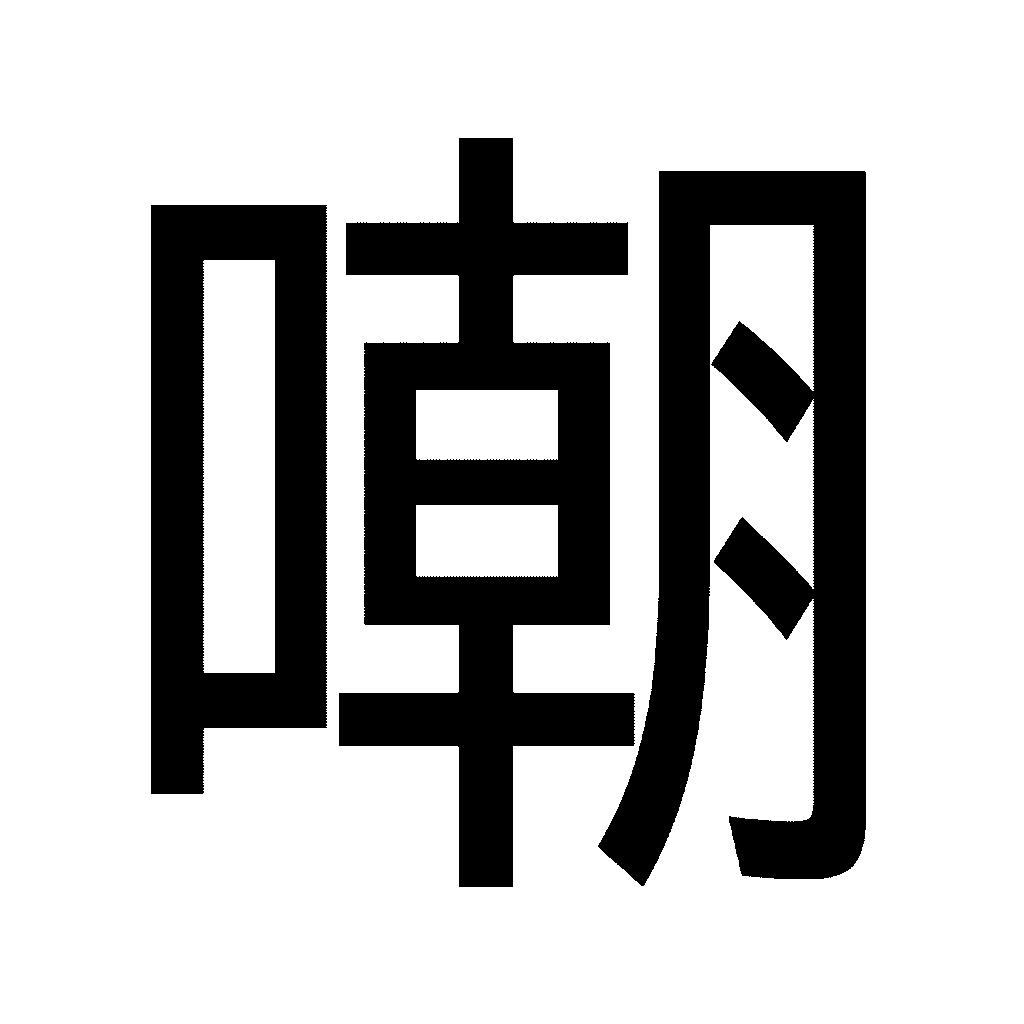} \vspace{-6pt}&
\includegraphics[width=0.75\linewidth]{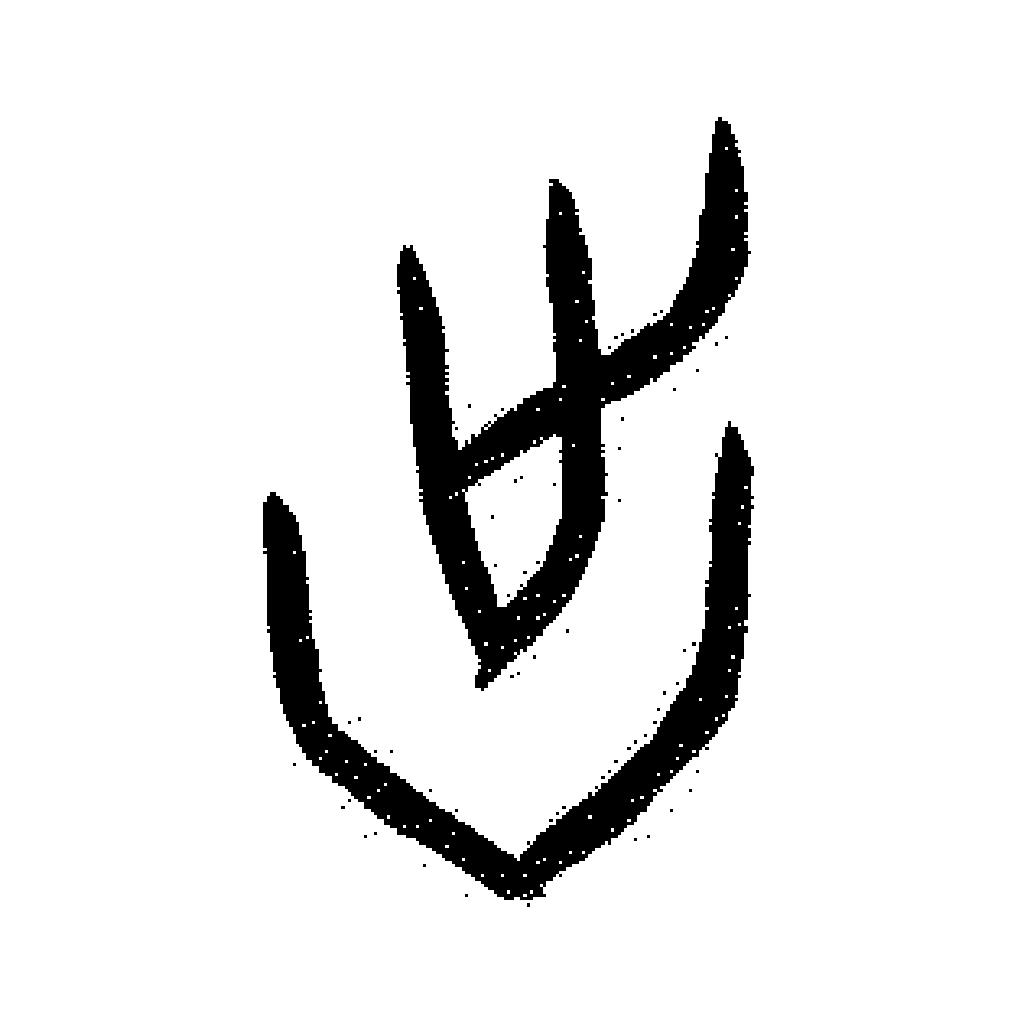} \vspace{-6pt}\\
1 & \methodname (ZH) &
\includegraphics[width=0.75\linewidth]{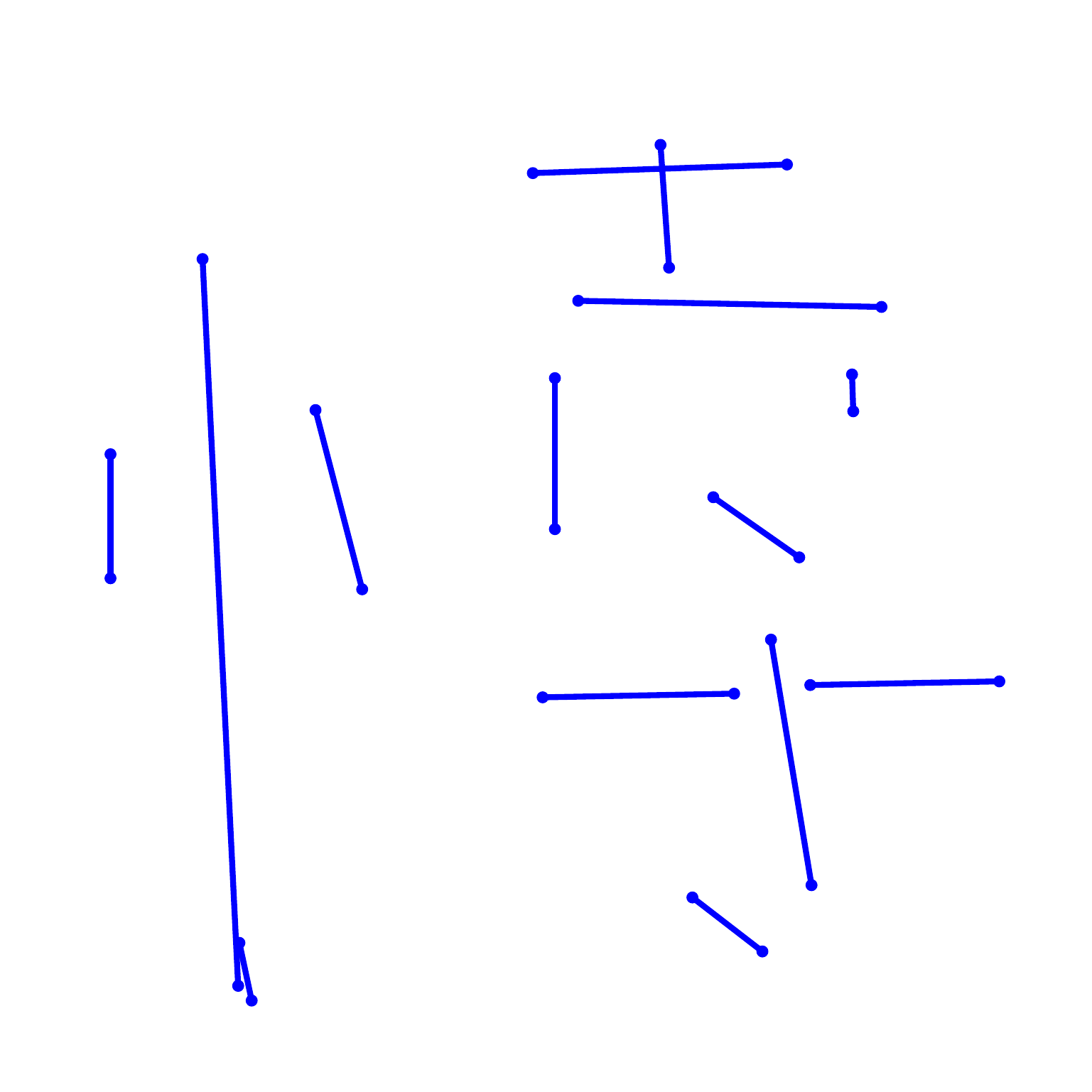}\vspace{-6pt} &
\includegraphics[width=0.75\linewidth]{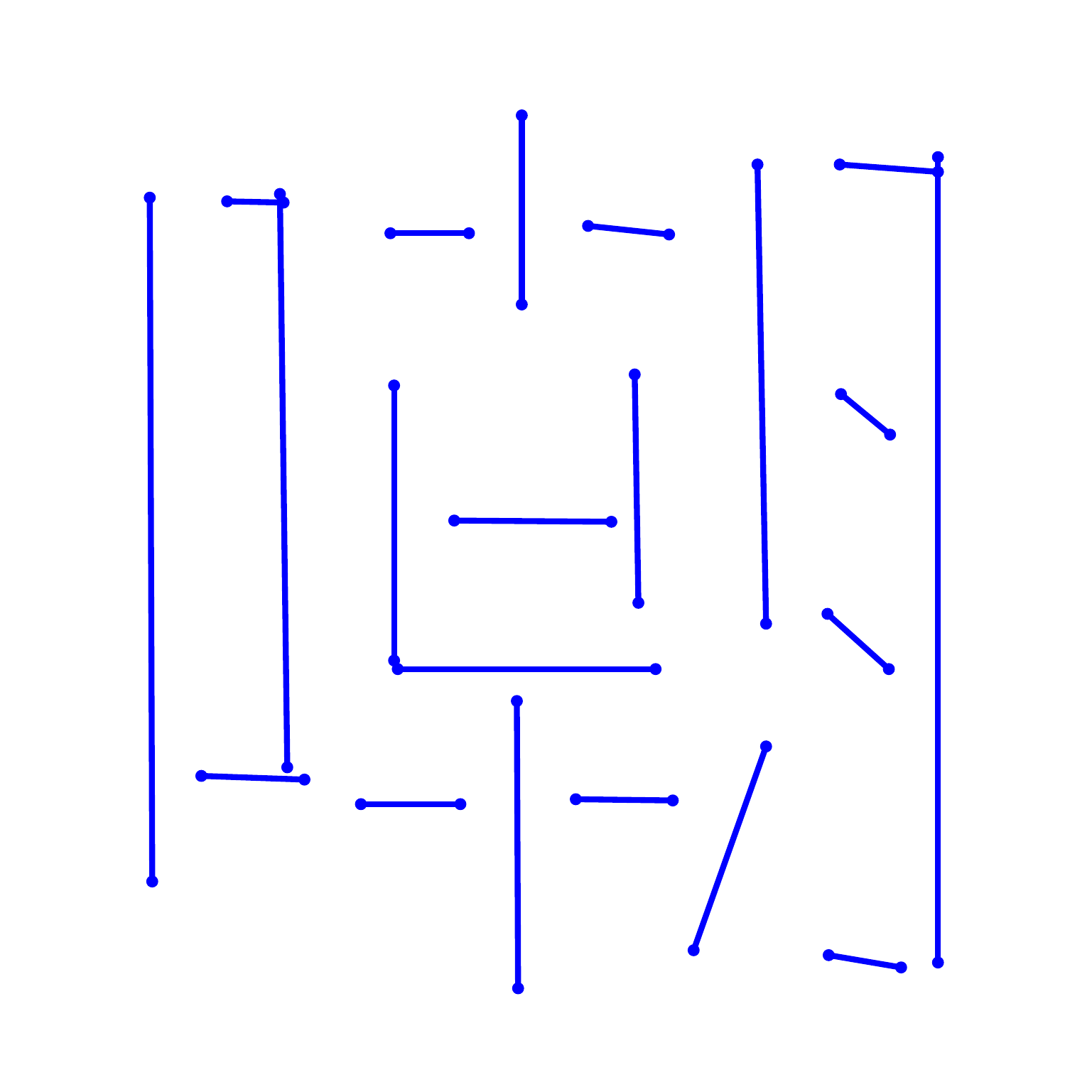}\vspace{-6pt} &
\includegraphics[width=0.75\linewidth]{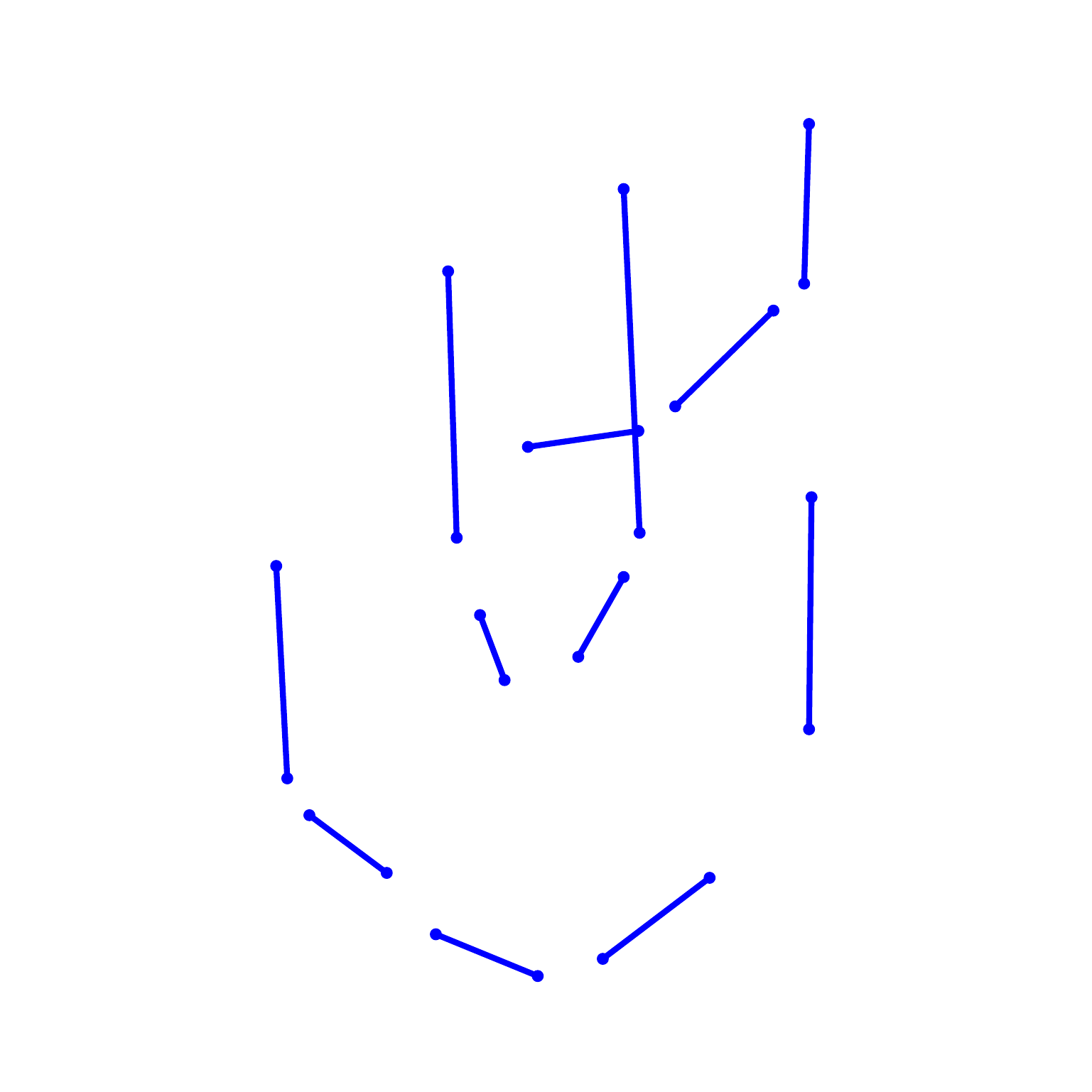}\vspace{-6pt} \\
2 & \methodname (JA) &
\includegraphics[width=0.75\linewidth]{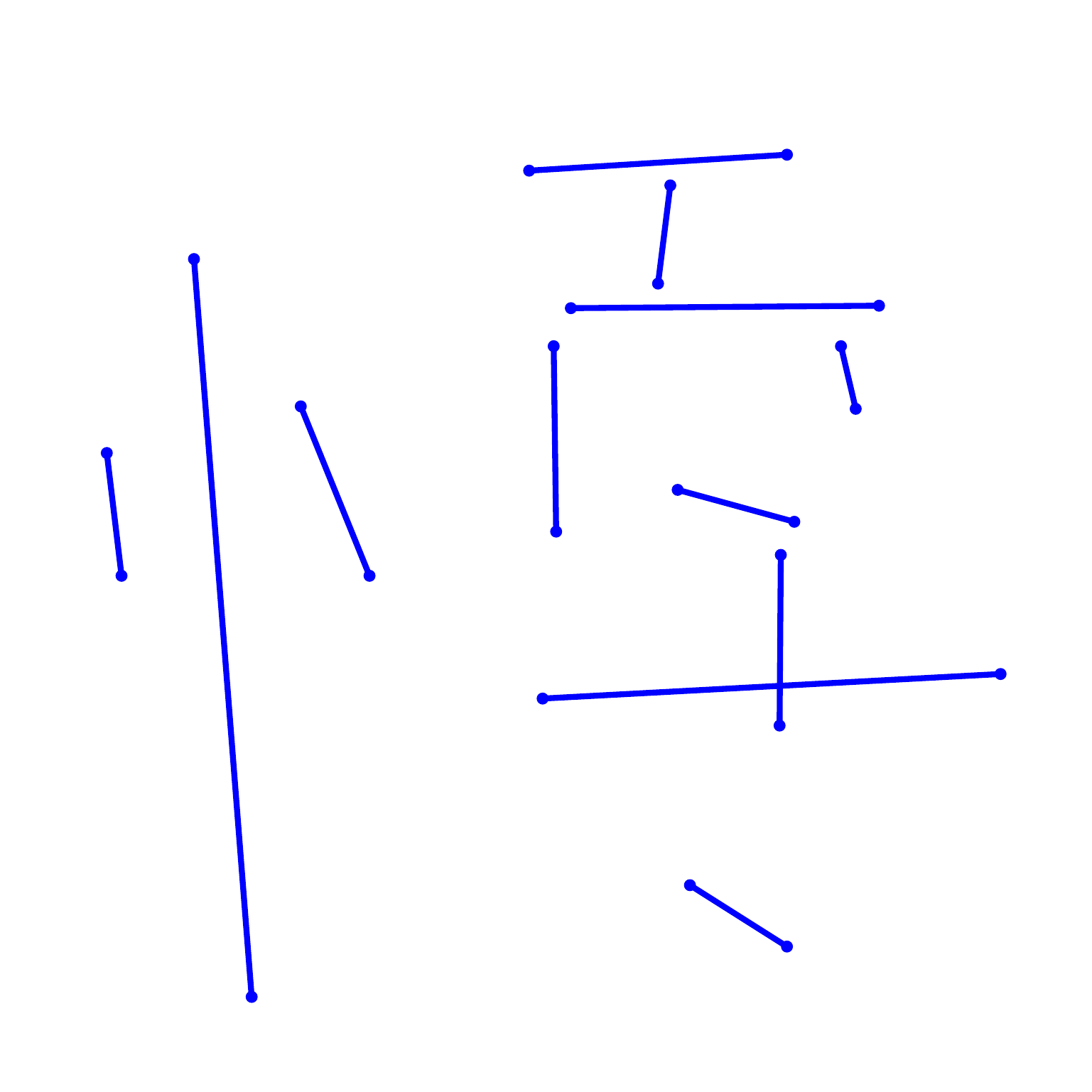}\vspace{-6pt} &
\includegraphics[width=0.75\linewidth]{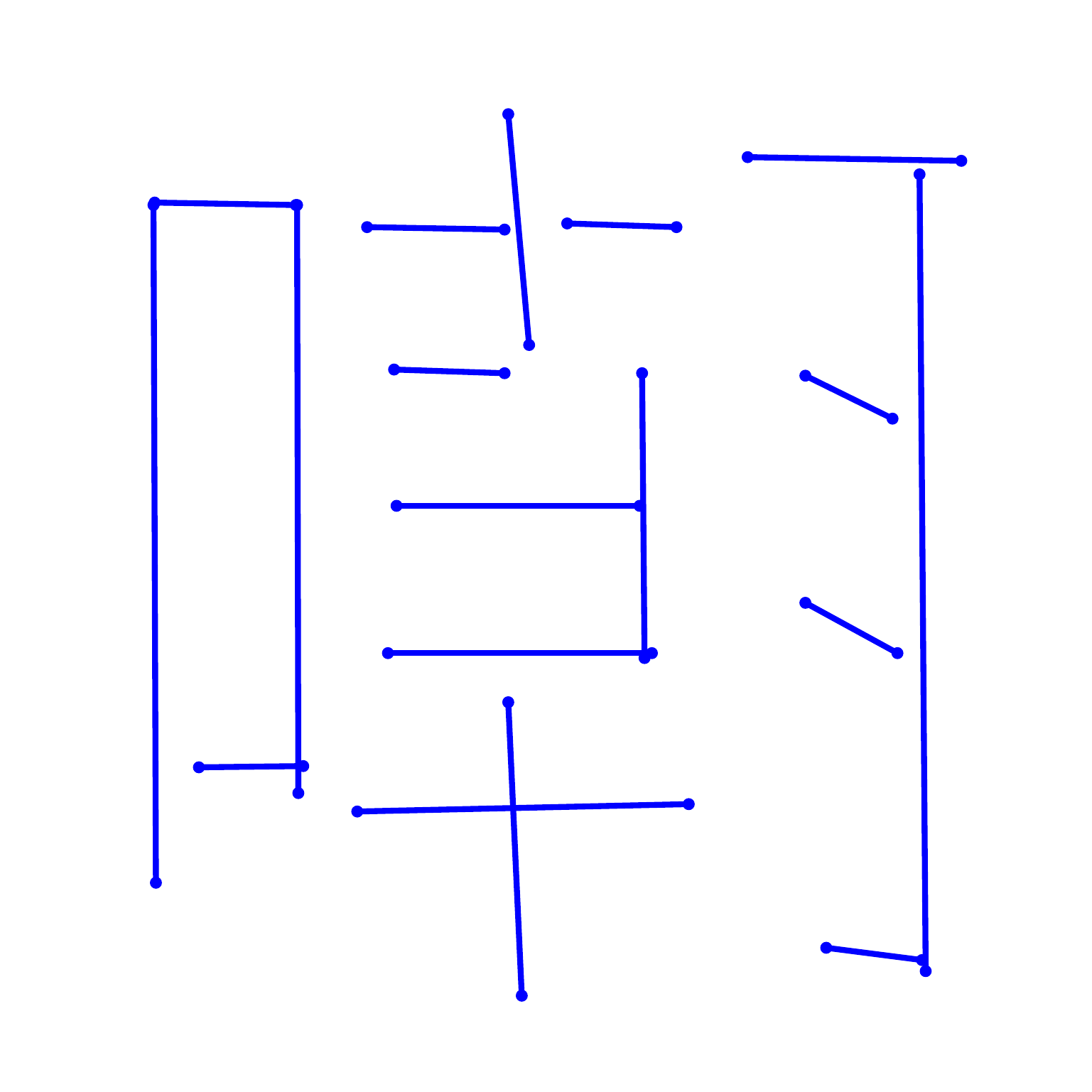}\vspace{-6pt} &
\includegraphics[width=0.75\linewidth]{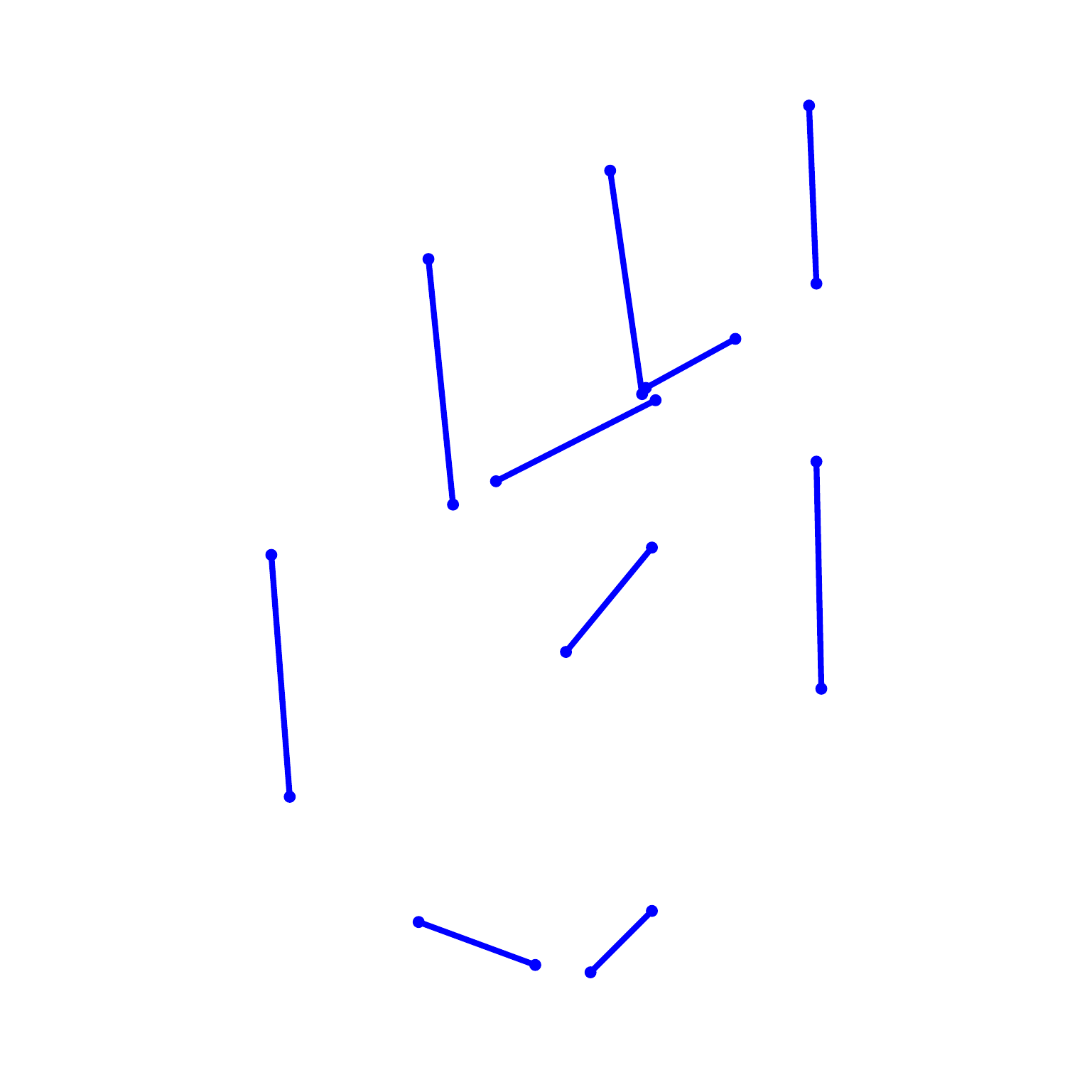}\vspace{-6pt} \\
3 & \methodname (OBS) &
\includegraphics[width=0.75\linewidth]{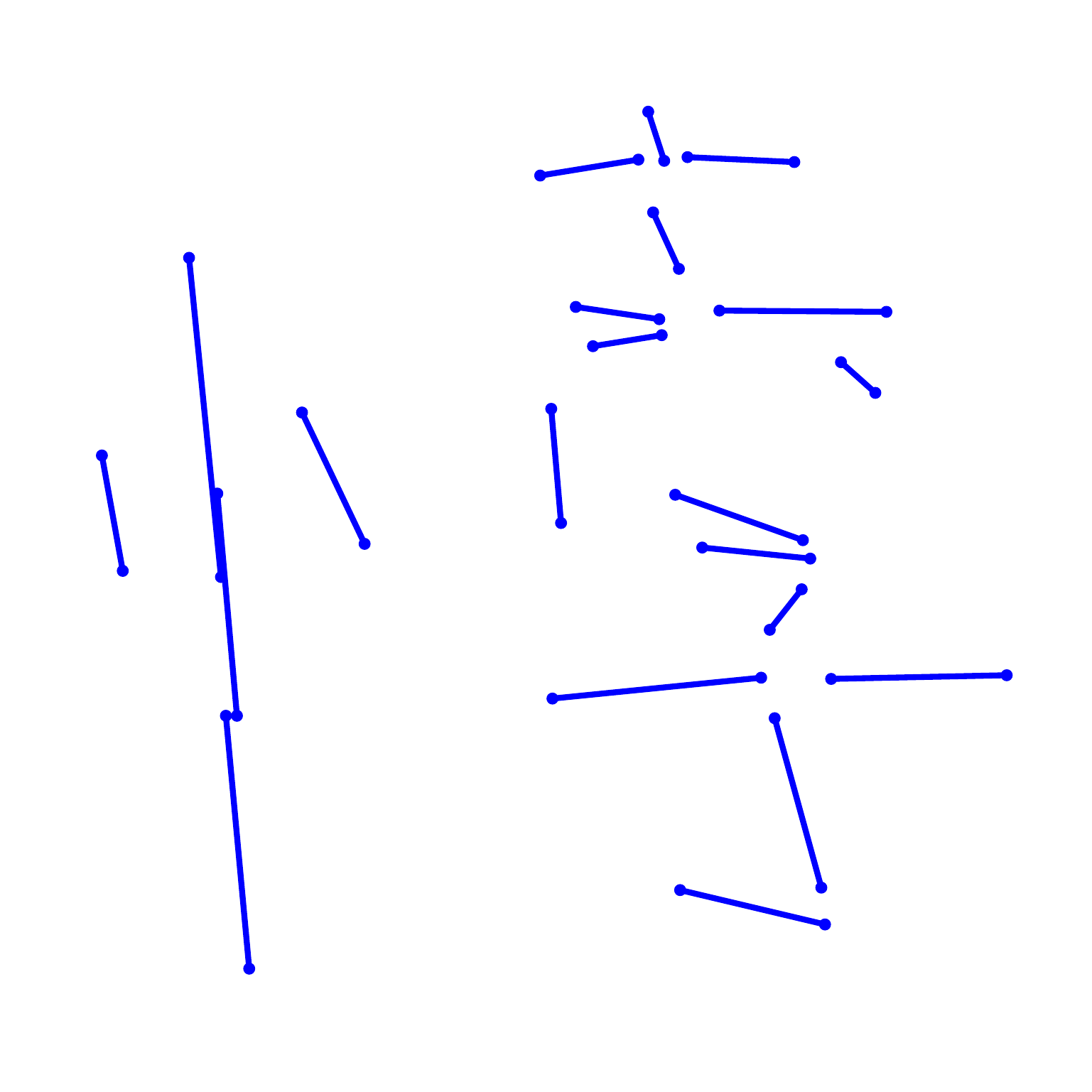} \vspace{-6pt}&
\includegraphics[width=0.75\linewidth]{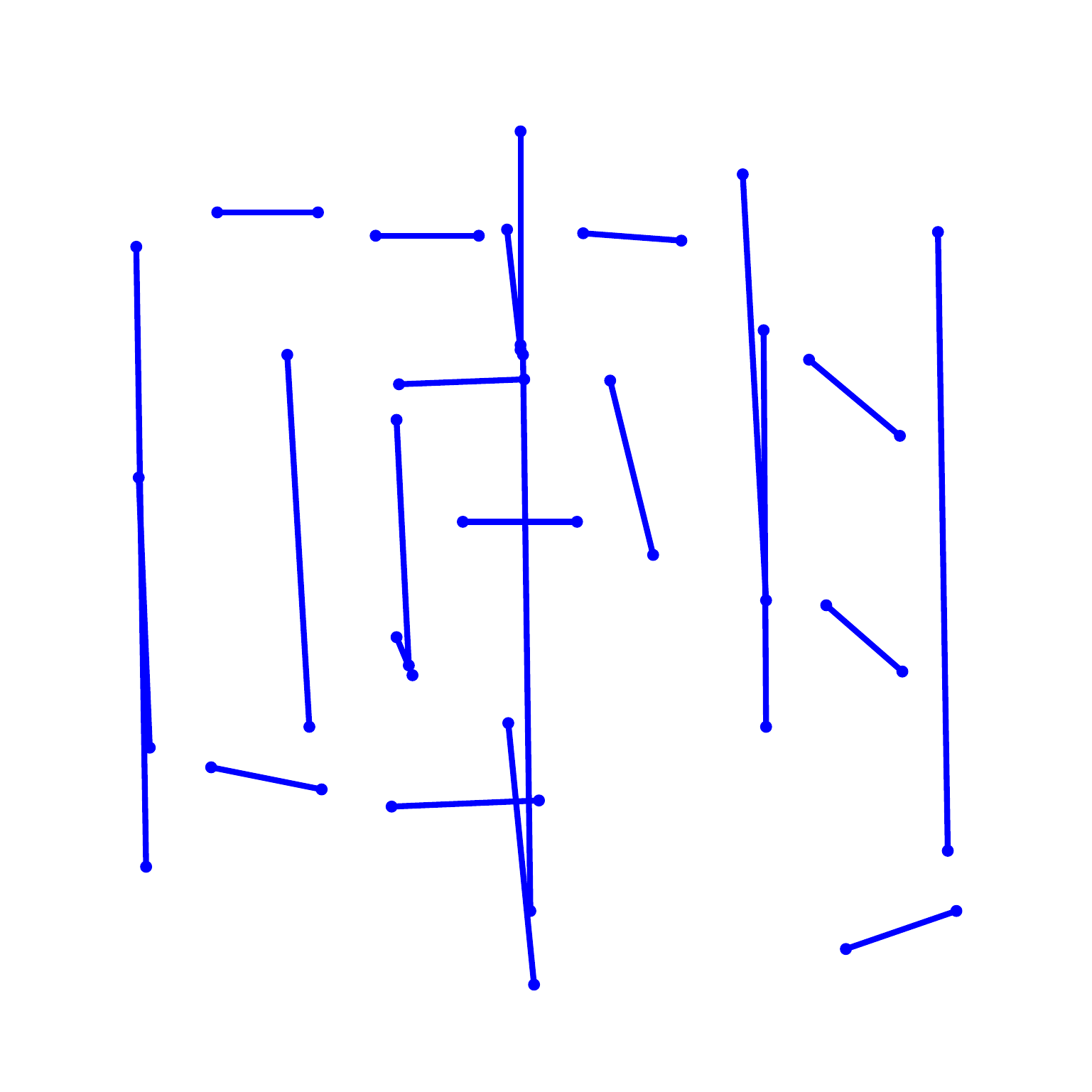} \vspace{-6pt}&
\includegraphics[width=0.75\linewidth]{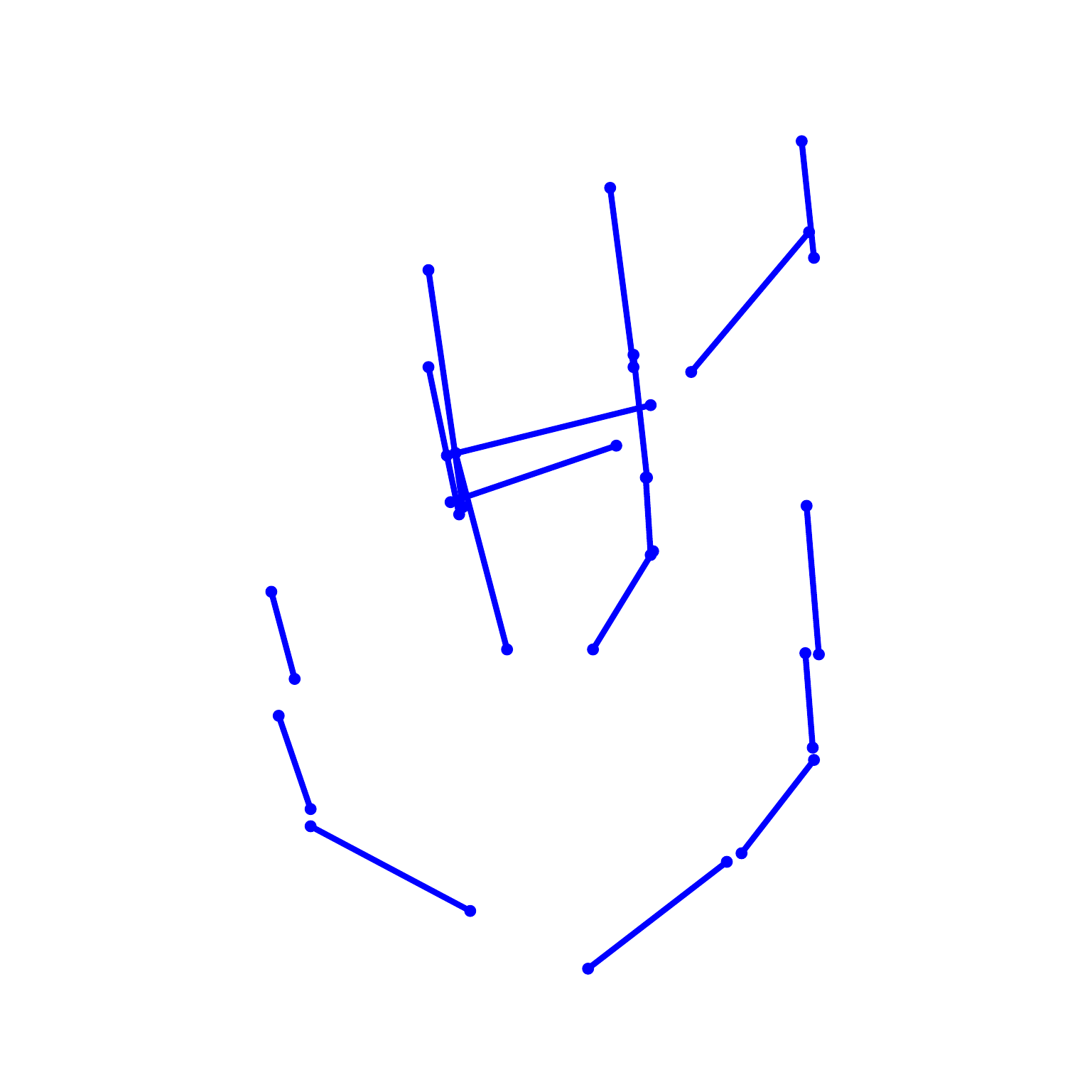}\vspace{-6pt} \\
\bottomrule
\end{tabular}}
\vspace{-6pt}
\caption{Qualitative visualization of stroke-structure parsing results across scripts and training settings. Rows correspond to models trained on different scripts, and columns correspond to evaluation scripts.}
\label{tab:qualitative_results}
\vspace{-15pt}
\end{table}

\subsection{Implementations}\label{sec:implementations}

The training data is curated from two sources: bitmap images generated from SVG glyphs in font files, and publicly available datasets of hieroglyphic character images, without requiring any additional annotations. Our training dataset covers three writing systems: contemporary Chinese characters~(Chinese; CH), Japanese Kanji~(Japanese; JA), and Oracle Bone Scripts~(OBS), an ancient Chinese hieroglyphic writing system with distinctive pictographic forms.

We adopt Qwen3-VL-4B-Instruct~\cite{bai2025qwen3vl} as the base model and train a separate model for each script for two epochs. During training, each image is augmented with an overlaid coordinate system to enhance coordinate prediction accuracy and spatial localization capability. Details of data curation are provided in Appendix~\ref{app:training_set_curation}, with training configurations described in Appendix~\ref{app:training_details}. Effectiveness of the coordinate system is further validated through ablation studies in Appendix~\ref{app:ablation_coordinate}.

\subsection{Baselines and Evaluation Metrics}\label{sec:baselines_and_evaluation_metrics}

We compare our method with current strong MLLMs, GPT-5 (gpt-5-2025-08-07) and Claude Sonnet 4 (claude-sonnet-4-20250514). In addition, we report the performance of Qwen3-VL-4B (Qwen3-VL-4B-Instruct), which serves as the base model of HieroSA. Furthermore, to compare against computer vision approaches for line segmentation, we also include ELSED~\cite{suarez2022elsed} and DeepLSD~\cite{pautrat2023deeplsd} as supplementary baselines.

For evaluation, we consider three metrics reported in Table~\ref{tab:results}: RE (reward), CO (coverage), and IS (invalid strokes). RE is the reward score computed on the test set using the same reward function adopted during training, and serves as an overall measure of how well the predicted stroke decomposition aligns with the training objective. CO evaluates structural completeness by measuring the percentage of the black-pixel region $\Omega_B$ that is covered by the predicted strokes, reflecting whether the full character structure is adequately captured. IS denotes the percentage of invalid strokes among all predicted strokes, quantifying incorrect, redundant, or structurally implausible outputs. Higher RE and CO, together with lower IS, indicate better overall performance.

\subsection{Main Results}\label{sec:main_results}

We evaluate the stroke-structure parsing performance of our model on Chinese, Japanese, and OBS characters. Given an input character image, the model predicts a set of stroke segments that represent the underlying stroke structure of the character. To assess the stroke parsing quality, we measure the spatial overlap between the polygonal coverage induced by the predicted strokes and the foreground region of the original character image. Results are reported in Table~\ref{tab:results}. Details of the test set are provided in Appendix~\ref{app:test_set_curation}.

\vspace{-6pt}
\paragraph{Quantitative results.} Models trained with \methodname consistently outperform all baselines, including both powerful existing MLLMs (GPT-5 and Claude Sonnet 4) and specialized line-segmentation methods (ELSED and DeepLSD). They achieve higher reward (RE), indicating a better alignment with the training objective, while simultaneously yielding higher coverage (CO) and substantially lower invalid-stroke rates (IS). Improvements are evident in not only intra-script experiments, but also cross-script evaluation settings.

To have a more balanced view of model performance beyond the training-related metrics RE, CO, and IS, we further adopt the Make Me A Hanzi~\footnote{\url{https://github.com/skishore/makemeahanzi}} dataset as an additional test set, since it provides stroke median-line annotations that enable finer geometric evaluation. Based on these stroke-level ground-truth annotations, we introduce three complementary metrics: CD (stroke count difference), MD (mean distance between stroke centers), and LD (stroke length difference). These metrics respectively measure structural quantity, spatial alignment, and geometric proportion, thereby covering multiple fundamental aspects of stroke-level fidelity. As shown in Table~\ref{tab:results_make_me_a_hanzi}, HieroSA consistently achieves better CD, MD, and LD than other methods, which is consistent with its advantages under RE, CO, and IS. The qualitative results in Table 2 of our paper further demonstrate that our model generates structurally reasonable strokes.

\vspace{-6pt}
\paragraph{Qualitative results.} Qualitative results further illustrate the effectiveness of our approach. As shown in Table~\ref{tab:qualitative_results}, the predicted stroke structures align with the core stroke composition of the input characters across scripts. The models capture the main structural skeletons while preserving essential stroke connectivity, yielding visually coherent stroke-structure decompositions. These observations are consistent with the quantitative results.

\vspace{-6pt}
\paragraph{Performance differences across training languages.} We observe noticeable performance differences among models trained on different writing systems. To better understand this behavior, we analyze structural statistics of the corresponding training datasets (Table~\ref{tab:structural_statistics}), including the number of eight-connected components, foreground boundary length and area, and the boundary-to-area ratio. From Chinese to Japanese to OBS, the number of connected components, boundary length, and area decrease monotonically, indicating progressively simpler glyph structures. In contrast, the boundary-to-area ratio increases, suggesting more less smooth stroke boundaries.

These suggest a plausible reason for the observed performance gap: overly simple structures reduce structural diversity and weaken stroke-structure priors, whereas excessive boundary curvature introduces geometric noise during training. The Chinese characters, with more complex yet cleaner structures, enables models to learn more robust and transferable representations, leading to superior cross-script generalization.

\begin{table}[t]
\centering
\small
\begin{tabular}{l|ccc|c}
\toprule
\textbf{Training Set} & \textbf{CC} & \textbf{FB} & \textbf{FA} & \textbf{BAR} \\
\midrule
Chinese & 7.466 & 8.670 & 0.178 & 49.827 \\
Japanese & 4.389 & 7.371 & 0.138 & 55.566 \\
OBS & 2.680 & 6.154 & 0.098 & 63.922 \\
\bottomrule
\end{tabular}
\caption{Structural statistics of binarized images across training sets in different languages, including the average connected components (CC), foreground boundary length (FB), foreground area (FA), and boundary-to-area ratio (BAR).}
\label{tab:structural_statistics}
\end{table}

\begin{table}[t]
\centering
\small
\begin{tabular}{l|ccc}
\toprule
\textbf{OCR Setting} & \textbf{Chinese} & \textbf{Japanese} & \textbf{AVG} \\
\midrule
Zero-shot & 66.4 & 63.2 & 64.8 \\
Trained w/o Stroke Rep & 88.9 & 75.8 & 82.4 \\
Trained w/ Stroke Rep & \textbf{89.7} & \textbf{77.3} & \textbf{83.5} \\
\bottomrule
\end{tabular}
\caption{OCR accuracy under different training settings. Stroke representations are generated by \methodname~(Chinese) and are consistently applied to OCR training and test sets.}
\label{tab:ocr}
\end{table}

\begin{table*}[t]
\centering
\small
\resizebox{1.\textwidth}{!}{
{\setlength{\tabcolsep}{4pt}
\renewcommand{\arraystretch}{1.2}
\newcommand{\imgwithbox}[2][]{
  \begin{tikzpicture}[baseline=(img.base)]
    \node[inner sep=0pt] (img) {\includegraphics[#1]{#2}};
    \draw[red, line width=1pt] (img.south west) rectangle (img.north east);
  \end{tikzpicture}
}
\newcommand{\imgfade}[2][]{
  \begin{tikzpicture}[baseline=(img.base)]
    \node[inner sep=0pt, opacity=0.1] (img)
      {\includegraphics[#1]{#2}};
  \end{tikzpicture}
}
\begin{tabular}{>{\centering\arraybackslash}m{0.01\linewidth}||>{\centering\arraybackslash}m{0.07\linewidth}|>{\centering\arraybackslash}m{0.07\linewidth}>{\centering\arraybackslash}m{0.07\linewidth}>{\centering\arraybackslash}m{0.07\linewidth}>{\centering\arraybackslash}m{0.07\linewidth}>{\centering\arraybackslash}m{0.07\linewidth}|>{\centering\arraybackslash}m{0.07\linewidth}>{\centering\arraybackslash}m{0.07\linewidth}>{\centering\arraybackslash}m{0.07\linewidth}>{\centering\arraybackslash}m{0.07\linewidth}>{\centering\arraybackslash}m{0.07\linewidth}||>{\centering\arraybackslash}m{0.07\linewidth}}
\toprule
& \textbf{a} & \textbf{b} & \textbf{c} & \textbf{d} & \textbf{e} & \textbf{f} & \textbf{g} & \textbf{h} & \textbf{i} & \textbf{j} & \textbf{k} & \textbf{l} \\
\midrule\midrule
& \textbf{Query} & \multicolumn{5}{c|}{\textbf{Direct Image-based Matching Results}} & \multicolumn{5}{c||}{\textbf{Stroke-Derived Masked Image Matching Results}} & \textbf{Overlap} \\
\midrule\noalign{\vskip -3pt}\rowcolor{gray!20}\multicolumn{13}{c}{\textbf{Oracle Bone Scripts}} \\\noalign{\vskip -2pt}\midrule
1 &
\includegraphics[width=0.95\linewidth]{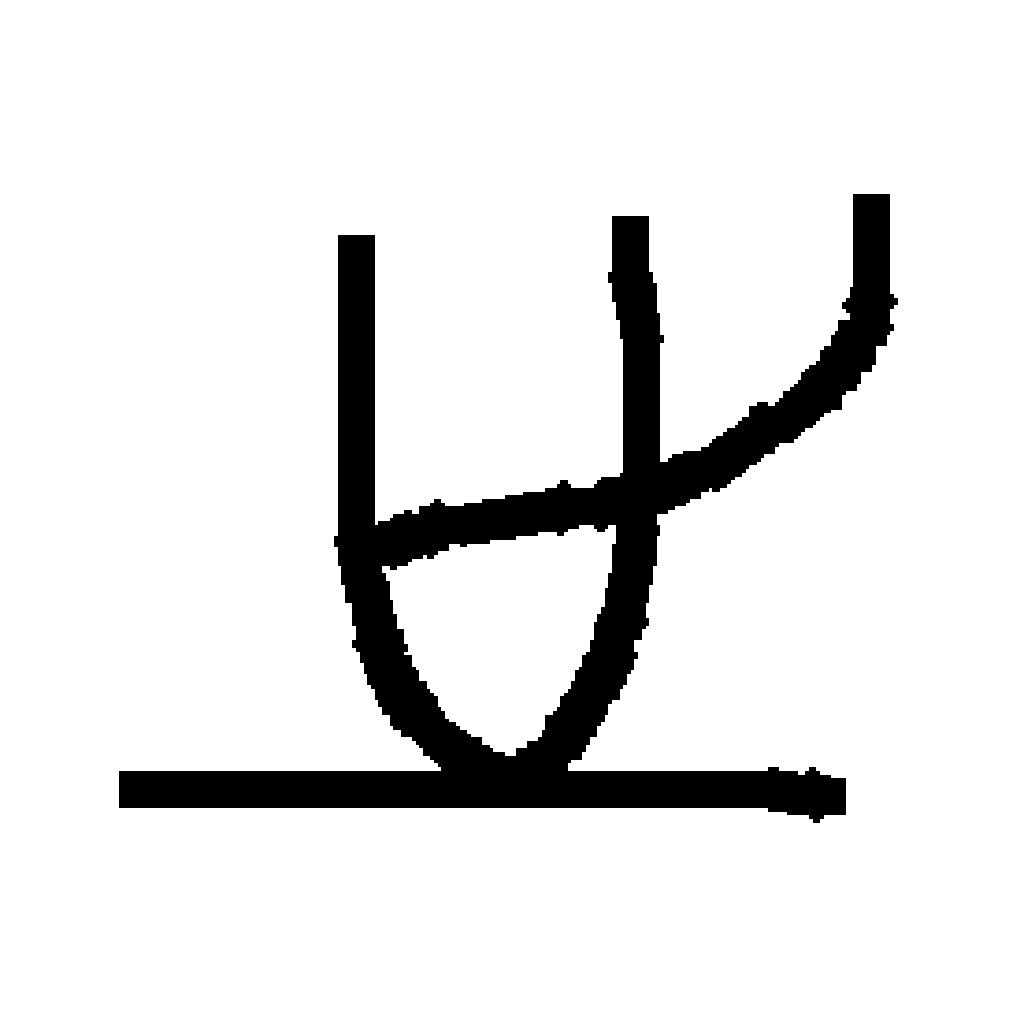} &
\imgfade[width=0.95\linewidth]{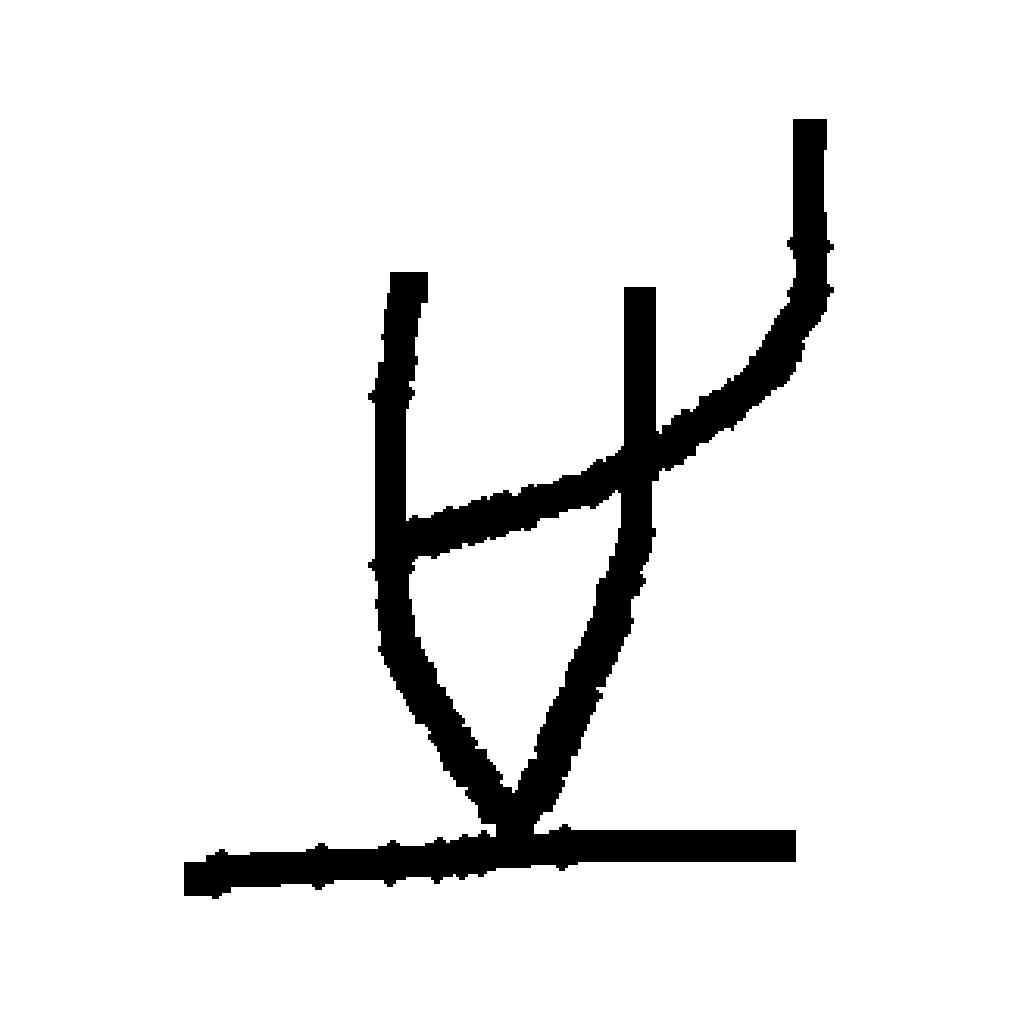} &
\imgfade[width=0.95\linewidth]{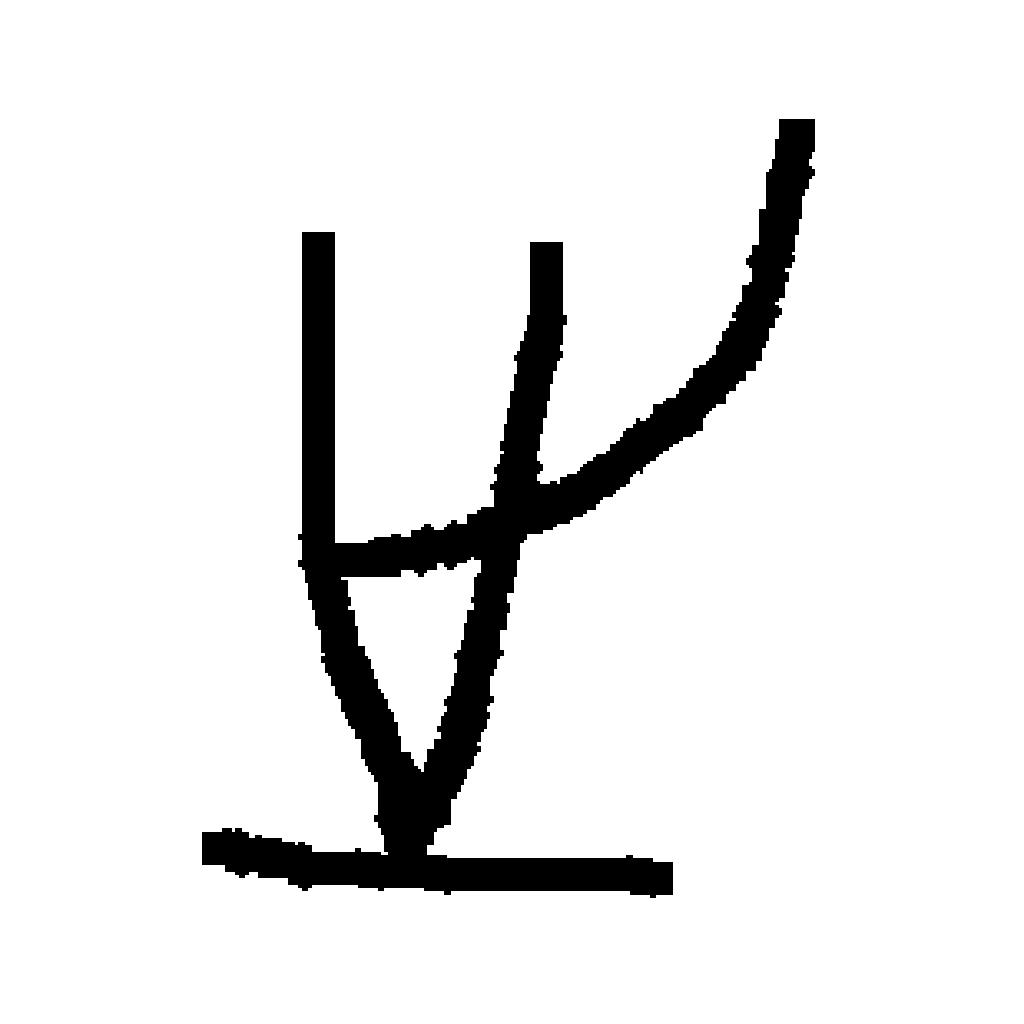} &
\imgfade[width=0.95\linewidth]{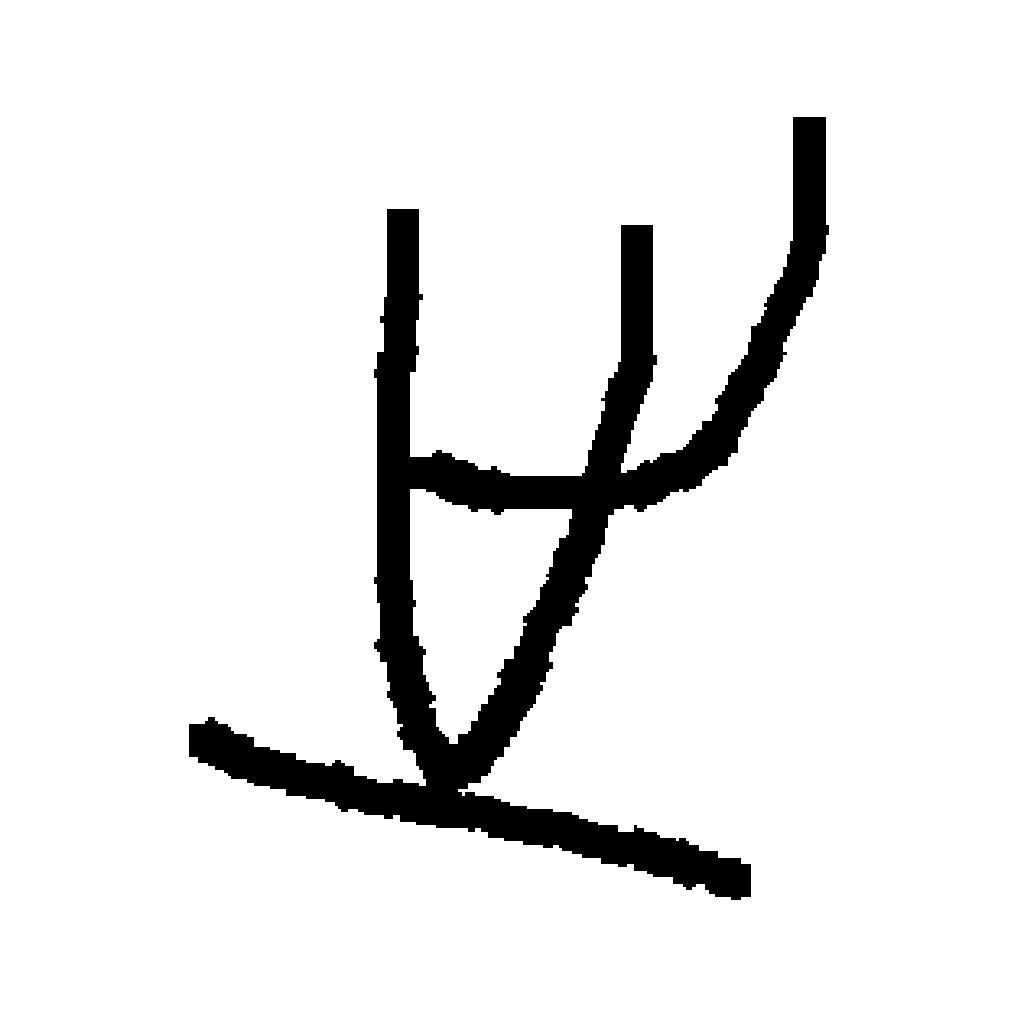} &
\imgfade[width=0.95\linewidth]{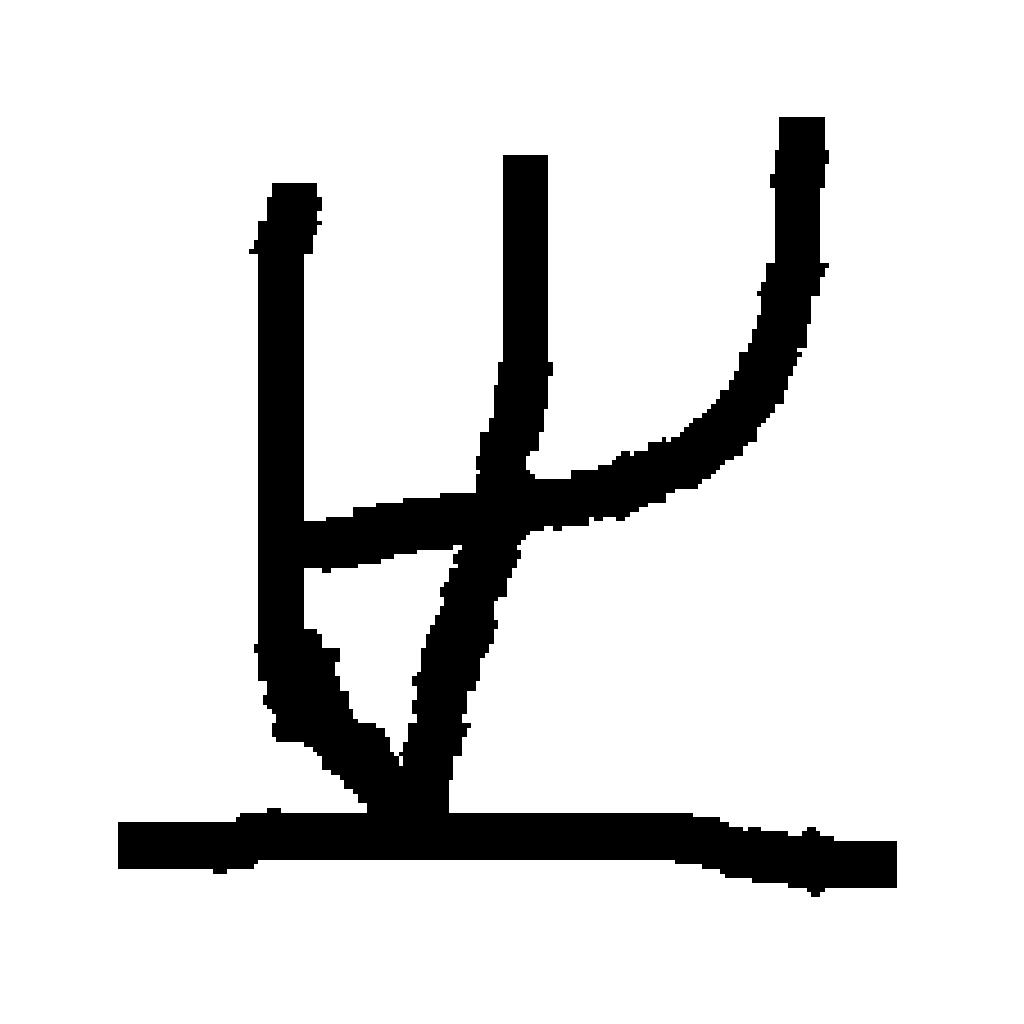} &
\imgfade[width=0.95\linewidth]{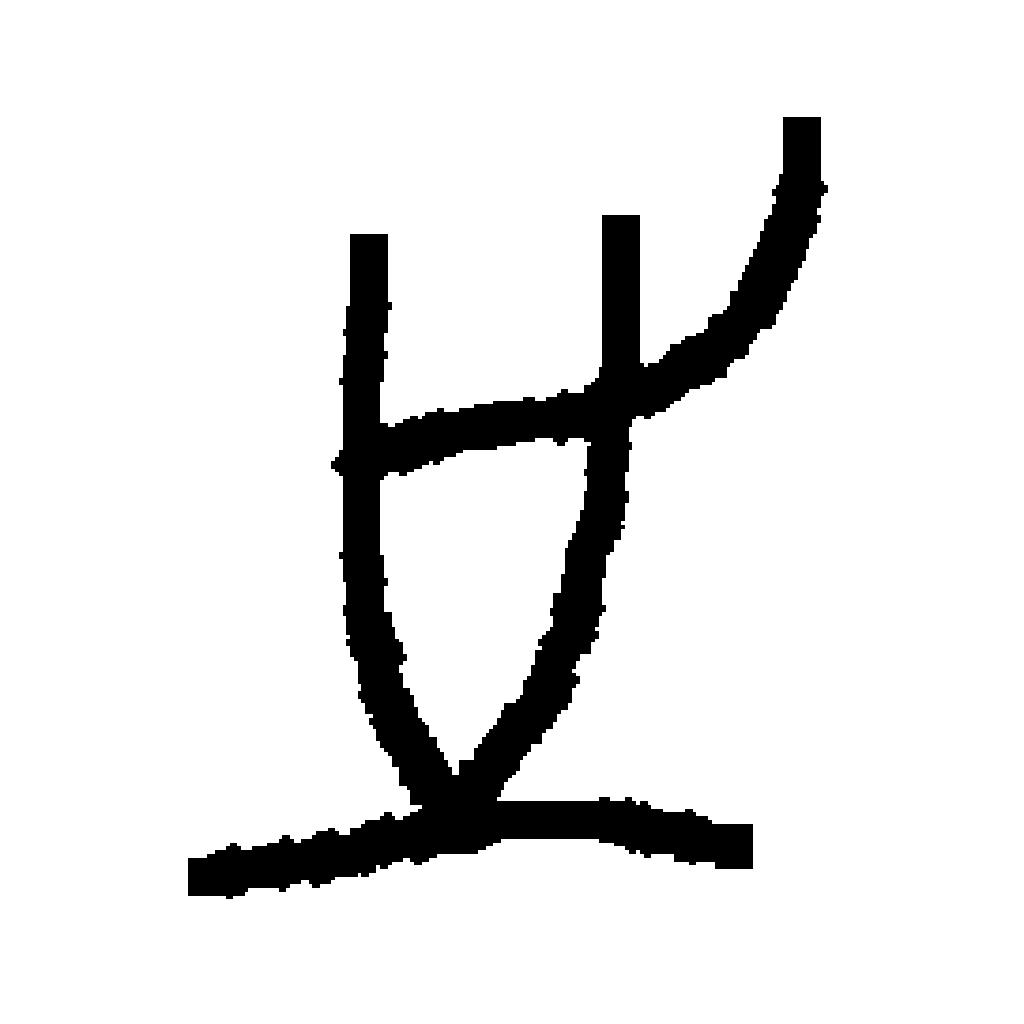} &
\imgfade[width=0.95\linewidth]{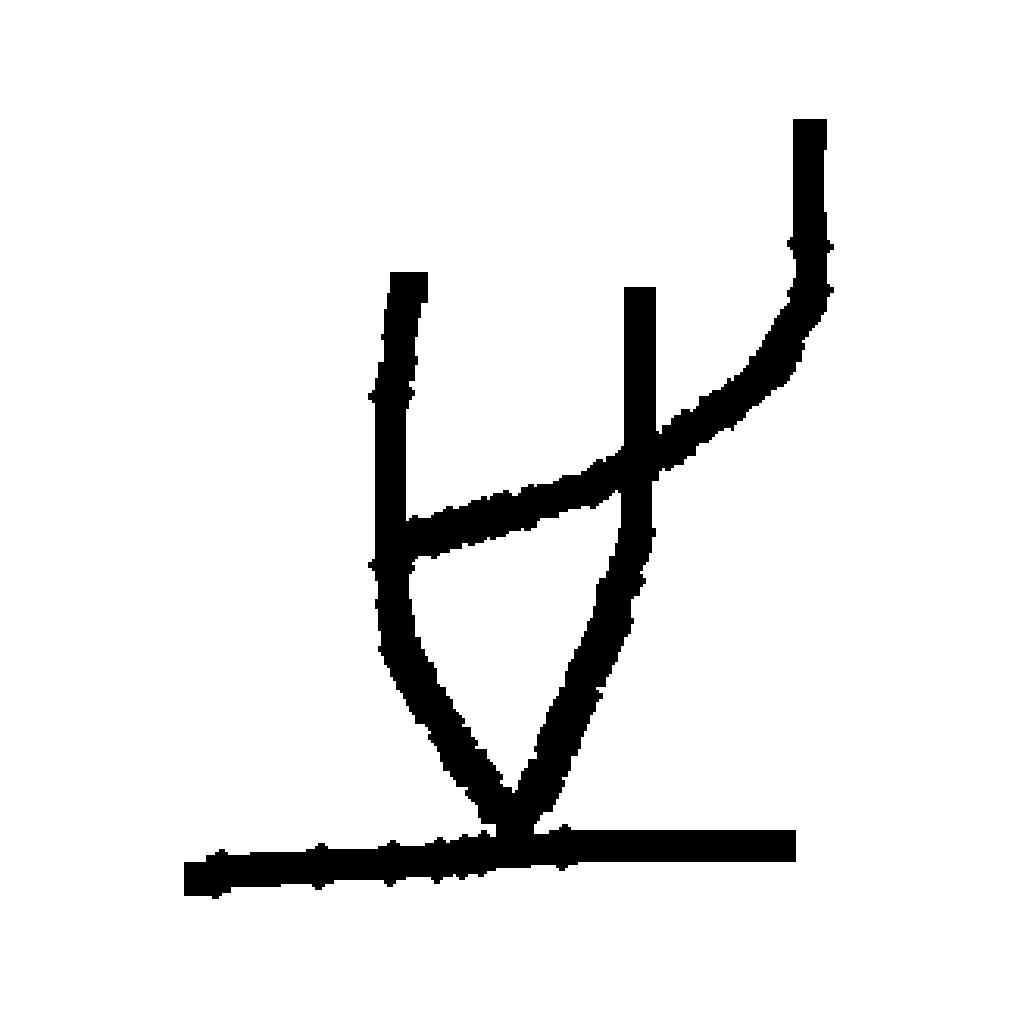} &
\includegraphics[width=0.95\linewidth]{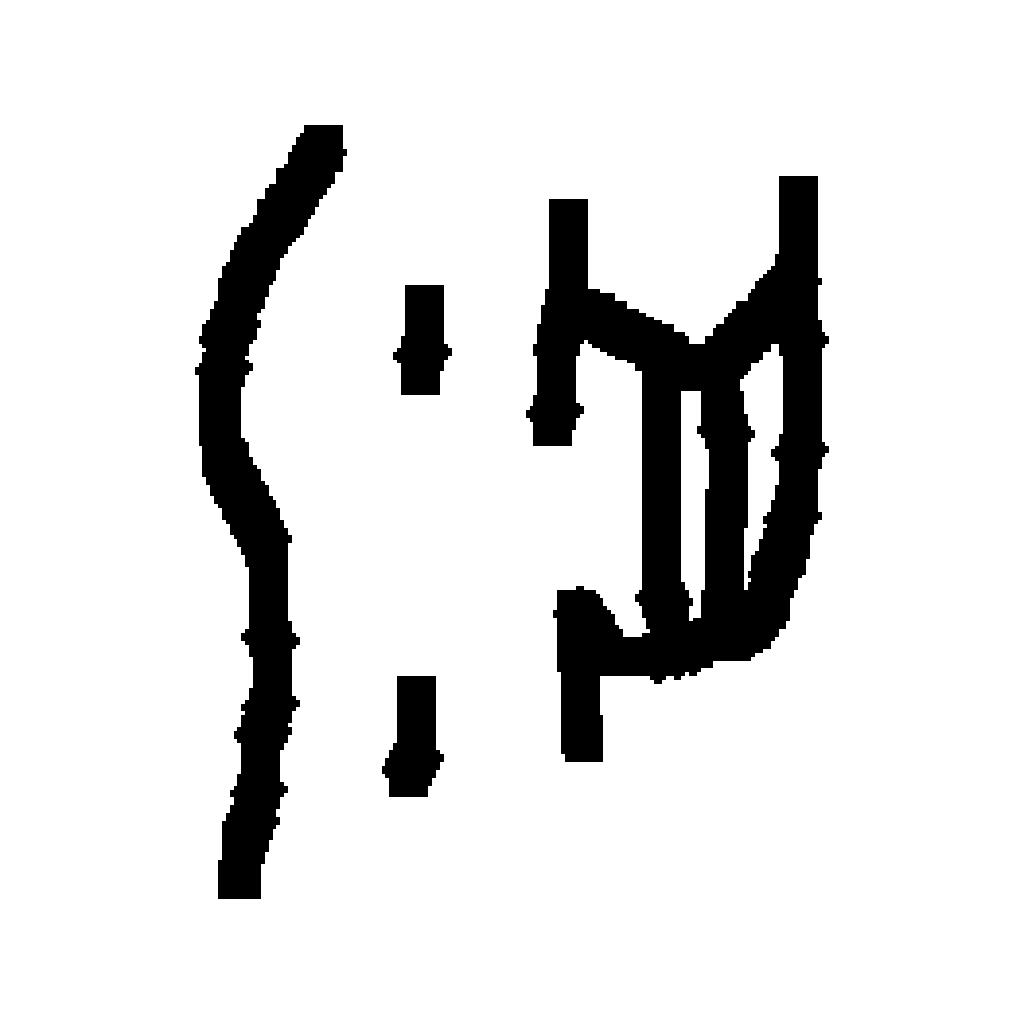} &
\includegraphics[width=0.95\linewidth]{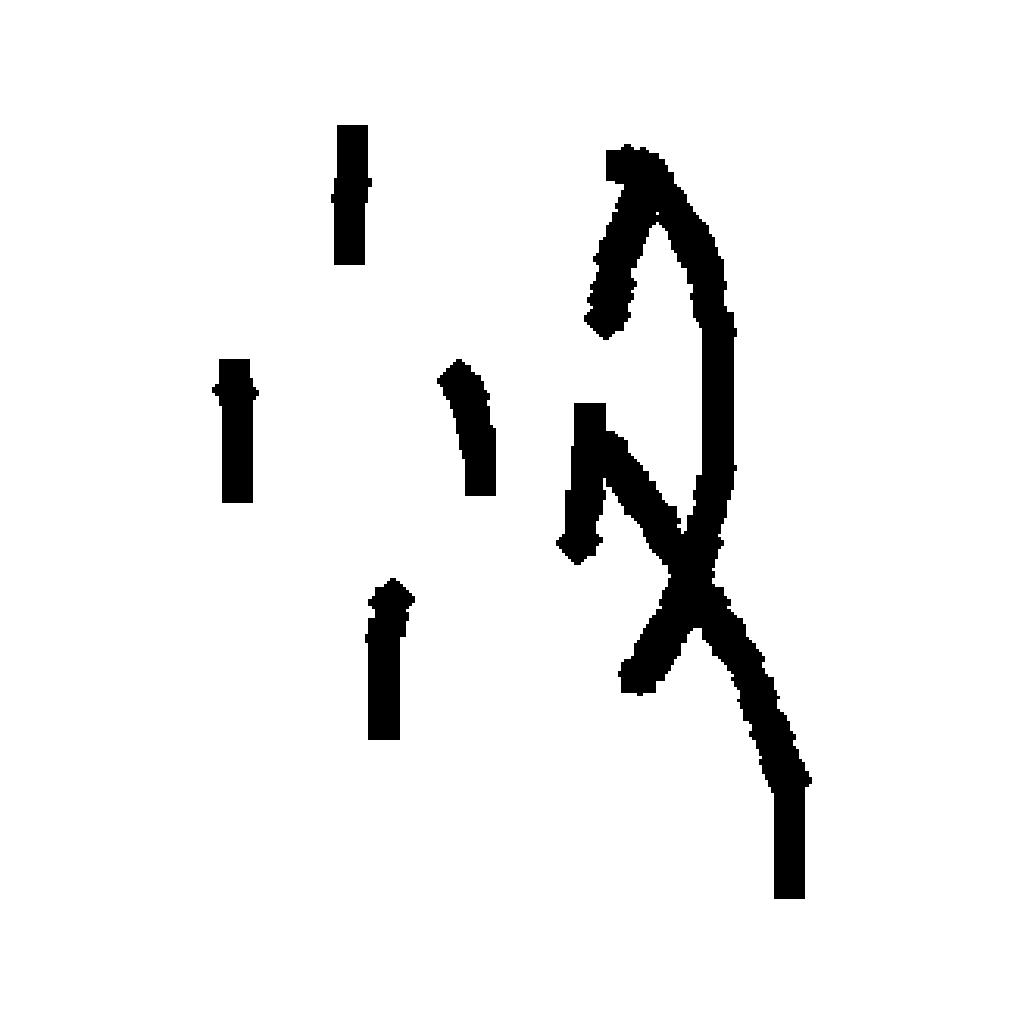} &
\imgwithbox[width=0.95\linewidth]{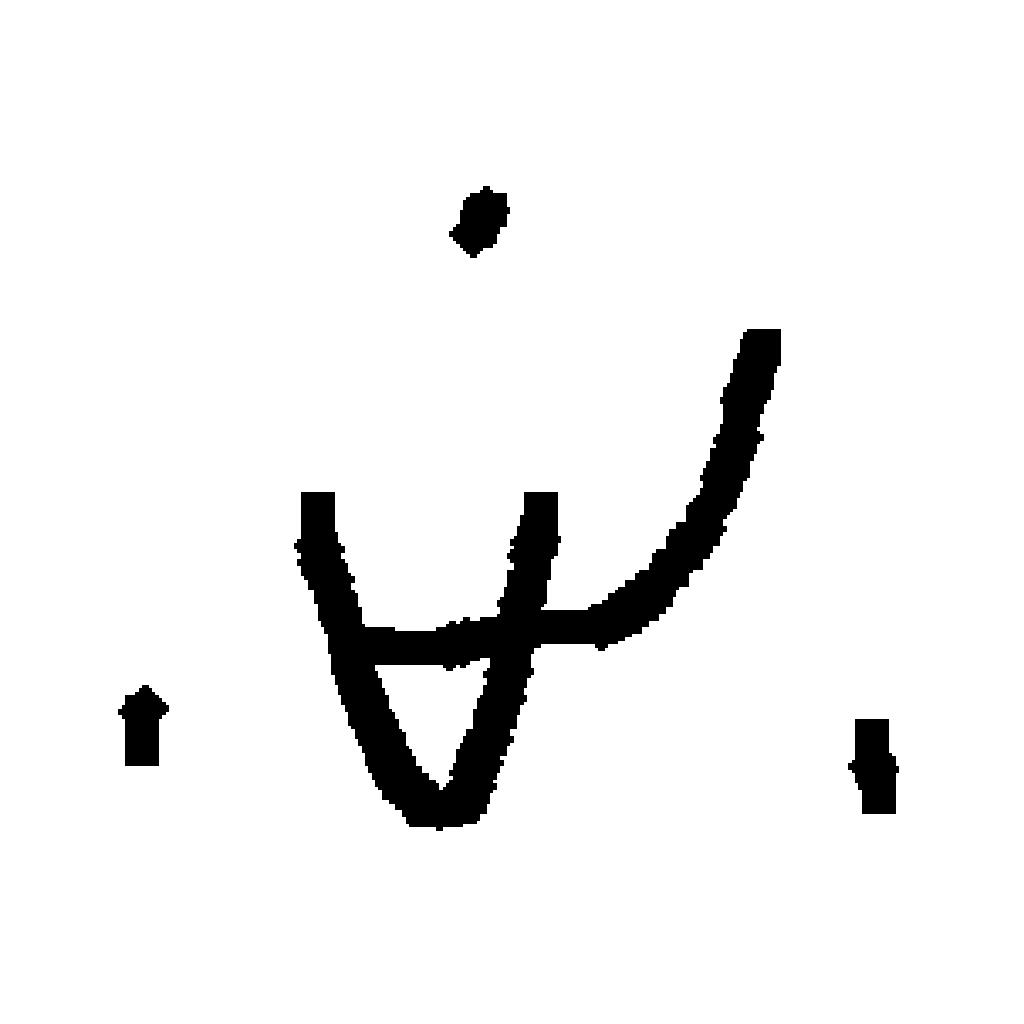} &
\includegraphics[width=0.95\linewidth]{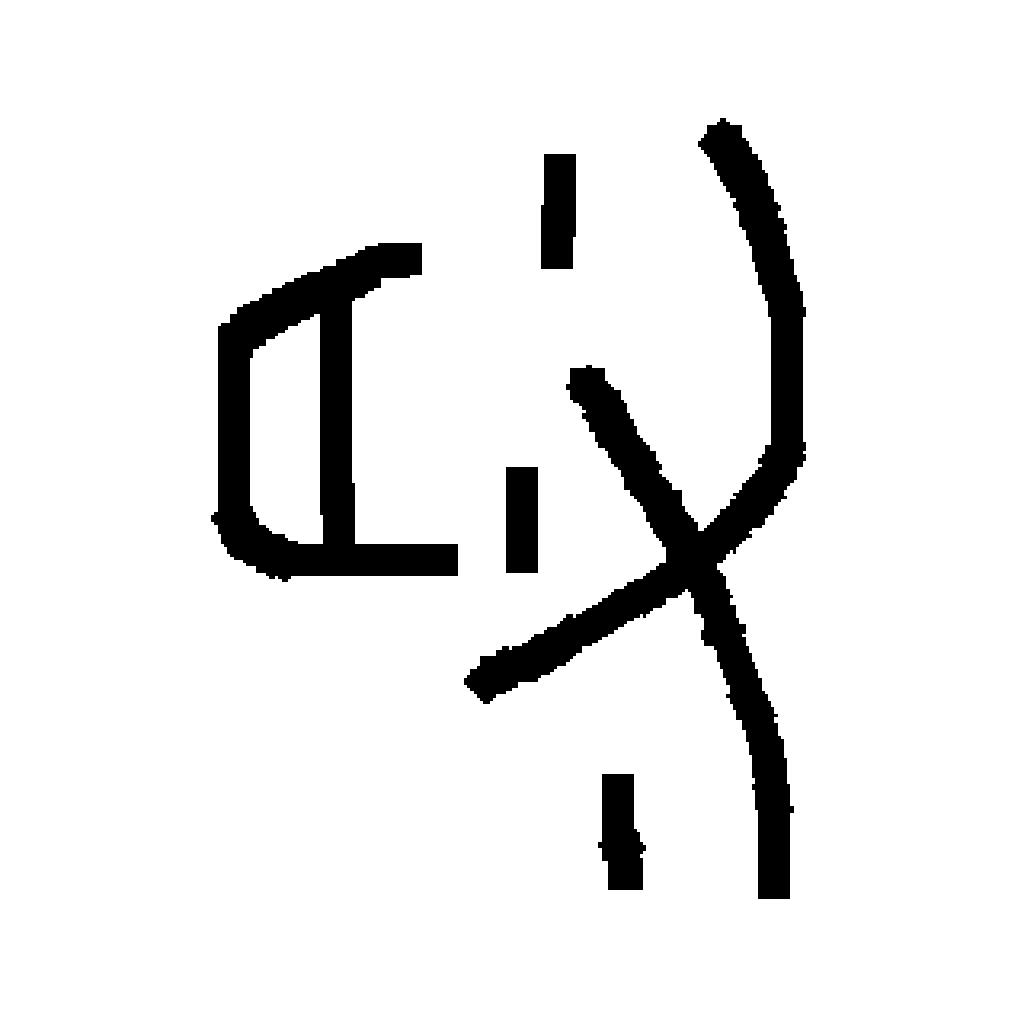} &
\includegraphics[width=0.95\linewidth]{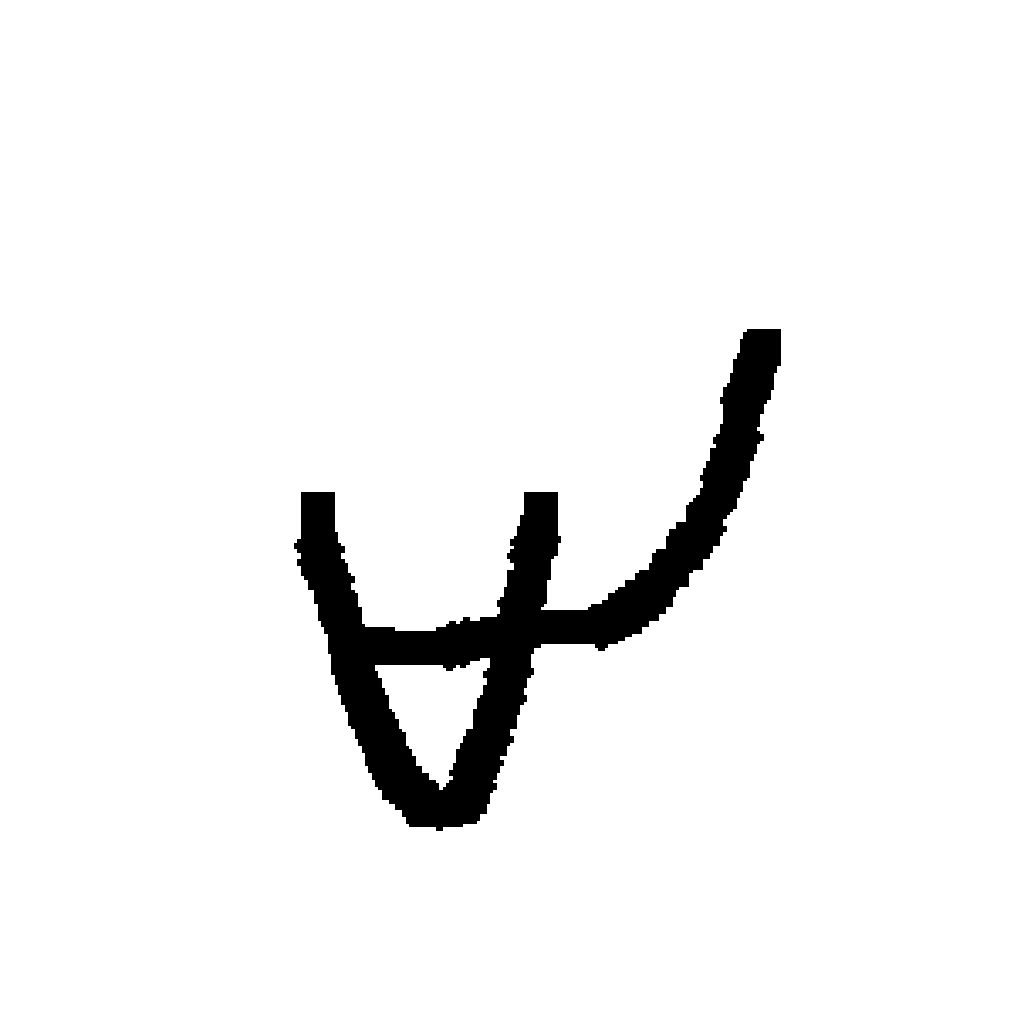} \\
2 &
\includegraphics[width=0.95\linewidth]{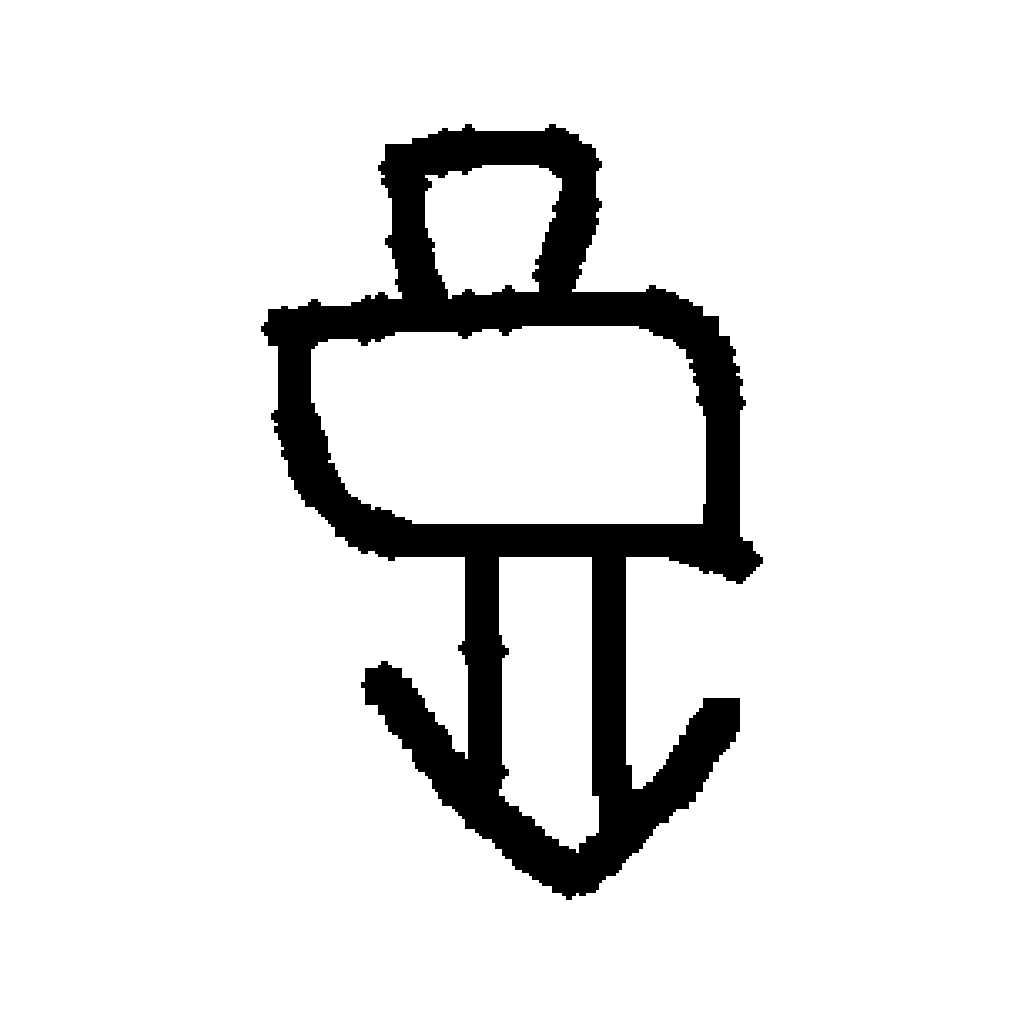} &
\imgfade[width=0.95\linewidth]{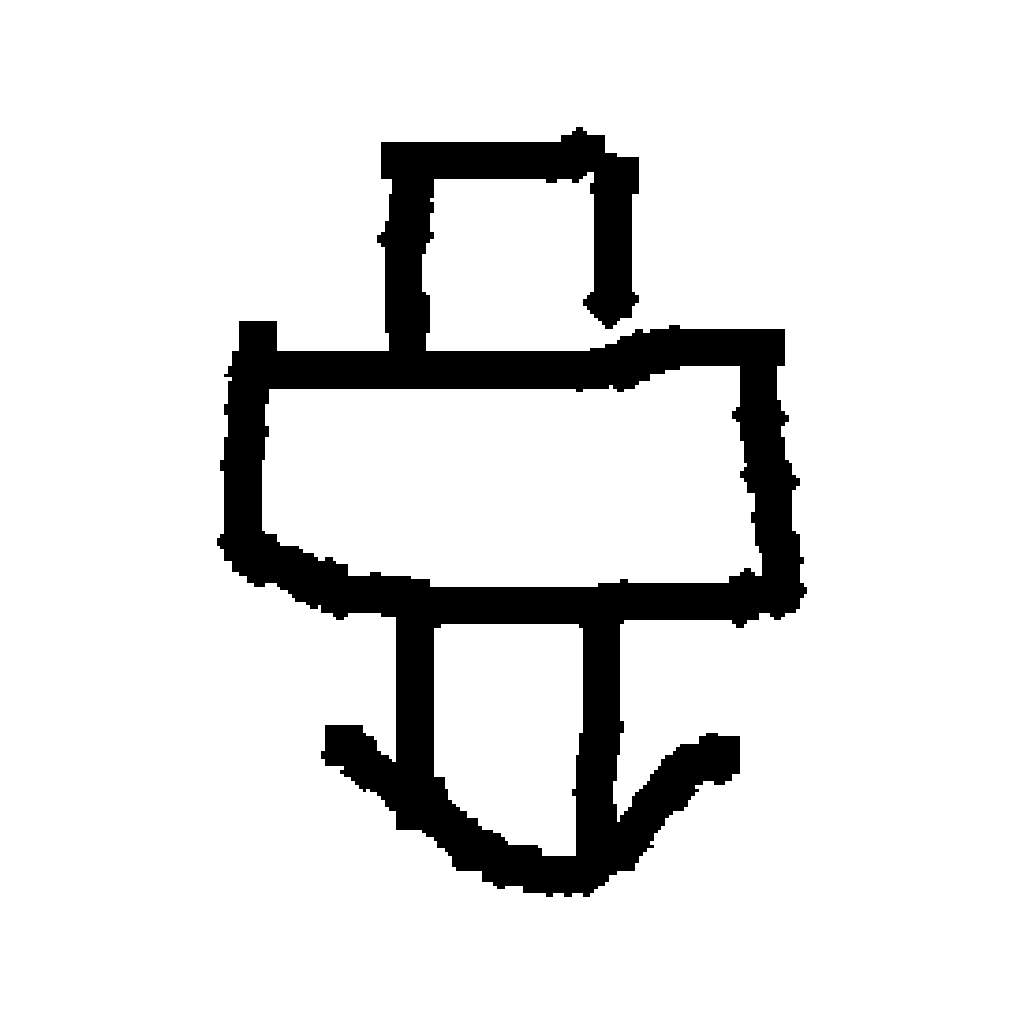} &
\imgfade[width=0.95\linewidth]{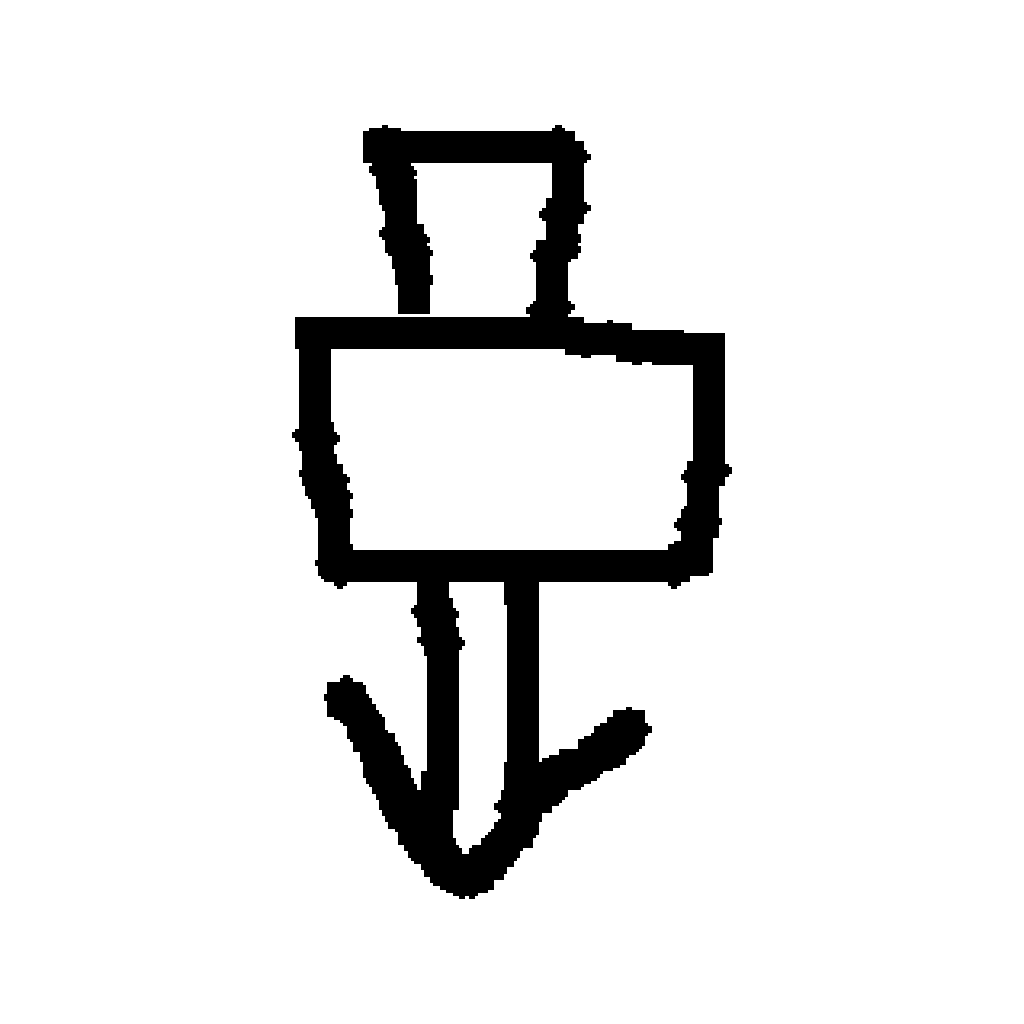} &
\imgfade[width=0.95\linewidth]{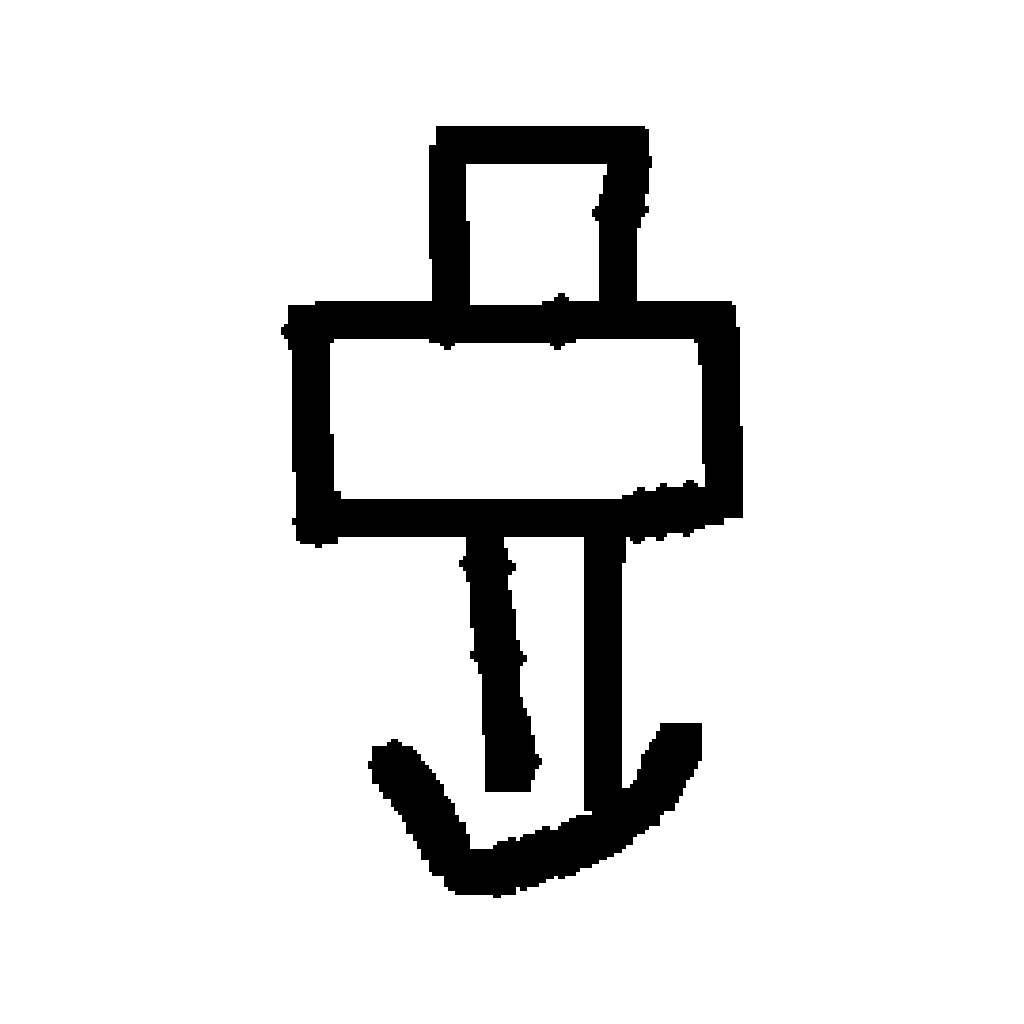} &
\imgfade[width=0.95\linewidth]{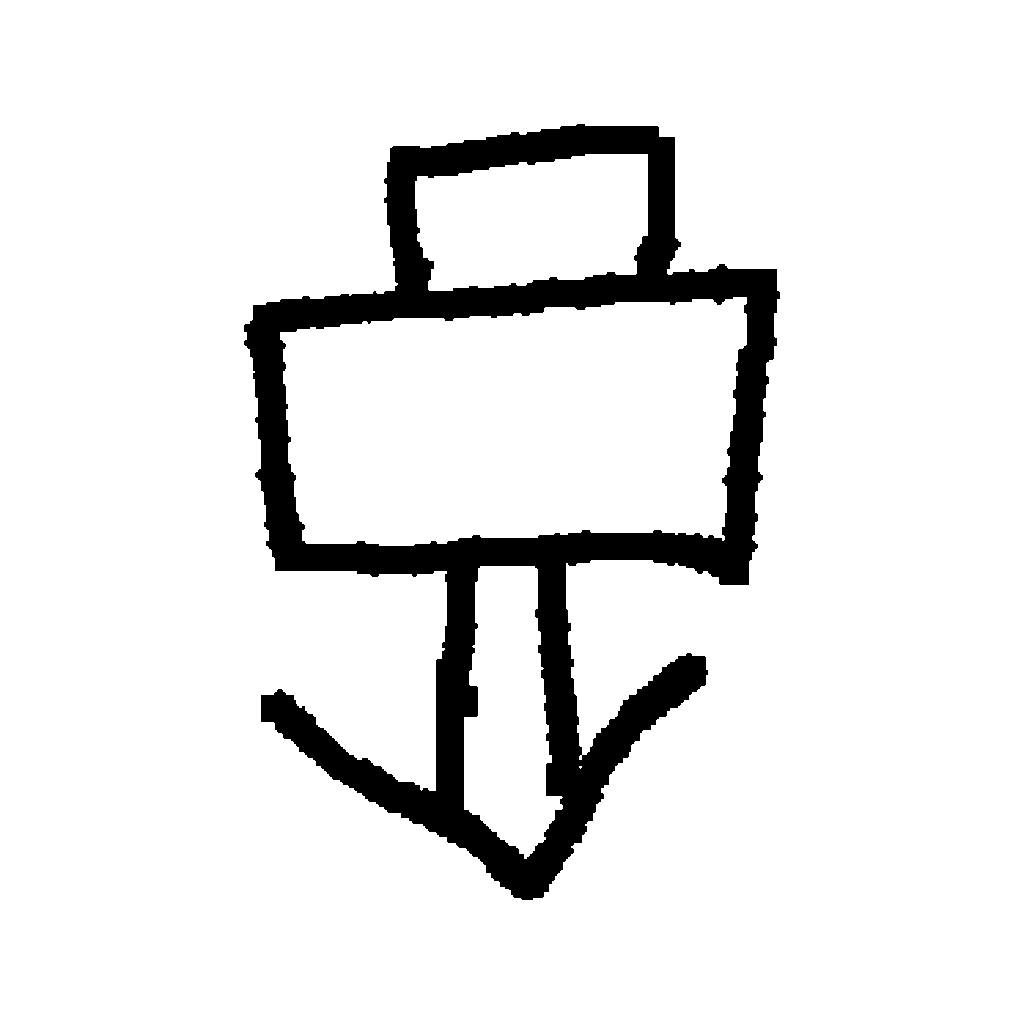} &
\imgfade[width=0.95\linewidth]{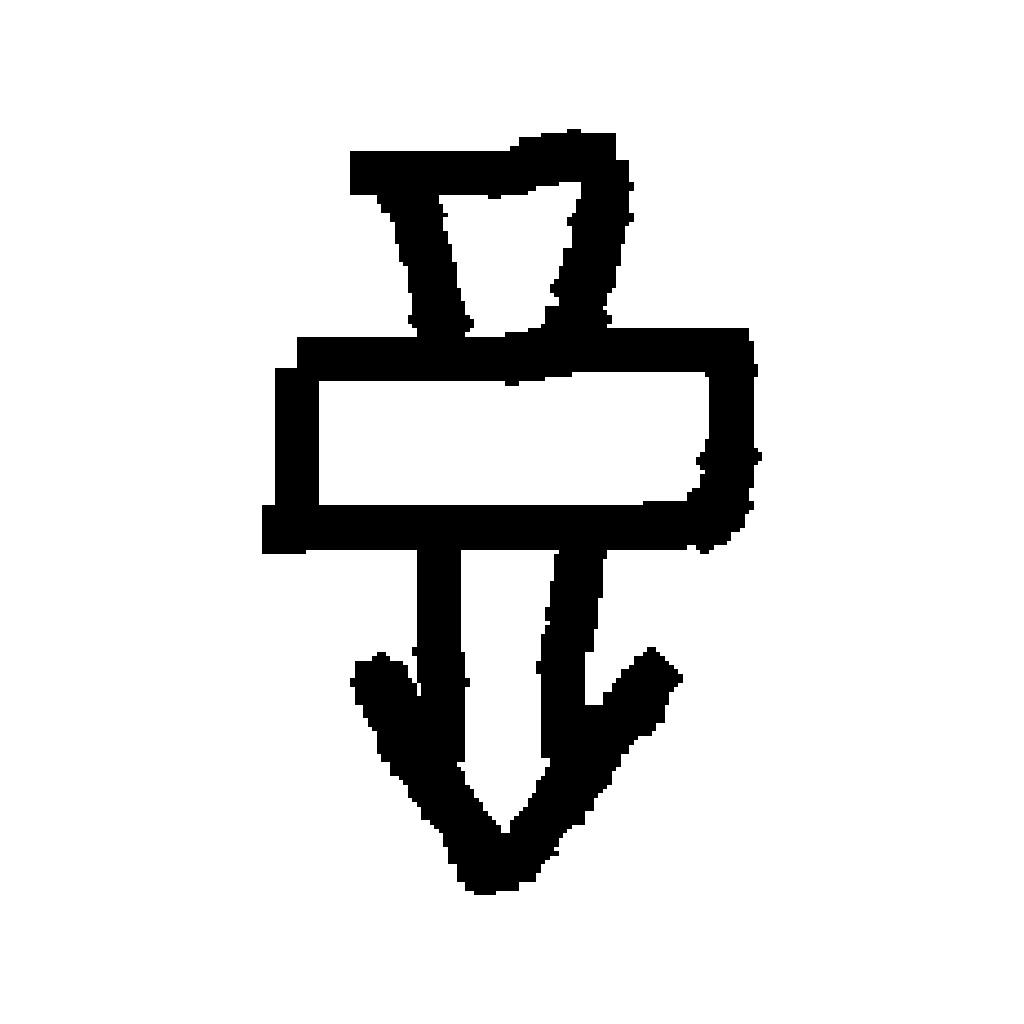} &
\imgfade[width=0.95\linewidth]{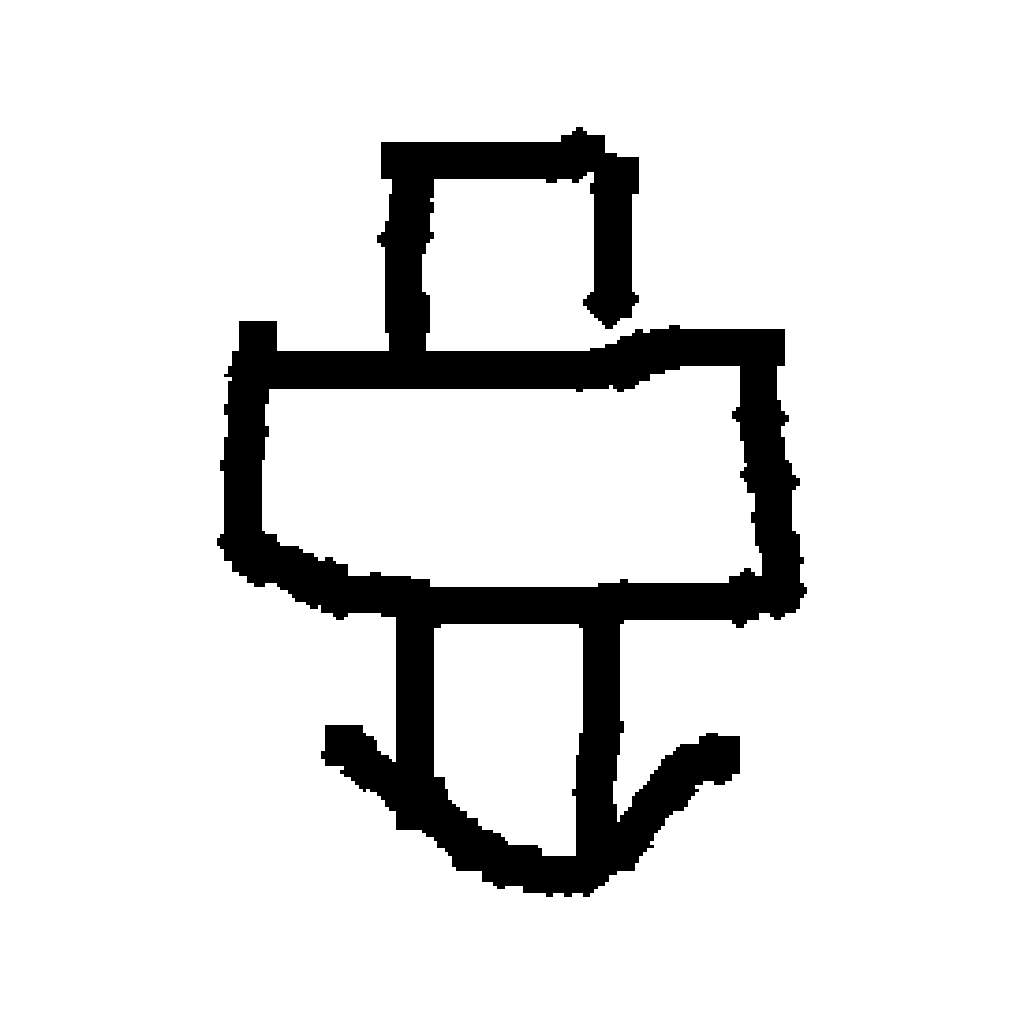} &
\imgfade[width=0.95\linewidth]{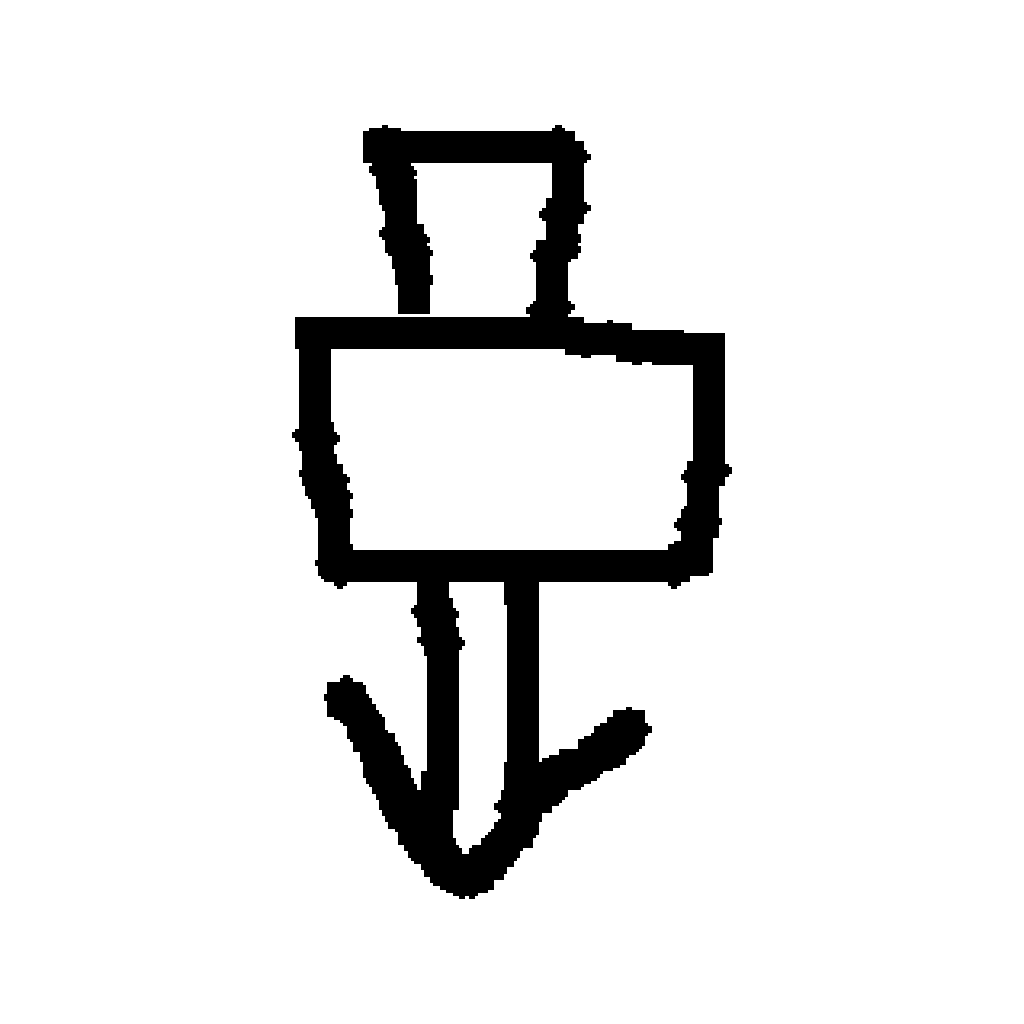} &
\imgfade[width=0.95\linewidth]{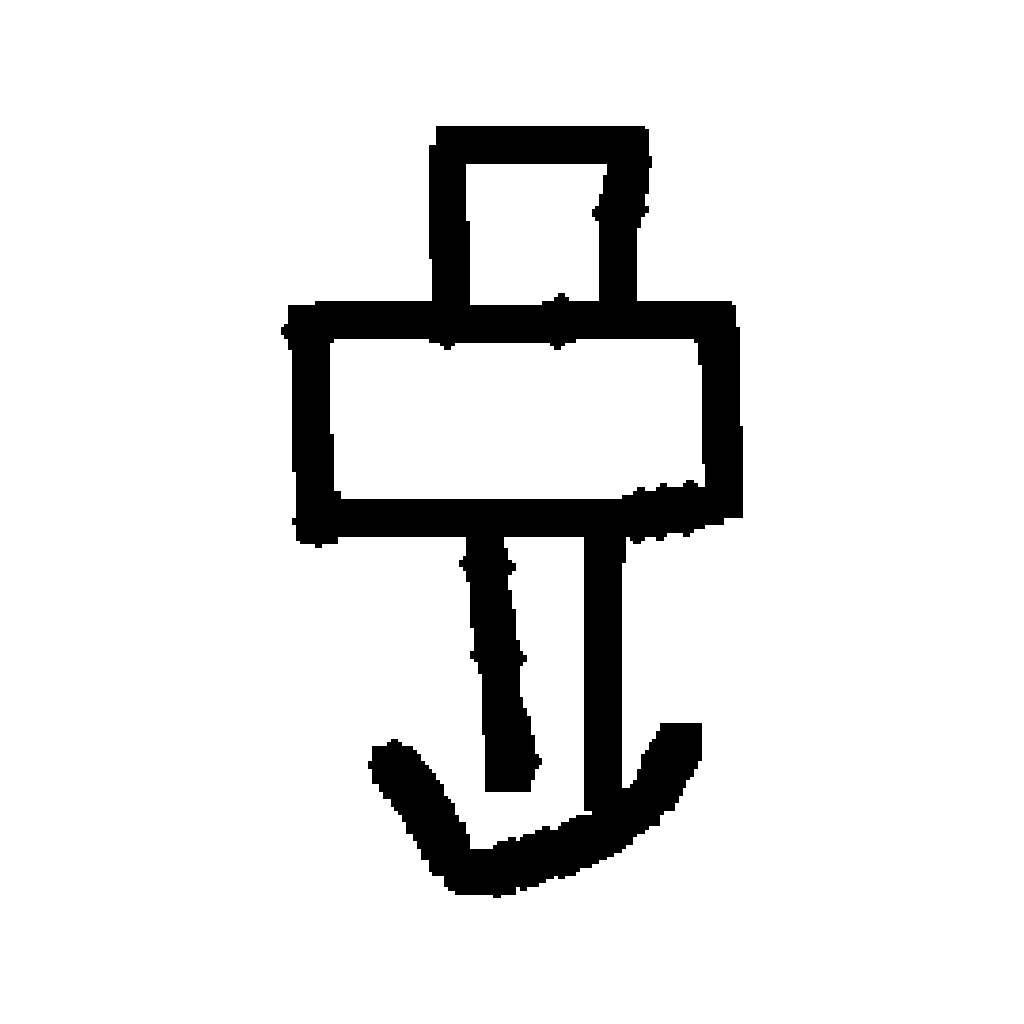} &
\imgwithbox[width=0.95\linewidth]{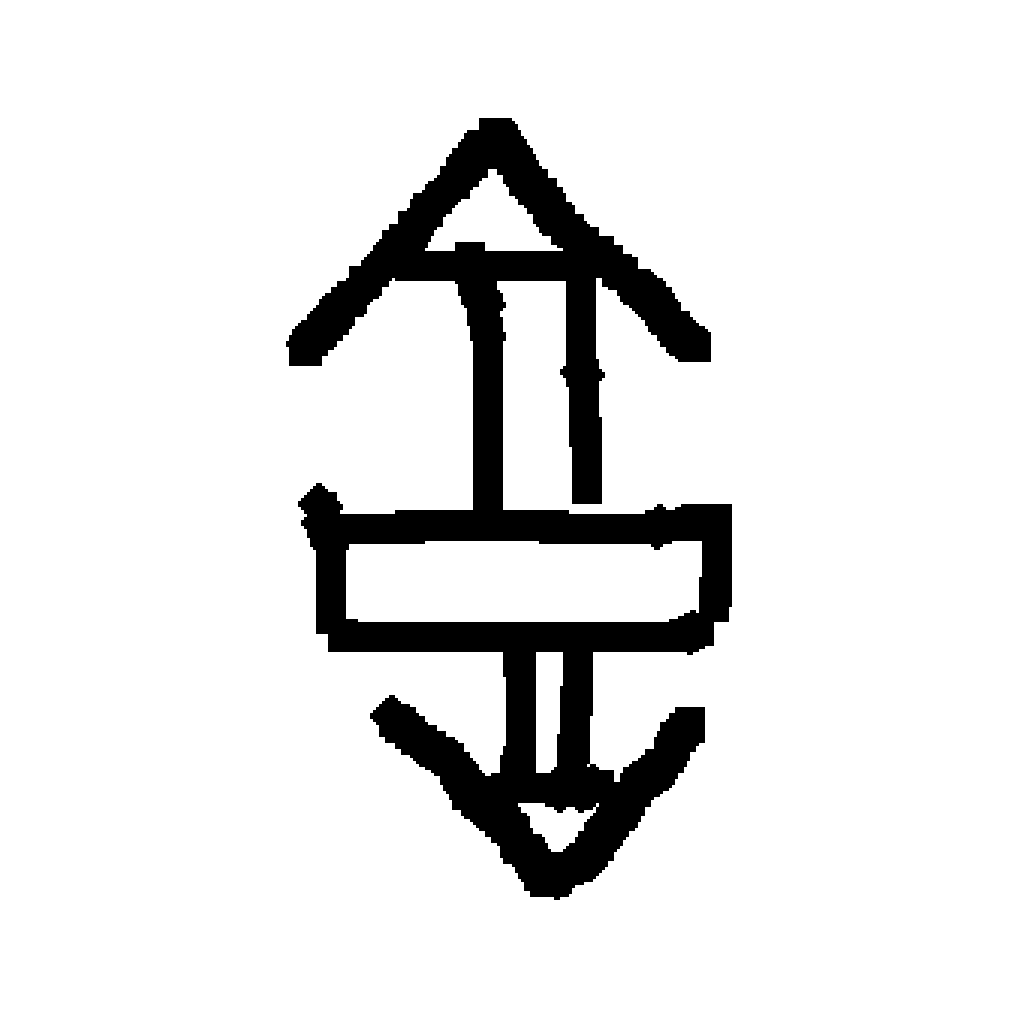} &
\imgfade[width=0.95\linewidth]{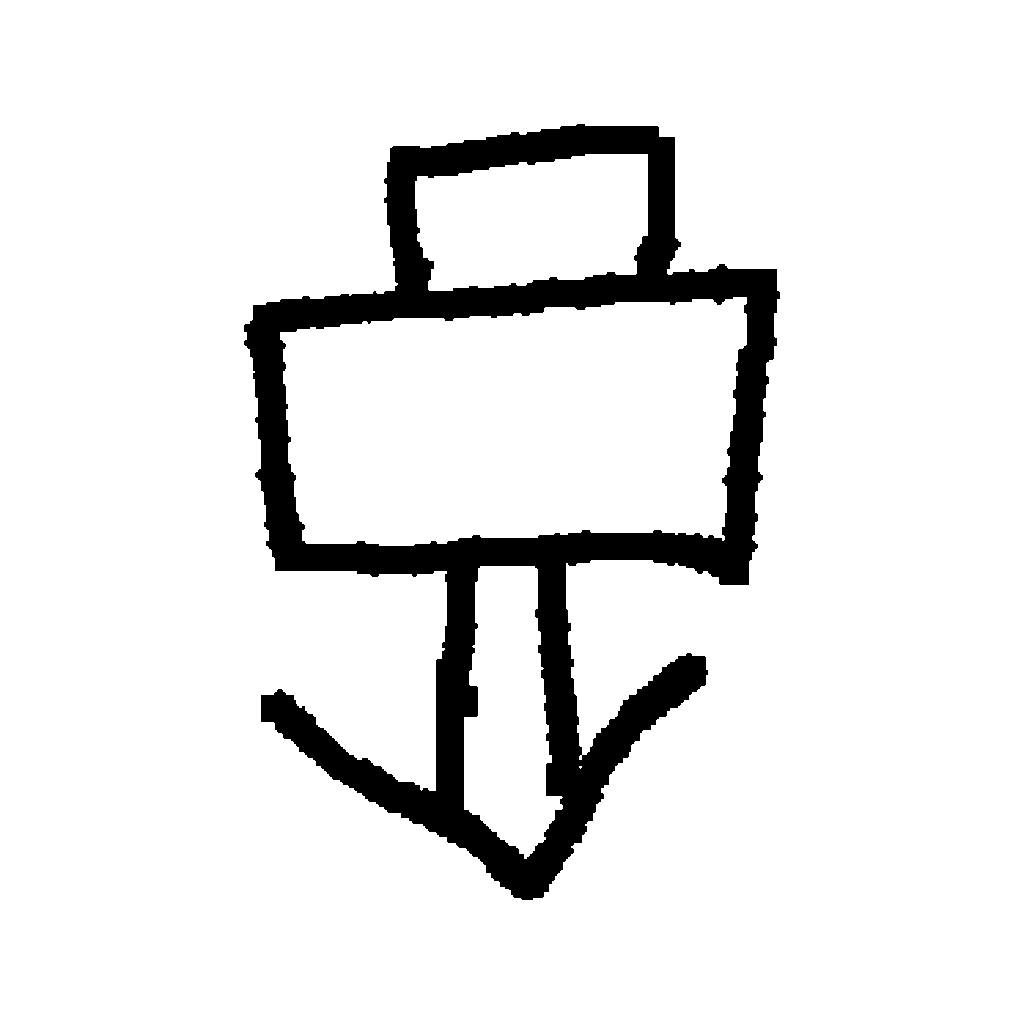} &
\includegraphics[width=0.95\linewidth]{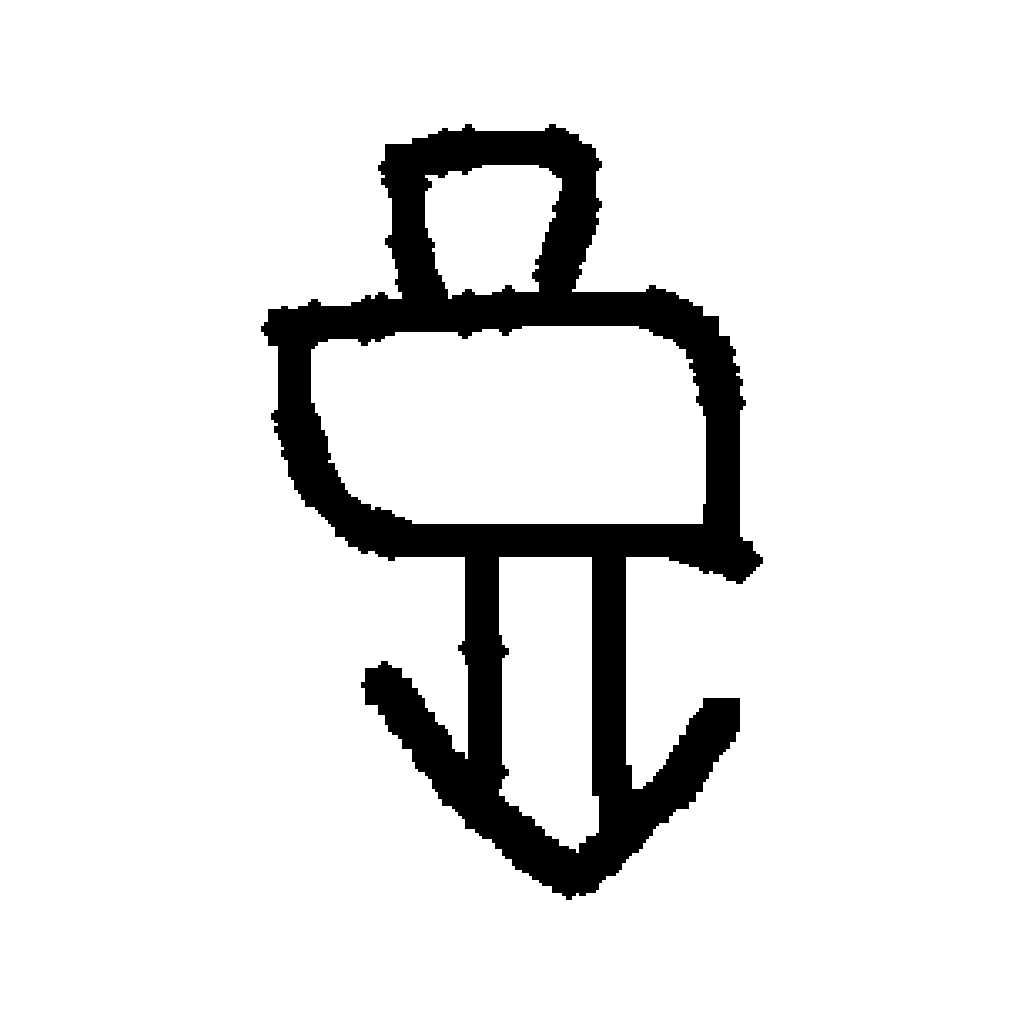} \\
\midrule\noalign{\vskip -3pt}\rowcolor{gray!20}\multicolumn{13}{c}{\textbf{Dongba Pictographs}} \\\noalign{\vskip -2pt}\midrule
3 &
\includegraphics[width=0.95\linewidth]{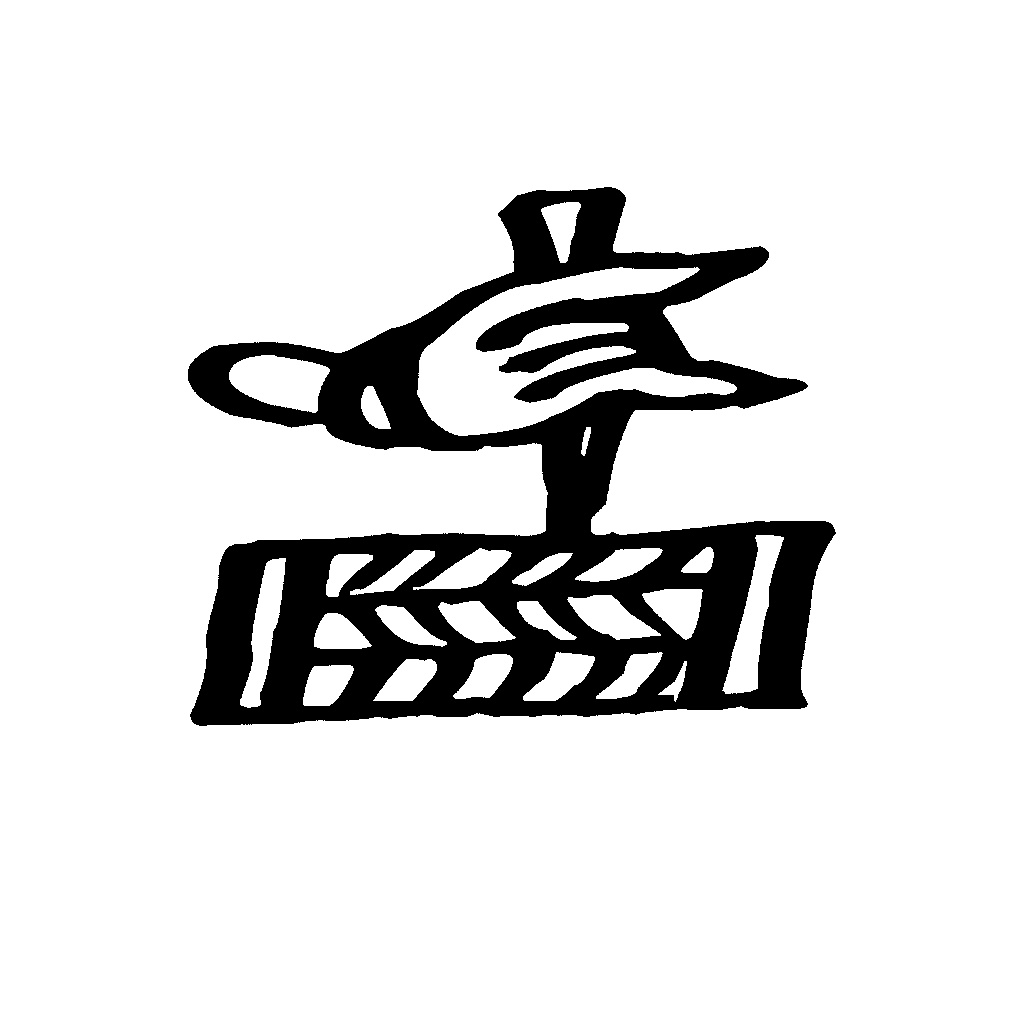} &
\includegraphics[width=0.95\linewidth]{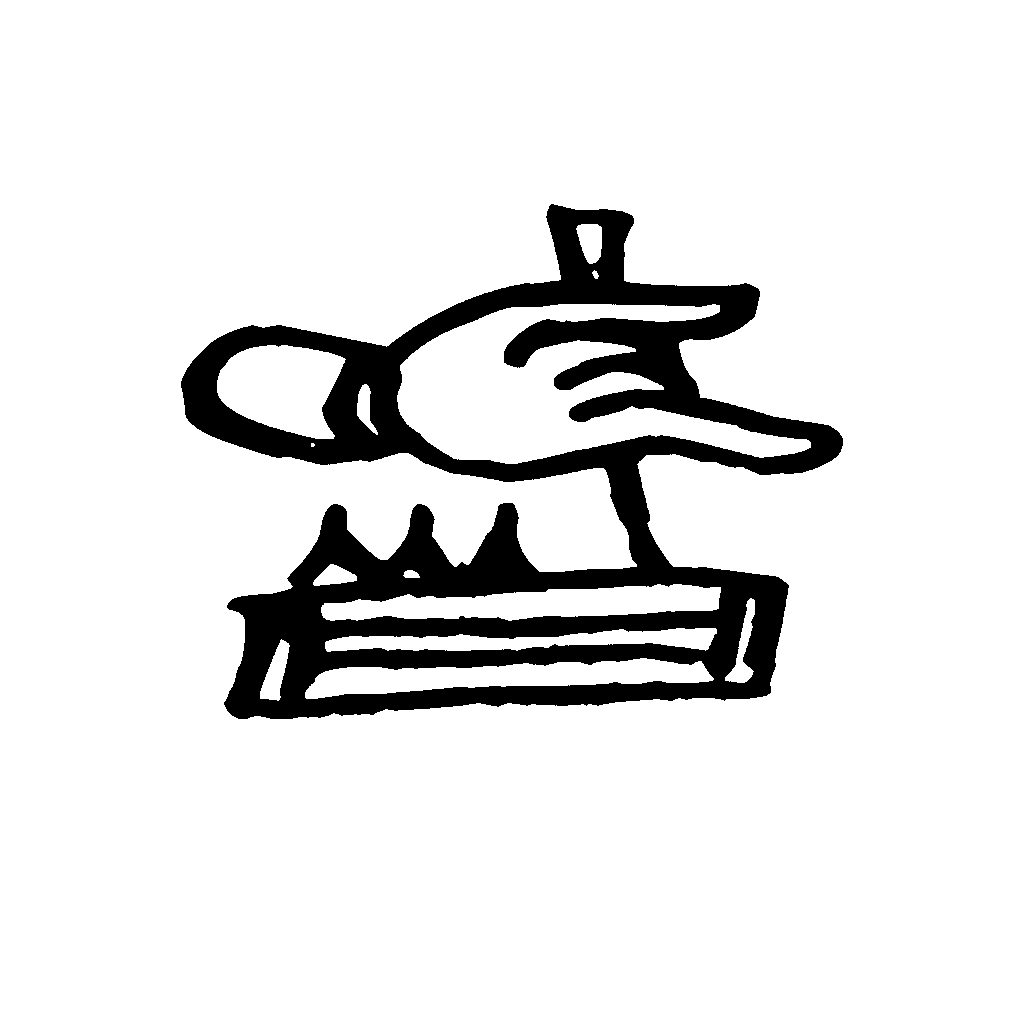} &
\imgwithbox[width=0.95\linewidth]{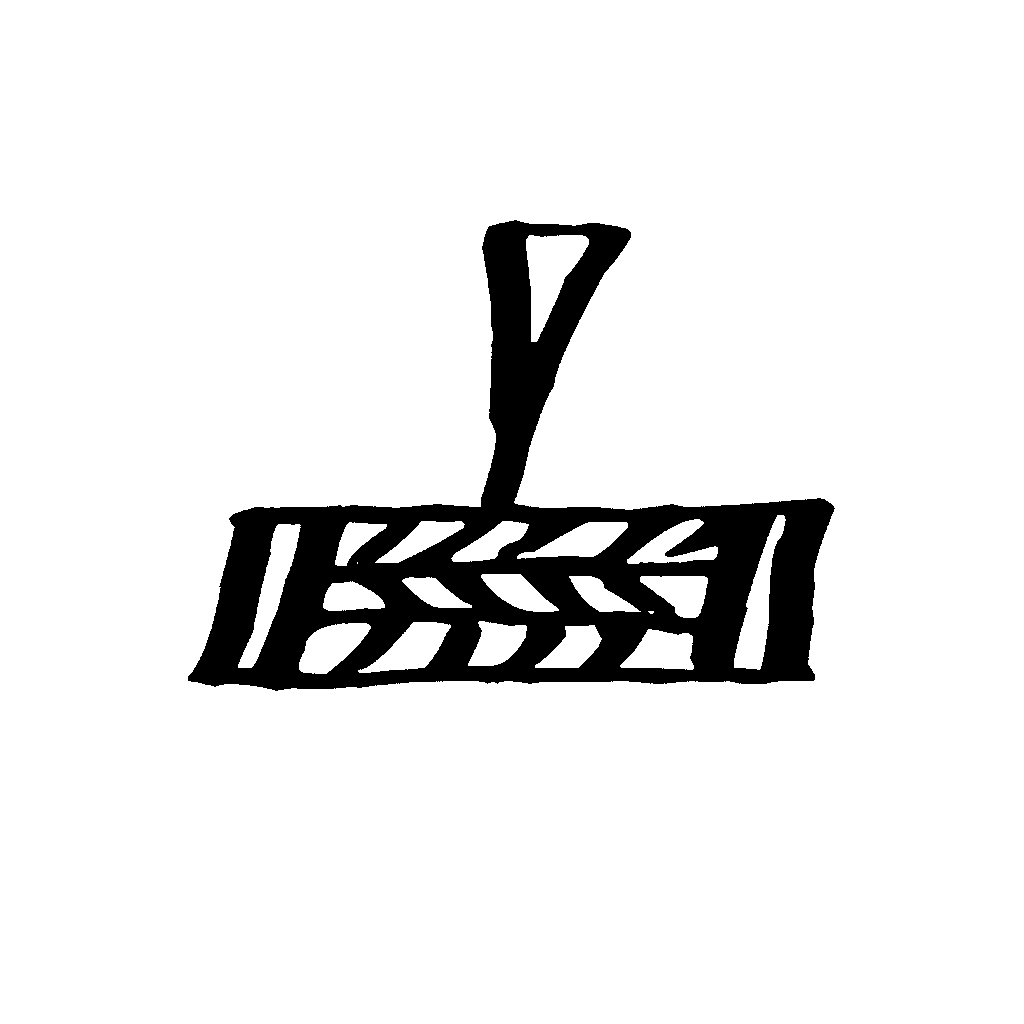} &
\includegraphics[width=0.95\linewidth]{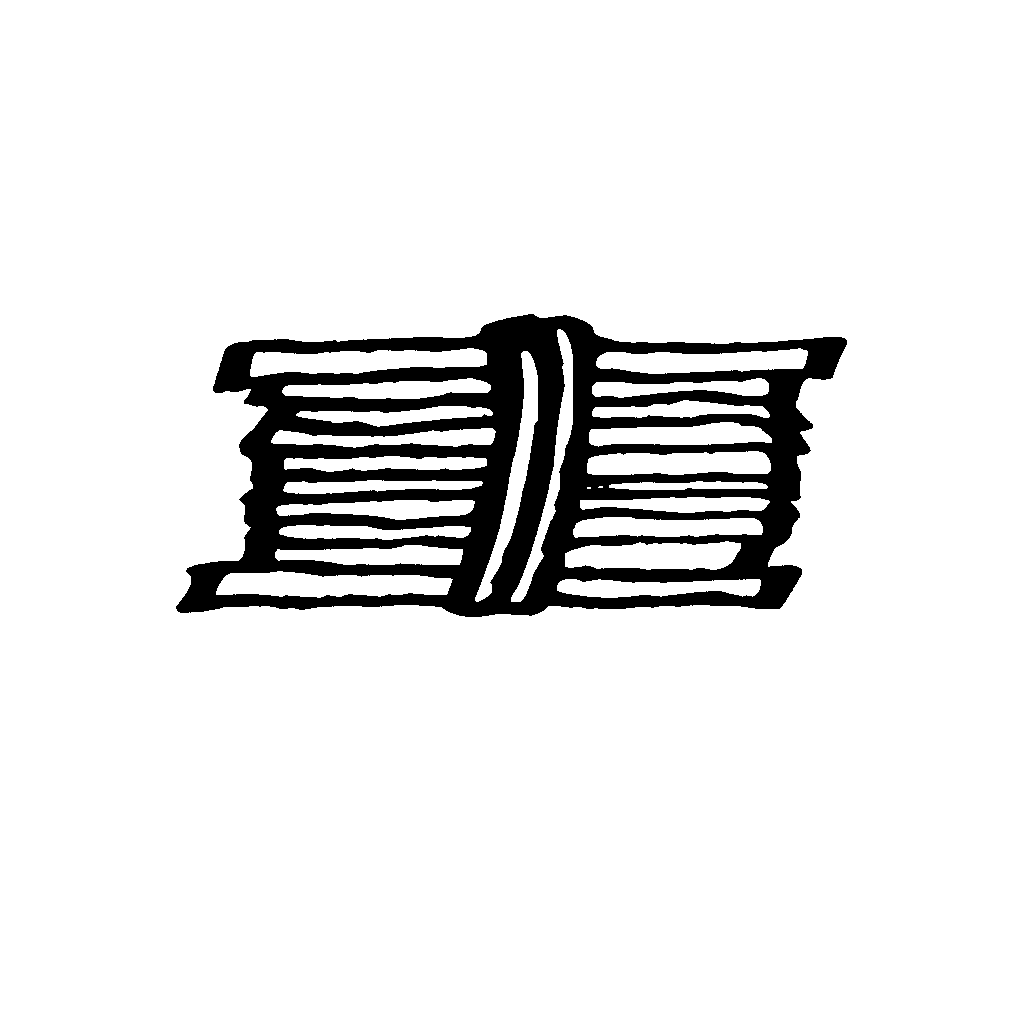} &
\includegraphics[width=0.95\linewidth]{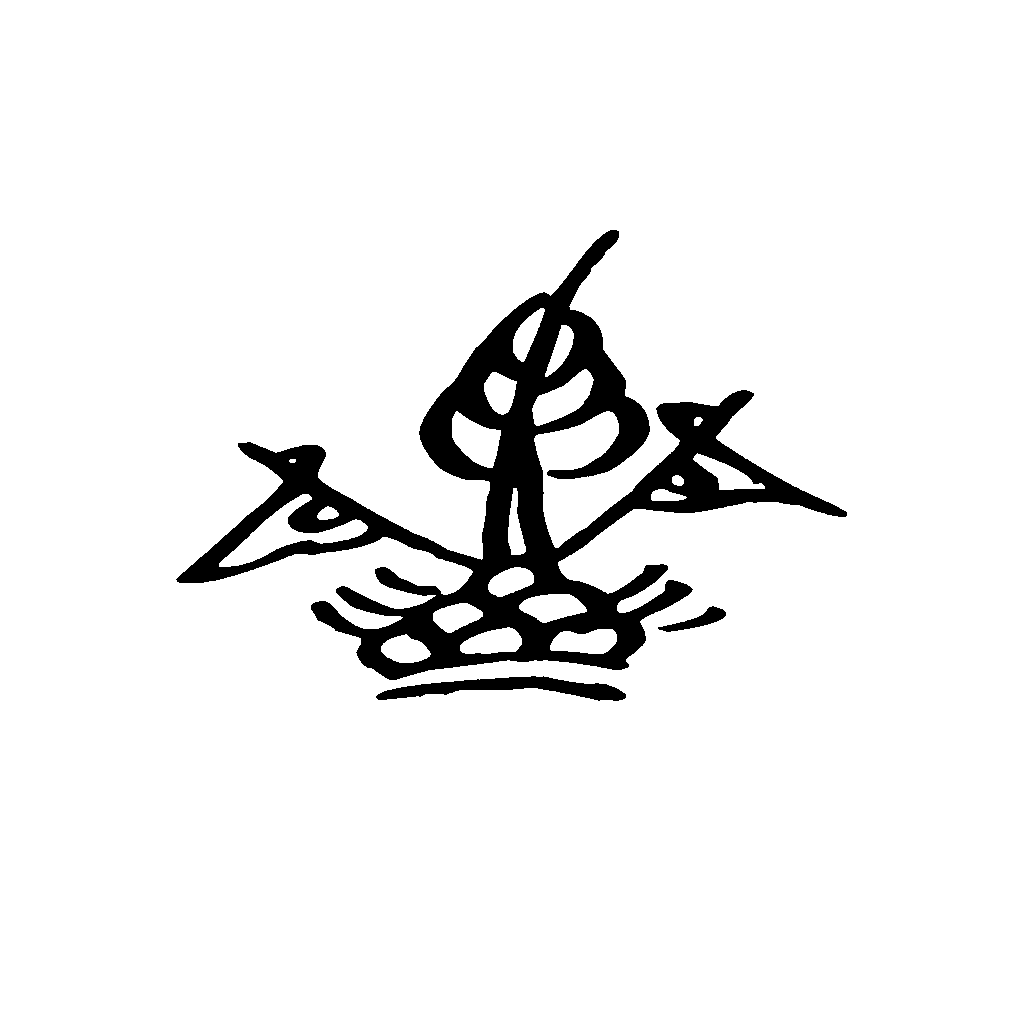} &
\includegraphics[width=0.95\linewidth]{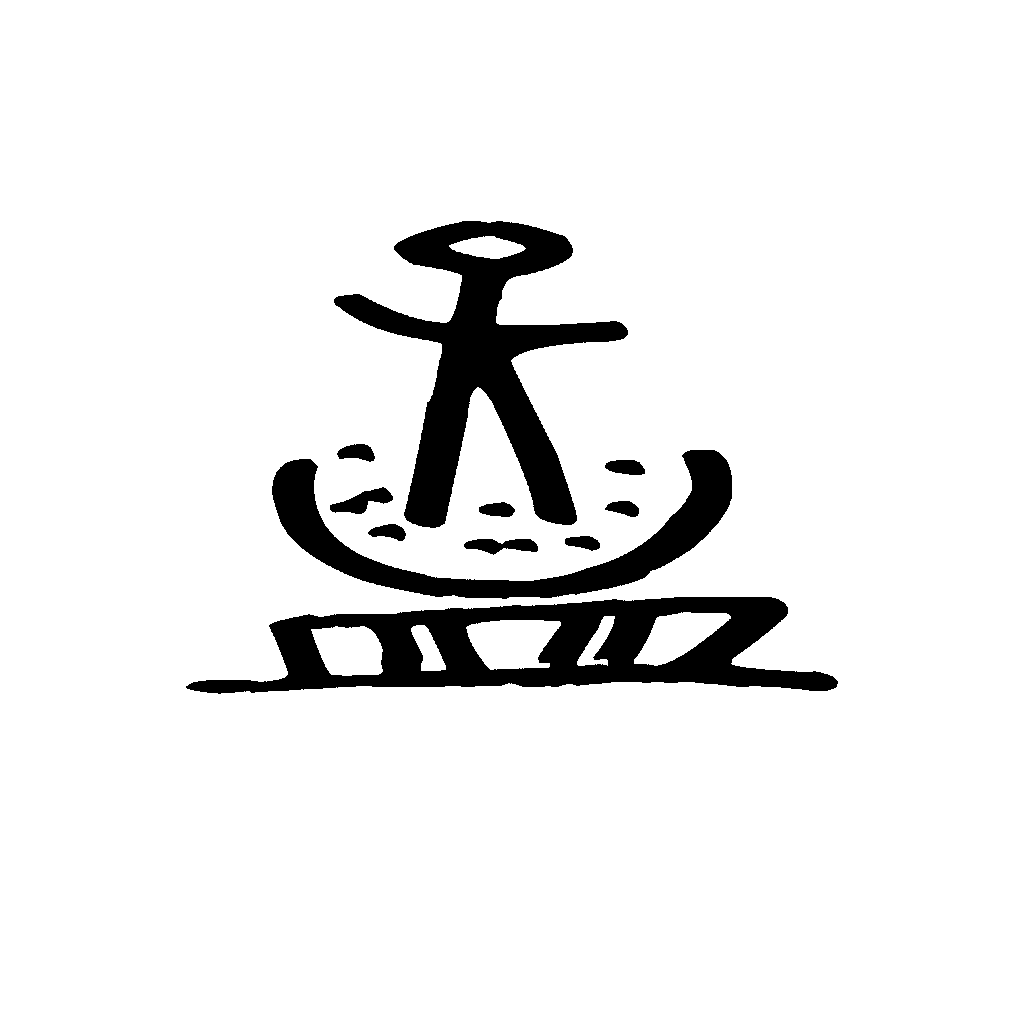} &
\includegraphics[width=0.95\linewidth]{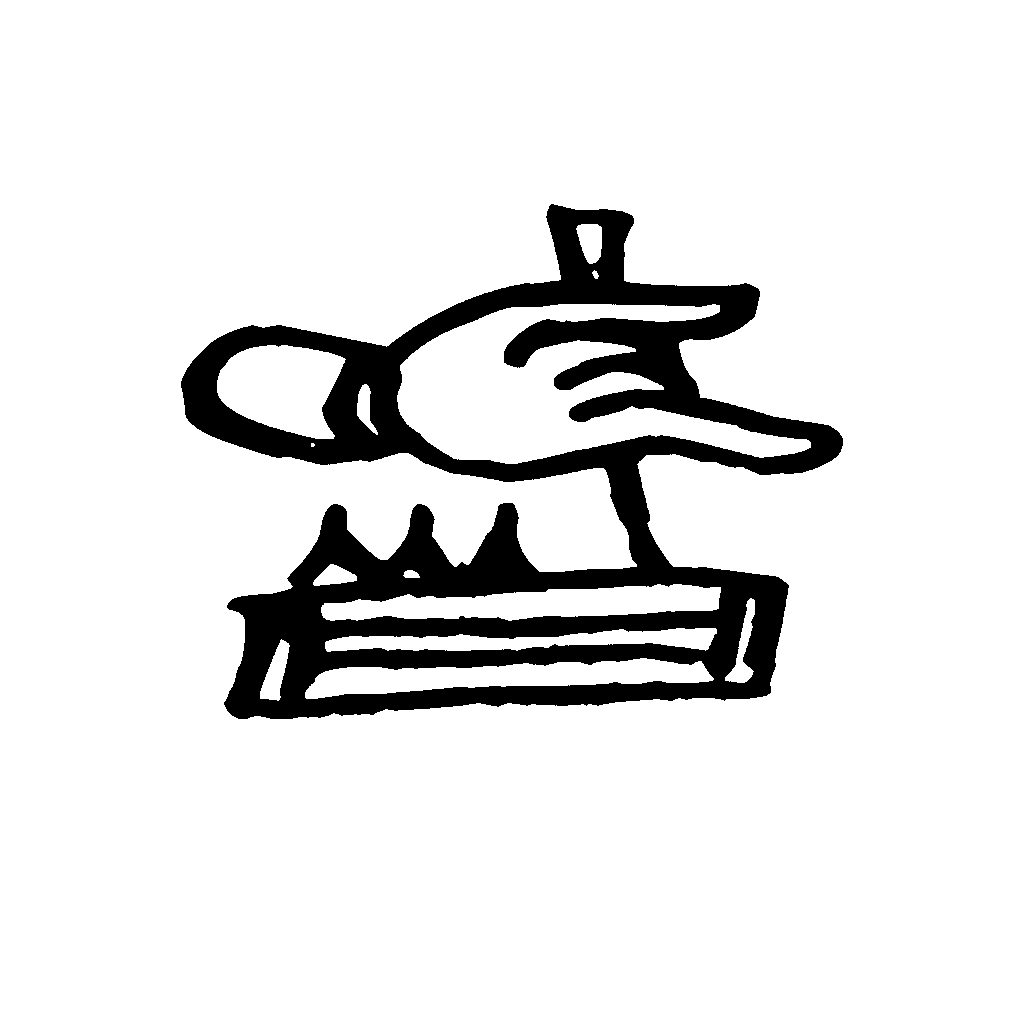} &
\imgwithbox[width=0.95\linewidth]{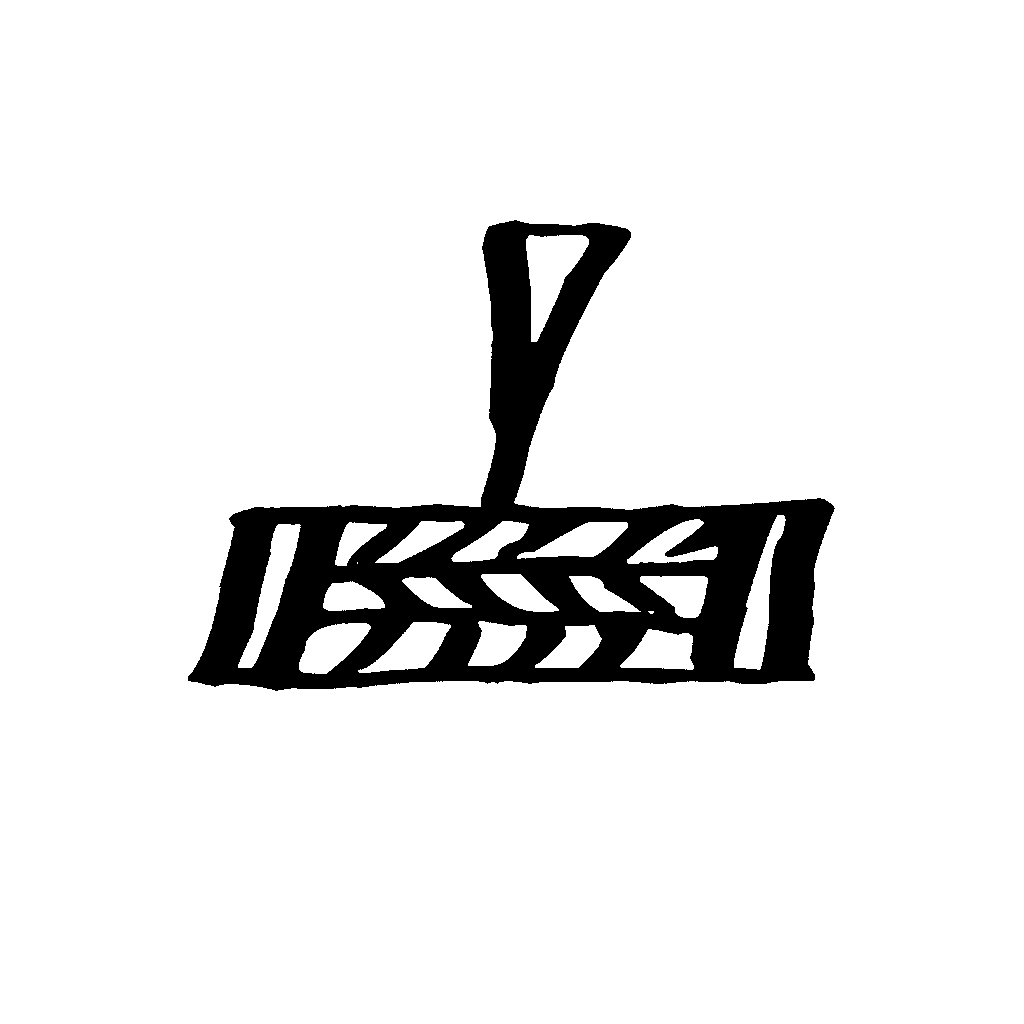} &
\includegraphics[width=0.95\linewidth]{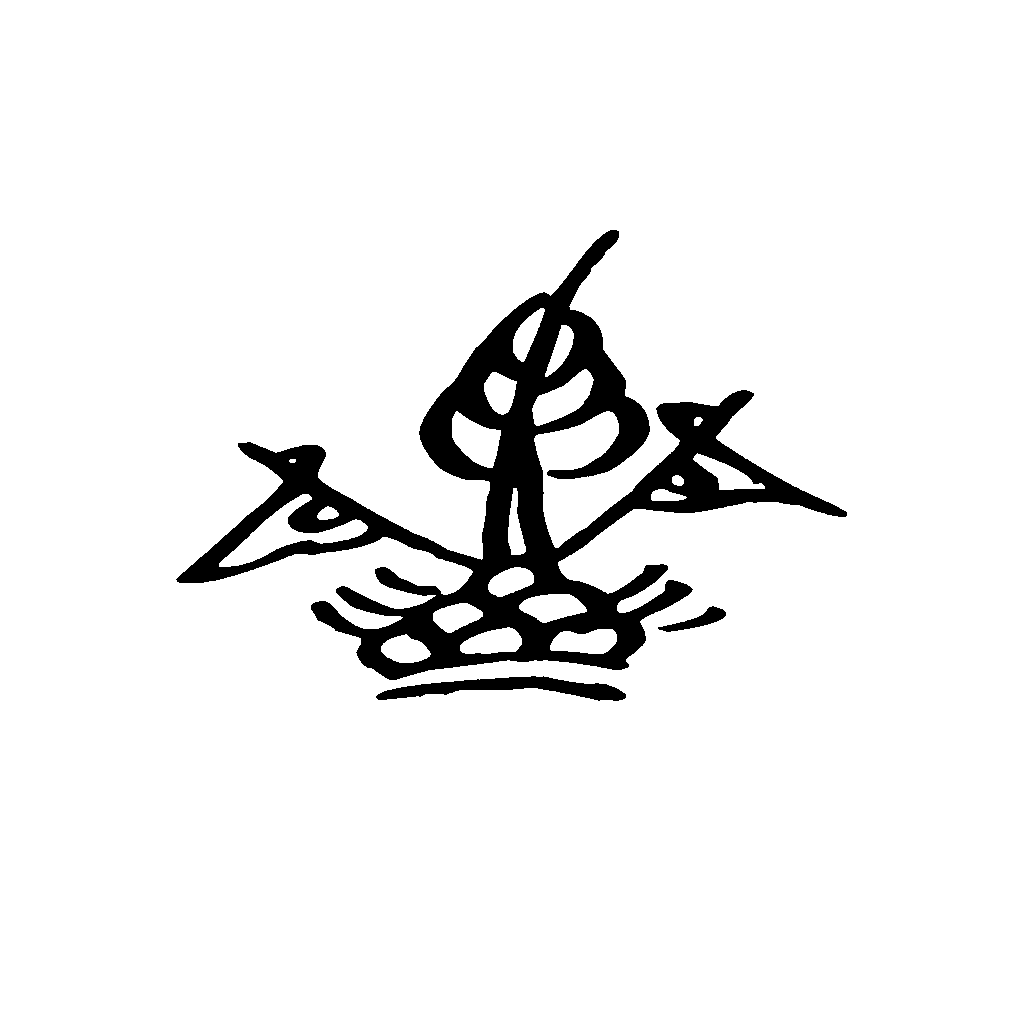} &
\imgwithbox[width=0.95\linewidth]{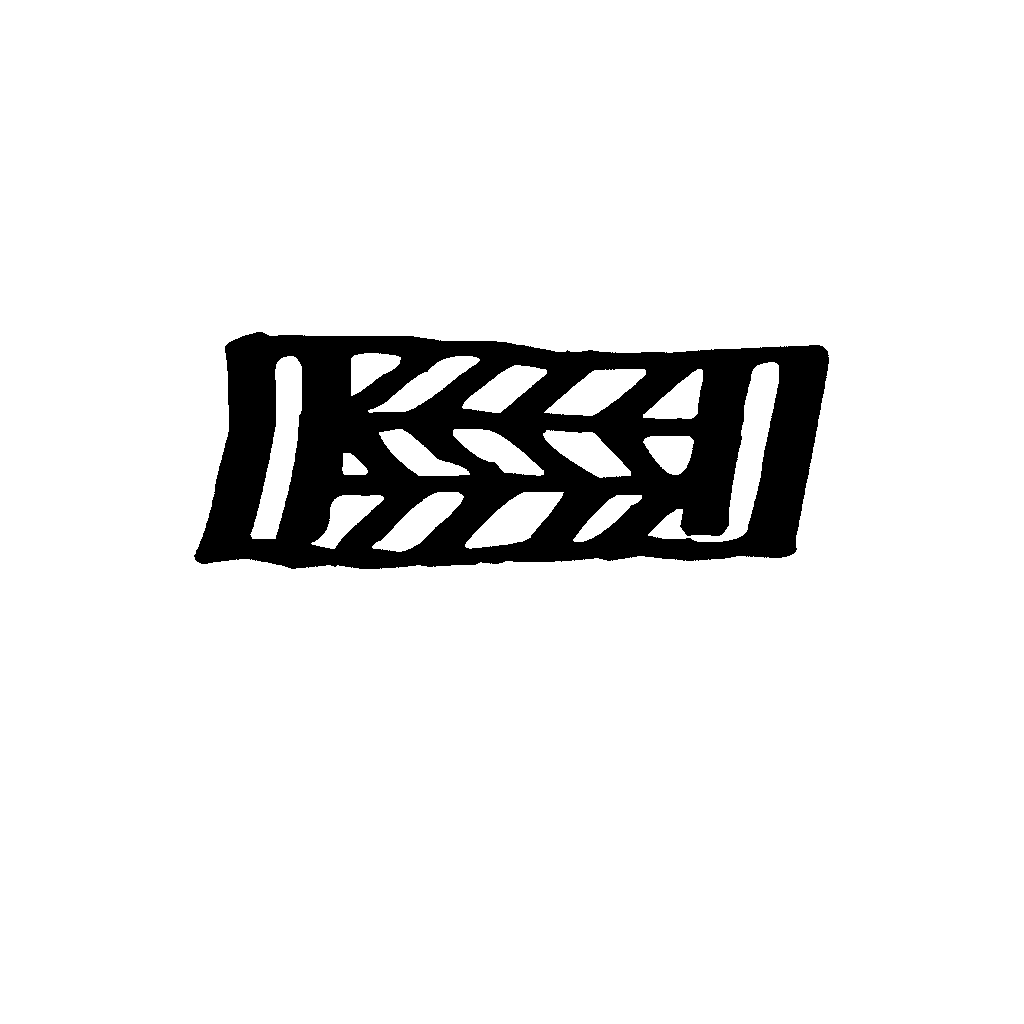} &
\includegraphics[width=0.95\linewidth]{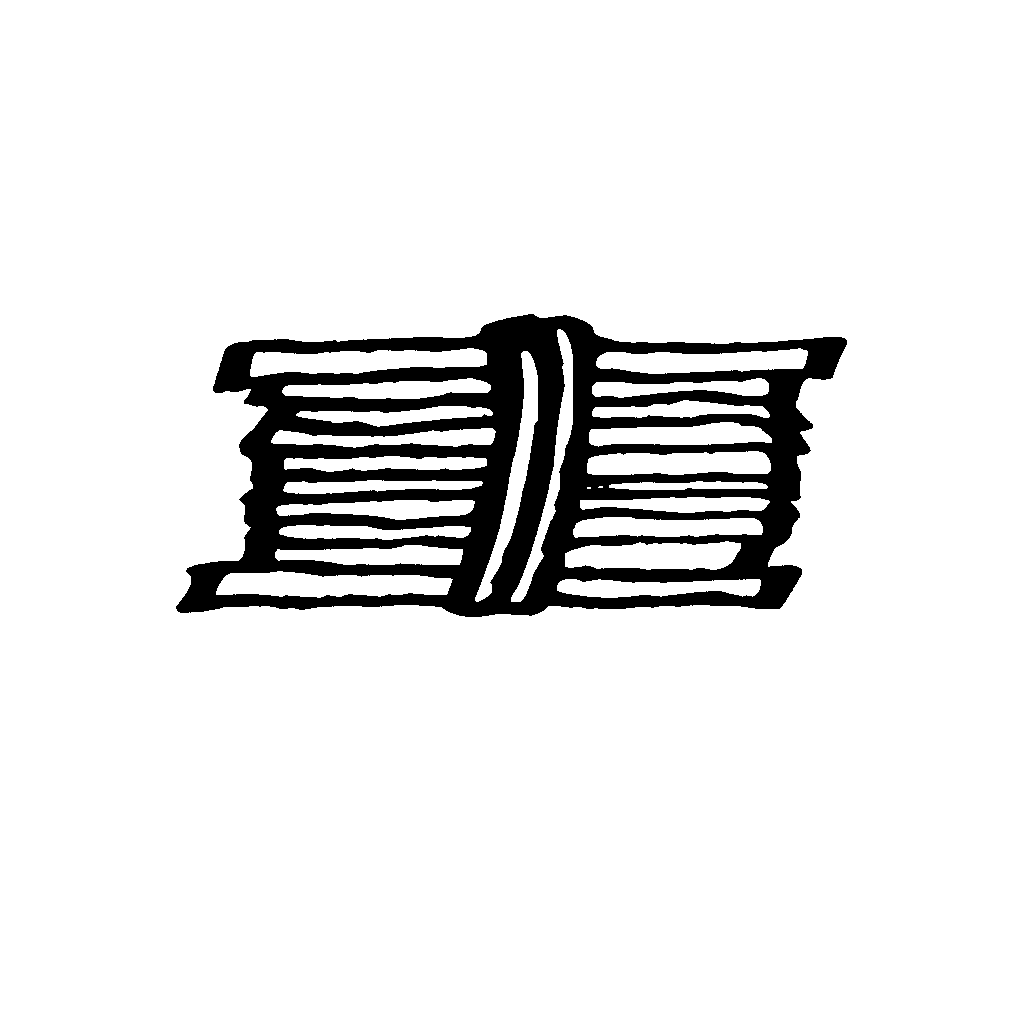} &
\includegraphics[width=0.95\linewidth]{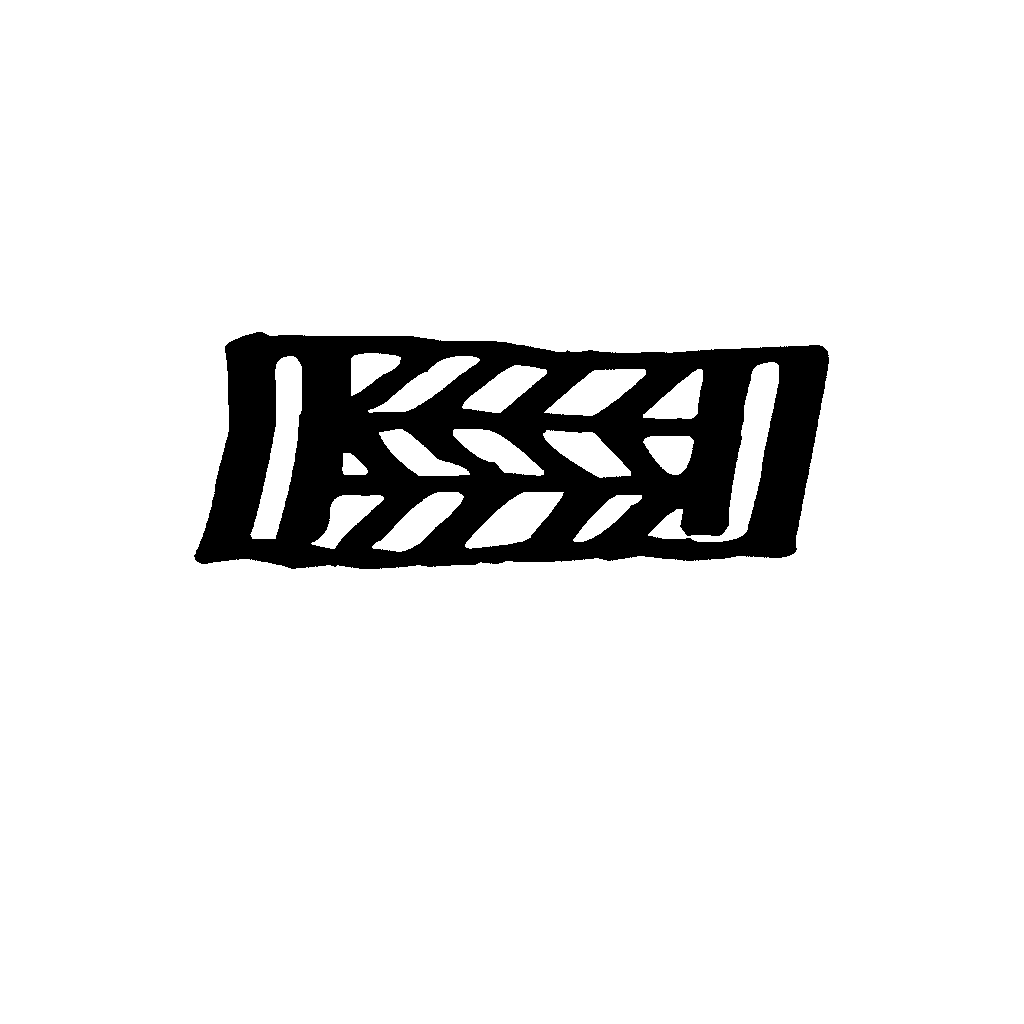} \\
4 &
\includegraphics[width=0.95\linewidth]{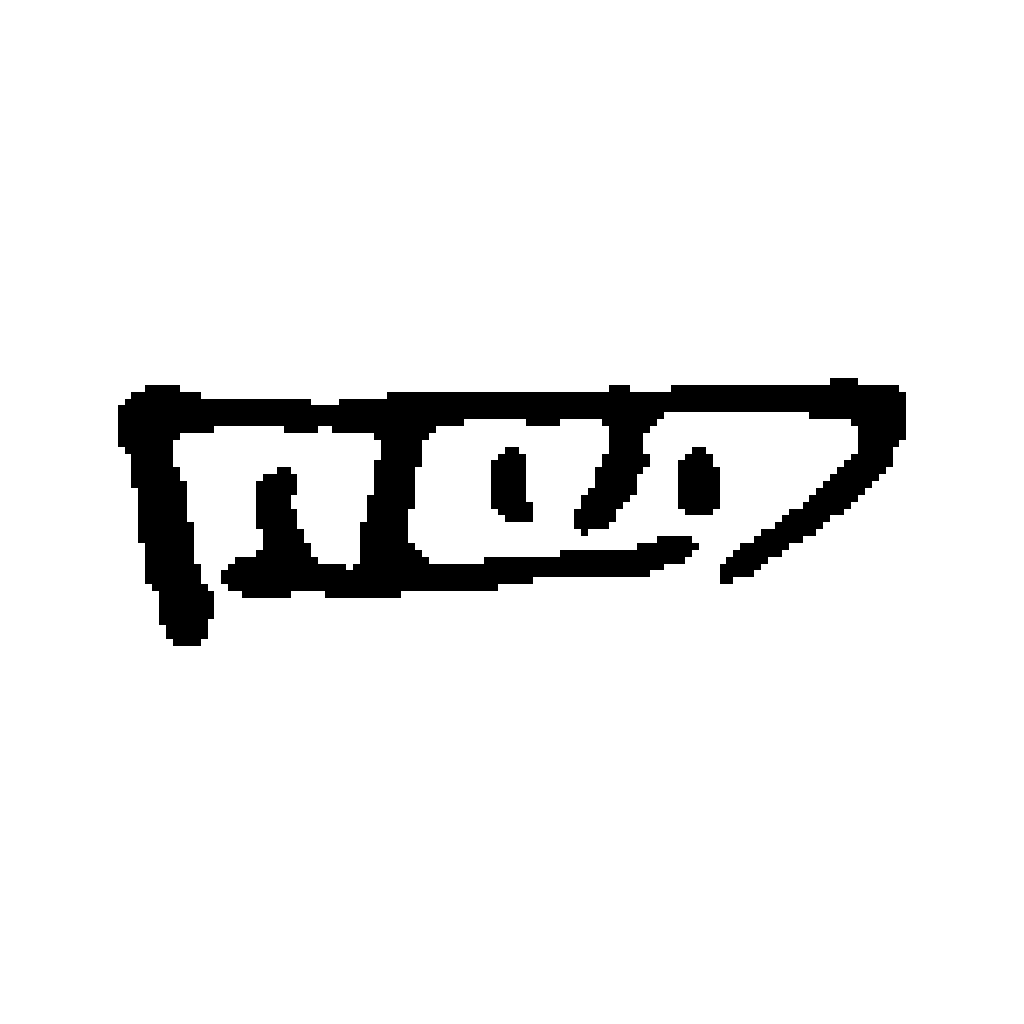} &
\includegraphics[width=0.95\linewidth]{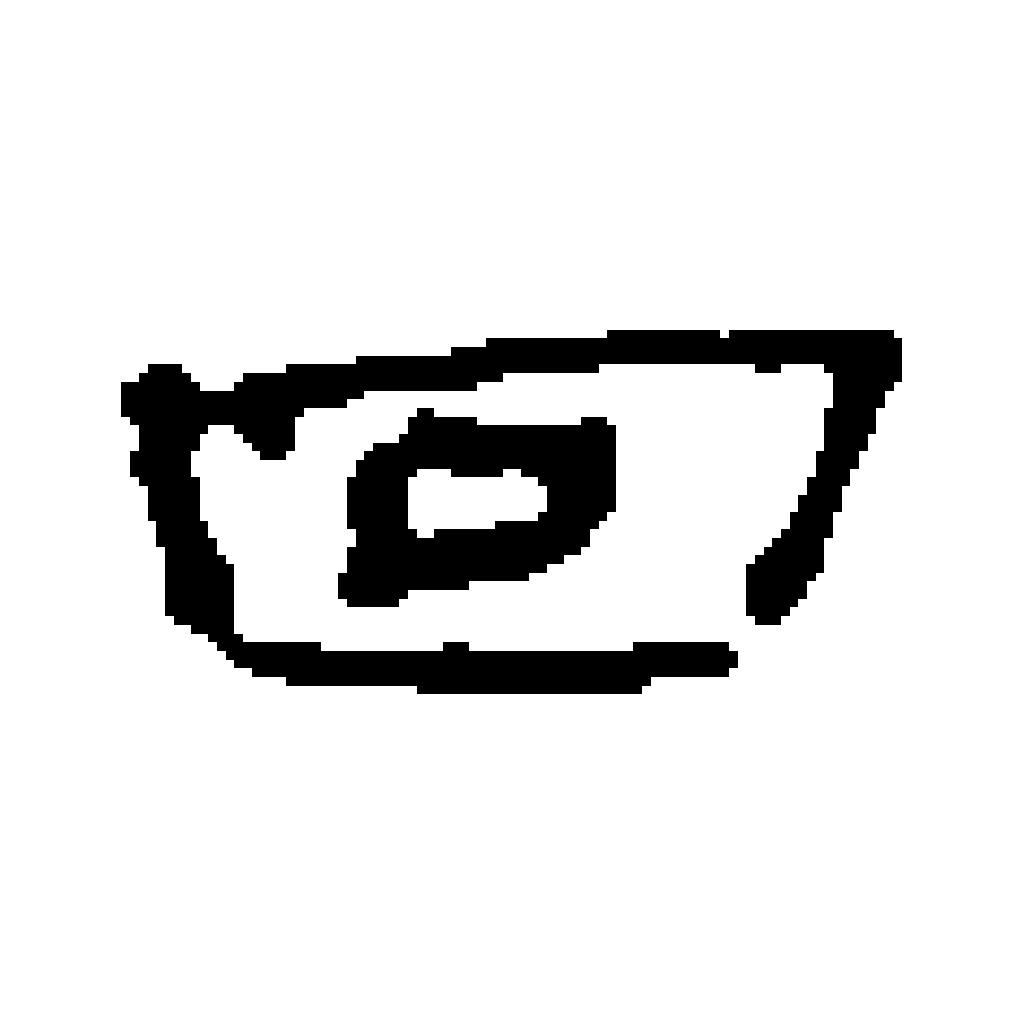} &
\includegraphics[width=0.95\linewidth]{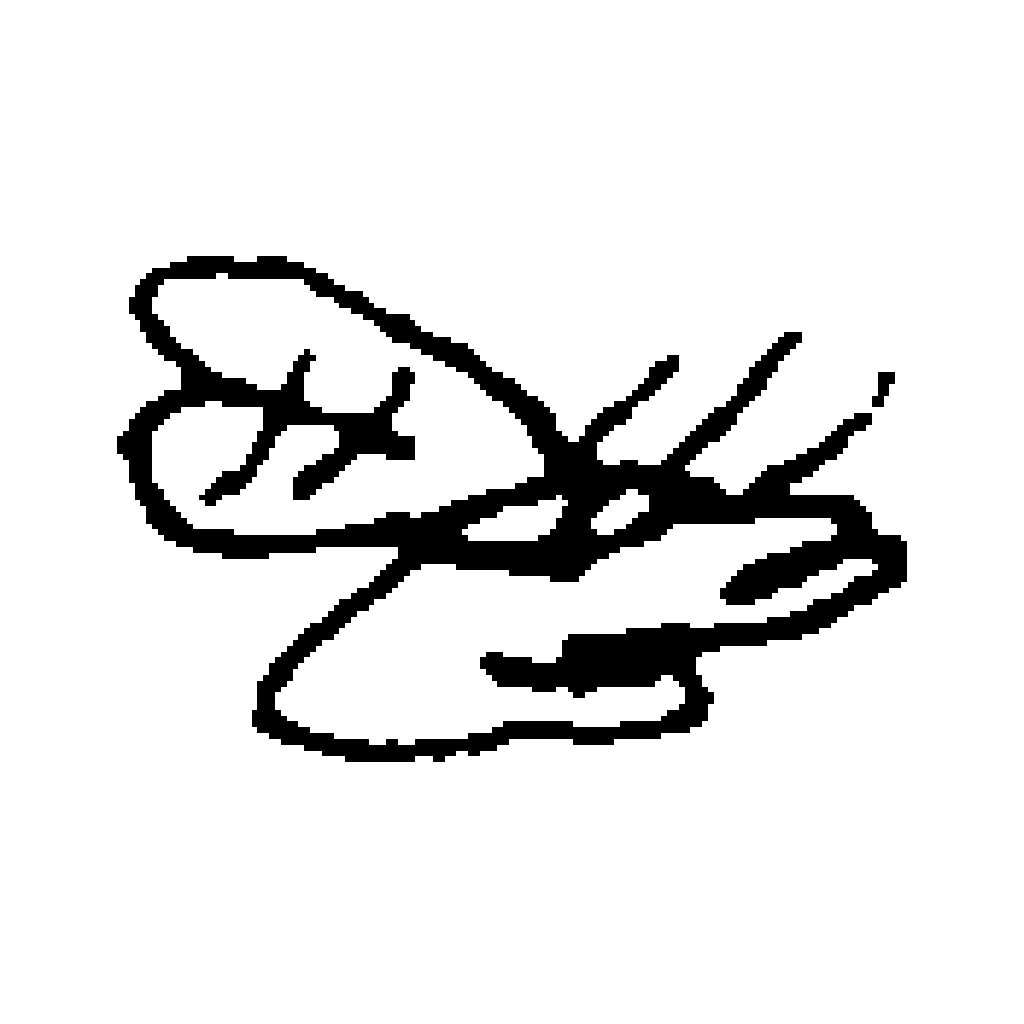} &
\includegraphics[width=0.95\linewidth]{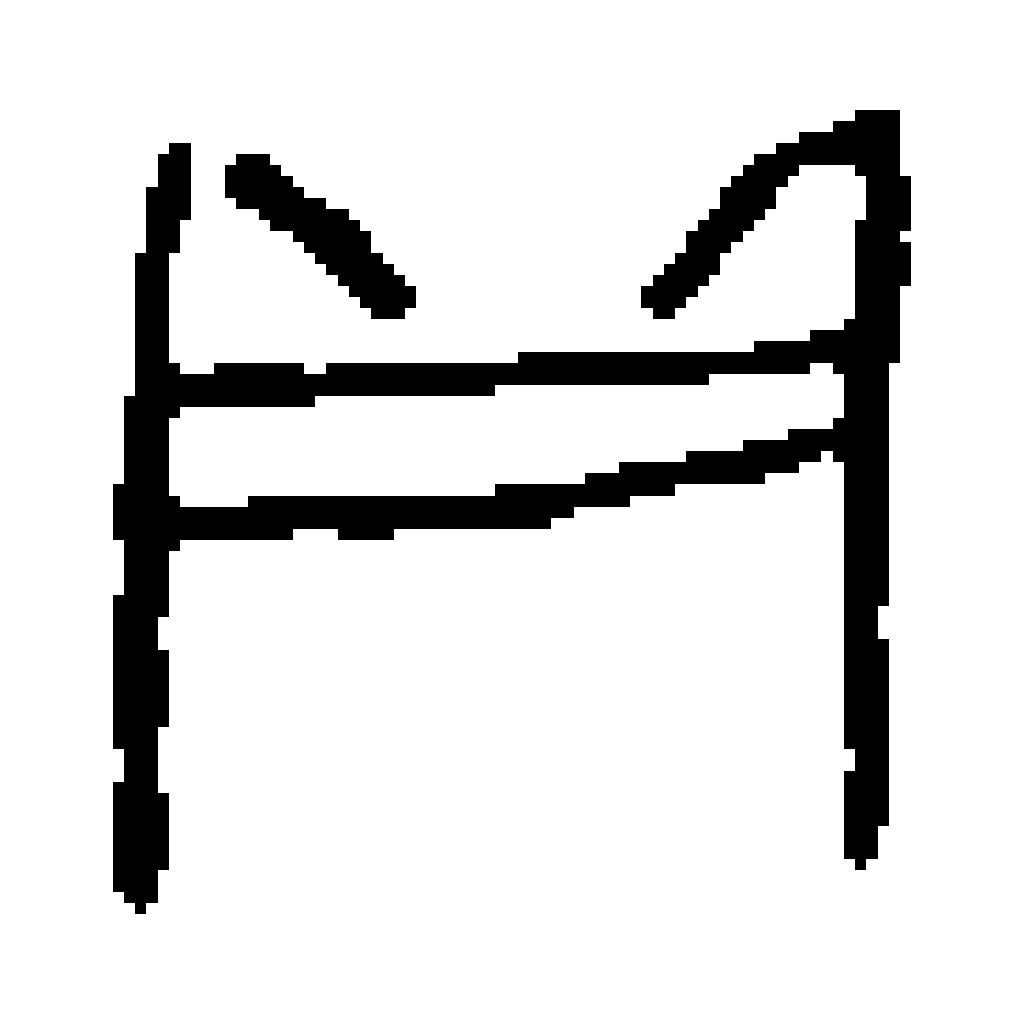} &
\includegraphics[width=0.95\linewidth]{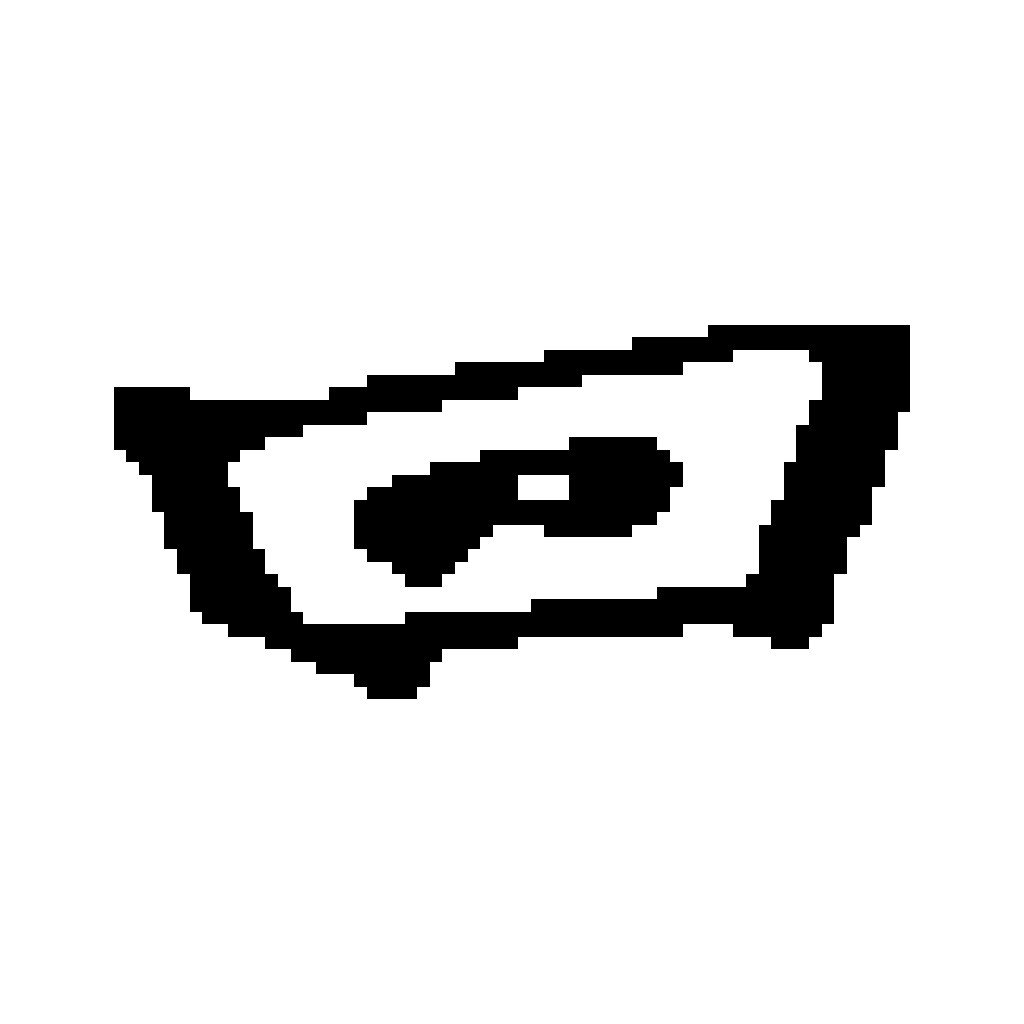} &
\includegraphics[width=0.95\linewidth]{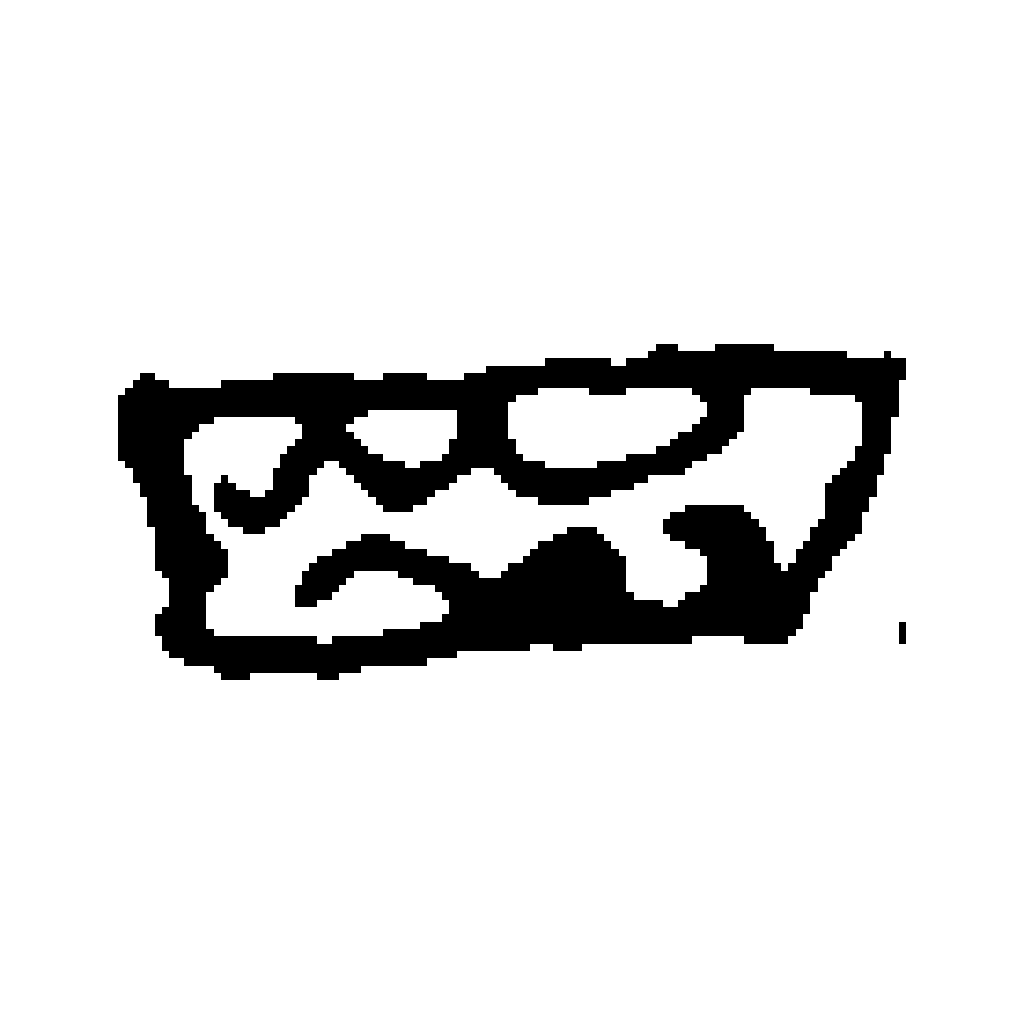} &
\includegraphics[width=0.95\linewidth]{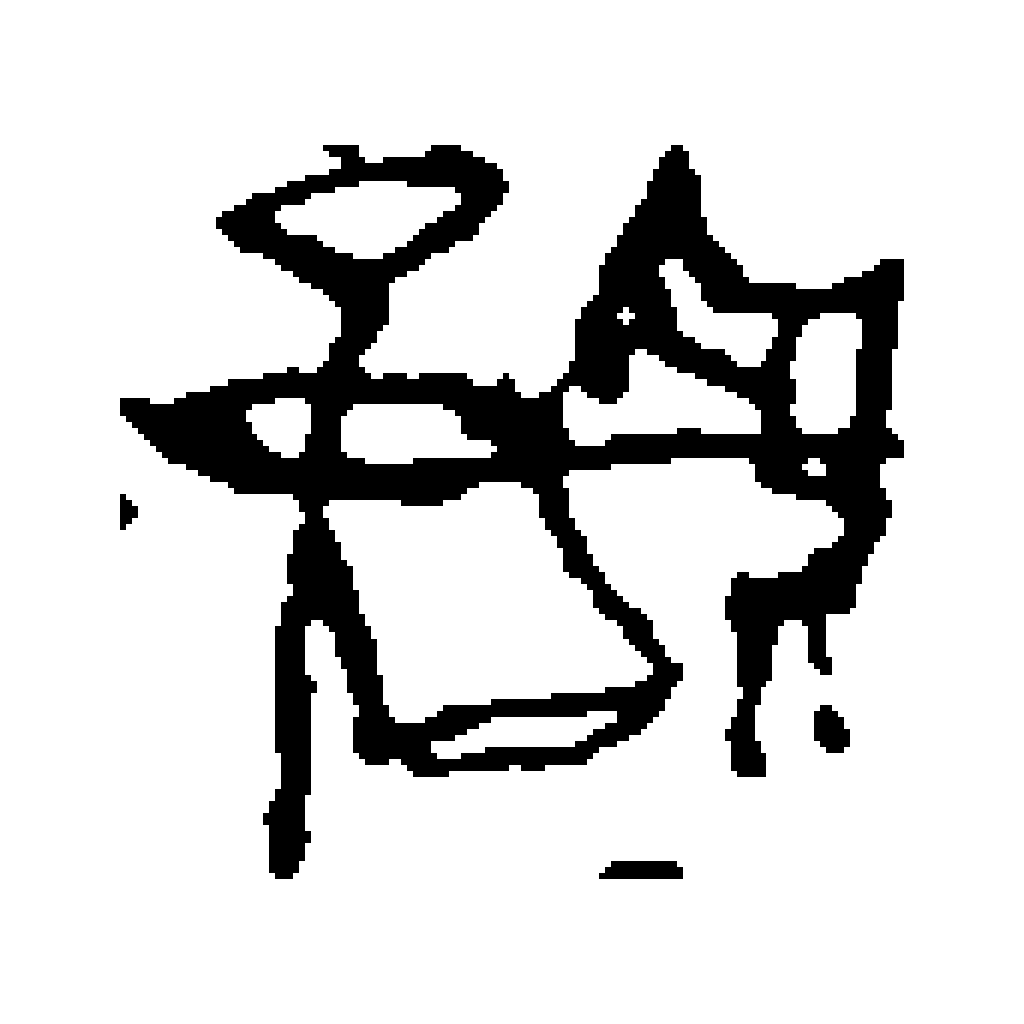} &
\imgwithbox[width=0.95\linewidth]{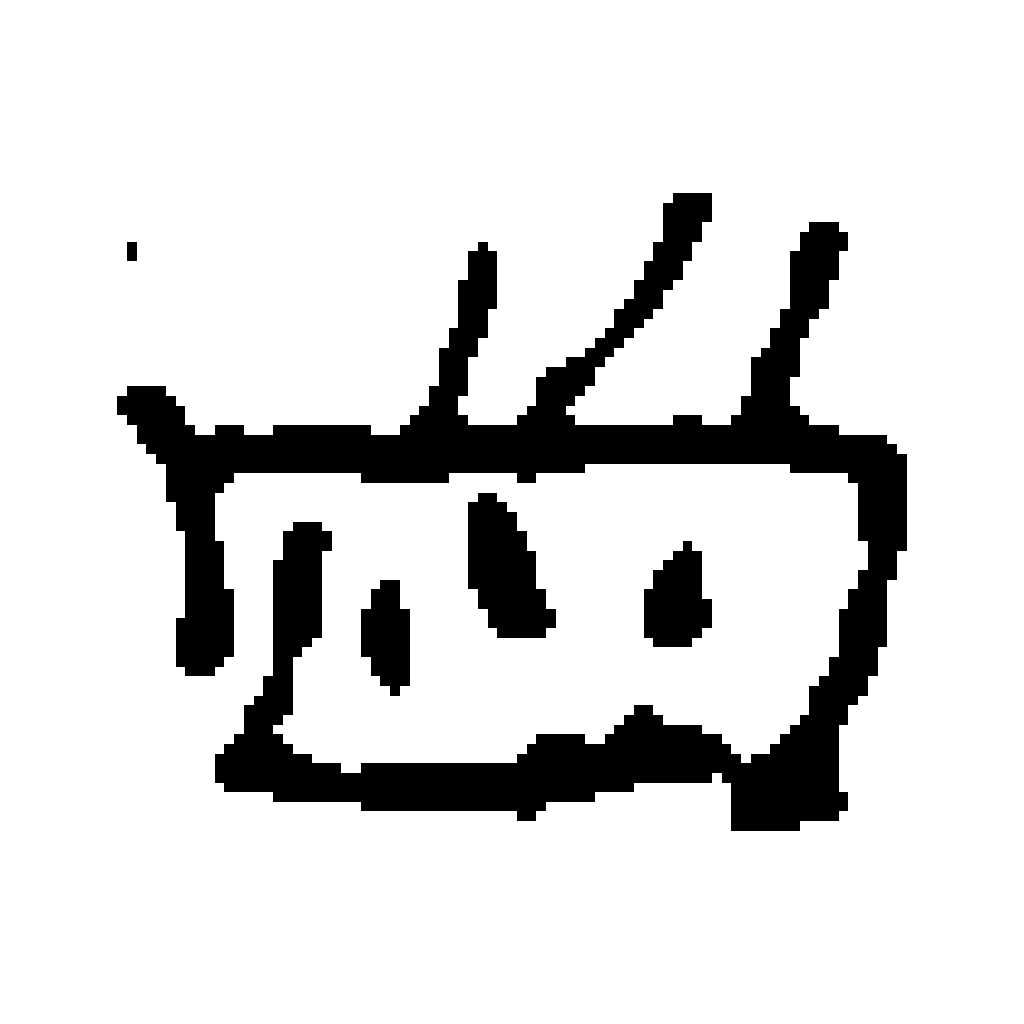} &
\includegraphics[width=0.95\linewidth]{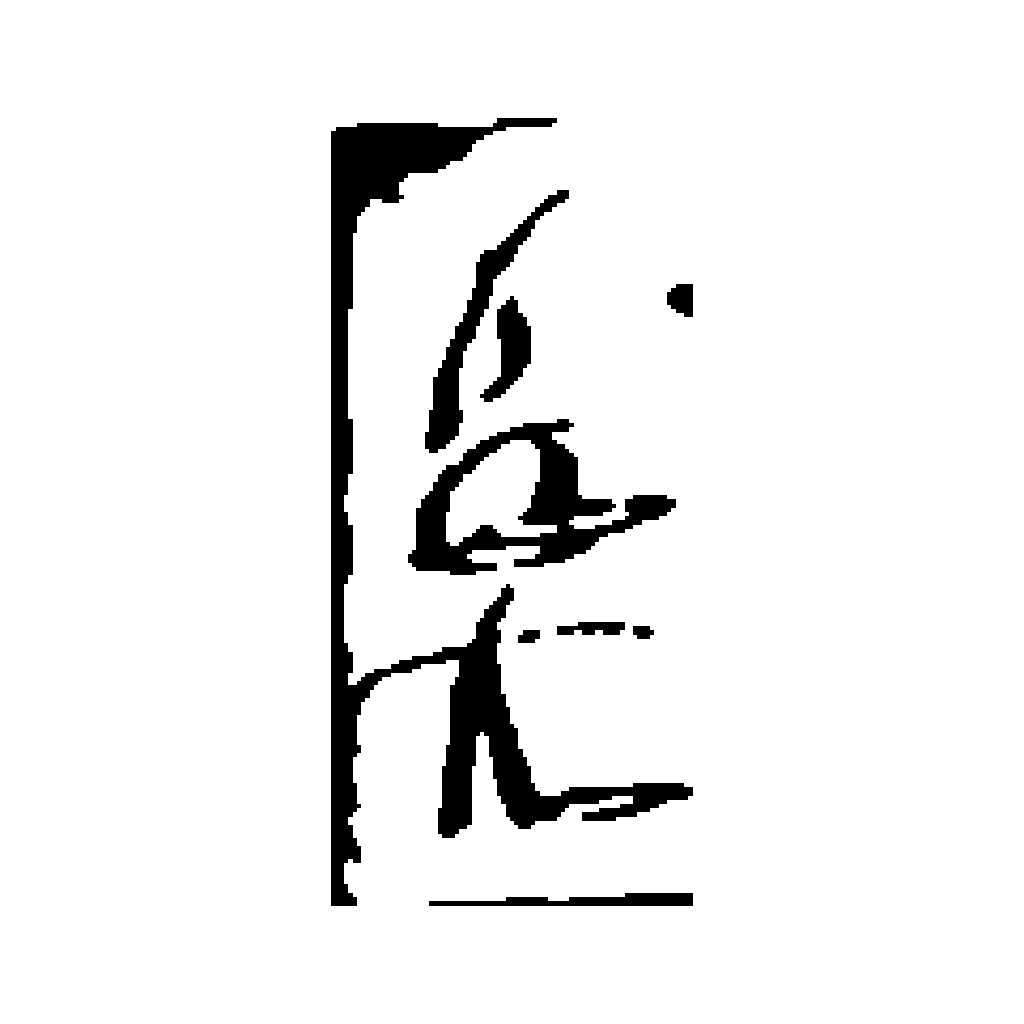} &
\includegraphics[width=0.95\linewidth]{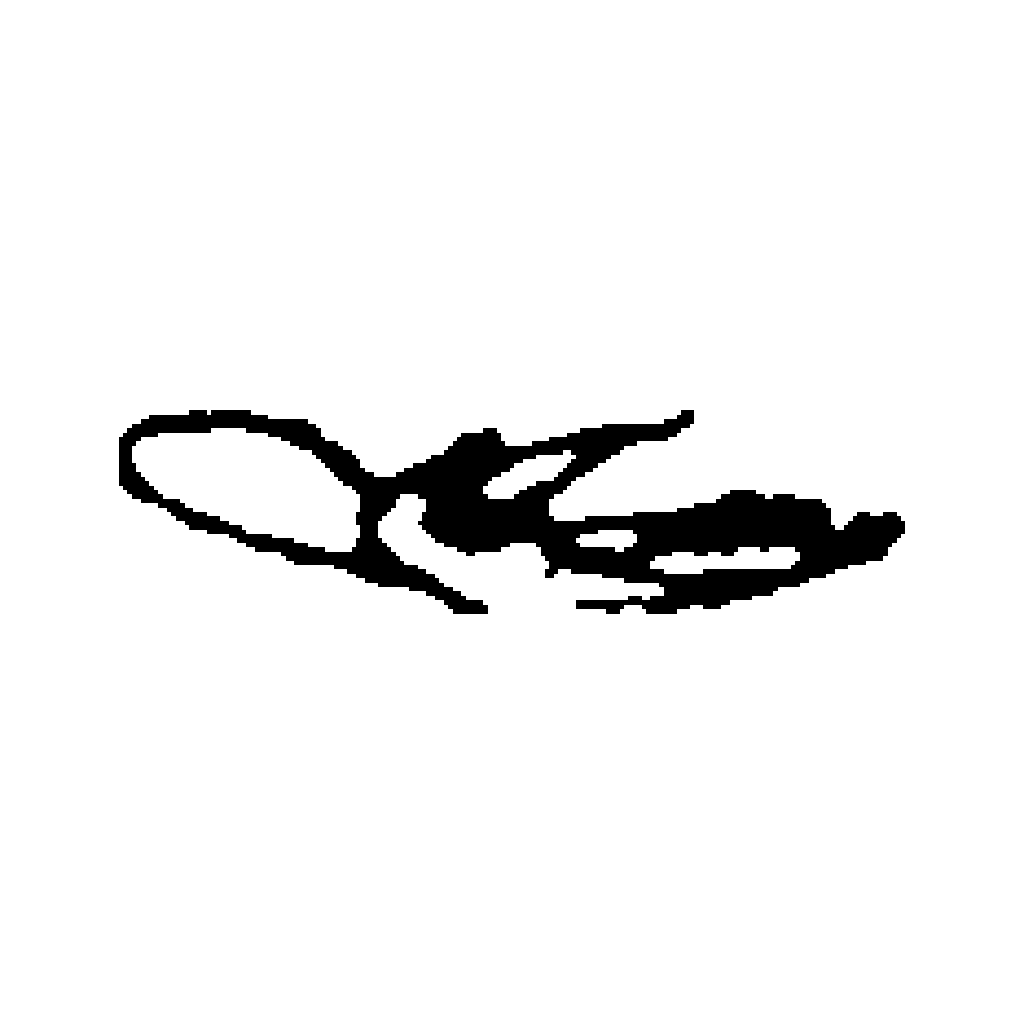} &
\includegraphics[width=0.95\linewidth]{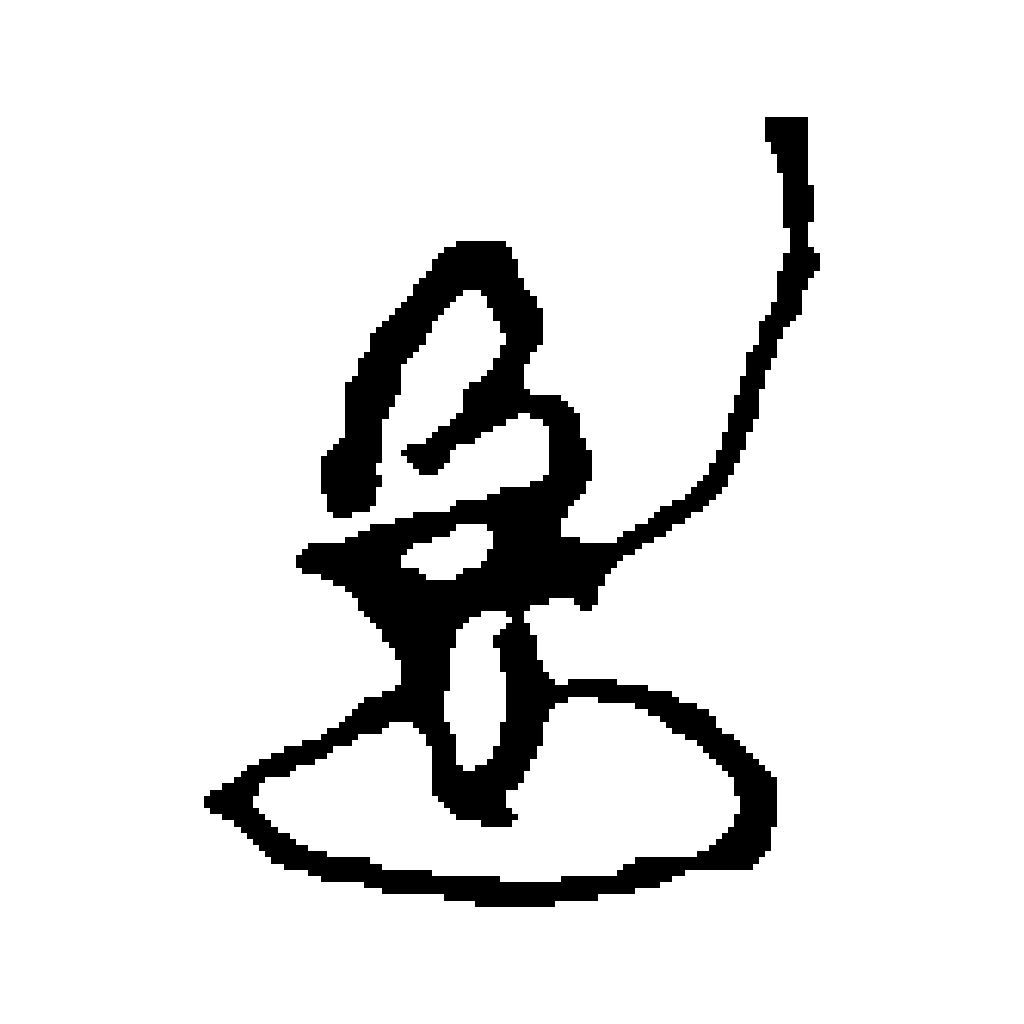} &
\includegraphics[width=0.95\linewidth]{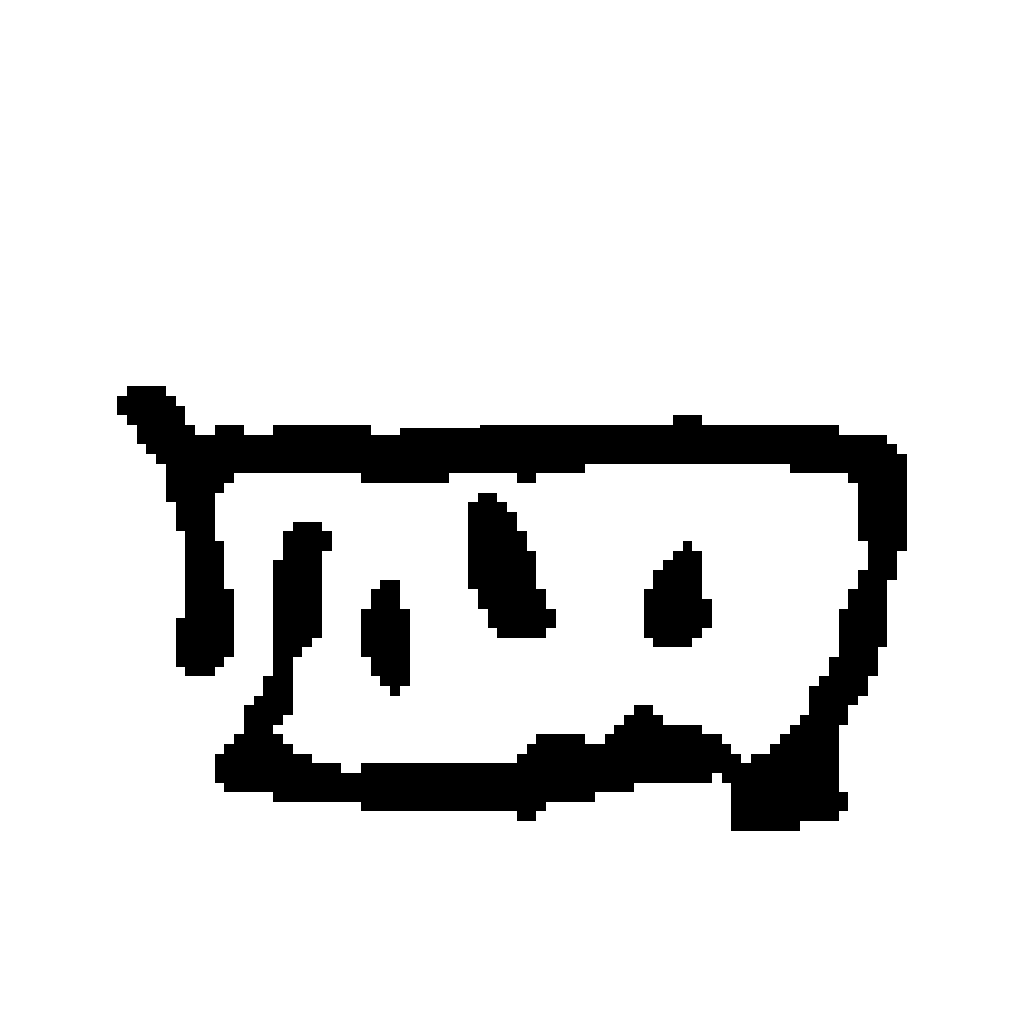} \\
\midrule\noalign{\vskip -3pt}\rowcolor{gray!20}\multicolumn{13}{c}{\textbf{Egyptian Hieroglyphs}} \\\noalign{\vskip -2pt}\midrule
5 &
\includegraphics[width=0.95\linewidth]{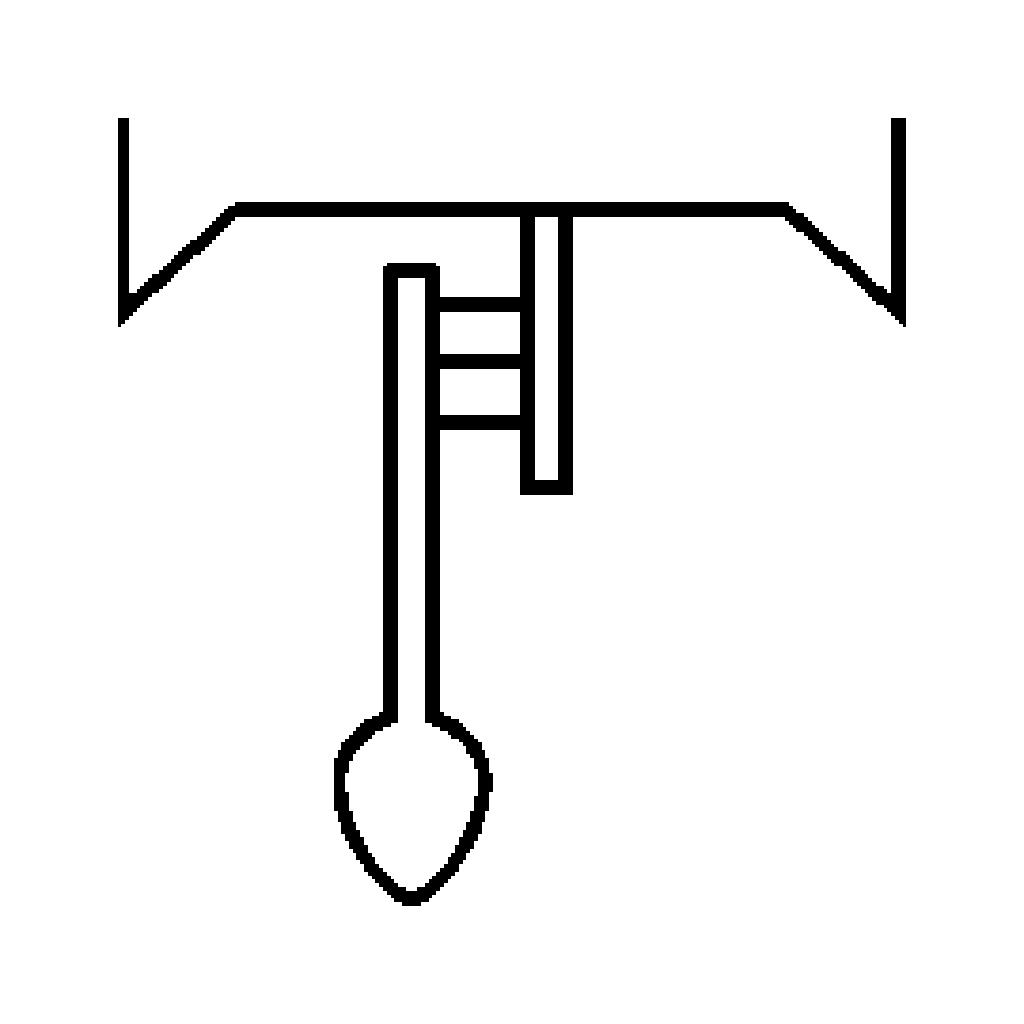} &
\includegraphics[width=0.95\linewidth]{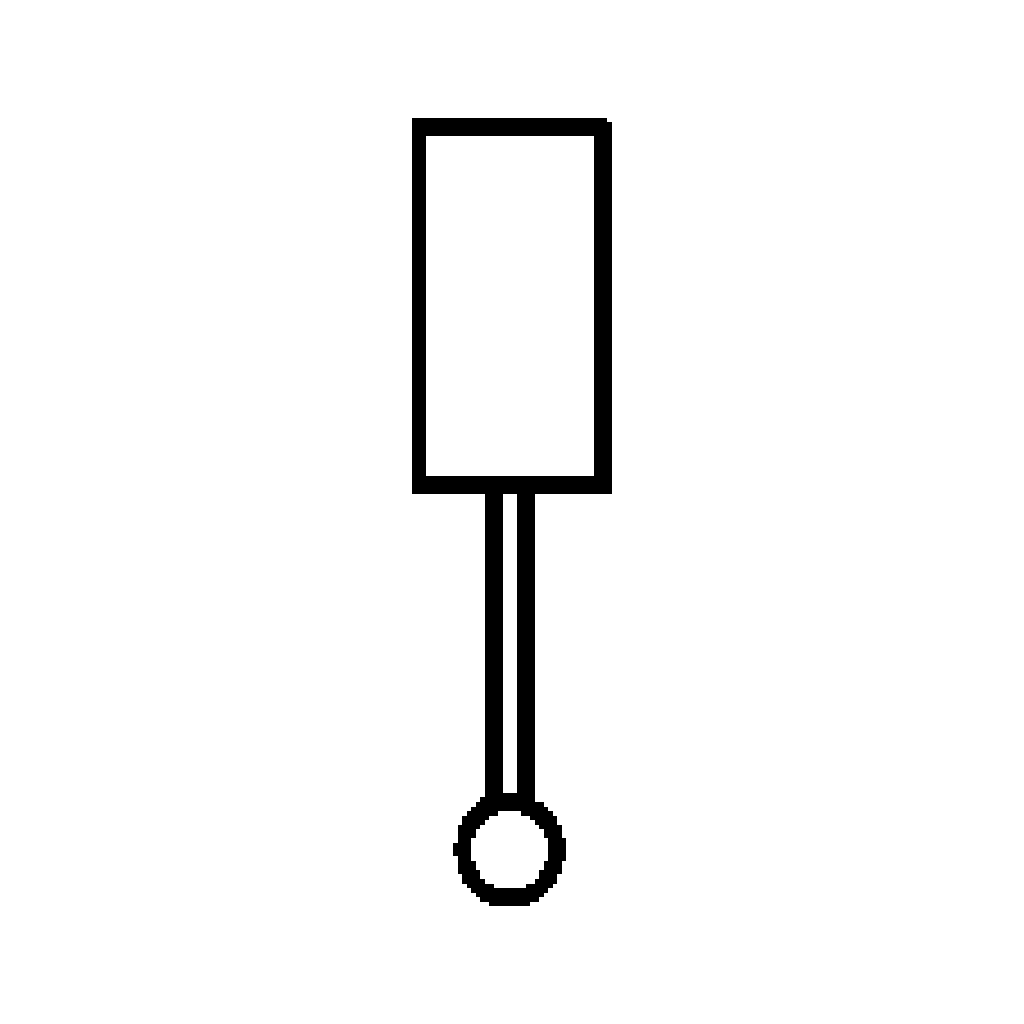} &
\includegraphics[width=0.95\linewidth]{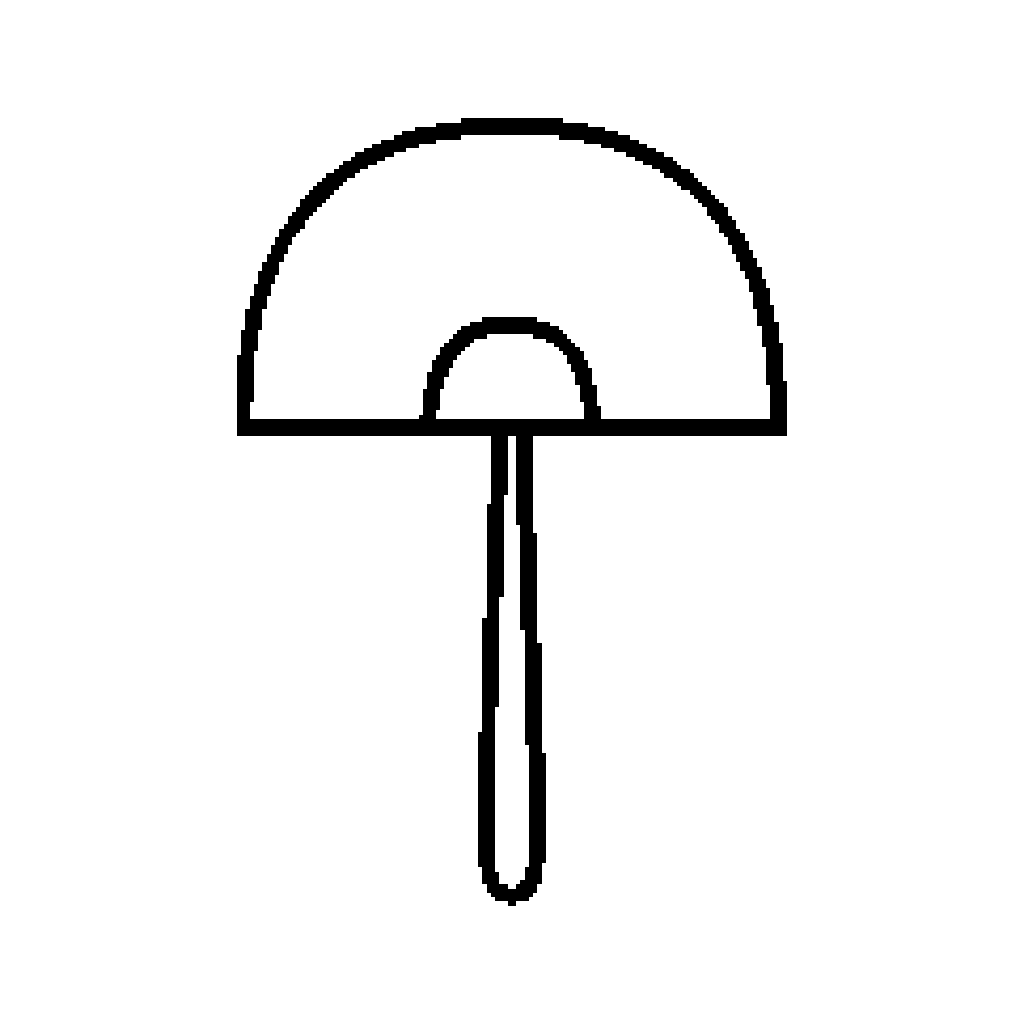} &
\includegraphics[width=0.95\linewidth]{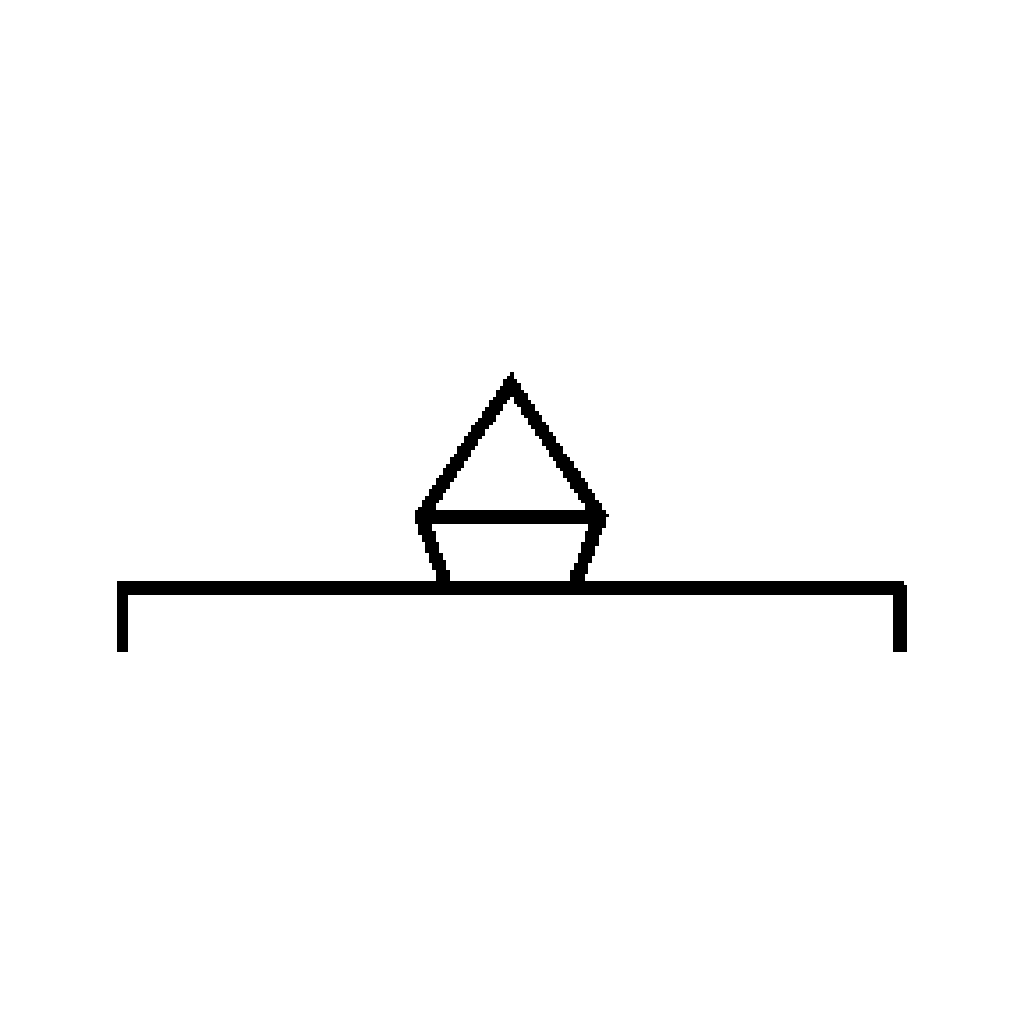} &
\includegraphics[width=0.95\linewidth]{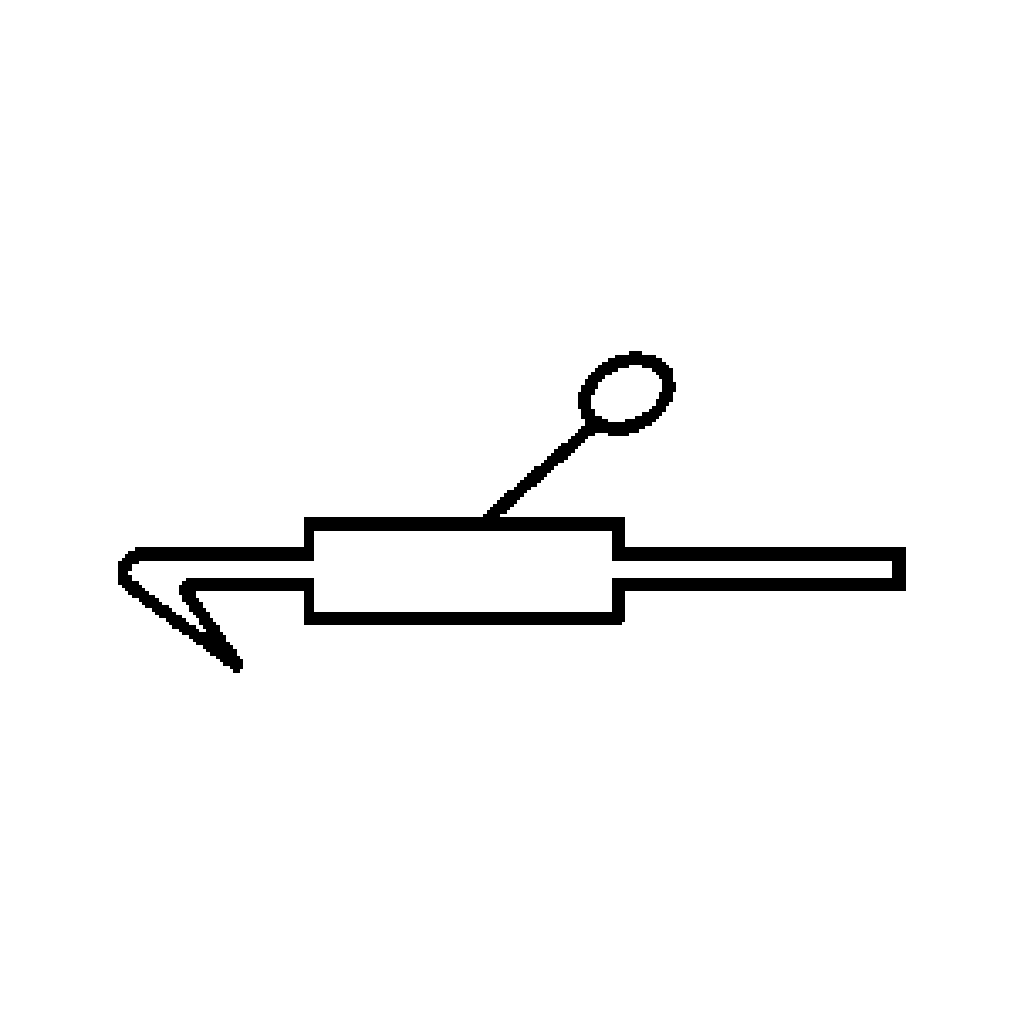} &
\includegraphics[width=0.95\linewidth]{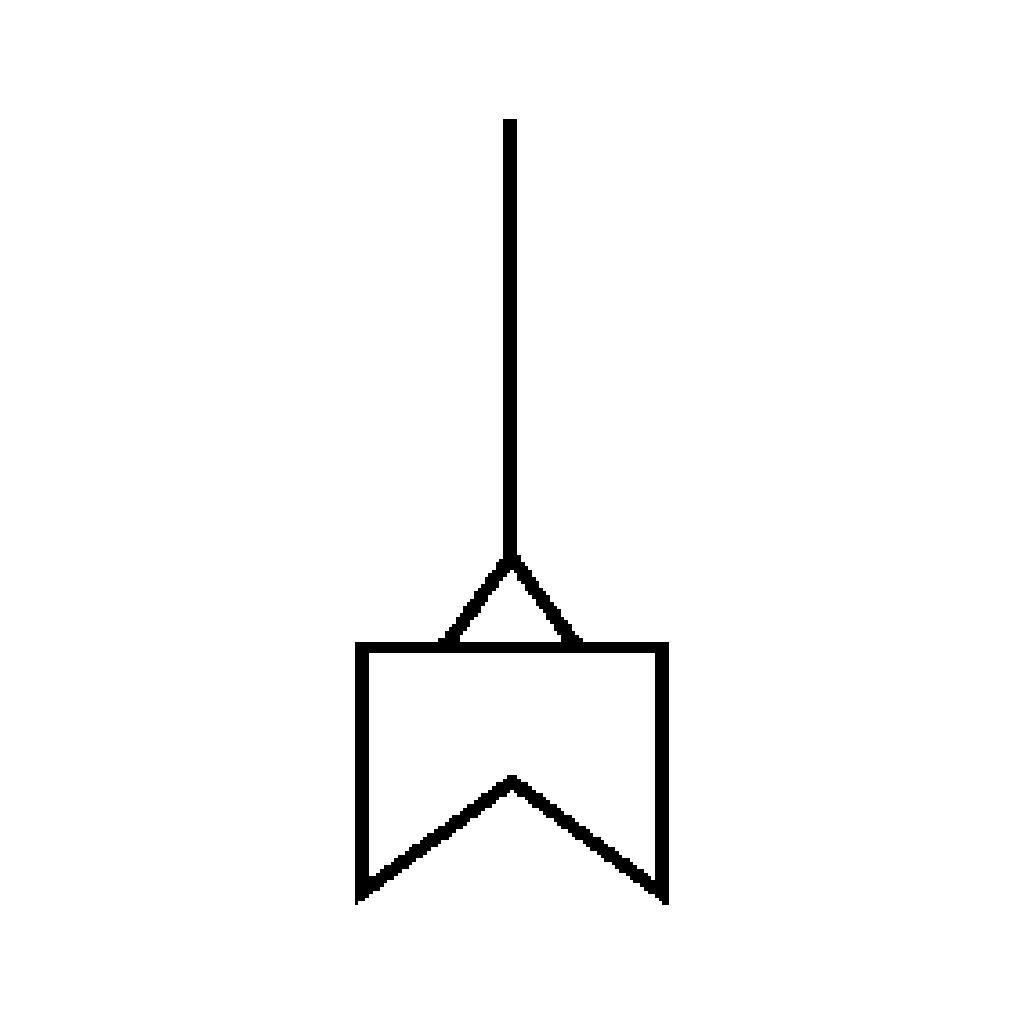} &
\includegraphics[width=0.95\linewidth]{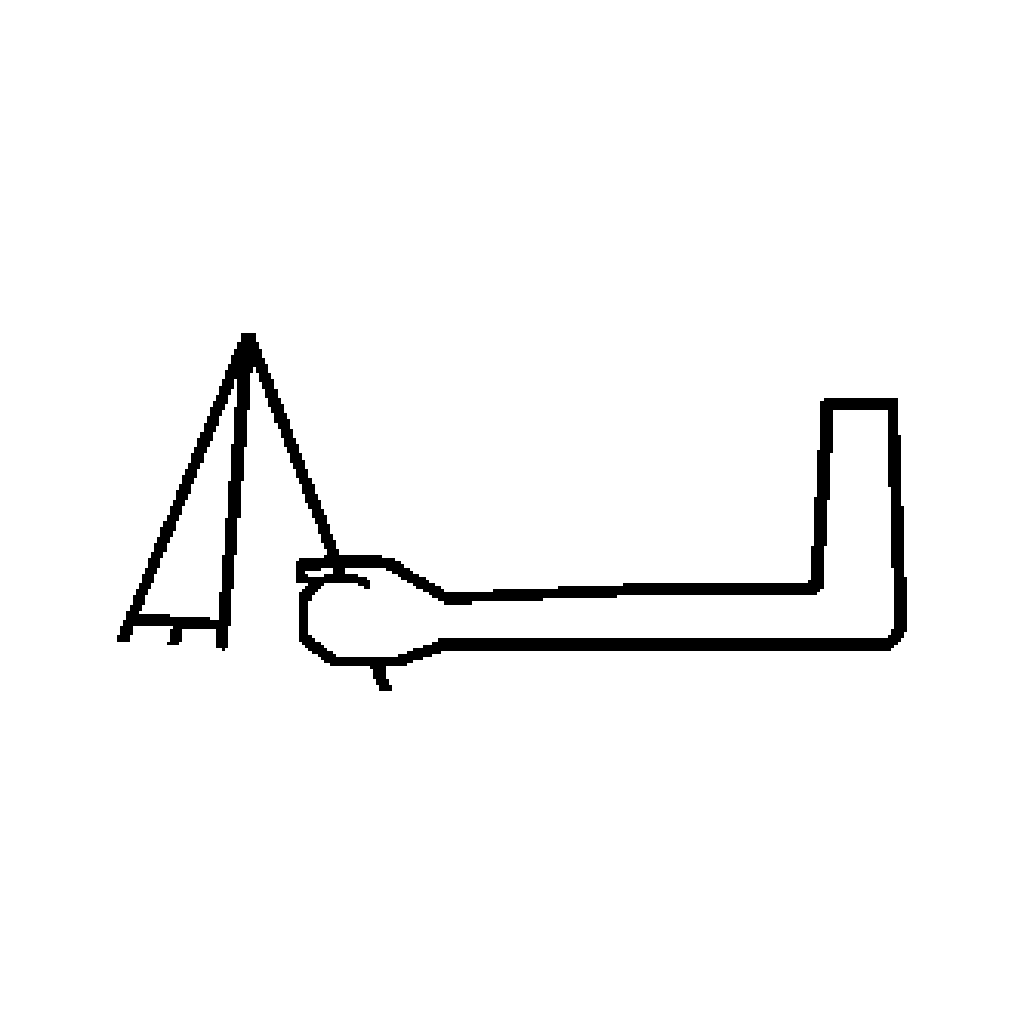} &
\imgwithbox[width=0.95\linewidth]{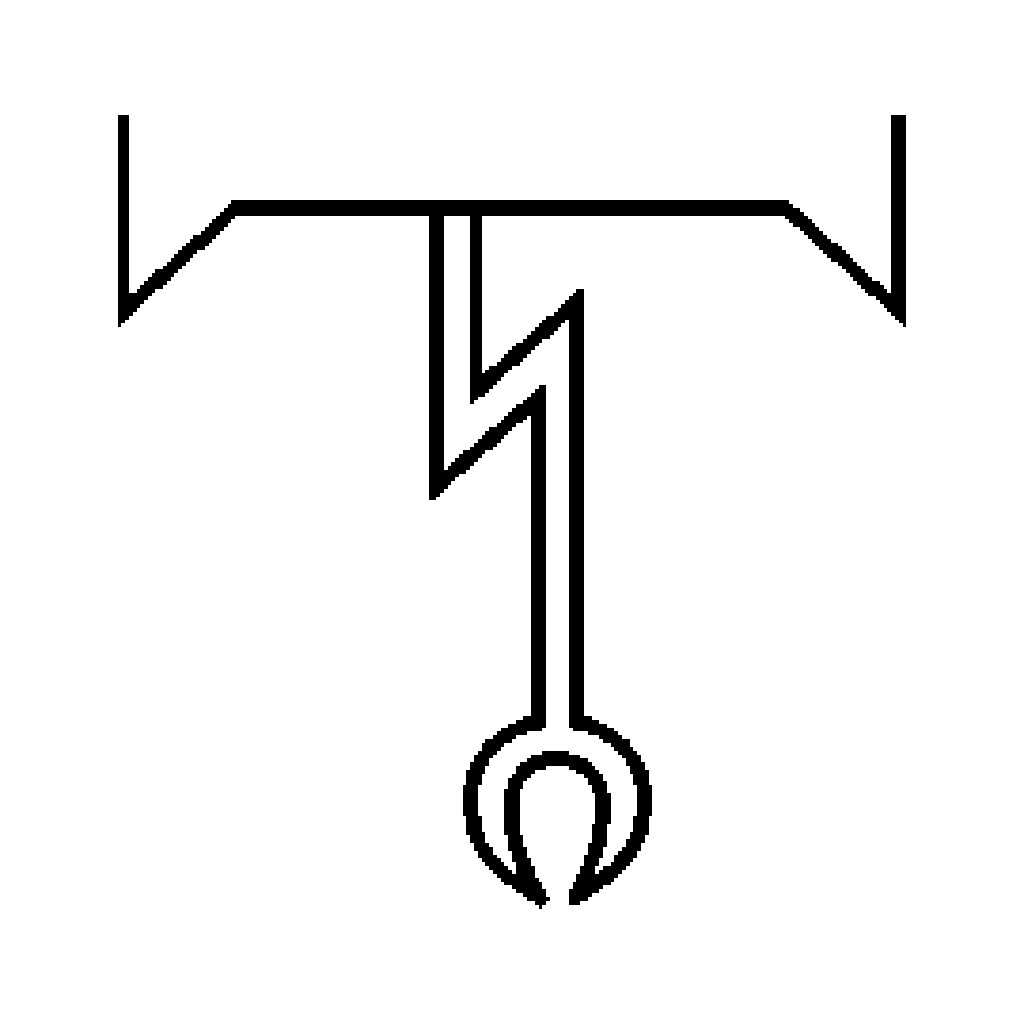} &
\includegraphics[width=0.95\linewidth]{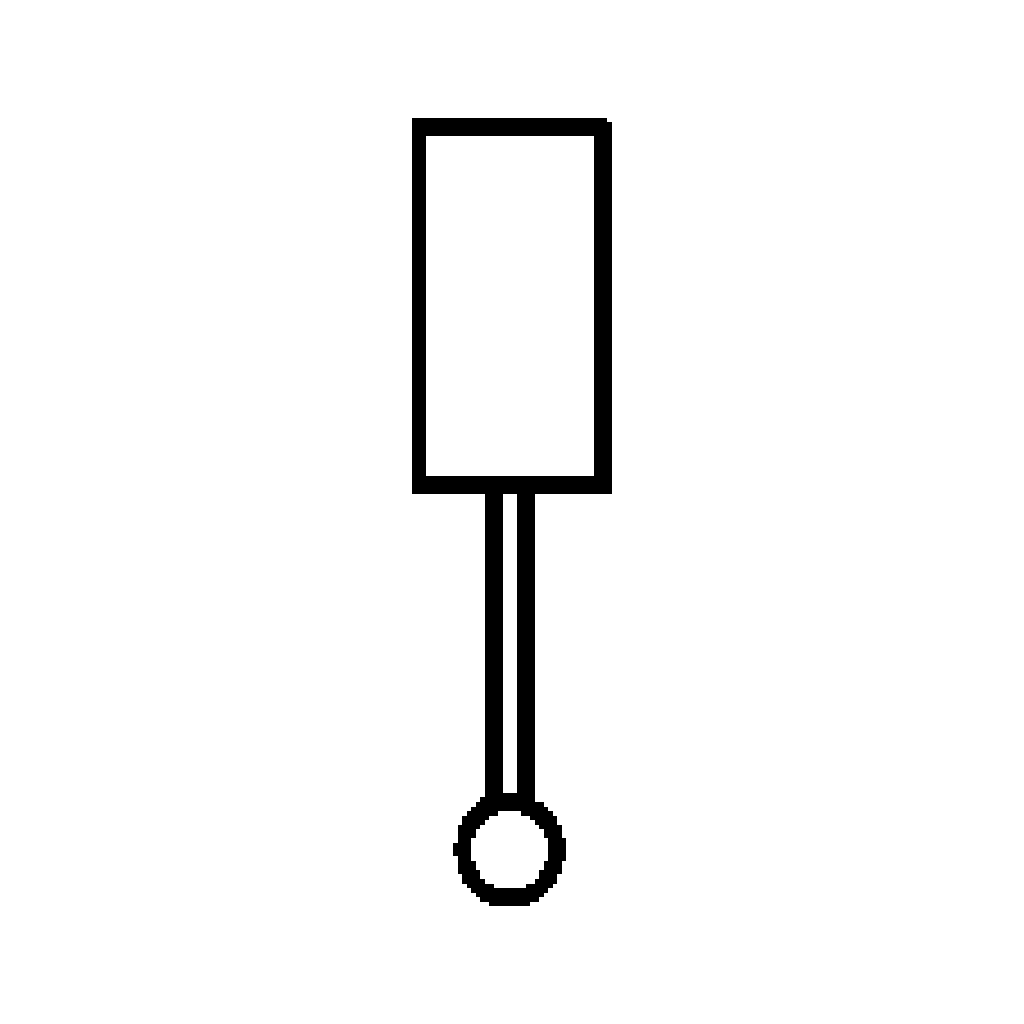} &
\includegraphics[width=0.95\linewidth]{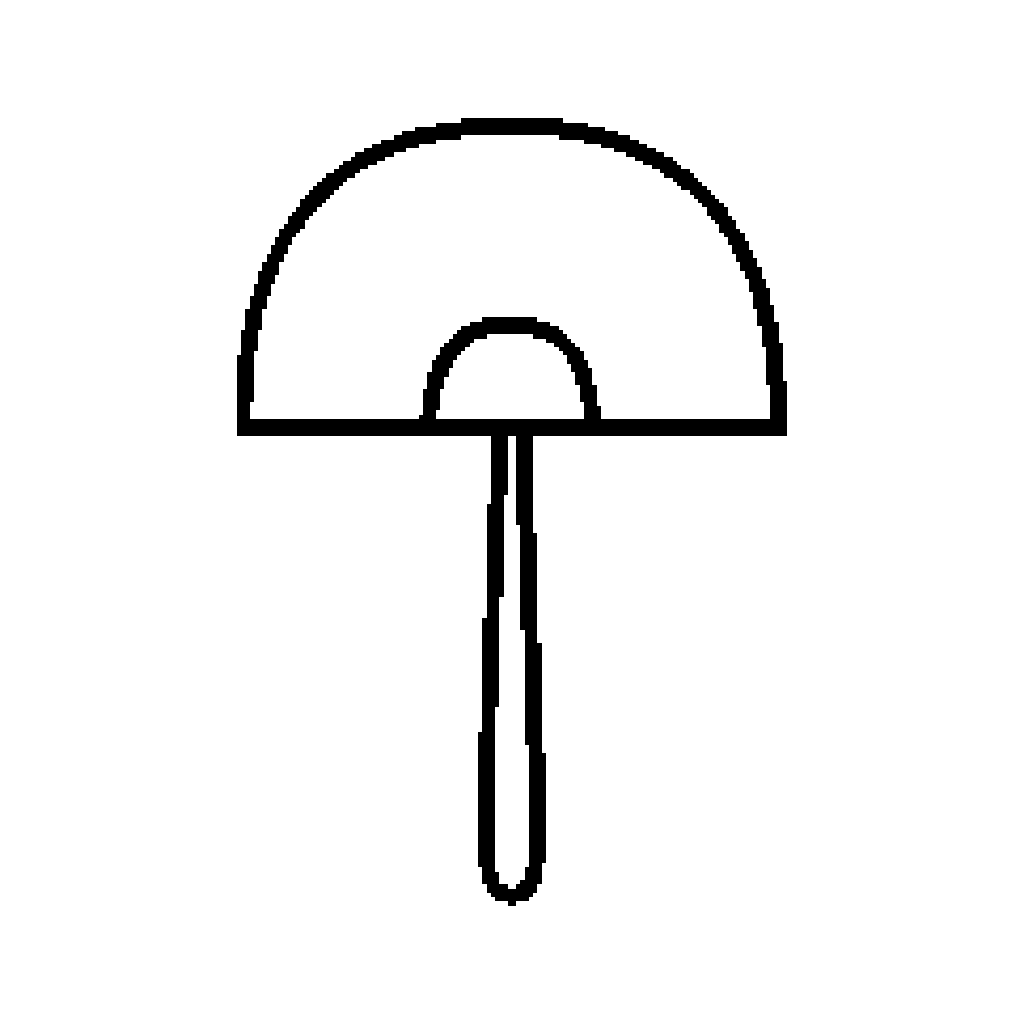} &
\includegraphics[width=0.95\linewidth]{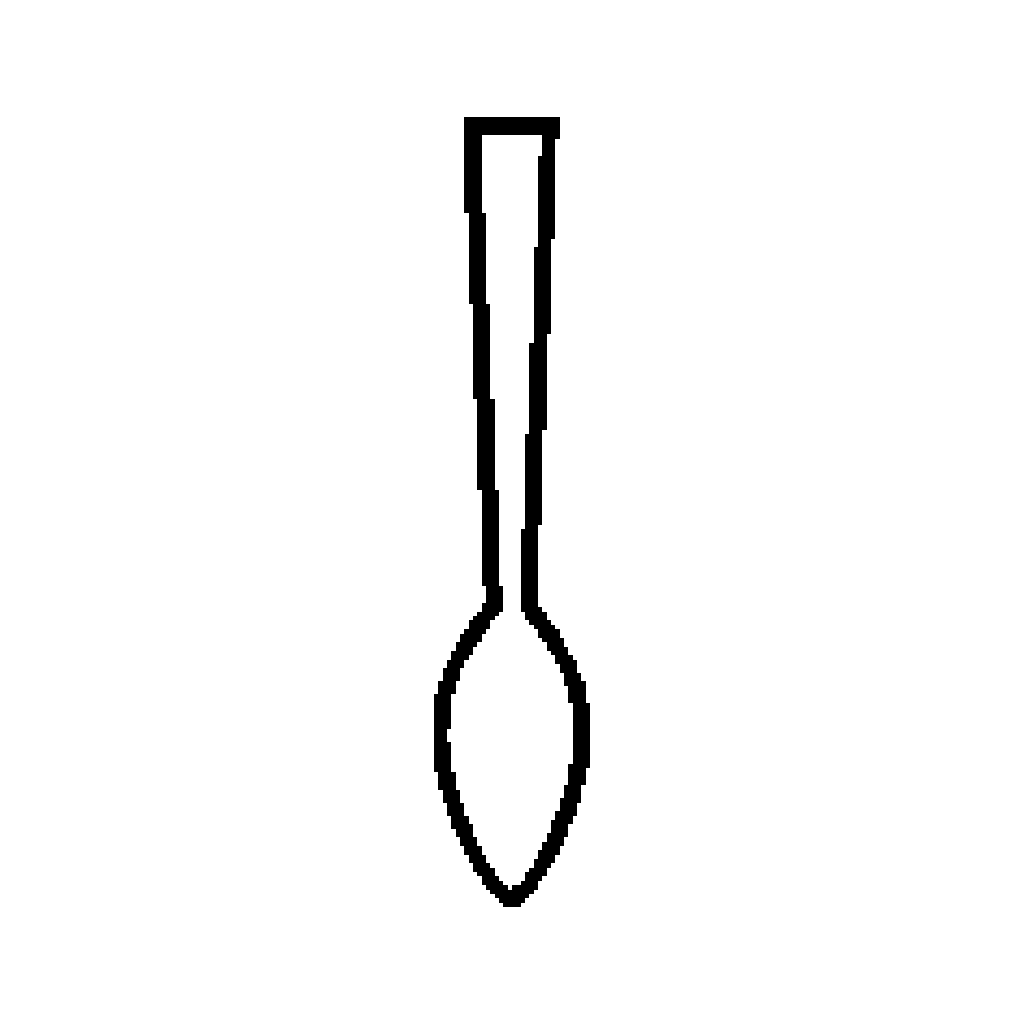} &
\includegraphics[width=0.95\linewidth]{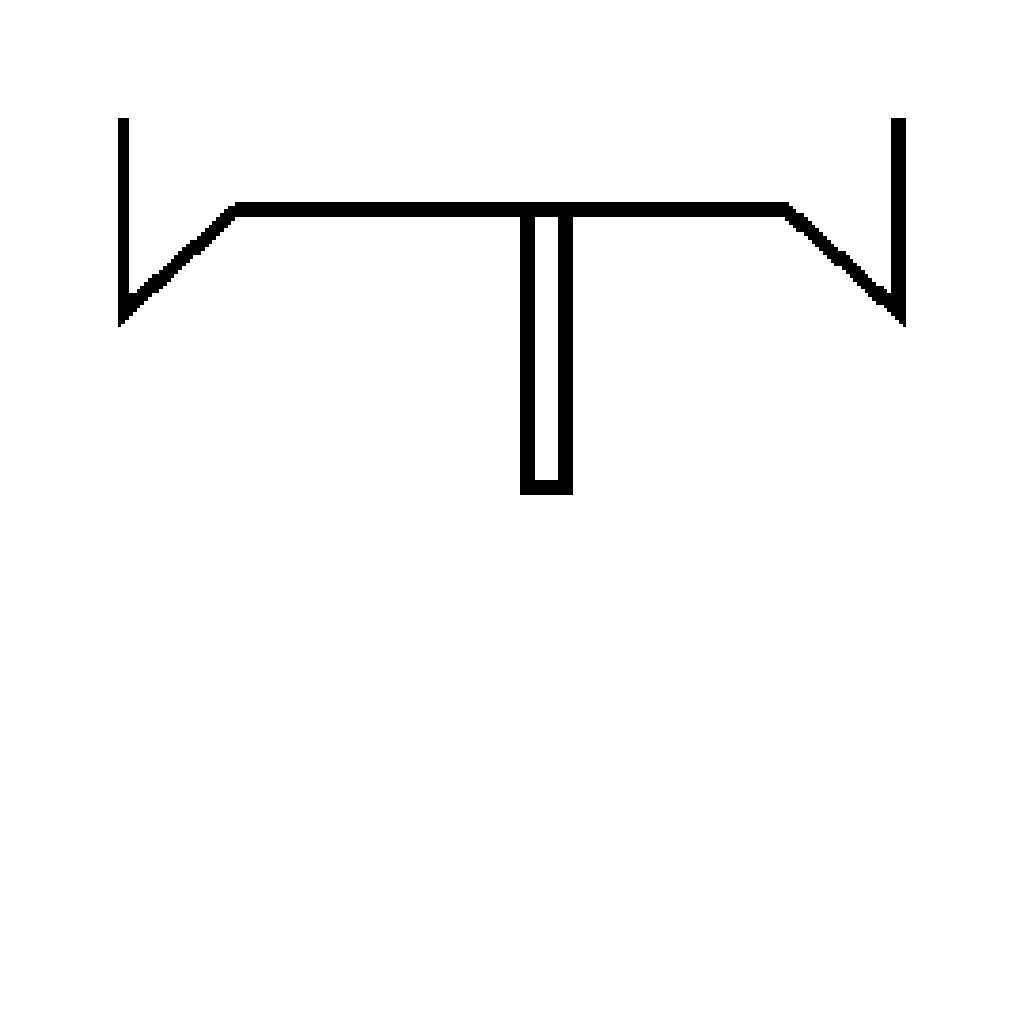} \\
6 &
\includegraphics[width=0.95\linewidth]{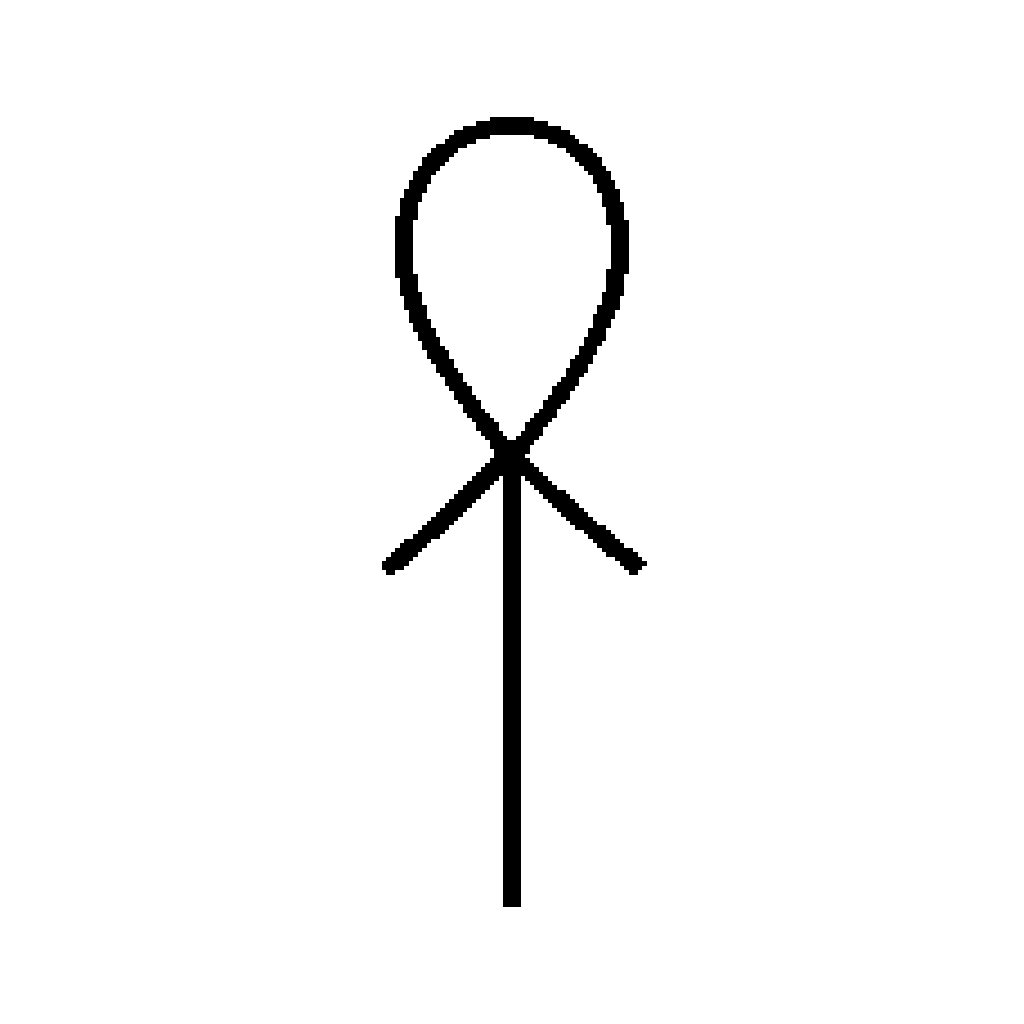} &
\includegraphics[width=0.95\linewidth]{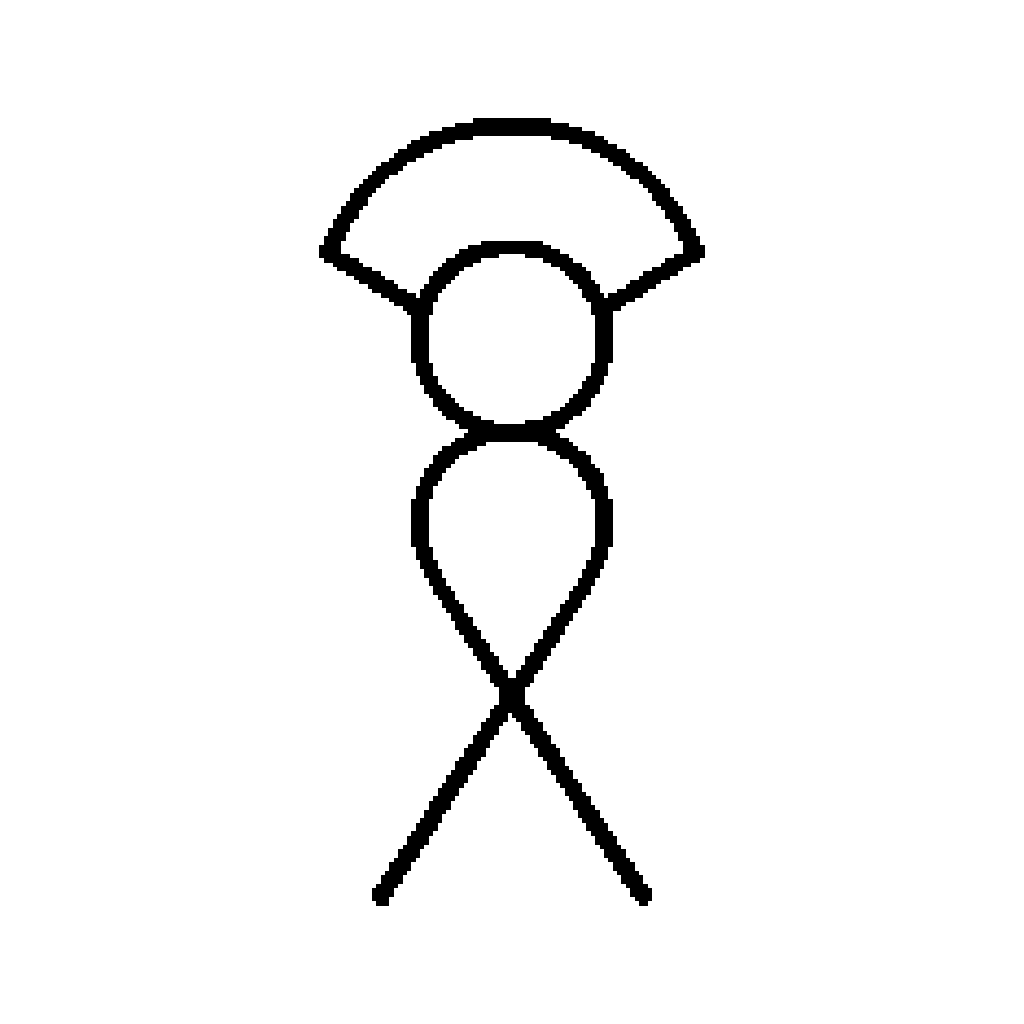} &
\includegraphics[width=0.95\linewidth]{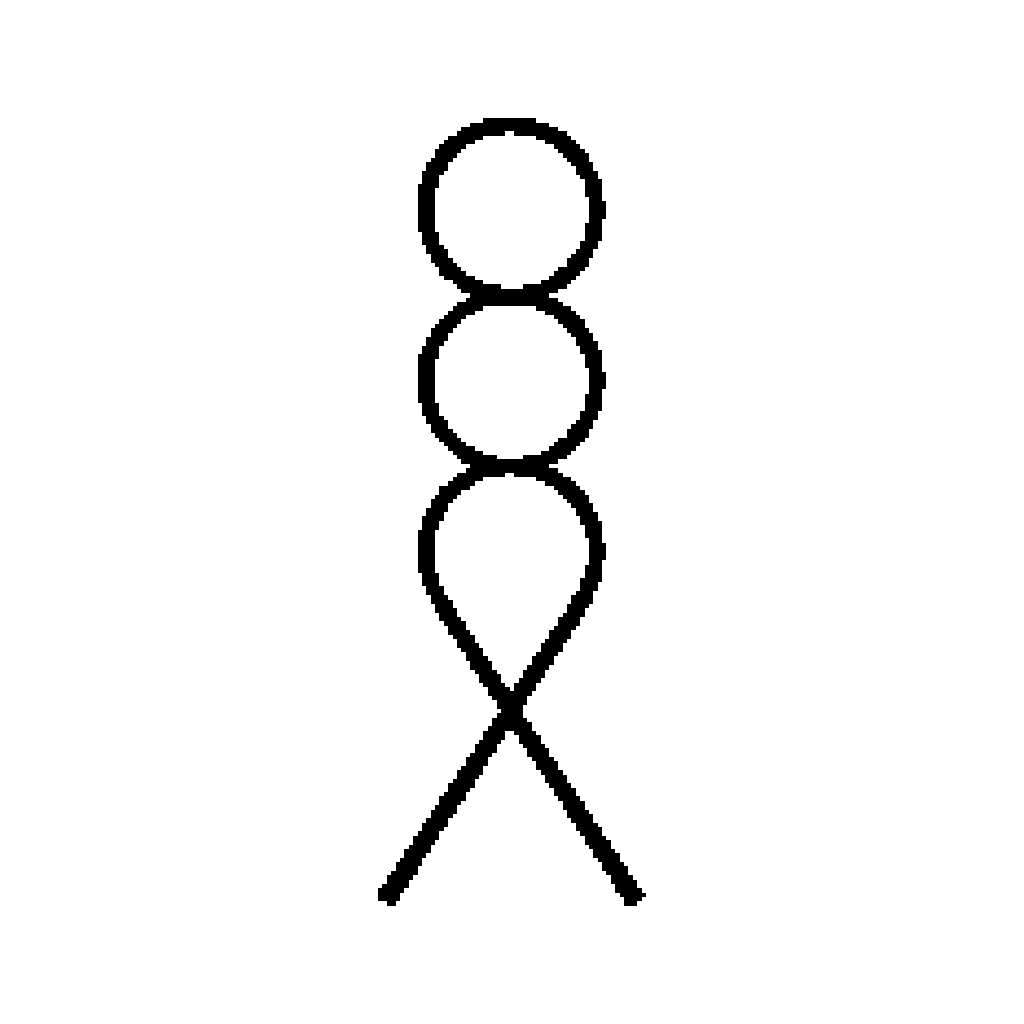} &
\includegraphics[width=0.95\linewidth]{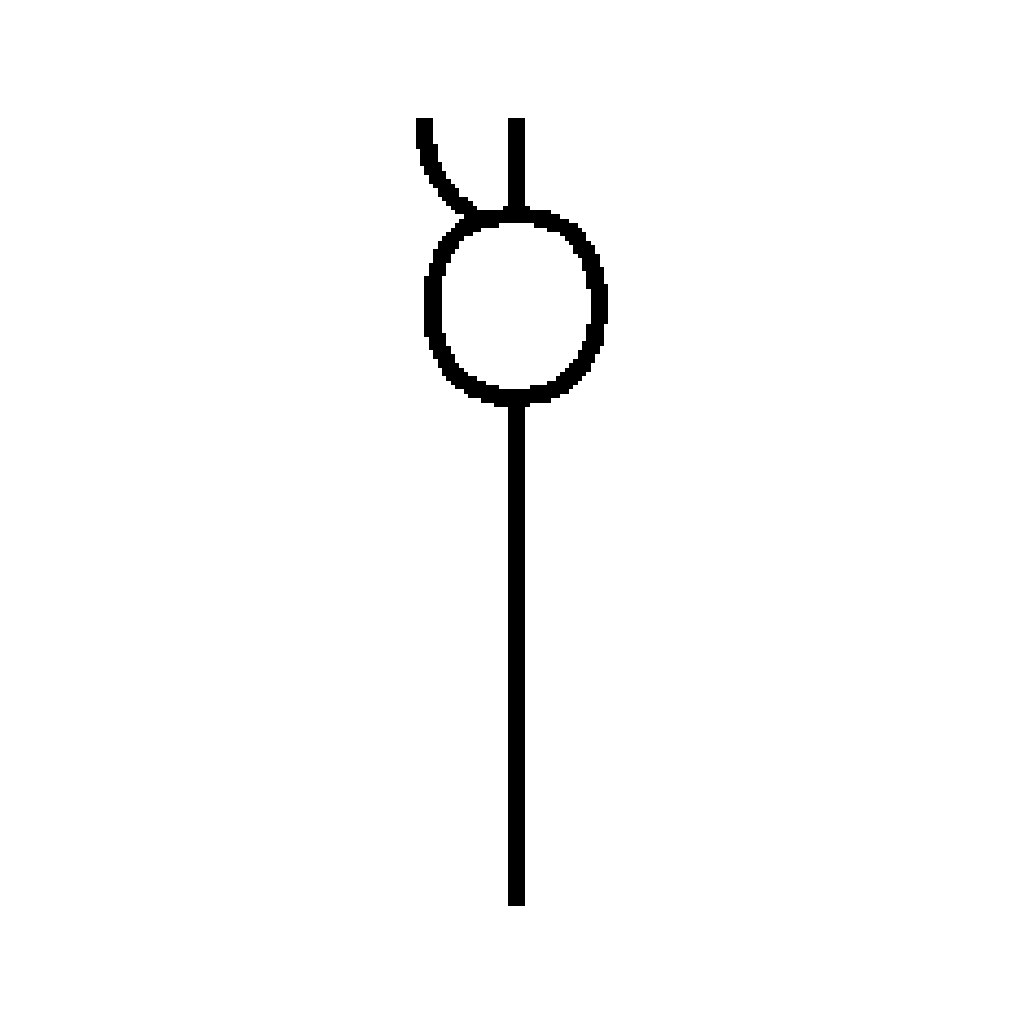} &
\includegraphics[width=0.95\linewidth]{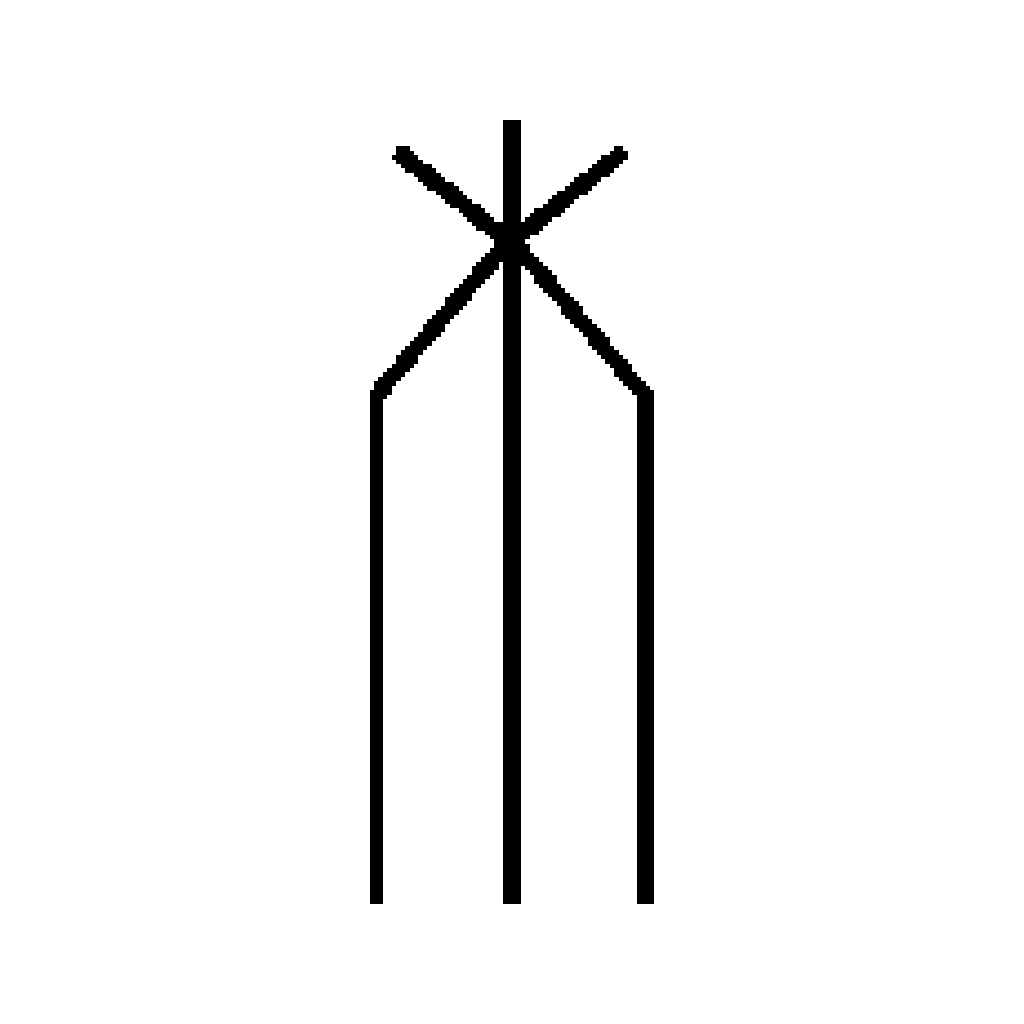} &
\includegraphics[width=0.95\linewidth]{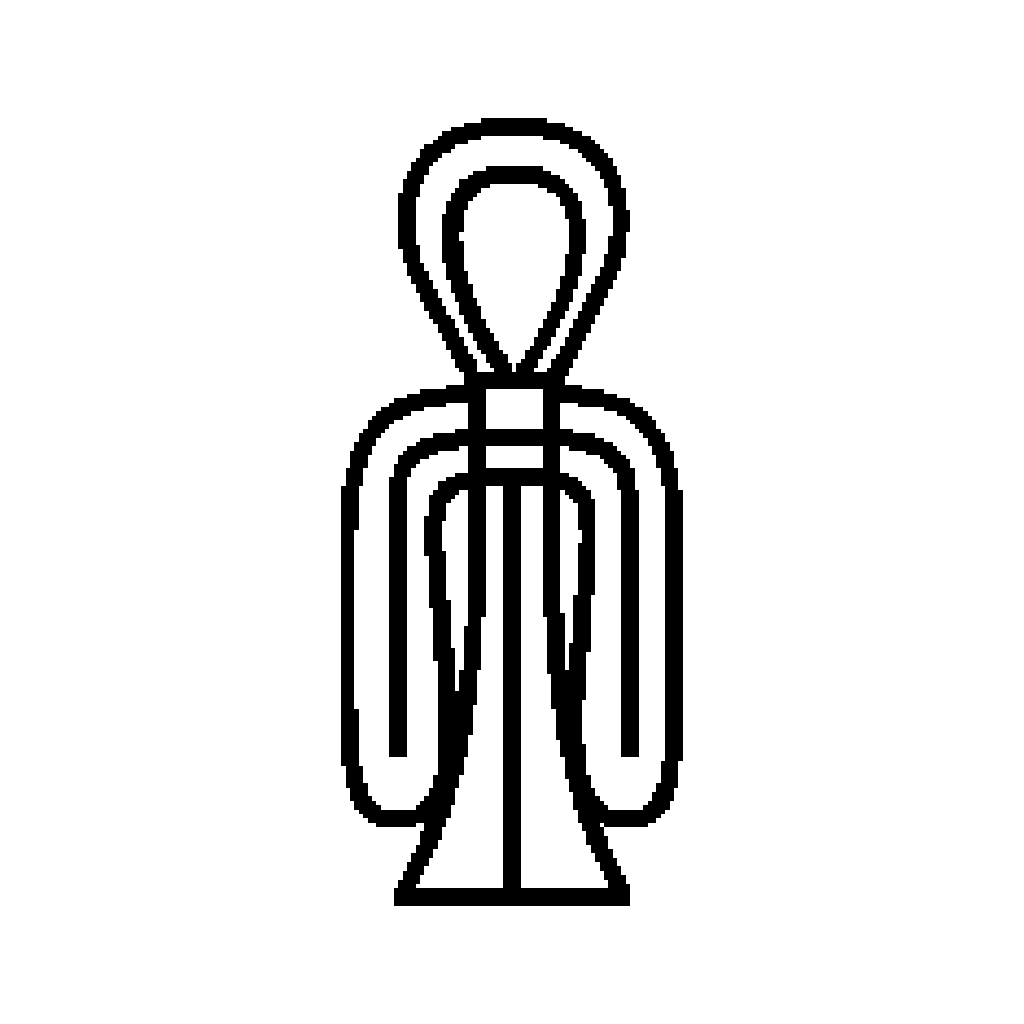} &
\imgwithbox[width=0.95\linewidth]{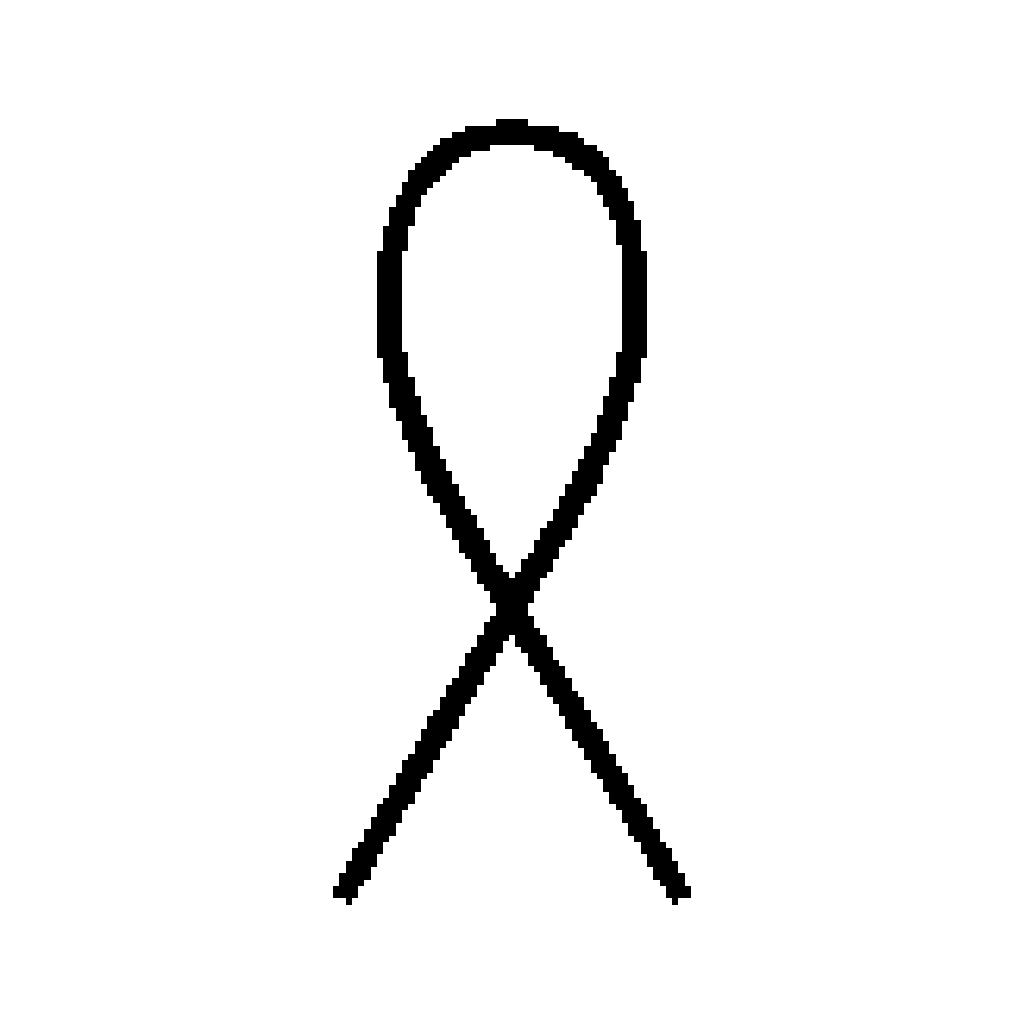} &
\includegraphics[width=0.95\linewidth]{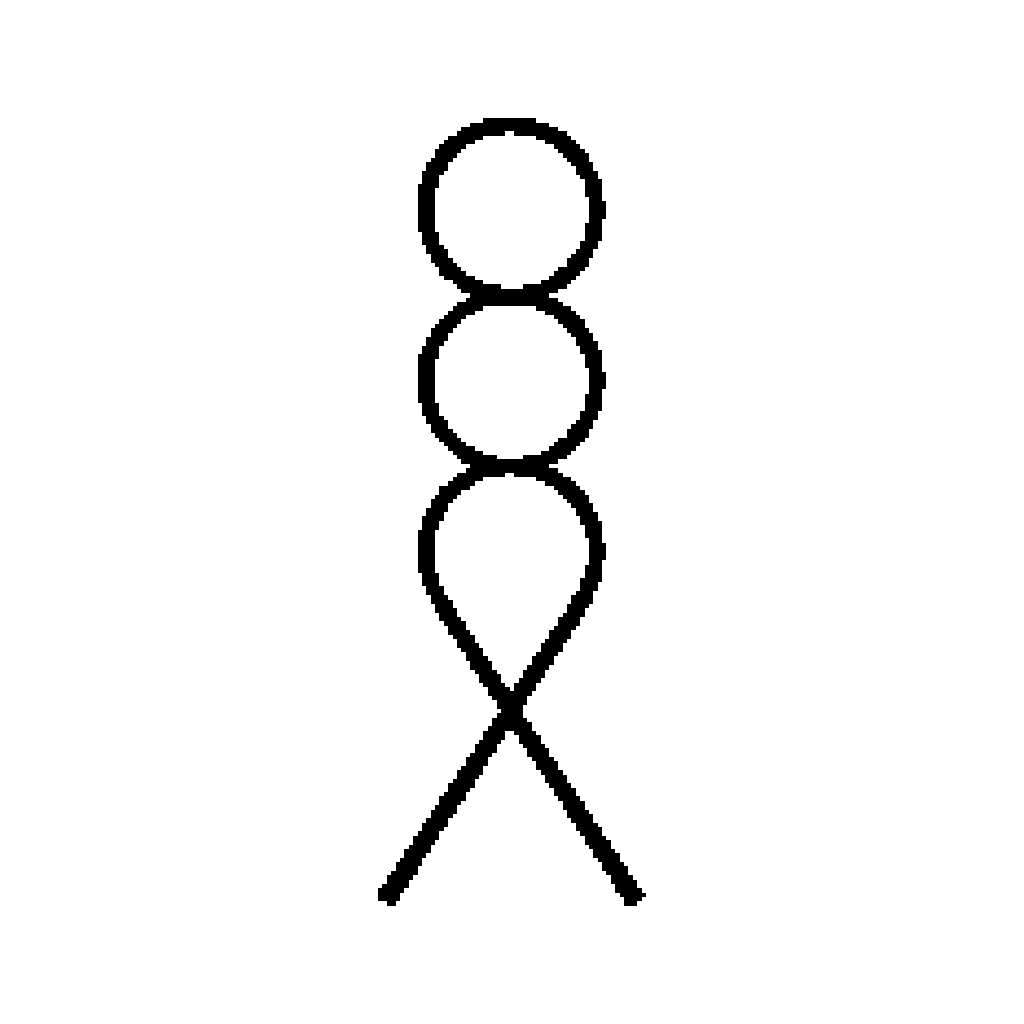} &
\imgwithbox[width=0.95\linewidth]{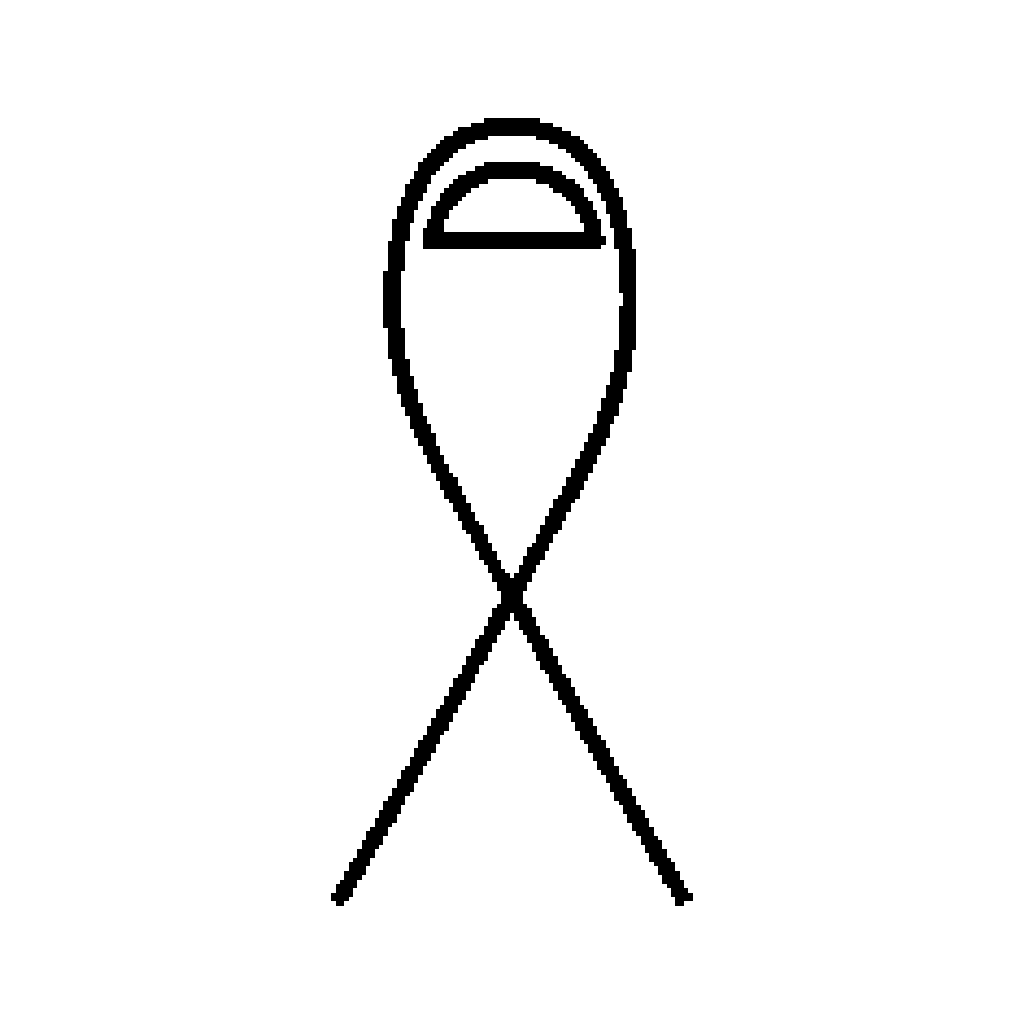} &
\imgwithbox[width=0.95\linewidth]{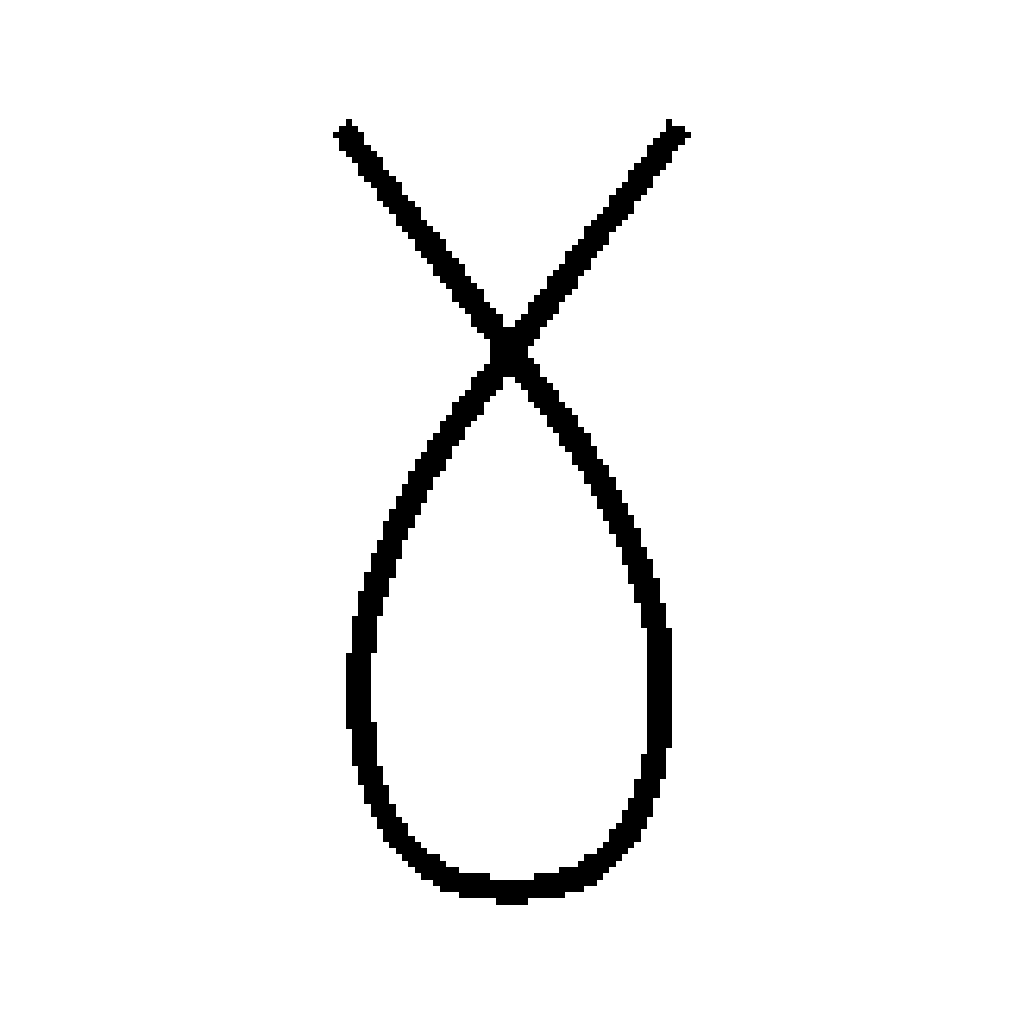} &
\includegraphics[width=0.95\linewidth]{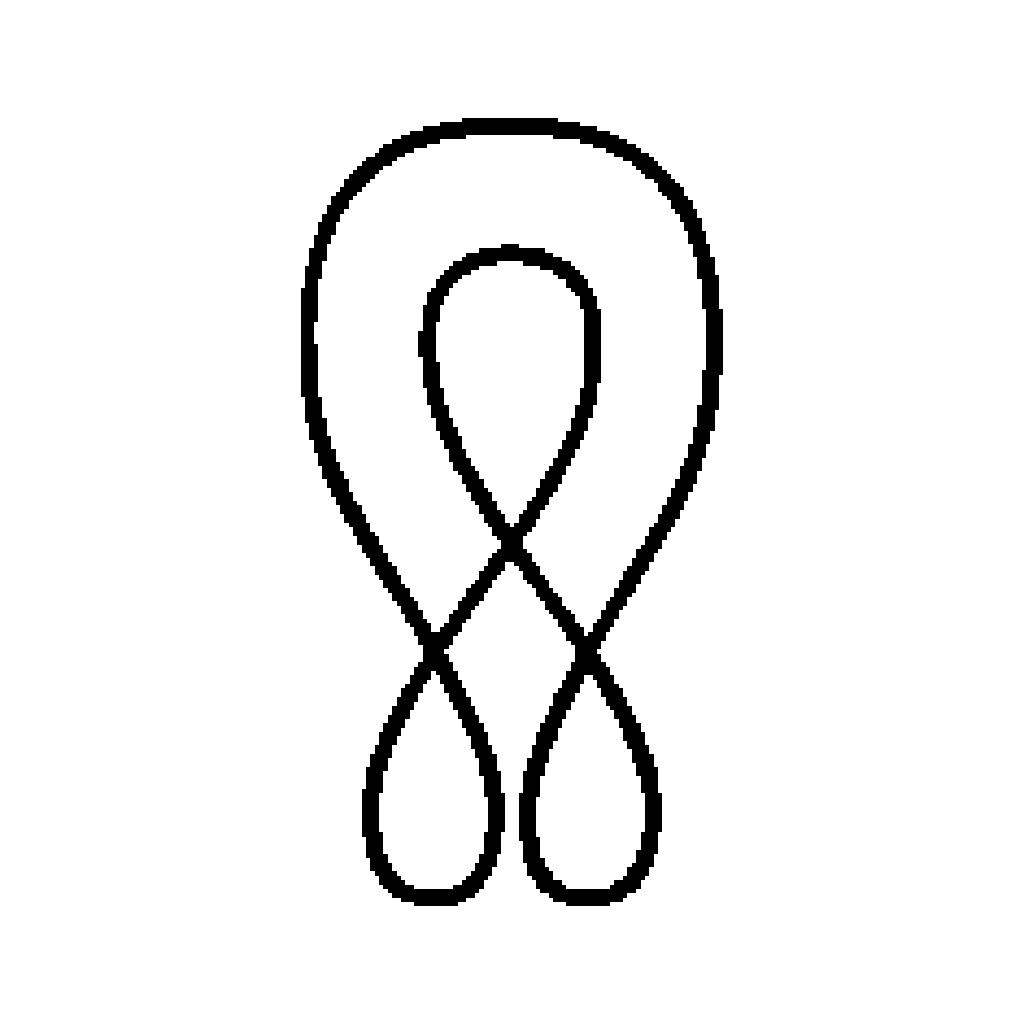} &
\includegraphics[width=0.95\linewidth]{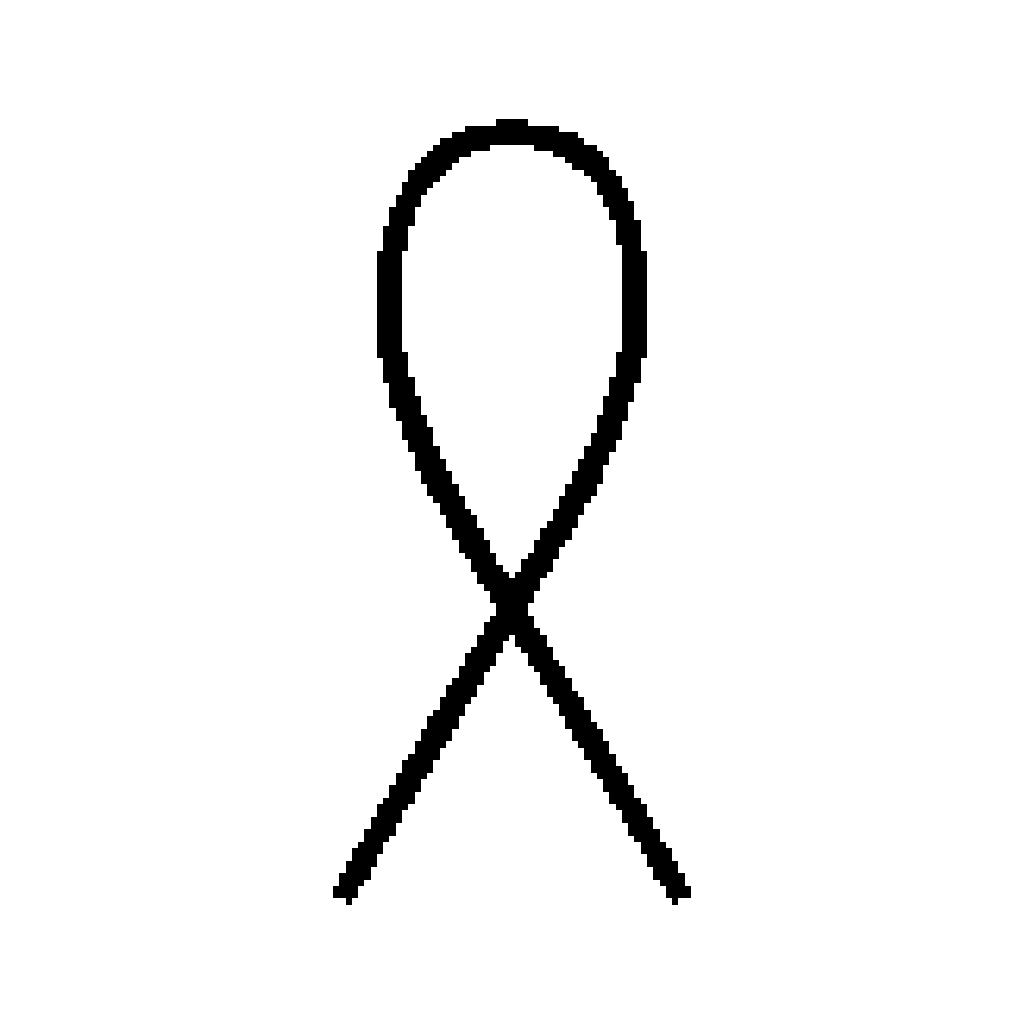} \\
\bottomrule
\end{tabular}}}
\vspace{-6pt}
\caption{Illustrative cases of structure-guided character exploration. We aim to find characters that \textbf{differ in identity but share structural similarity with queries}. For each query, we show top-5 results with and without stroke-guided masking; red boxes highlight cases \textbf{where the characters are structurally similar to the queries while clearly differing in identity}; \textcolor{gray}{gray characters} are the identical ones that discourage exploration. The ``Overlap'' column (\textbf{\textit{for illustration only}}) visualizes shared structures between queries and selected results. See more cases in Appendix~\ref{app:more_cases}.}
\label{tab:character_exploration}
\vspace{-9pt}
\end{table*}

\subsection{Results on Downstream Tasks}\label{sec:results_on_downstream_tasks}

We further demonstrate the practical utility of \methodname on downstream tasks, structure-guided hieroglyphic character exploration and optical character recognition, showing that the learned underlying logic of strokes can be effectively leveraged in practical character-related tasks.

\subsubsection{Optical Character Recognition}\label{sec:optical_character_recognition}

Optical Character Recognition~(OCR) is a common task that aims to recognize characters from visual inputs. For this task, we examine whether stroke-structure representations parsed by \methodname can improve OCR performance.

We curate datasets using the same font families as those in the Chinese and Japanese test sets. We select 1{,}000 characters rendered from \texttt{JinNianYeYaoJiaYouYa} (Chinese) and \texttt{As~Winter~Comes} (Japanese) for OCR training. For evaluation, we use the same 1{,}000 character identities rendered from \texttt{Source~Han~Sans} (Chinese) and \texttt{BIZ~UDPGothic} (Japanese). We ensure that the characters used for OCR are disjoint from those used for \methodname\ training. The OCR task uses paired image-character supervision, and the model outputs the corresponding character sequence in text form.

We fine-tune Qwen3-VL-4B for the OCR task under different settings, with or without additional stroke representations. All OCR models are trained for two epochs under the same configuration for fair comparison.

Quantitative results are reported in Table~\ref{tab:ocr}. As shown in the table, fine-tuning substantially improves OCR performance over the zero-shot setting. Furthermore, incorporating stroke-structure representations consistently leads to additional performance gains on both Chinese and Japanese datasets, indicating that the parsed stroke information provides complementary structural cues beyond raw visual features.

\subsubsection{Structure-guided Character Exploration}\label{sec:structure_guided_character_exploration}

We demonstrate that the stroke-structure representations produced by \methodname enable the exploration of relationships among hieroglyphic characters by facilitating the matching of structurally similar characters with different identities.

We conduct structure-guided exploration on three writing systems that are generally unfamiliar to contemporary readers: Oracle Bone Script (OBS), Dongba pictographs, an indigenous writing system traditionally used by Naxi priests in southern China, and Egyptian hieroglyphs, an ancient Egyptian writing system (see Appendix~\ref{app:introduction_scripts}).

We compare a structure-guided matching strategy using masks derived from predicted strokes with a direct image-based matching baseline. 

Specifically, given a query character image with detected stroke segments from \methodname, we randomly select a local region and probabilistically mask nearby strokes, producing structure-aware perturbations of the character. We then generate multiple masked variants of the query and compute their visual similarity to all candidate characters using CLIP image embeddings. For each candidate, the maximum similarity score across masking trials is retained, and the top-5 matches are returned. As a baseline, we directly rank candidates using CLIP similarity on the original unmasked query image. See additional experimental details in Appendix~\ref{app:experimental_details_character_exploration}. Representative examples are shown in Table~\ref{tab:character_exploration}.

Overall, we observe that structure-guided exploration yields a richer set of structurally related characters compared to image-based matching baseline. In addition, many of the identified characters exhibit not only stroke-level structural similarity but also semantic resemblance to the query character, enabling more informative character-level analysis.

\vspace{-6pt}
\paragraph{OBS character analysis.}\hspace{-1em}\footnote{For OBS characters, we mainly refer to \emph{Yinxu Jiagu Wen Shiyong Zidian} (A Practical Dictionary of Yinxu Oracle Bone Script) by Rusen Ma (2008).} The OBS characters at positions (1a) and (1j) share a common structural component in the central part of the glyph, which is illustrated in (1l). It depicts human feet, which reflects a pictographic origin related to foot movement or stepping actions. The character at position (1a) conveys the meaning of ``arrival'' or ``reaching a destination'', where the foot component directly encodes the act of movement. In contrast, the character at position (1j) denotes a small landform within water (the three surrounding dots), where the same foot-related component can be interpreted as indicating a place of standing or stable footing within the surrounding environment.

\vspace{-6pt}
\paragraph{Dongba pictographs analysis.}\hspace{-1em}\footnote{For Dongba pictographs, we mainly refer to \emph{Naxi Xiangxing Wenzi Pu} (A Corpus of Naxi Pictographic Script) by Guoyu Fang (2017).} A similar pattern is observed for Dongba pictographs. The characters at positions (3a), (3c), (3h), and (3j) exhibit a shared structural pattern centered on the depiction of a book-like object. Characters at positions (3a), (3c), and (3h) represent variant forms of a pictograph denoting the act of writing records, whereas the character at position (3j) denotes the book itself as an object. Despite these semantic distinctions, the shared pictographic structure reveals a common conceptual grounding in writing practices.

\vspace{-6pt}
\paragraph{Egyptian hieroglyphs analysis.}\hspace{-1em}\footnote{For Egyptian hieroglyphs, we mainly refer to the Egyptian Hieroglyphs block (U+13000--U+1342F) in the Unicode Explorer, based on Gardiner's sign list; see \url{https://unicode-explorer.com/b/13000}.} Consistent evidence is also found in Egyptian hieroglyphs. Stroke-guided masking identifies alternative character variants, such as the pair at positions (5a) and (5h), both of which convey meanings related to ``night''. In addition, the method links multiple variants of the same phonemogram, as illustrated by the correspondence between characters at positions (6a), (6g), and (6i), as well as a closely related phonemogram at position (6j).

Finally, even for characters with undocumented meanings, such as positions (2a) and (4a), stroke-guided masking remains effective in identifying structurally related forms. Our proposed method surfaces plausible structurally related characters. Notably, character (2j) corresponds to a deciphered OBS character meaning ``the outer city wall''.

Taken together, these qualitative results suggest that \textit{structure-guided exploration provides a principled way to uncover latent structural and semantic relationships across diverse, under-documented writing systems.}

\section{Analyses}\label{sec:analyses}

\subsection{Effect of Penalty for Invalid Strokes}\label{sec:ablation_penalty}

\begin{table}[t]
\small
\centering
\begin{tabular}{c|cccc}
\toprule
\textbf{$\alpha$} & \textbf{CO (\%)$\uparrow$} & \textbf{IS (\%)$\downarrow$} & \textbf{CS(\%)$\uparrow$} & \textbf{TS} \\
\midrule
0.00 & 41.7 & 64.6 & 1.8 & 26.6 \\
0.01 & 21.7 & 46.0 & 4.8 & \hphantom{0}4.8 \\
0.10 & 22.0 & 45.2 & 4.8 & \hphantom{0}4.8 \\
0.50 & \hphantom{0}2.7 & 75.7 & 1.4 & \hphantom{0}2.0 \\
\bottomrule
\end{tabular}
\vspace{-6pt}
\caption{Results under different values of the penalty coefficient $\alpha$ in the reward function. Experiments are conducted on the Chinese training set for 100 training steps. CO measures black pixel region coverage, CS indicates the average coverage per stroke, IS denotes the percentage of invalid strokes, and TS indicates the total number of generated strokes. The definitions of CO and IS follow those in Table~\ref{tab:results}.}
\label{tab:penalty_coefficient}
\vspace{-9pt}
\end{table}

We analyze the effect of the penalty coefficient $\alpha$ in the reward aggregation described in Section~\ref{sec:reward_aggregation}, with results summarized in Table~\ref{tab:penalty_coefficient}.

When $\alpha=0$, no penalty is applied to invalid strokes, resulting in excessive stroke generation, low average coverage per stroke, and a high proportion of invalid strokes. Although overall coverage remains relatively high, the generated characters lack coherent and structurally valid shapes, indicating the necessity of the penalty term.

As $\alpha$ increases, both the total number of strokes and the proportion of invalid strokes decrease, demonstrating that the penalty effectively suppresses invalid stroke generation. However, overly large values of $\alpha$ (e.g., $\alpha = 0.5$) impose excessive constraints, reducing coverage and degrading performance.

A moderate penalty coefficient provides a good balance between suppressing invalid strokes and preserving sufficient coverage. In our experiments, $\alpha = 0.1$ yields reasonable stroke counts, low invalid-stroke ratios, and sufficient coverage, and is therefore adopted.

\subsection{Effect of Number of Stroke Points}\label{sec:ablation_number_of_strokes}

\begin{table}[t]
\small
\centering
\begin{tabular}{c|ccc}
\toprule
\textbf{Number of Points} & \textbf{RE$\uparrow$} & \textbf{CO (\%)$\uparrow$} & \textbf{IS (\%)$\downarrow$} \\
\midrule
2 & 0.311 & 22.0 & 45.2 \\
3 & 0.237 & 13.8 & 51.2 \\
4 & 0.230 & 13.6 & 52.1 \\
\bottomrule
\end{tabular}
\vspace{-6pt}
\caption{Effect of the number of points used to represent a single stroke. Experiments are conducted on the Chinese training set for 100 training steps. Metric definitions follow those used in Table~\ref{tab:results}.}
\label{tab:number}
\vspace{-6pt}
\end{table}

We study the effect of stroke representation granularity on model performance. In the main experiments, each stroke is represented by two points corresponding to its start and end locations. We vary the number of points per stroke to evaluate whether a finer representation is beneficial.

Results in Table~\ref{tab:number} show that using two points consistently achieves the best performance in terms of reward and coverage, while yielding fewer invalid strokes. Increasing the number of points degrades performance across all metrics.

Qualitative inspection indicates that \textit{representing a stroke with more points does not improve geometric modeling}, but instead increases the difficulty of learning stable and valid stroke structures. Therefore, we adopt the two-point representation and model complex characters through the composition of multiple strokes rather than increasing the complexity of individual strokes.

\section{Related Work}\label{sec:related_work}

\subsection{Computational Analysis of Hieroglyphs}\label{sec:computational_analysis_of_hieroglyphs}

Hieroglyphs, such as OBS and Dongba, encode meaning primarily through structural composition, distinguishing them from purely phonetic characters (see Appendix \ref{app:introduction_scripts} for details). Early computational approaches often treated them as discrete text via standardized codes~\cite{morioka2015multiple, lu2019computers}, while more recent work shifts towards visual-centric methods that often formulate the problem as image classification to address pixel-level patterns via Convolutional Neural Networks (CNNs)~\cite{liu2020oracle, liu2024ancient, zhou2025oraclenet} or Vision Transformers (ViT)~\cite{rust2023pixel, kesen2025pixelm4}.

The emphasis on structures naturally aligns hieroglyph analysis with sketch learning, which models abstract line drawings and skeleton patterns~\cite{ha2017neural,singer2020representation,aksan2020cose,suarez2022elsed,pautrat2023deeplsd}. In both domains, semantic context is primarily conveyed by compositional layouts and positioning of strokes, driving methods to prioritize line-based input over texture or appearance~\cite{yu2017sketch,jia2020coupling}. These motivate us to further explore stroke-structure representations for hieroglyph and logograph analysis.

\subsection{Stroke-based Language Modeling}

Stroke-based modeling captures internal character structures in hieroglyphic and logographic writing systems to provide structural and compositional inductive biases for language modeling~\cite{8910369, jiang2024finding}. Previous work encodes characters as stroke sequences or sub-character embeddings via neural encoders~\cite{cao2018cw2vec, xiong2021learning}, demonstrating advantages over purely symbolic representations, and proving effective for downstream tasks such as Named Entity Recognition (NER)~\cite{cao2018cw2vec,yi2023medical}, character recognition~\cite{chen2023stroke,huang2025strokenet}, and Neural Machine Translation (NMT)~\cite{wang-etal-2022-breaking}. Building on these, recent work further integrates stroke-related visual features with MLLMs for ancient hieroglyph recognition~\cite{chen-etal-2025-multi}.

However, these methods largely depend on handcrafted stroke annotations and predefined stroke labels~\cite{assael2022restoring, sommerschield2023machine}, limiting scalability and generalization across diverse writing systems. Contrastly, our \textbf{\methodname} is annotation-free, scalable, and generalizable across languages.

\section{Conclusion}
In this paper, we propose \textbf{\methodname}, a novel and generalizable framework that enables MLLMs to autonomously derive stroke-level structures from character bitmaps without manual annotations. We conduct extensive experiments on modern logographic and ancient hieroglyphic writing systems and demonstrate that \methodname robustly generalizes across diverse writing systems without linguistic priors. It also achieves consistent advancements in downstream tasks such as OCR and structure-based character exploration. Results and case study highlight the potential of \methodname as a scalable tool for the computational linguistic analysis and decipherment of hieroglyphic scripts.

\section*{Limitations}

While our framework demonstrates strong performance across multiple hieroglyphic scripts, several limitations remain. First, the quality of the learned stroke-level representations can be affected by variability in structural complexity and geometric noise in the training data. Developing more effective data curation and structure-aware denoising strategies could further improve robustness. Second, due to computational constraints, our experiments are conducted with moderate-scale models and datasets. Training larger models with more diverse data may further enhance representation quality and generalization. Third, our current framework adopts a single-model setting; incorporating ensemble-based strategies could help stabilize predictions and improve overall performance. Finally, while our method enables structure-guided character exploration, the identification of the most informative exploration results still relies on a limited degree of manual inspection. Future work could investigate automatic filtering or ranking mechanisms to streamline exploration and improve usability.

\section*{Acknowledgements}

This work is supported by Fundamental and Interdisciplinary Disciplines Breakthrough Plan of the Ministry of Education of China (No. JYB2025XDXM101), the National Natural Science Foundation of China (No. 62276152, 62236011), Key Laboratory of Ethnic Language Intelligent Analysis and Security Governance of MOE, Minzu University of China, Beijing, China, and funding from Wuxi Research Institute of Applied Technologies, Tsinghua University under Grant 20242001120.

\bibliographystyle{thunlp_cvpr}
\bibliography{references}

\thunlplogooff
\clearpage
\appendix

\section{Examples of Stroke Coverage Estimation}\label{app:examples_of_stroke_coverage_estimation}

Table~\ref{tab:stroke_coverage} presents examples of polygonal coverage regions estimated for individual strokes. Across Chinese, Japanese, and Oracle Bone Script characters, the estimated regions accurately reflect the spatial extent of individual strokes. The results indicate that the proposed method remains effective even in the presence of complex stroke geometries, such as curved strokes and stroke intersections.

\section{Data Curation}\label{app:data_curation}

Our datasets consist solely of bitmap images generated from SVG glyphs in font files, as well as publicly available datasets of hieroglyphic character images. No additional annotations are used. In this section, we describe the construction of the training set and the test set.

\subsection{Training Set Curation}\label{app:training_set_curation}

Our training dataset includes Chinese characters, Japanese Kanji, and OBS. For Chinese, we select 2{,}000 characters. For each character, bitmap images are rendered from SVG glyphs of six Chinese font families: \texttt{SimHei}, \texttt{KaiTi}, \texttt{SimLi}, \texttt{Microsoft~YaHei}, \texttt{SimYou}, and \texttt{SimSun}, resulting in a total of 12{,}000 images. For Japanese Kanji, we follow the same procedure using six Japanese font families: \texttt{M~PLUS~1p}, \texttt{Zen~Maru~Gothic}, \texttt{Klee~One}, \texttt{Zen~Kurenaido}, \texttt{Noto~Sans~Japanese}, and \texttt{Noto~Serif~Japanese}. For OBS, we collect images from the undeciphered subset of the HUST-OBC dataset~\cite{wang2024open}.

\subsection{Test Set Curation}\label{app:test_set_curation}

Our test set is curated following the same procedure as the training set, with all font families and character identities disjoint from those used for training. For Chinese, test images are rendered from SVG glyphs of two font families, \texttt{JinNianYeYaoJiaYouYa} and \texttt{Source~Han~Sans}, with 500 characters sampled from each font, resulting in a total of 1{,}000 test images. For Japanese Kanji, we follow the same setting, using \texttt{As~Winter~Comes} and \texttt{BIZ~UDPGothic}, with 500 characters per font and 1{,}000 images in total. For OBS, the test set consists of 1{,}000 images collected from the deciphered subset of the HUST-OBC dataset~\cite{wang2024open}.

\begin{table}[t]
\small
\centering
{\setlength{\tabcolsep}{4pt}
\renewcommand{\arraystretch}{1.2}
\newcommand{\imgwithbox}[2][]{
  \begin{tikzpicture}[baseline=(img.base)]
    \node[inner sep=0pt] (img) {\includegraphics[#1]{#2}};
    \draw[gray, line width=1pt] (img.south west) rectangle (img.north east);
  \end{tikzpicture}
}
\begin{tabular}{>{\raggedright\arraybackslash}m{0.07\linewidth}|>{\centering\arraybackslash}m{0.25\linewidth}>{\centering\arraybackslash}m{0.25\linewidth}>{\centering\arraybackslash}m{0.25\linewidth}}
\toprule
\multicolumn{4}{c}{\textbf{Examples of Estimated Coverages of Strokes}} \\
\midrule
ZH &
\imgwithbox[width=0.95\linewidth]{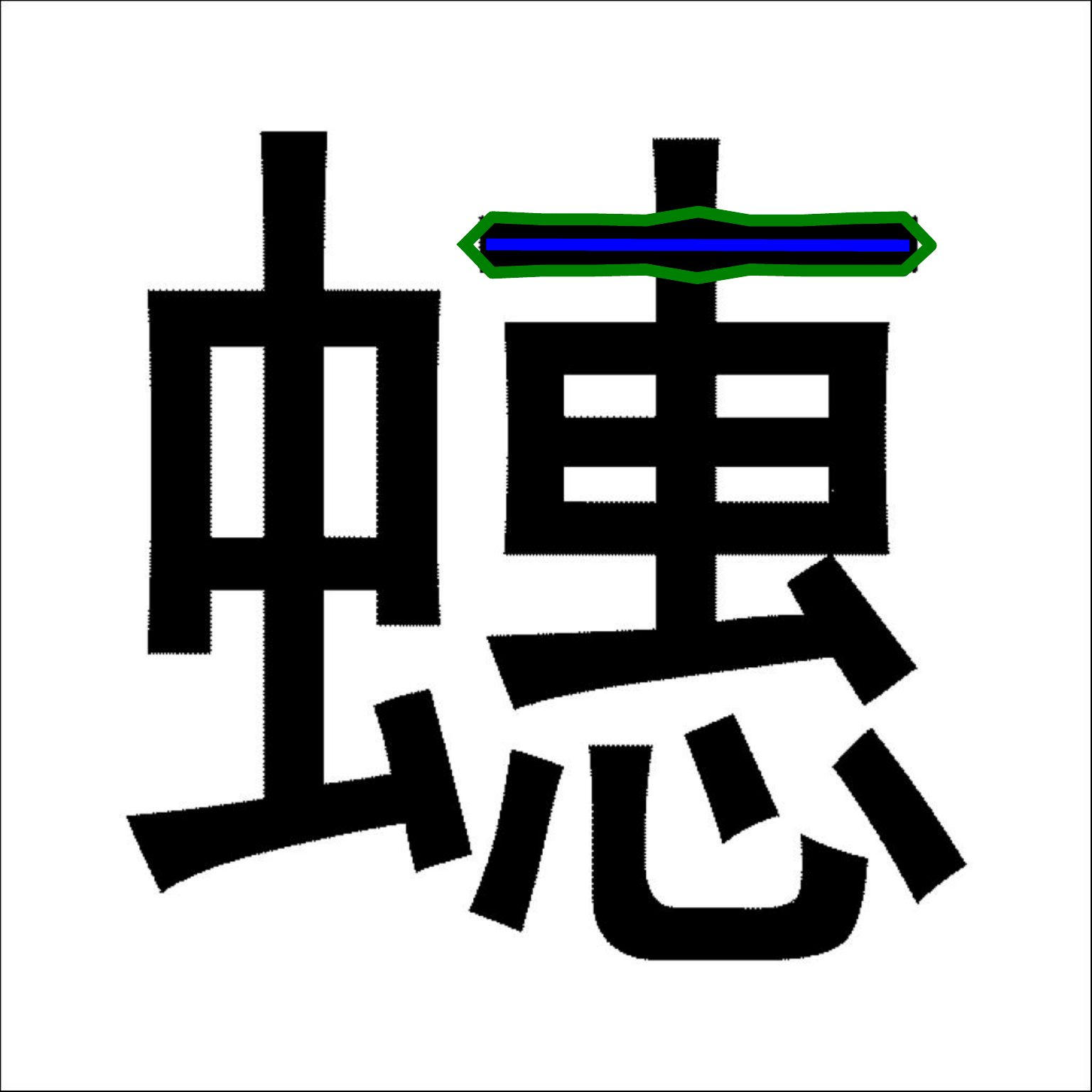} &
\imgwithbox[width=0.95\linewidth]{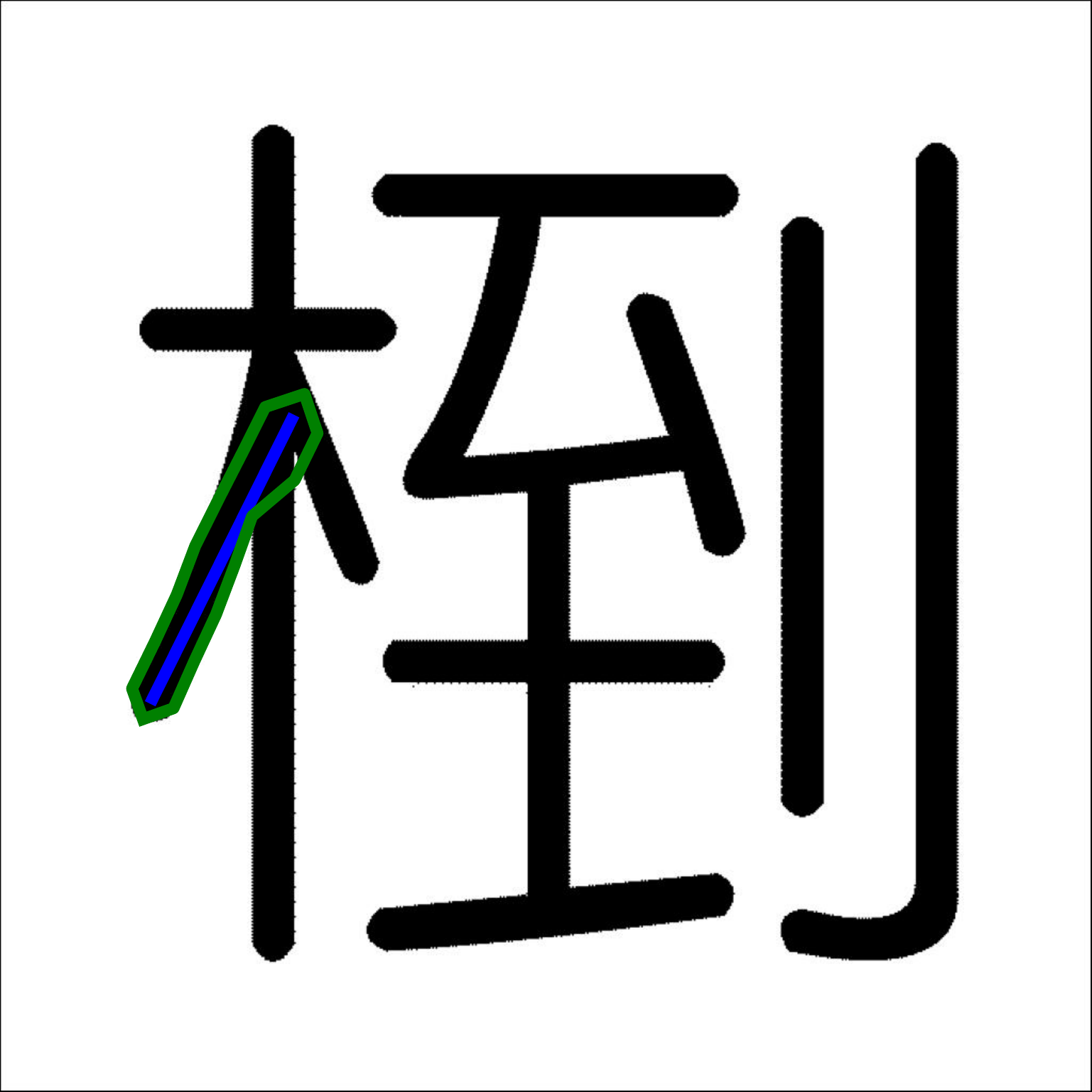} &
\imgwithbox[width=0.95\linewidth]{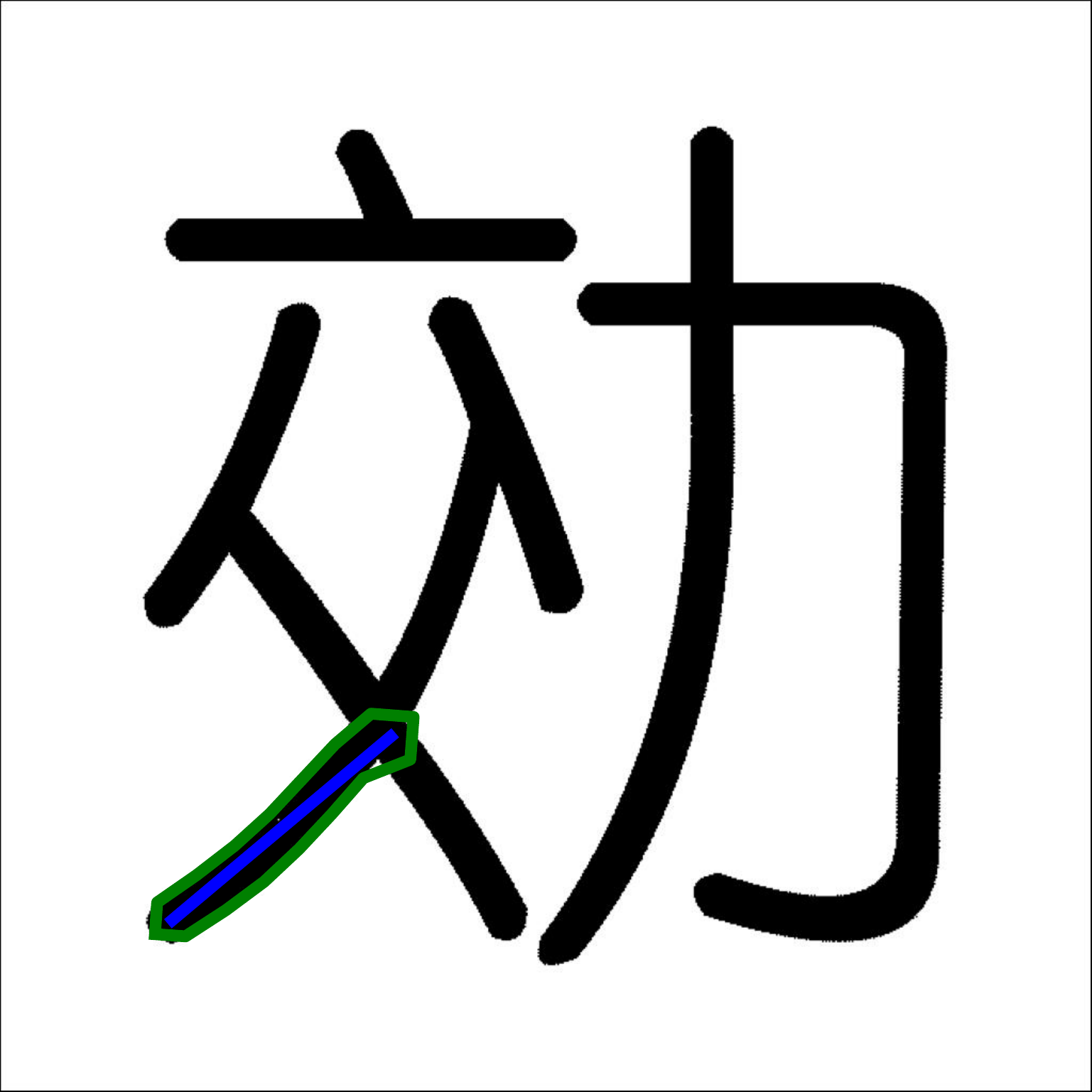} \\
JA &
\imgwithbox[width=0.95\linewidth]{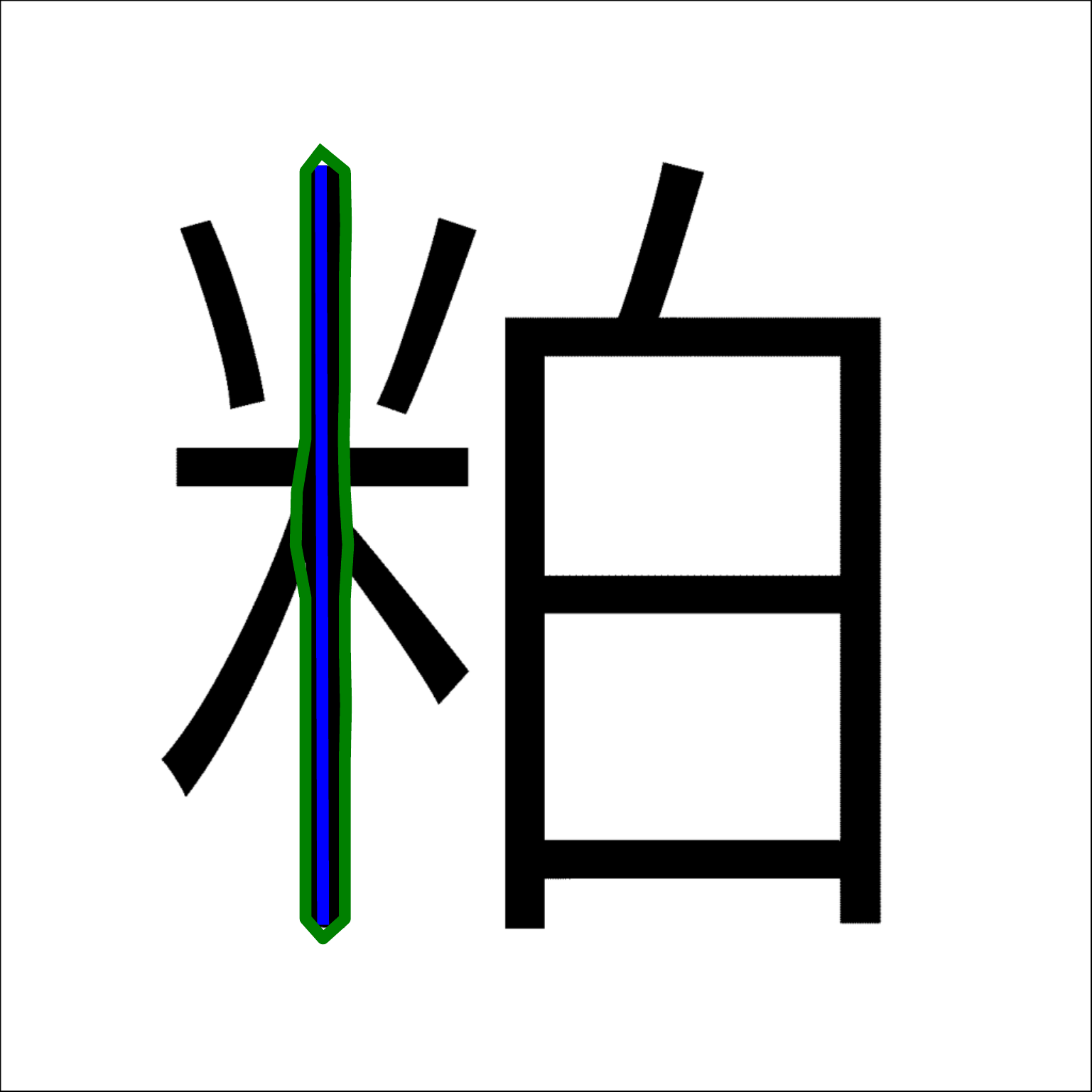} &
\imgwithbox[width=0.95\linewidth]{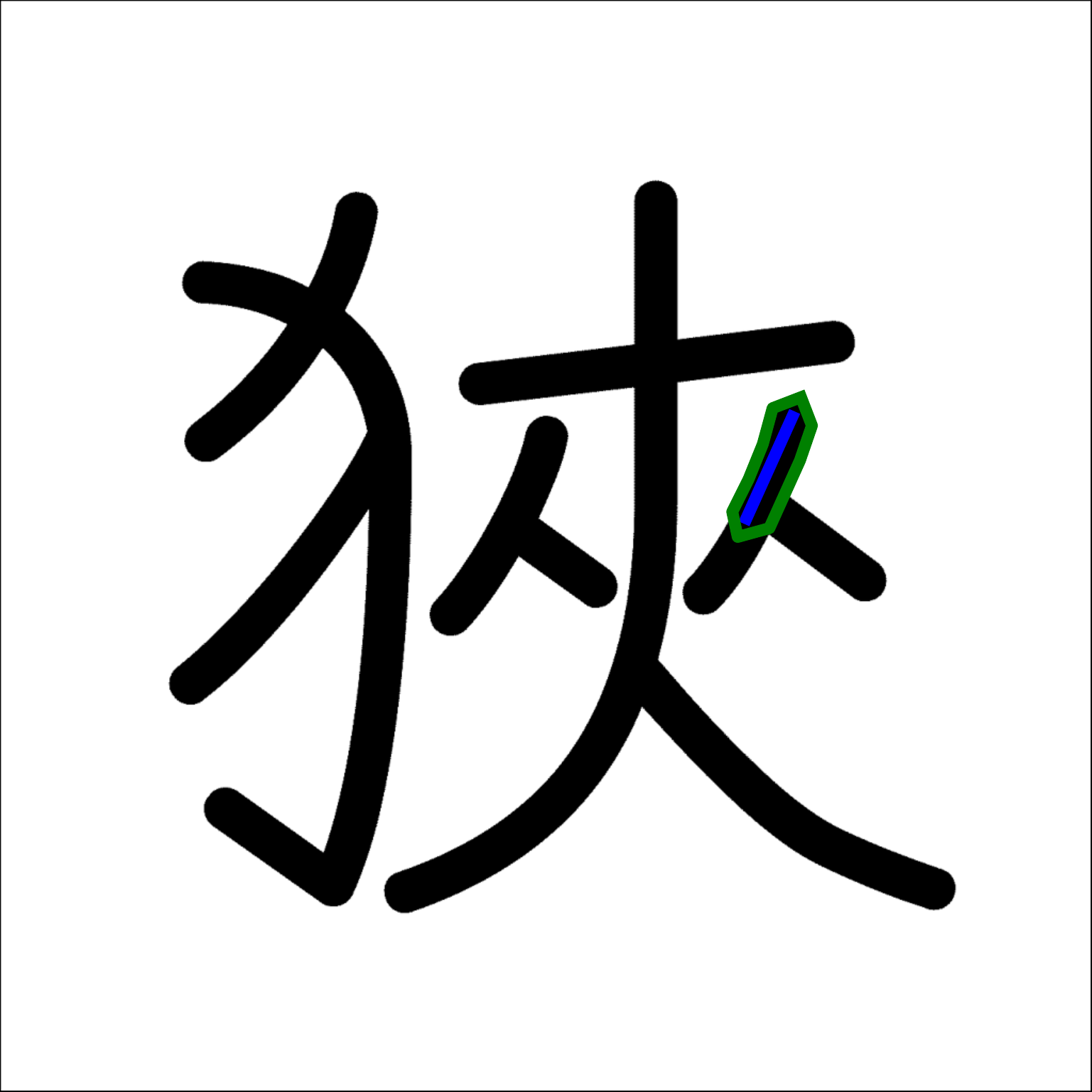} &
\imgwithbox[width=0.95\linewidth]{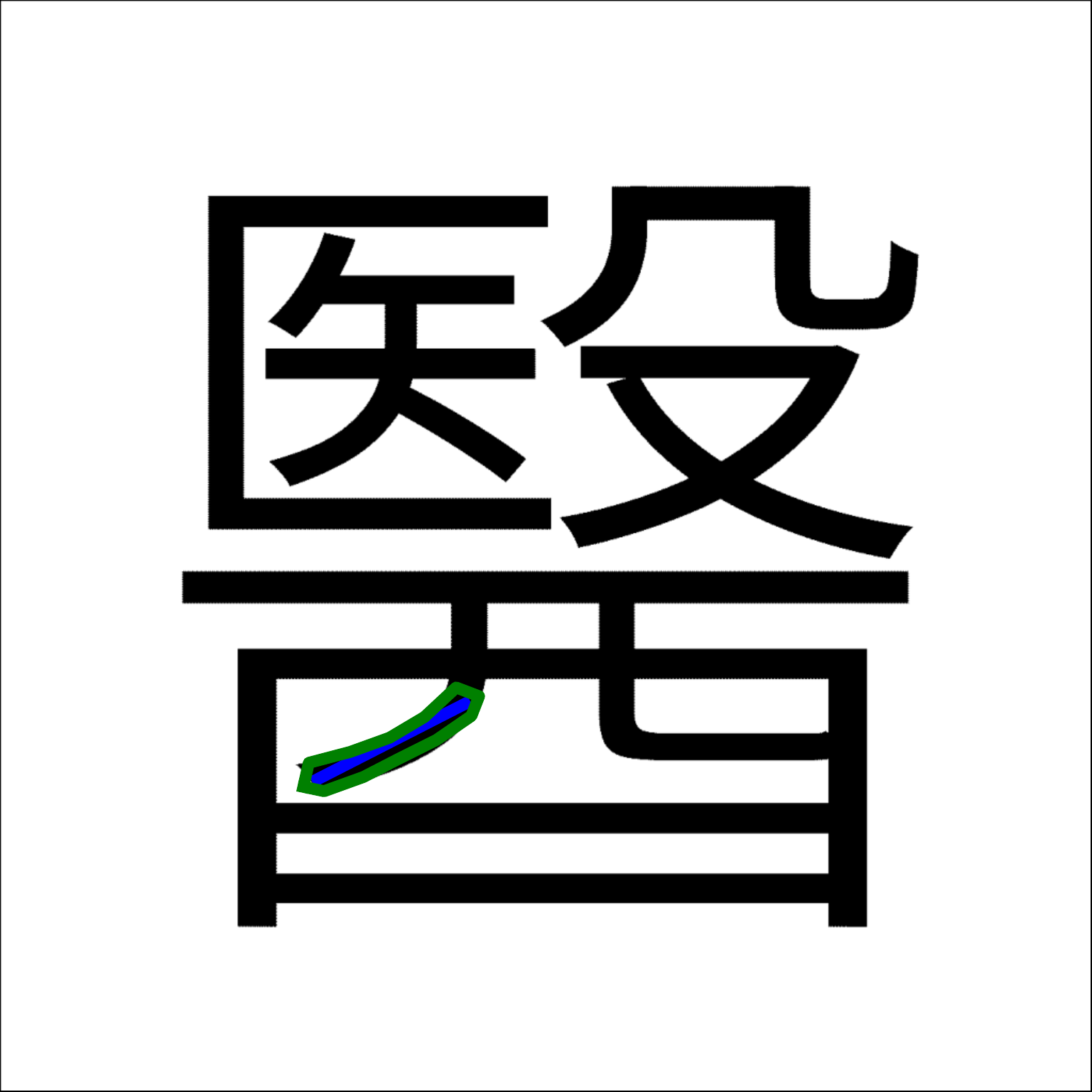} \\
OBS &
\imgwithbox[width=0.95\linewidth]{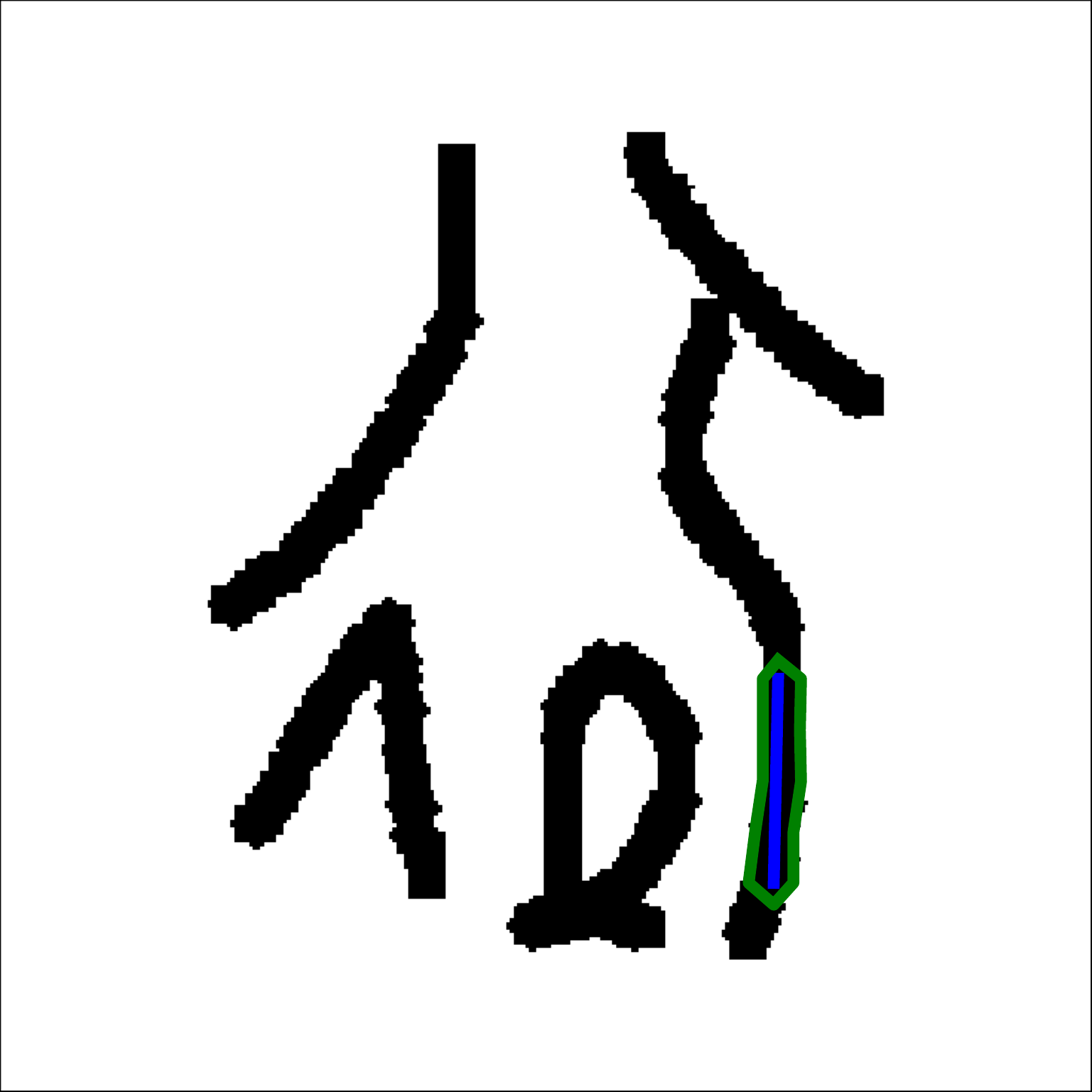} &
\imgwithbox[width=0.95\linewidth]{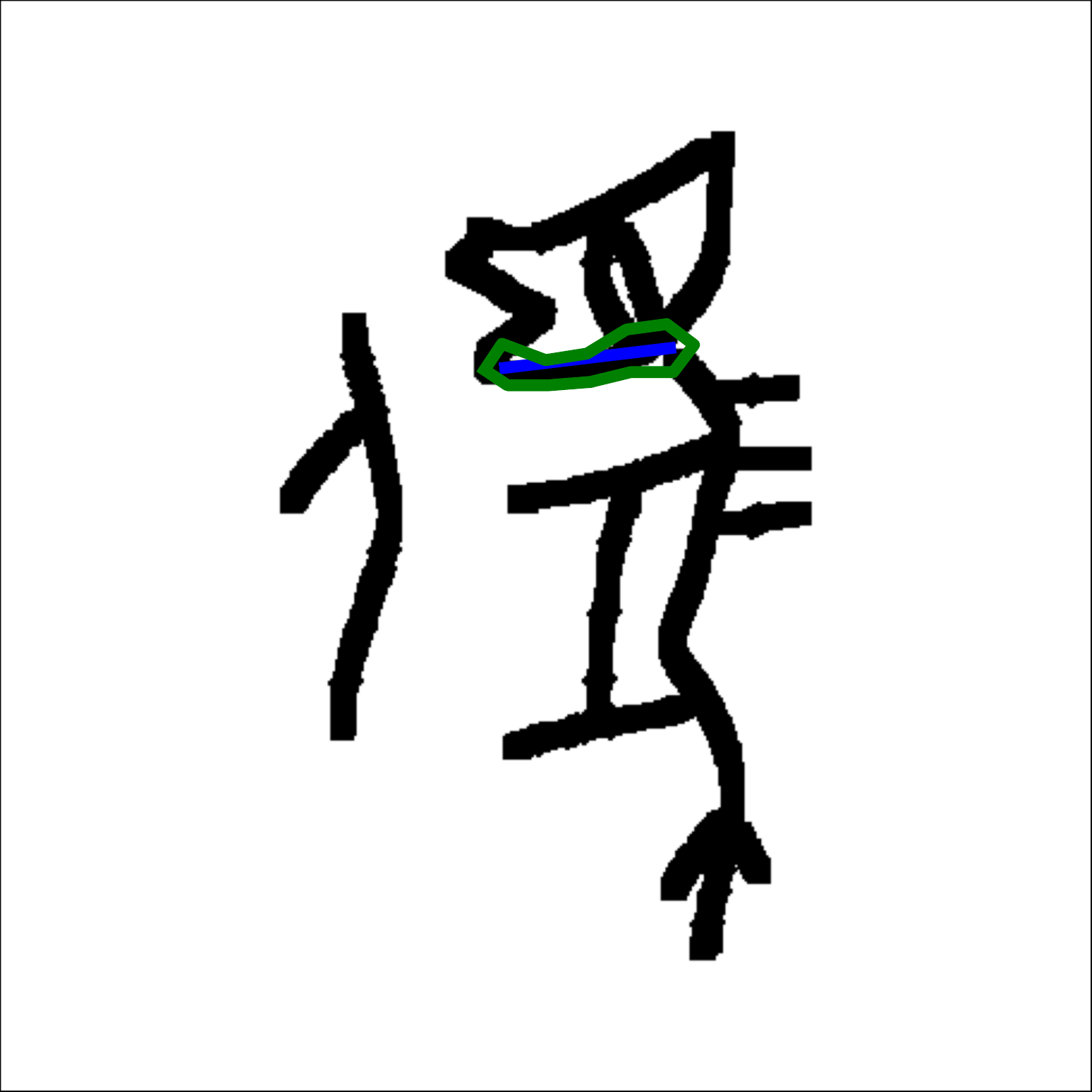} &
\imgwithbox[width=0.95\linewidth]{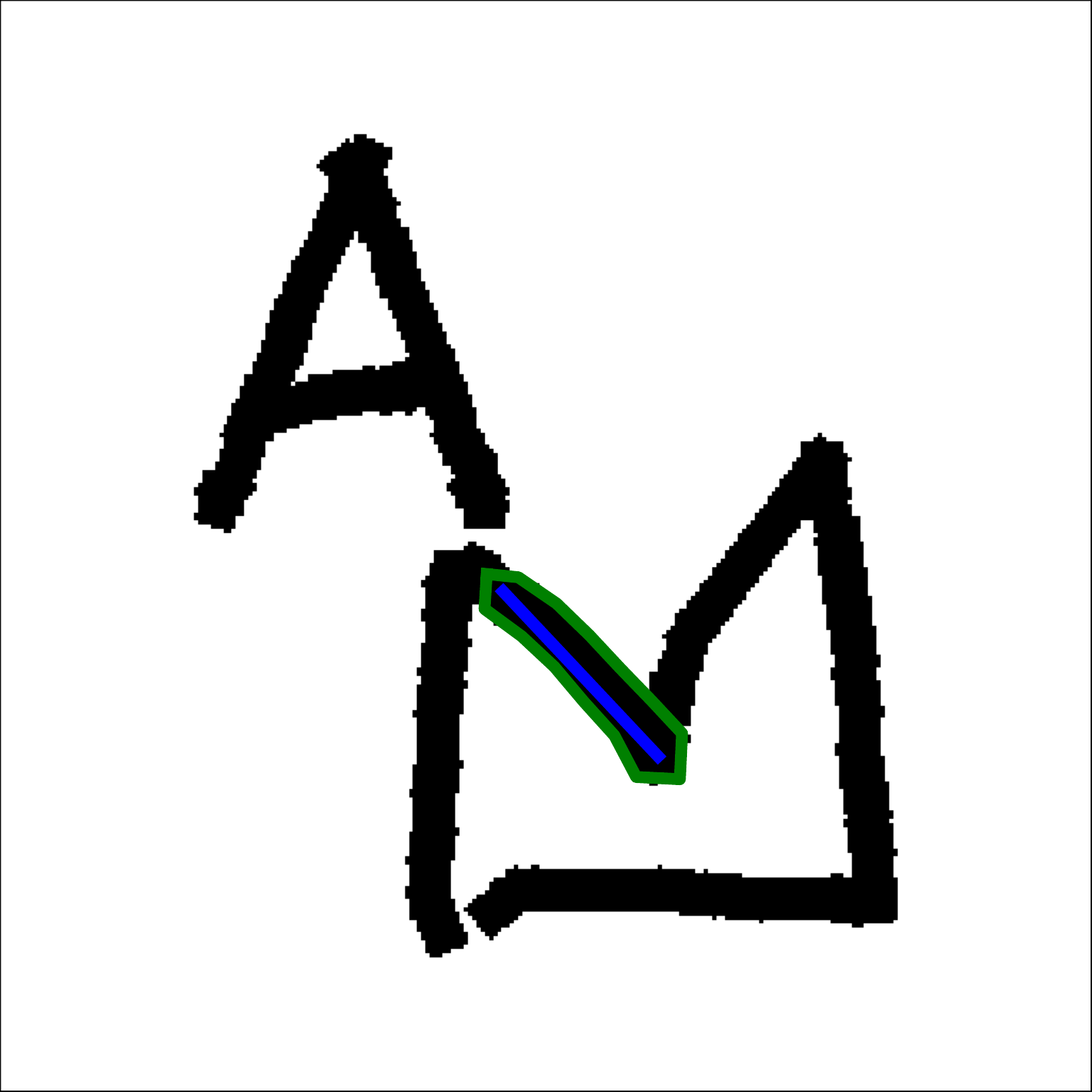} \\
\bottomrule
\end{tabular}}
\vspace{-6pt}
\caption{Examples of estimated polygonal coverage regions for individual strokes across Chinese (ZH), Japanese (JA), and Oracle Bone Script (OBS) characters. For visualization clarity, the original images are cropped to highlight the target regions. Blue curves indicate the strokes, while green polygons denote the corresponding estimated coverage regions.}
\label{tab:stroke_coverage}
\end{table}

\begin{table}[t]
\small
\centering
\begin{tabular}{c|ccc}
\toprule
\textbf{$\beta$} & \textbf{$r_s\uparrow$} & \textbf{CO (\%)$\uparrow$} & \textbf{IS (\%)$\downarrow$} \\
\midrule
0.0625 & 0.172 & 19.3 & 34.3 \\
0.1250 & 0.186 & 22.0 & 45.2 \\
0.2500 & 0.166 & 19.4 & 42.6 \\
0.5000 & 0.158 & 17.8 & 47.1 \\
1.0000 & 0.130 & 14.8 & 60.6 \\
\bottomrule
\end{tabular}
\vspace{-6pt}
\caption{Effect of the format reward coefficient $\beta$. Experiments are conducted on the Chinese training set for 100 training steps. $r_s$ denotes the stroke representation reward defined in
Section~\ref{sec:reward_aggregation}. Other metrics follow the definitions used in Table~\ref{tab:results}.}
\label{tab:format_reward}
\vspace{-6pt}
\end{table}

\begin{table}[t]
\small
\centering
\begin{tabular}{c|ccc}
\toprule
\textbf{$\beta$} & \textbf{RE$\uparrow$} & \textbf{CO (\%)$\uparrow$} & \textbf{IS (\%)$\downarrow$} \\
\midrule
w/o Coordinate System & 0.277 & 19.8 & 50.9 \\
w/ Coordinate System & 0.311 & 22.0 & 45.2 \\
\bottomrule
\end{tabular}
\vspace{-6pt}
\caption{Effect of the overlaid coordinate system. Experiments are conducted on the Chinese training set for 100 training steps. Metric definitions follow those used in Table~\ref{tab:results}.}
\label{tab:coordinate}
\vspace{-6pt}
\end{table}

\section{Training Details}\label{app:training_details}

We train models with a batch size of 32, a rollout size of 8, and a learning rate of $1\times10^{-6}$. The training hyperparameters are fixed to $D=0.05$, $\lambda=1.3$, $\alpha=0.1$, and $\beta=0.125$. All models are trained on each training set for approximately 22 hours using 8 $\times$ NVIDIA A800 (80GB) GPUs.

Ablation studies for $\alpha$ and $\beta$ are presented in Section~\ref{sec:ablation_penalty} and Appendix~\ref{app:ablation_format_reward}, respectively. Ablations for $D$ and $\lambda$ are omitted due to the lack of directly comparable quantitative metrics, while qualitative examples are provided in Appendix~\ref{app:examples_of_stroke_coverage_estimation}.

\section{Effect of Format Reward Coefficient}\label{app:ablation_format_reward}

We analyze the effect of the format reward coefficient $\beta$ in GRPO training, as described in Section~\ref{sec:training_paradigm}. Results under different values of $\beta$ are reported in Table~\ref{tab:format_reward}. As shown in the table, both overly small and large values of $\beta$ lead to degraded training performance. In practice, we select $\beta=0.125$, which achieves the highest stroke representation reward $r_s$.

\section{Effect of Overlaid Coordinate System}\label{app:ablation_coordinate}

We analyze the effect of incorporating the overlaid coordinate system, as described in Section~\ref{sec:implementations}. The results are reported in Table~\ref{tab:coordinate}. As shown in the table, introducing the coordinate system consistently improves performance across all metrics.

\section{Robustness to Rotation}\label{app:robustness_to_rotation}

To further evaluate the generalization ability of the proposed method, we conduct an additional robustness experiment on the Chinese test set under random image rotations. Specifically, for each test image, a rotation angle is independently sampled from $\{0^\circ,90^\circ,180^\circ,270^\circ\}$ , and the corresponding rotation is applied to generate a randomly rotated version of the test set. This setting simulates practical scenarios in which scanned documents or handwritten inputs may appear with inconsistent orientations. The quantitative results are reported in Table~\ref{tab:rotation_robustness}. Overall, HieroSA maintains highly consistent performance under random rotations without using any rotation-based data augmentation during training. This suggests that our method possesses strong robustness at test time.

\section{Failure Modes}\label{app:failure_modes}

Although our method achieves strong overall performance, several failure modes can still be observed. Representative examples can also be found in Table~\ref{tab:qualitative_results}. Repetitive strokes can be observed in (3a), (3b) and (3c), where visually similar stroke patterns are generated multiple times within a character. Abnormally short strokes are illustrated in (1a), suggesting inaccuracies in geometric length or endpoint control. Missing strokes appear in (1b), (2b) and (3b), where certain structural components are not fully generated. These examples indicate that further improvements are still needed in local stroke coordination, fine-grained geometric consistency, and complete structural composition.

\begin{table}[t]
\centering
\small
\resizebox{1.\linewidth}{!}{
\begin{tabular}{l|ccc|ccc}
\toprule
\multirow{2}{*}[-3pt]{\textbf{Model}} & \multicolumn{3}{c|}{\textbf{Chinese (ZH)}} & \multicolumn{3}{c}{\textbf{Rotation (RO)}} \\
\cmidrule(rl){2-4}\cmidrule(rl){5-7}
& \textbf{RE$\uparrow$} & \textbf{CO (\%)$\uparrow$} & \textbf{IS (\%)$\downarrow$} & \textbf{RE$\uparrow$} & \textbf{CO (\%)$\uparrow$} & \textbf{IS (\%)$\downarrow$} \\\midrule
Qwen3-VL-4B & 0.032 & \hphantom{0}0.5 & 97.9 & 0.038 & \hphantom{0}0.7 & 97.3 \\
\methodname~(ZH) & 0.837 & 78.5 & \hphantom{0}6.1 & 0.830 & 78.2 & \hphantom{0}6.6 \\
\methodname~(JA) & 0.756 & 72.2 & 10.2 & 0.760 & 72.9 & 10.3 \\
\methodname~(OBS) & 0.446 & 64.6 & 23.1 & 0.427 & 64.6 & 24.2 \\
\bottomrule
\end{tabular}}
\vspace{-9pt}
\caption{Performance comparison under random image rotations on the Chinese test set.}
\label{tab:rotation_robustness}
\vspace{-6pt}
\end{table}

\section{Early Hieroglyphic Writing Systems}\label{app:introduction_scripts}

Early hieroglyphic writing systems vary widely in form and function across historical and cultural contexts. Compared to modern writing systems, these scripts preserve rich pictographic elements, making visual structure central to their interpretation. In this work, we focus on Oracle Bone Script, Dongba script, and Egyptian Hieroglyphs—three representative writing systems originating from ancient Chinese and Egyptian civilizations (see Table~\ref{tab:script_comparison} for a comparison).

\subsection{Oracle Bone Script}\label{app:introduction_obs}

Oracle Bone Script (OBS), dating back to the Shang Dynasty, is one of the earliest known writing systems in ancient China. OBS is characterized by a distinctive visual and topological structure, with symbols that closely reflect pictorial representations of objects, actions, and concepts~\cite{guan2024deciphering}. The script exhibits a high degree of visual variability and structural complexity, where meaning is often conveyed through the spatial arrangement and configuration of strokes. These properties make visual structure a fundamental component in the interpretation of OBS glyphs.

\subsection{Dongba Script}\label{app:introduction_dongba}

The Dongba script, used by the Naxi people in Yunnan Province, China, is widely regarded as the only living pictographic writing system in the world \cite{fuquan2012}. Dongba manuscripts are organized as spatially structured visual compositions rather than strictly sequential inscriptions, with glyphs arranged across a two-dimensional plane to convey meaning. Interpretation therefore depends on the relative spatial placement and visual relationships among symbols, rather than on a fixed reading order. A defining property of Dongba is its visual compositionality, whereby semantic distinctions are expressed through graphic variation at the stroke level. Glyph meaning can be modified by adding, removing, or altering visual components, allowing related concepts to be represented through systematic pictorial transformations.

\subsection{Egyptian Hieroglyphs}\label{app:introduction_egyptian_hieroglyphs}

Egyptian hieroglyphs served as the formal writing system of ancient Egypt and constitute a complex logophonetic system \cite{allen1951egyptian}, rather than a collection of simple drawings. Although hieroglyphs are organized under a standardized classification scheme known as Gardiner's Sign List, this system primarily serves as a descriptive taxonomy of glyph forms. The interpretation of signs and inscriptions depends not only on categorical membership, but also on visual context, including shape, orientation, and spatial arrangement of glyphs.

\begin{table*}[t]
\centering
\small
\begin{tabular}{lccll}
\toprule
\textbf{Script} & \textbf{Origin} & \textbf{Region} & \textbf{Type} & \textbf{Status} \\
\midrule
Oracle Bone Script & $\sim$1600 BC & China (Shang) & Pictograph & Extinct \\
Dongba Script & $\sim$7th Century & China (Yunnan) & Pictograph & \textbf{Living} \\
Egyptian Hieroglyphs & $\sim$3200 BC & Egypt & Logophonetic & Extinct \\
\bottomrule
\end{tabular}
\caption{Comparison of the early hieroglyphic writing systems used in our experiments.}
\label{tab:script_comparison}
\end{table*}

\section{Details for Structure-guided Character Exploration Experiment}\label{app:details_character_exploration}

\subsection{Data Curation}\label{app:data_curation_character_exploration}

We conduct character exploration experiments on Oracle Bone Script (OBS), Dongba script, and Egyptian hieroglyphs. For OBS, we use all images in the HUST-OBC dataset~\cite{wang2024open} sourced from HWOBC, which provides glyph images with high visual clarity. For the Dongba script, we render glyph images from the \texttt{BabelStone~Naxi}\footnote{\url{https://www.babelstone.co.uk/Fonts/Naxi.html}} font and additionally include 100 images manually cropped from DongbaMIE~\cite{bi2025dongbamie}. DongbaMIE mainly provides paragraph-level images, from which we manually select and crop clear single-glyph instances, excluding samples that contain multiple overlapping glyphs. For Egyptian hieroglyphs, we render character images using the \texttt{Noto~Sans~Egyptian~Hieroglyphs}\footnote{\url{https://fonts.google.com/noto/specimen/Noto+Sans+Egyptian+Hieroglyphs}} font.

\subsection{Experimental Details}\label{app:experimental_details_character_exploration}

Our character exploration process consists of two stages: generating perturbed query images by masking strokes and matching characters based on cosine similarity of CLIP~\cite{radford2021clip} image embeddings.

Each query is defined by an image $I$ and a set of valid strokes $\mathcal{S}=\left\{\left(\mathbf{p}_s^k,\mathbf{p}_e^k\right)\right\}_{k=1}^n$, where $\mathbf{p}_s^k,\mathbf{p}_e^k\in\mathbb{R}^2$ denote the start and end points of the $k$-th stroke, generated by \methodname~(Chinese).

We first compute the midpoint of each stroke as
\begin{equation}
    \mathbf{p}_m^k=\frac{\mathbf{p}_s^k+\mathbf{p}_e^k}{2}.
\end{equation}
To generate a stochastic masking pattern, we sample a masking center $\mathbf{c}$ uniformly from $\left\{\mathbf{p}_m^k\right\}_{k=1}^n$. For each stroke, we compute its Euclidean distance to the center,
\begin{equation}
    d_k=\left\|\mathbf{p}_m^k-\mathbf{c}\right\|_2,
\end{equation}
and define a distance-decayed weight
\begin{equation}
    w_k=\exp\!\left(-\frac{d_k}{\tau}\right),
\end{equation}
where $\tau>0$ is a temperature parameter controlling the spatial locality of masking. We set $\tau=0.4$ throughout our experiments. To normalize the overall masking strength, we rescale the weights by their mean,
\begin{equation}
    \tilde{w}_k=w_k/\left(\frac{1}{n}\sum_{k=1}^n w_k\right),
\end{equation}
and obtain the stroke-wise discard probability
\begin{equation}
    p_k=\mathrm{clip}\!\left(\rho\,\tilde{w}_k,\,0,\,1\right),
\end{equation}
with base discard rate $\rho\in(0,1)$. We set $\rho=0.5$ throughout our experiments. Each stroke is then independently masked according to $z_k\sim\mathrm{Bernoulli}(p_k)$, where $z_k=1$ indicates that the stroke $\left(\mathbf{p}_s^k,\mathbf{p}_e^k\right)$ is masked. We mask the corresponding polygonal coverage regions of the selected strokes to obtain a perturbed image $I\left(\mathbf{z}\right)$.

For a candidate pool of character images $\mathcal{D}=\{J_m\}_{m=1}^M$, which consists of all characters from the same writing system in our experiment, we compute cosine similarity between masked query and each candidate using CLIP image embeddings\footnote{We employ the version \texttt{clip-vit-large-patch14-336}.},
\begin{equation}
    s_m^t = \mathrm{sim}\!\left(I(\mathbf{z}^t), J_m\right),
\end{equation}
where $t$ indexes the masking trial. We repeat the masking-and-scoring process for $T=3$ independent trials and aggregate the similarities as
\begin{equation}
    \bar{s}_m = \max\left(\left\{s_m^t\right\}_{t=1}^T\right).
\end{equation}
The final matched results are obtained by ranking candidates by $\bar{s}_m$ and selecting the top-$5$ matches.

Our baseline ranks the candidate pool $\mathcal{D}=\{J_m\}_{m=1}^M$ by the CLIP cosine similarity $\mathrm{sim}(I, J_m)$ and selects the top-$5$ candidates. Compared to the baseline, our proposed stochastic masking procedure introduces localized, structure-aware perturbations at the stroke level. As a result, the matching process is biased toward identifying characters that differ in identity while sharing salient structural properties with the query, as observed in our experimental results.

\begin{table*}[t]
\centering
\small
\resizebox{1.\textwidth}{!}{
{\setlength{\tabcolsep}{4pt}
\renewcommand{\arraystretch}{1.2}
\newcommand{\imgwithbox}[2][]{
  \begin{tikzpicture}[baseline=(img.base)]
    \node[inner sep=0pt] (img) {\includegraphics[#1]{#2}};
    \draw[red, line width=1pt] (img.south west) rectangle (img.north east);
  \end{tikzpicture}
}
\newcommand{\imgfade}[2][]{
  \begin{tikzpicture}[baseline=(img.base)]
    \node[inner sep=0pt, opacity=0.1] (img)
      {\includegraphics[#1]{#2}};
  \end{tikzpicture}
}
\begin{tabular}{>{\centering\arraybackslash}m{0.02\linewidth}|>{\centering\arraybackslash}m{0.07\linewidth}|>{\centering\arraybackslash}m{0.07\linewidth}>{\centering\arraybackslash}m{0.07\linewidth}>{\centering\arraybackslash}m{0.07\linewidth}>{\centering\arraybackslash}m{0.07\linewidth}>{\centering\arraybackslash}m{0.07\linewidth}|>{\centering\arraybackslash}m{0.07\linewidth}>{\centering\arraybackslash}m{0.07\linewidth}>{\centering\arraybackslash}m{0.07\linewidth}>{\centering\arraybackslash}m{0.07\linewidth}>{\centering\arraybackslash}m{0.07\linewidth}}
\toprule
& \textbf{a} & \textbf{b} & \textbf{c} & \textbf{d} & \textbf{e} & \textbf{f} & \textbf{g} & \textbf{h} & \textbf{i} & \textbf{j} & \textbf{k} \\
\midrule
& \textbf{Query} & \multicolumn{5}{c|}{\textbf{Direct Image-based Matching Results}} & \multicolumn{5}{c}{\textbf{Stroke-Derived Masked Image Matching Results}} \\
\midrule\noalign{\vskip -3pt}\rowcolor{gray!20}\multicolumn{12}{c}{\textbf{Oracle Bone Scripts}} \\\noalign{\vskip -2pt}\midrule
1 &
\includegraphics[width=0.95\linewidth]{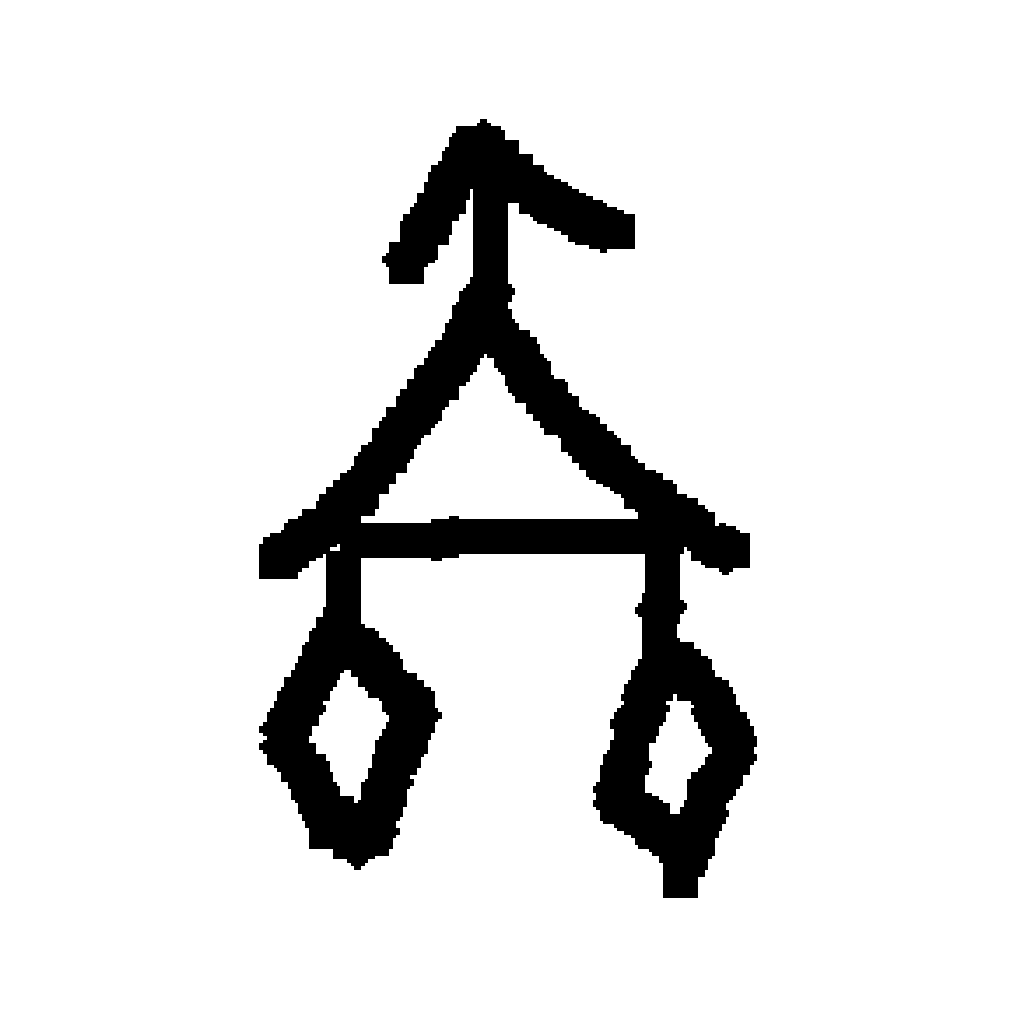} &
\imgfade[width=0.95\linewidth]{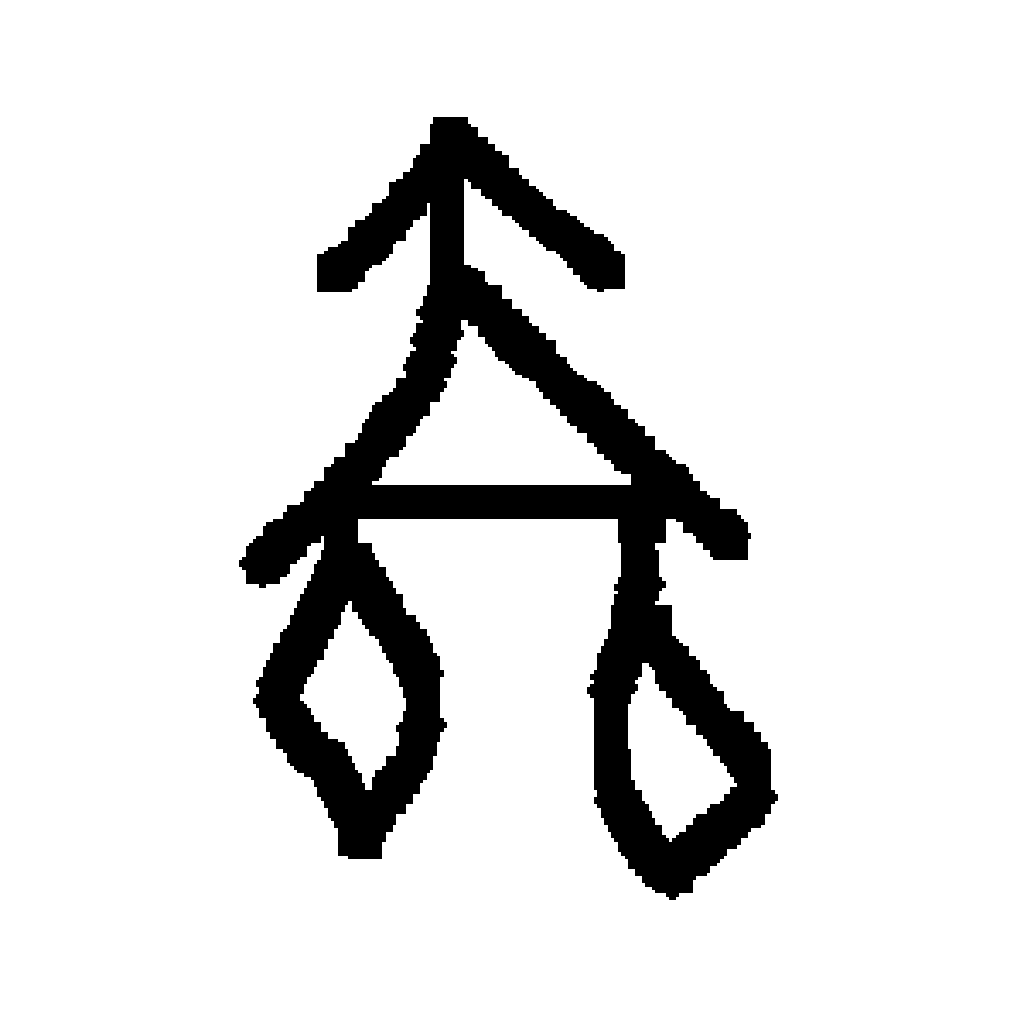} &
\imgfade[width=0.95\linewidth]{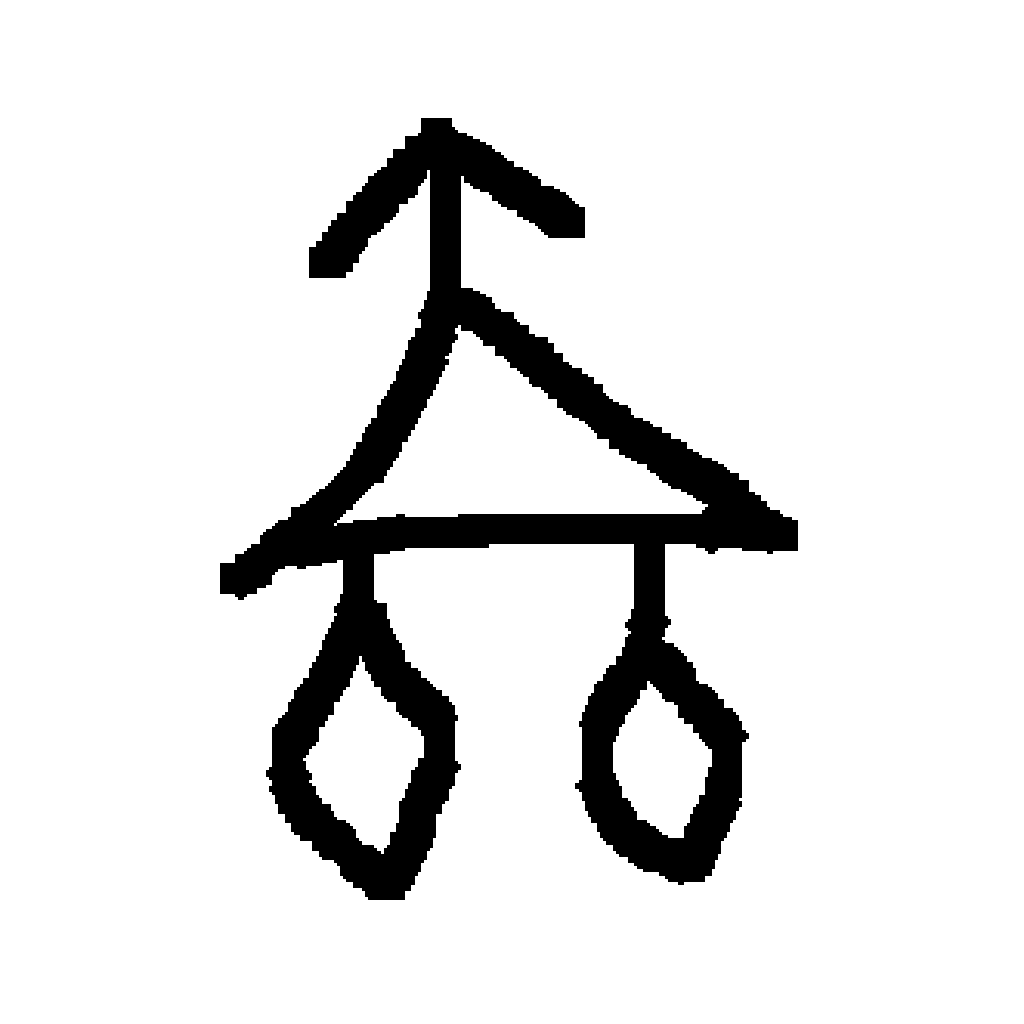} &
\imgfade[width=0.95\linewidth]{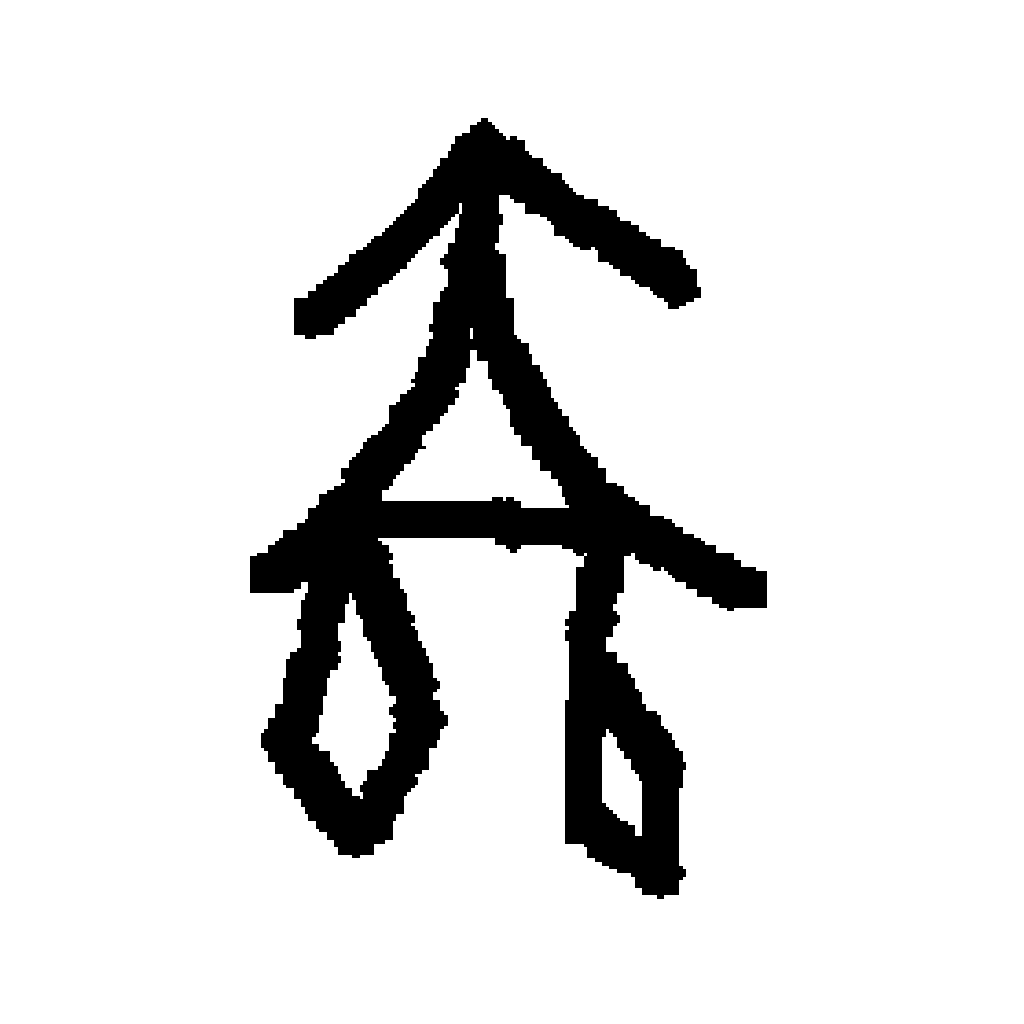} &
\imgfade[width=0.95\linewidth]{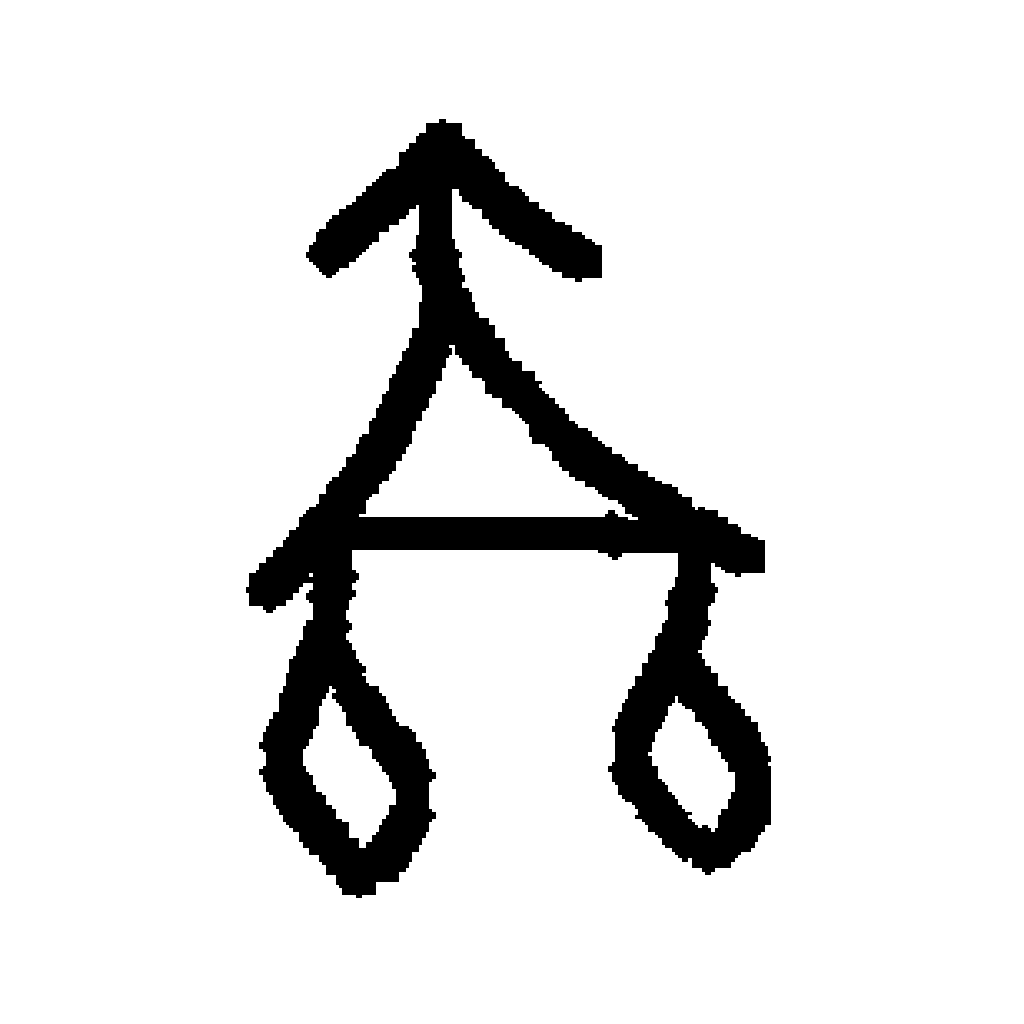} &
\imgfade[width=0.95\linewidth]{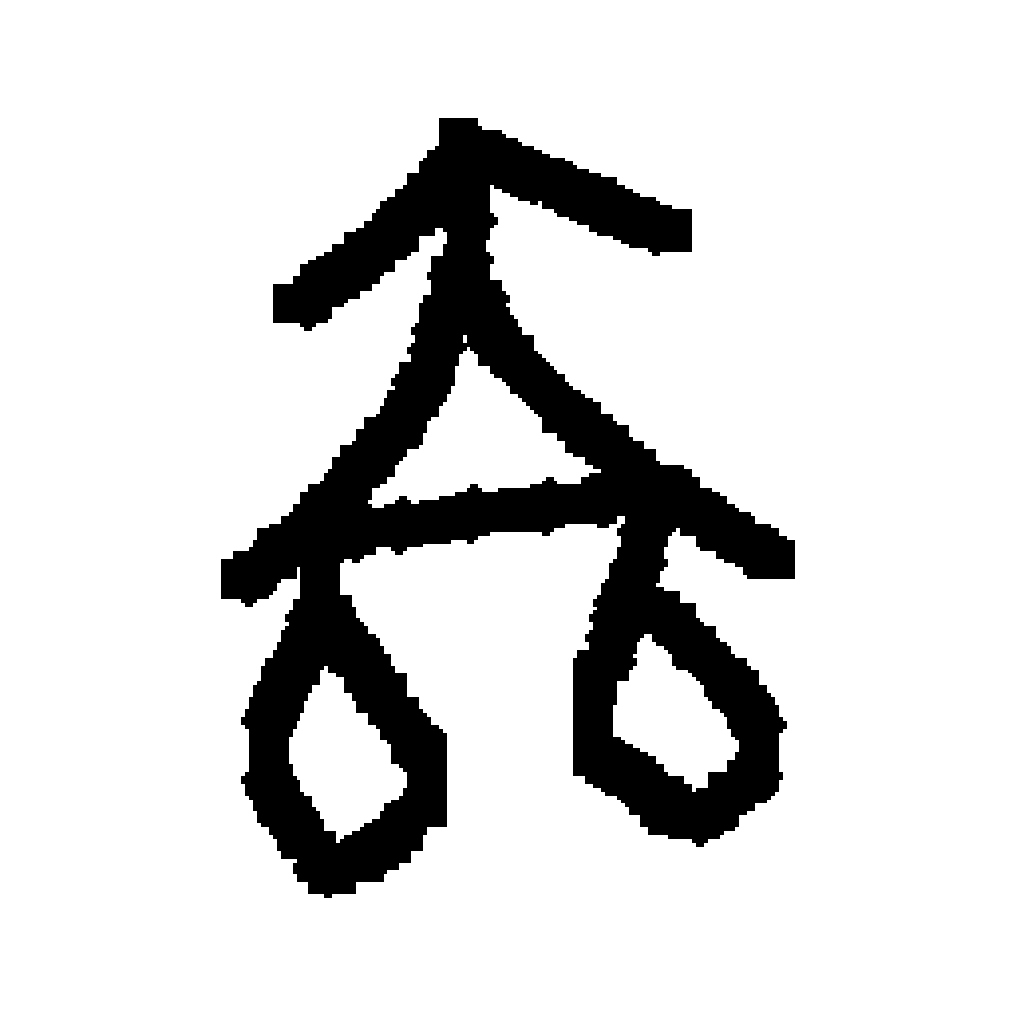} &
\imgfade[width=0.95\linewidth]{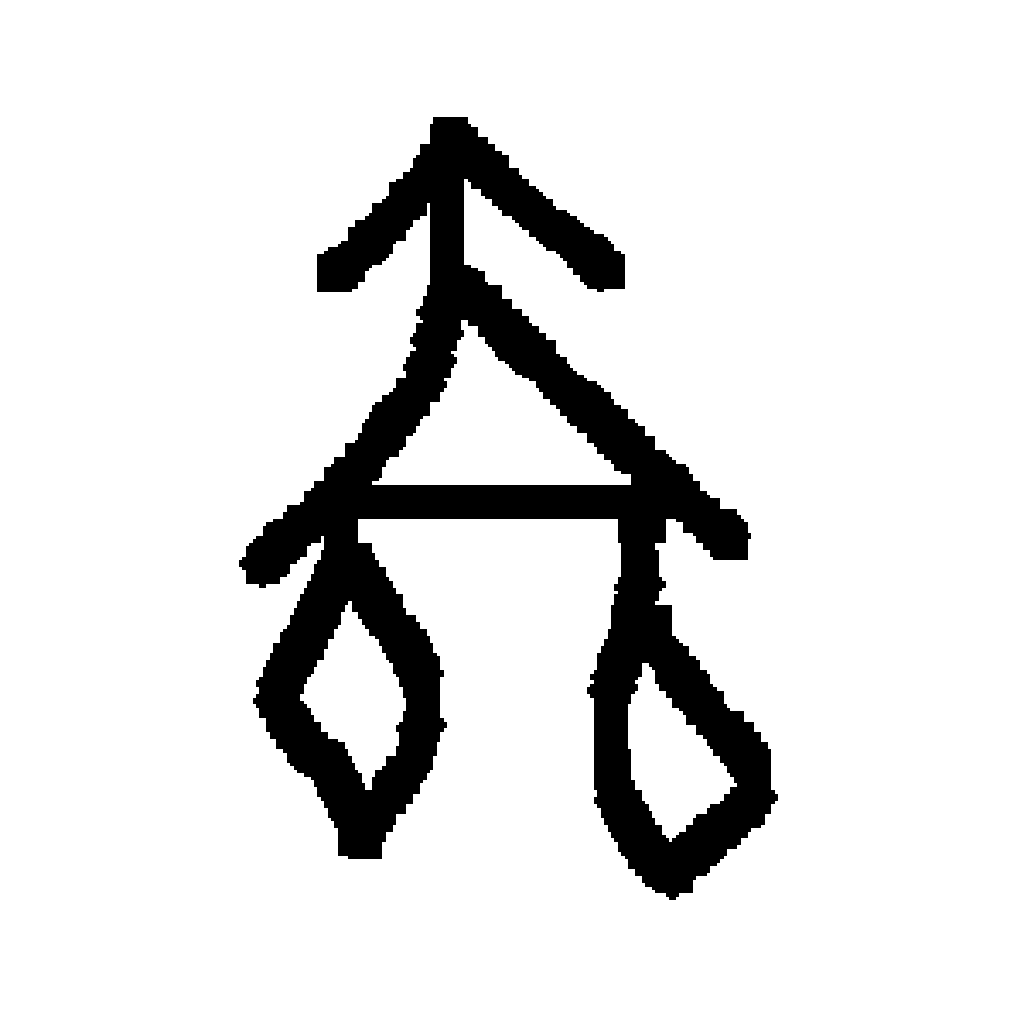} &
\imgfade[width=0.95\linewidth]{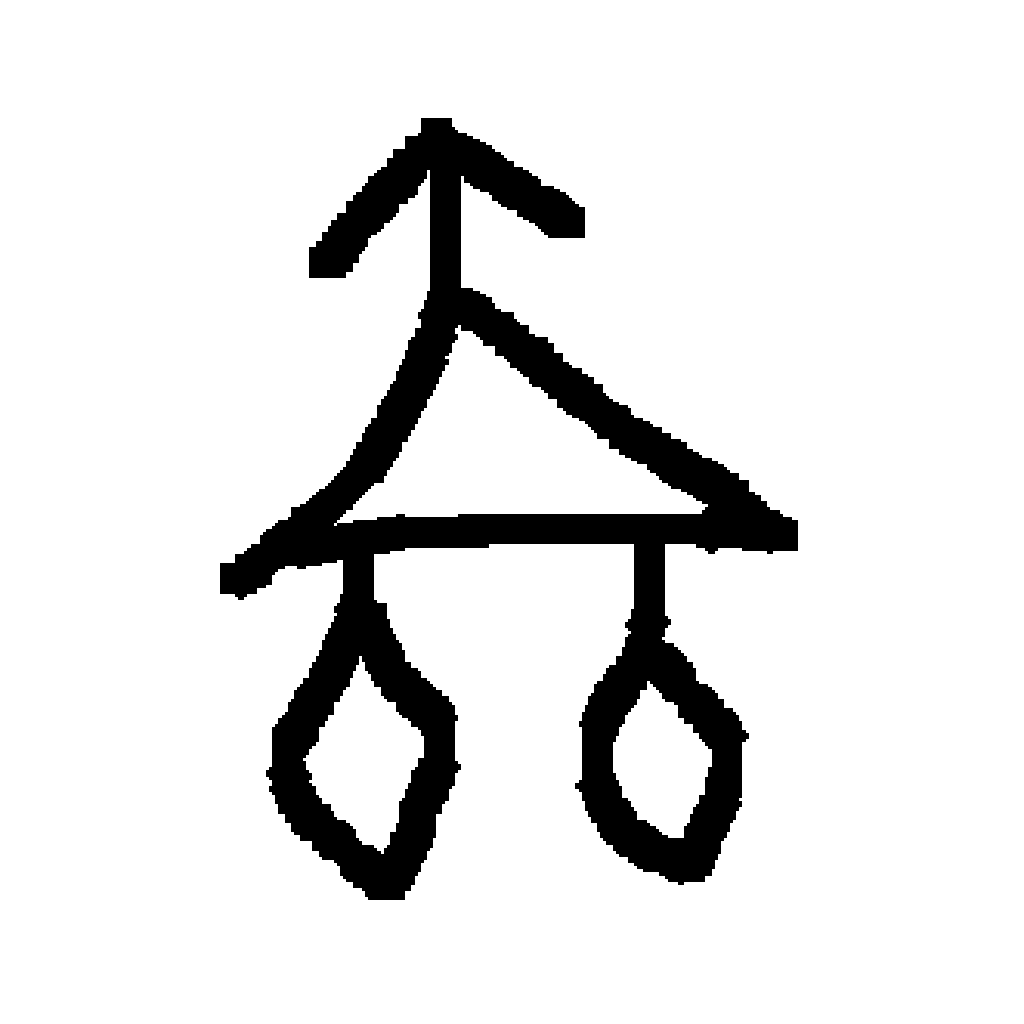} &
\imgfade[width=0.95\linewidth]{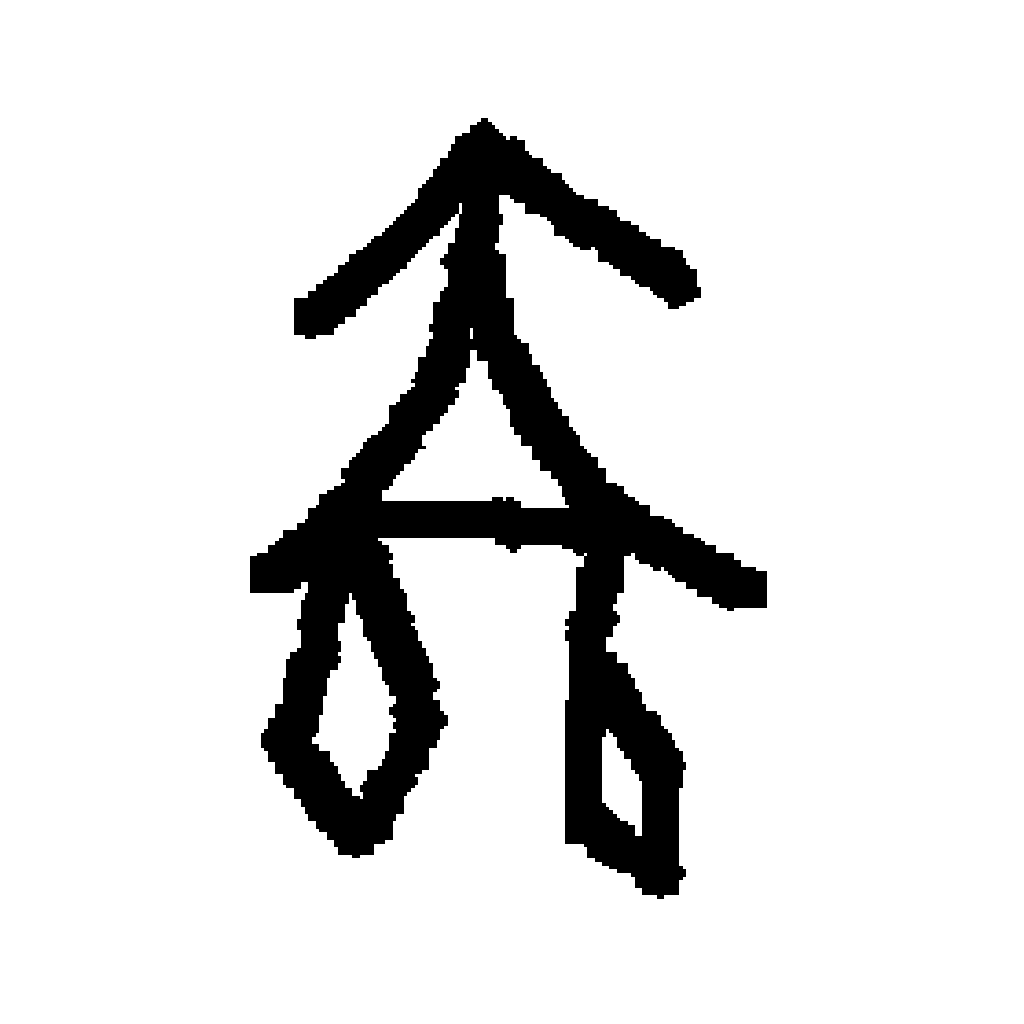} &
\imgfade[width=0.95\linewidth]{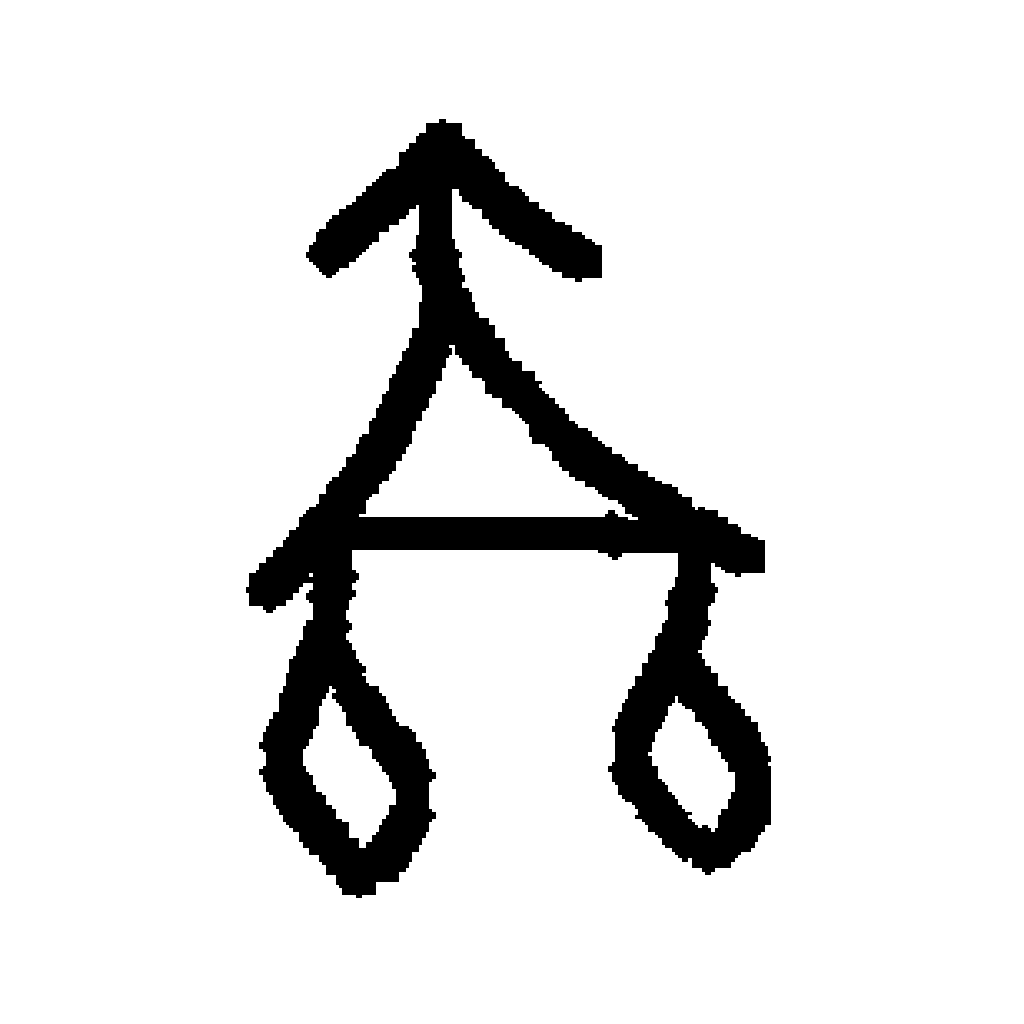} &
\imgwithbox[width=0.95\linewidth]{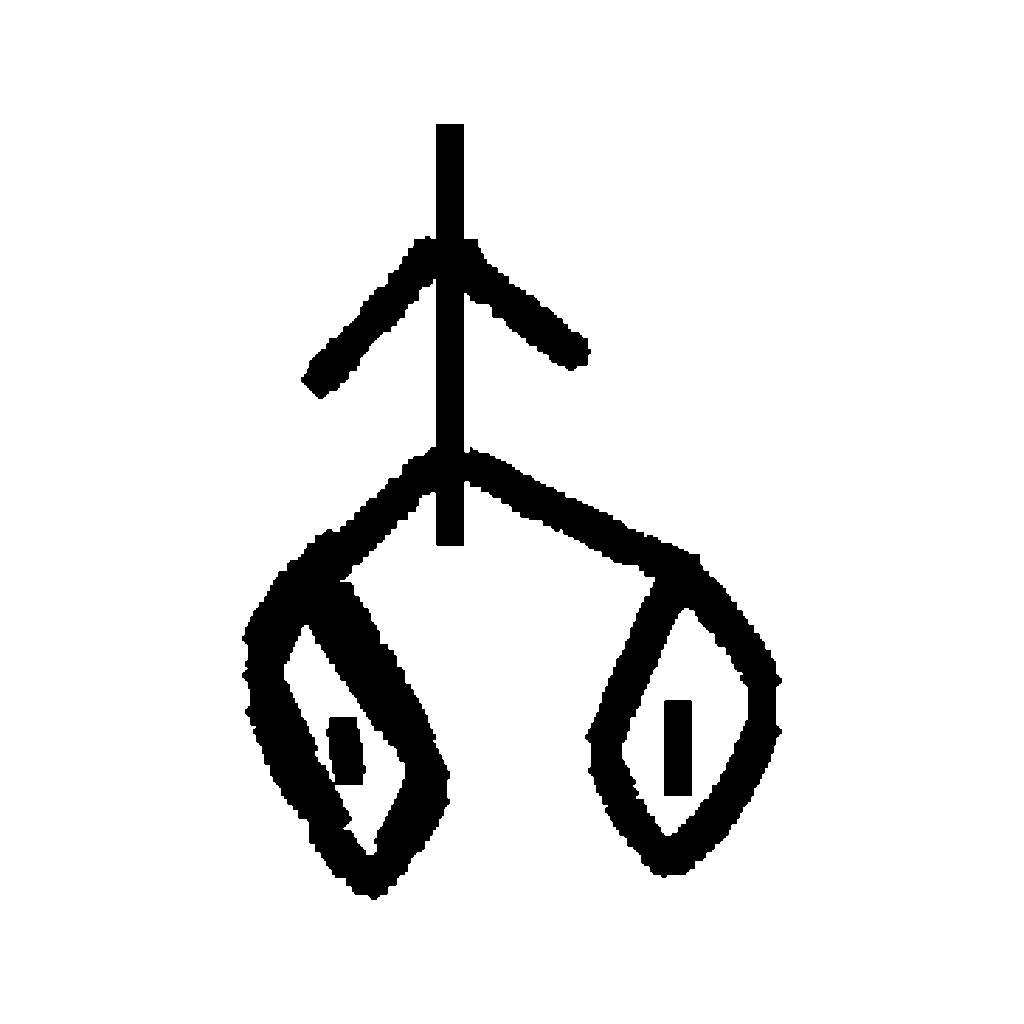} \\
2 &
\includegraphics[width=0.95\linewidth]{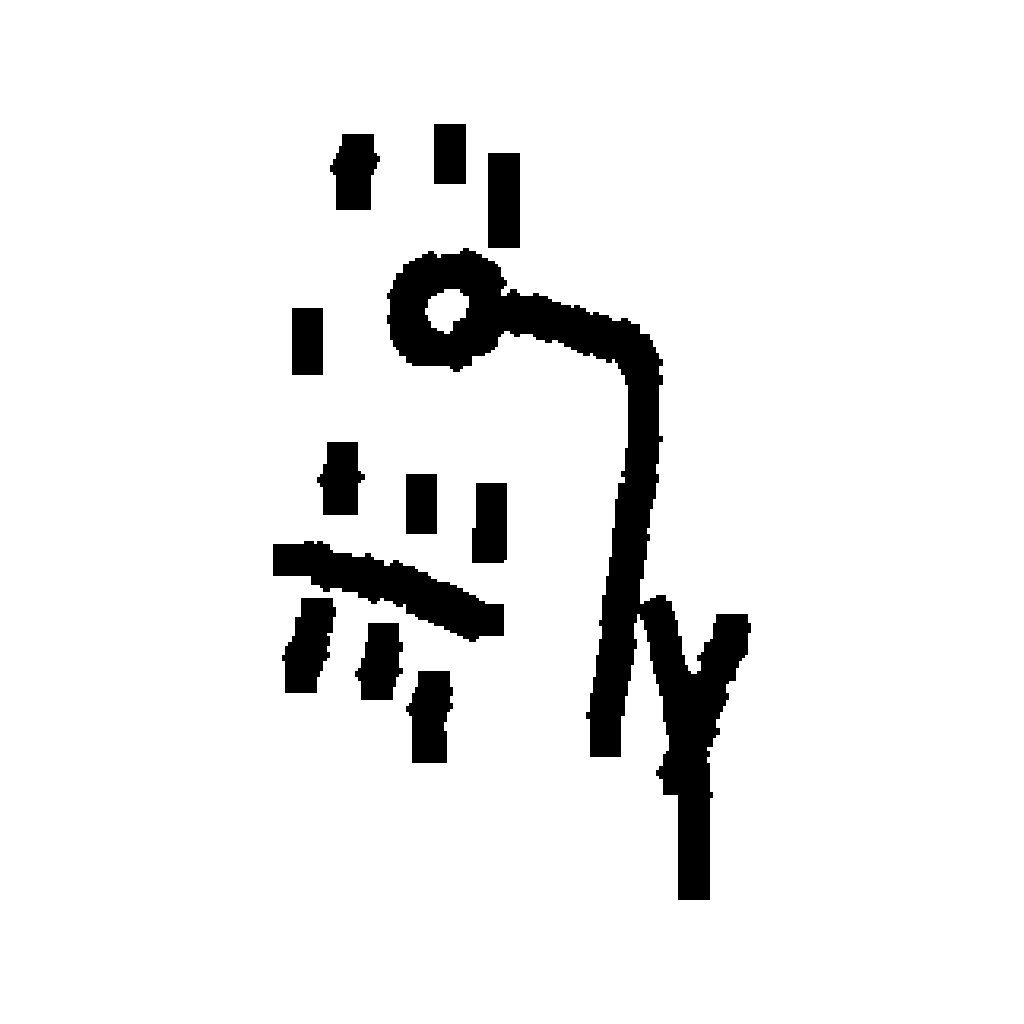} &
\imgfade[width=0.95\linewidth]{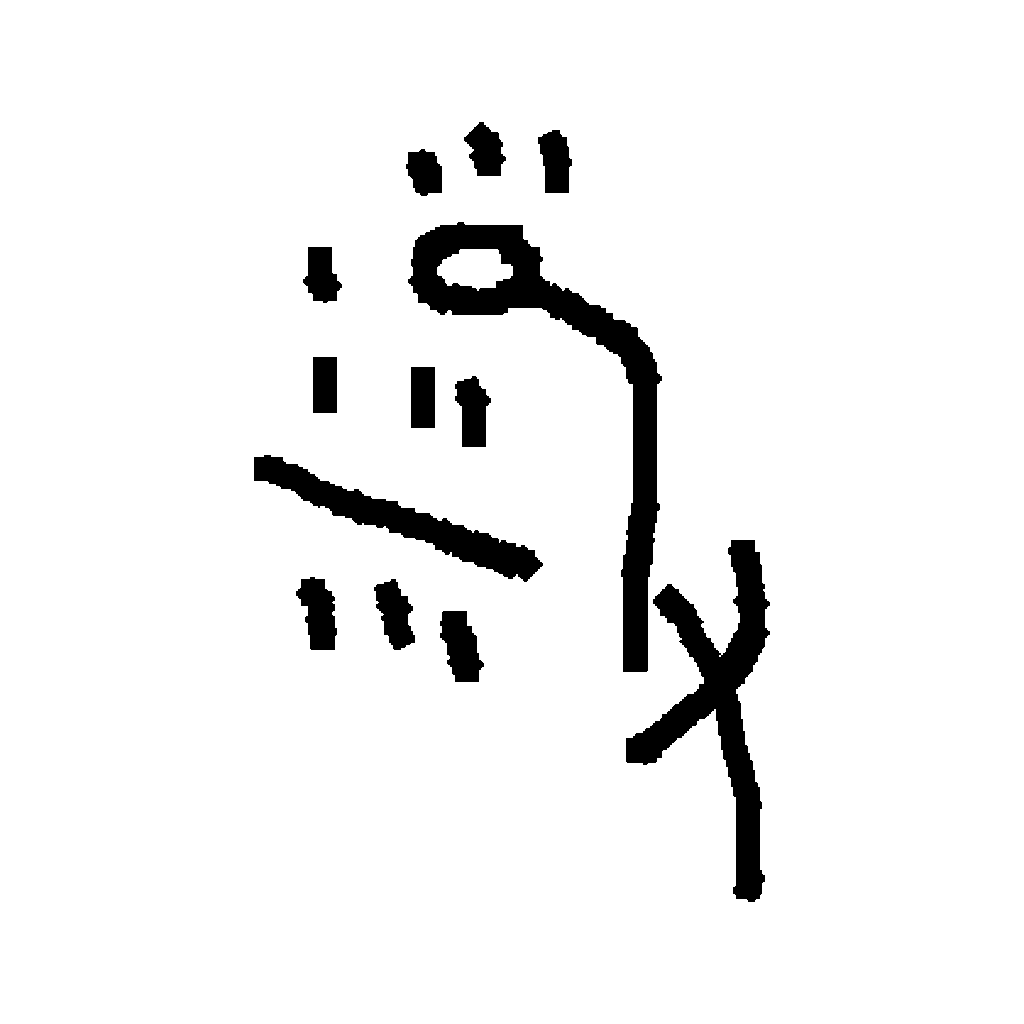} &
\imgfade[width=0.95\linewidth]{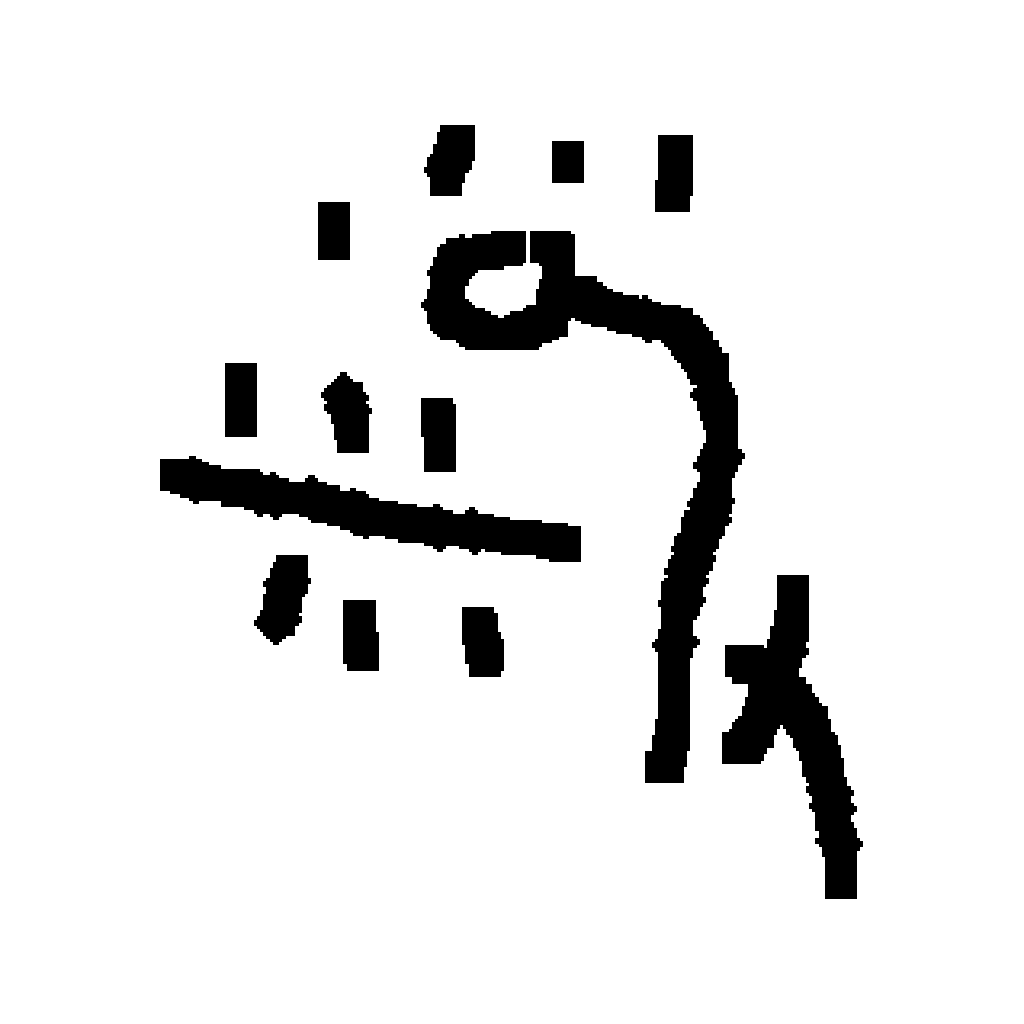} &
\imgfade[width=0.95\linewidth]{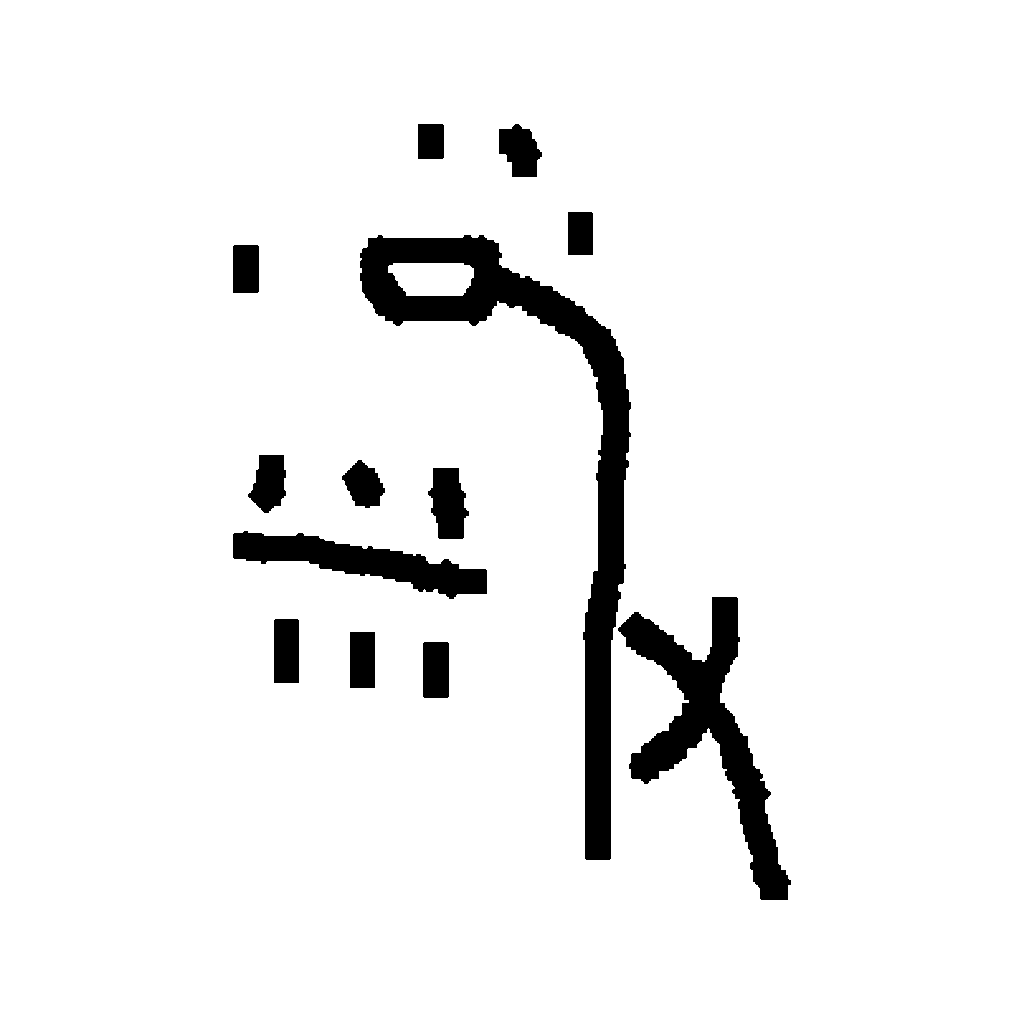} &
\imgfade[width=0.95\linewidth]{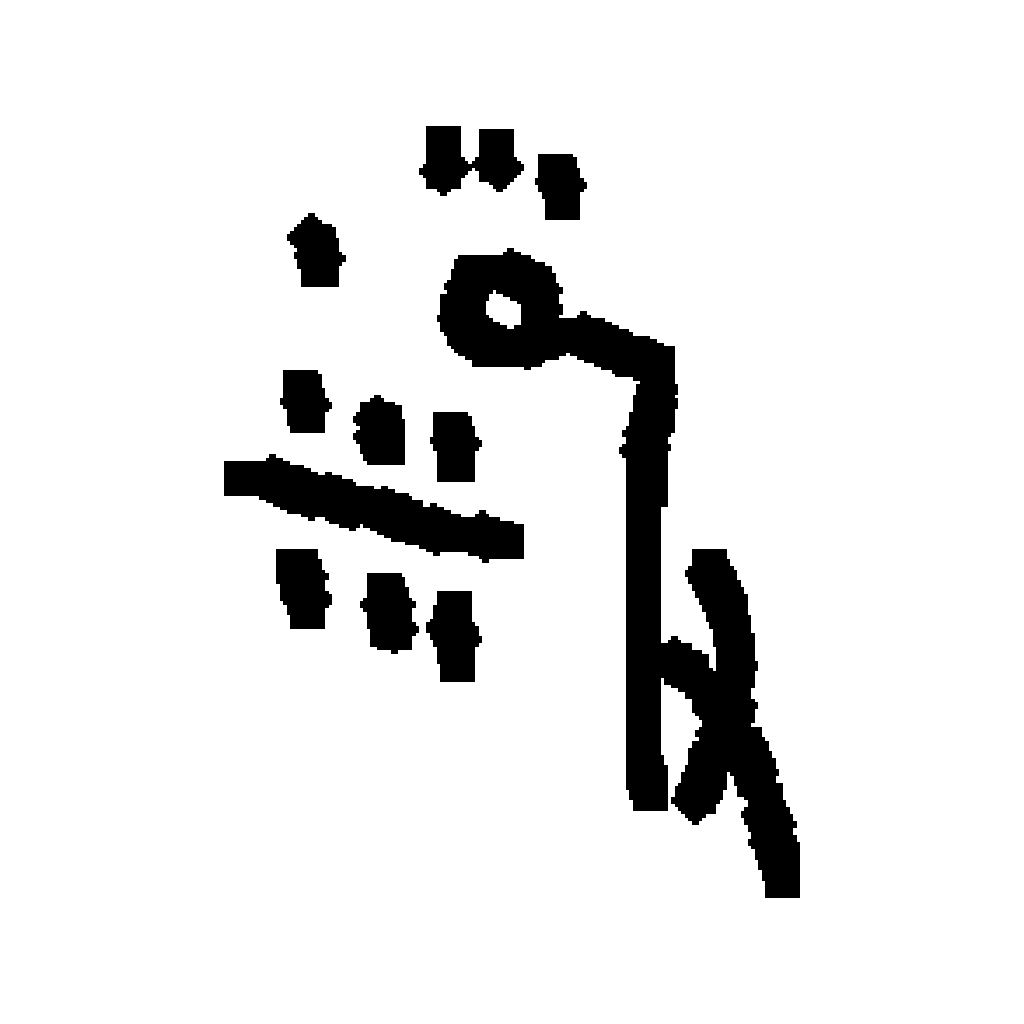} &
\imgfade[width=0.95\linewidth]{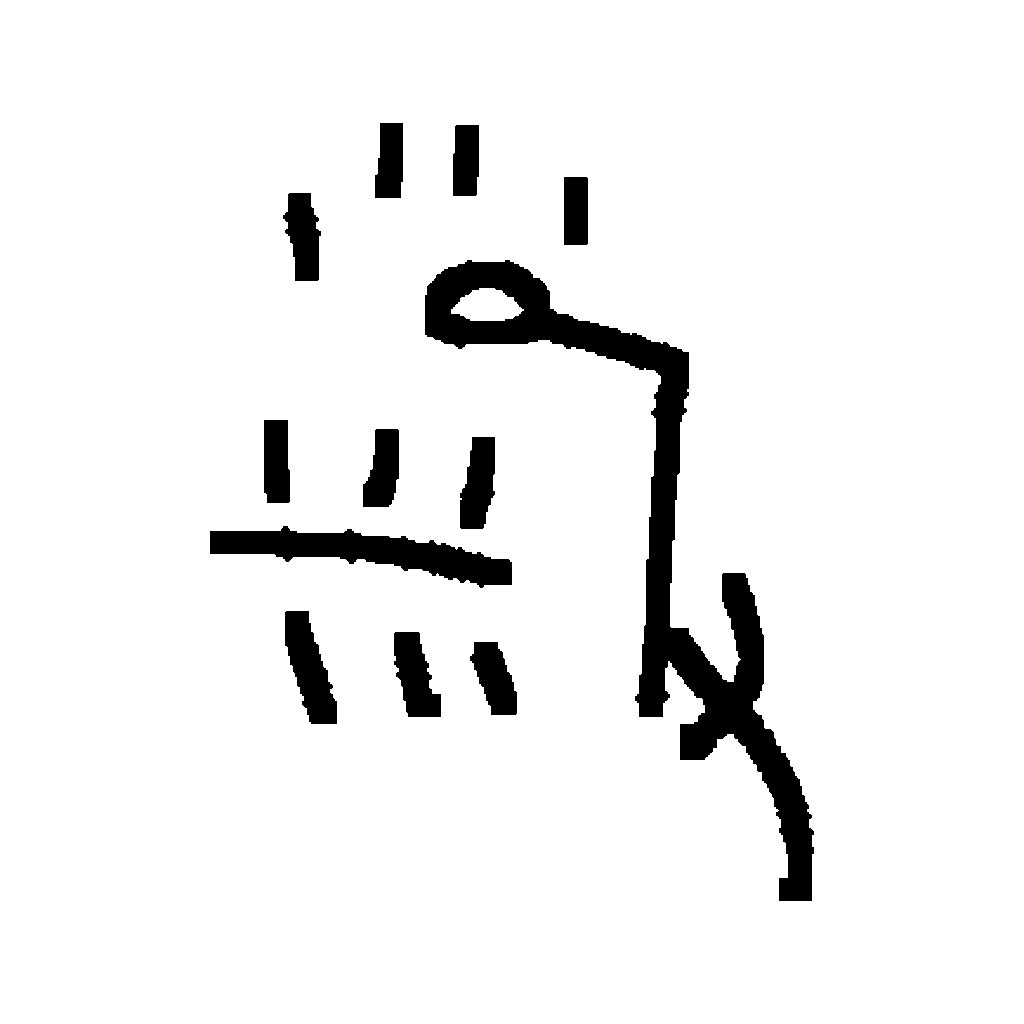} &
\imgfade[width=0.95\linewidth]{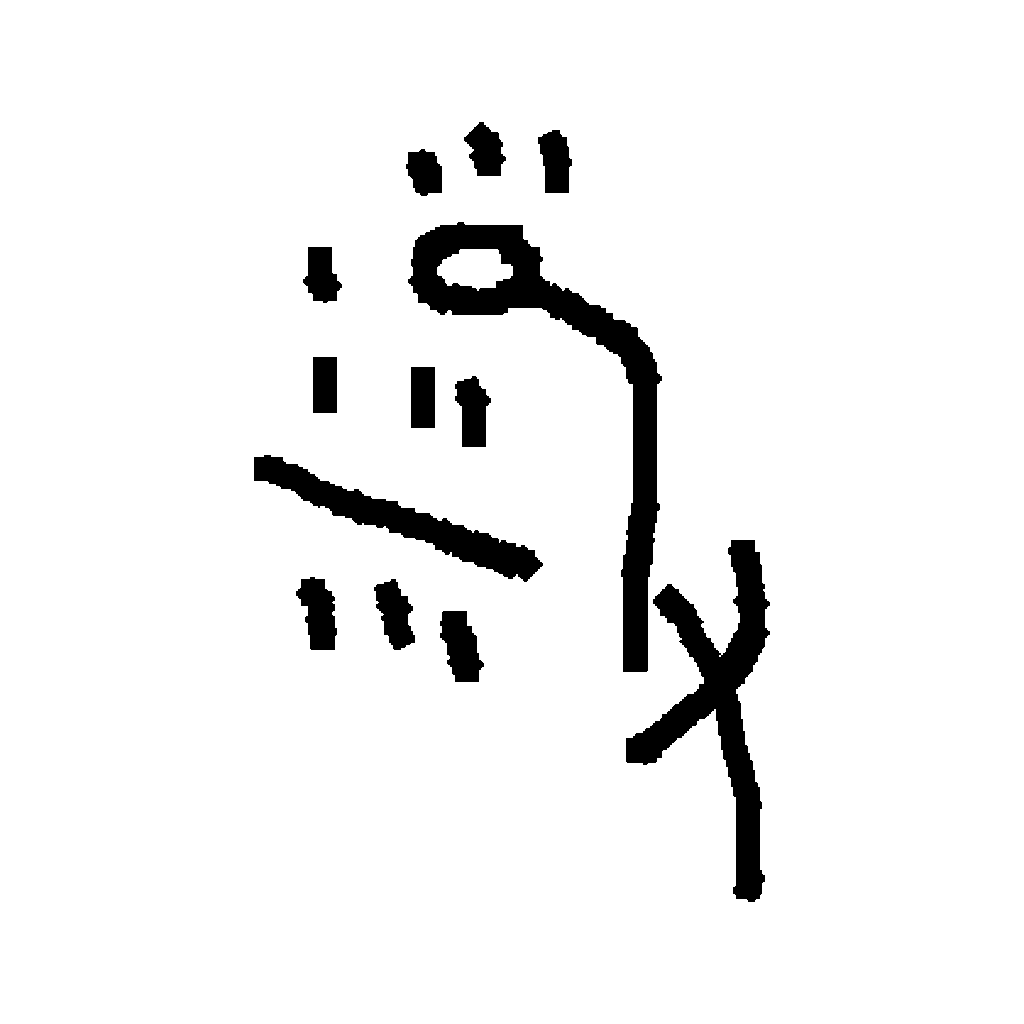} &
\imgfade[width=0.95\linewidth]{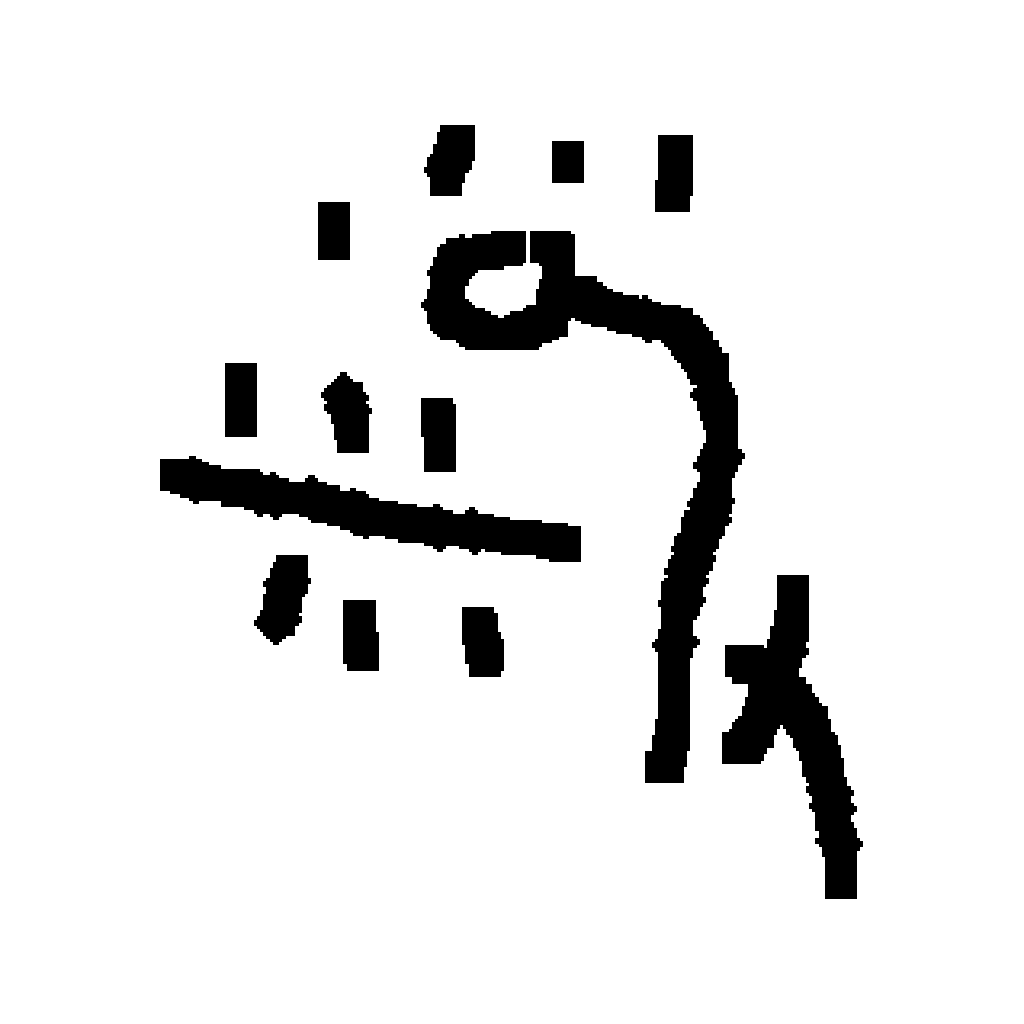} &
\imgfade[width=0.95\linewidth]{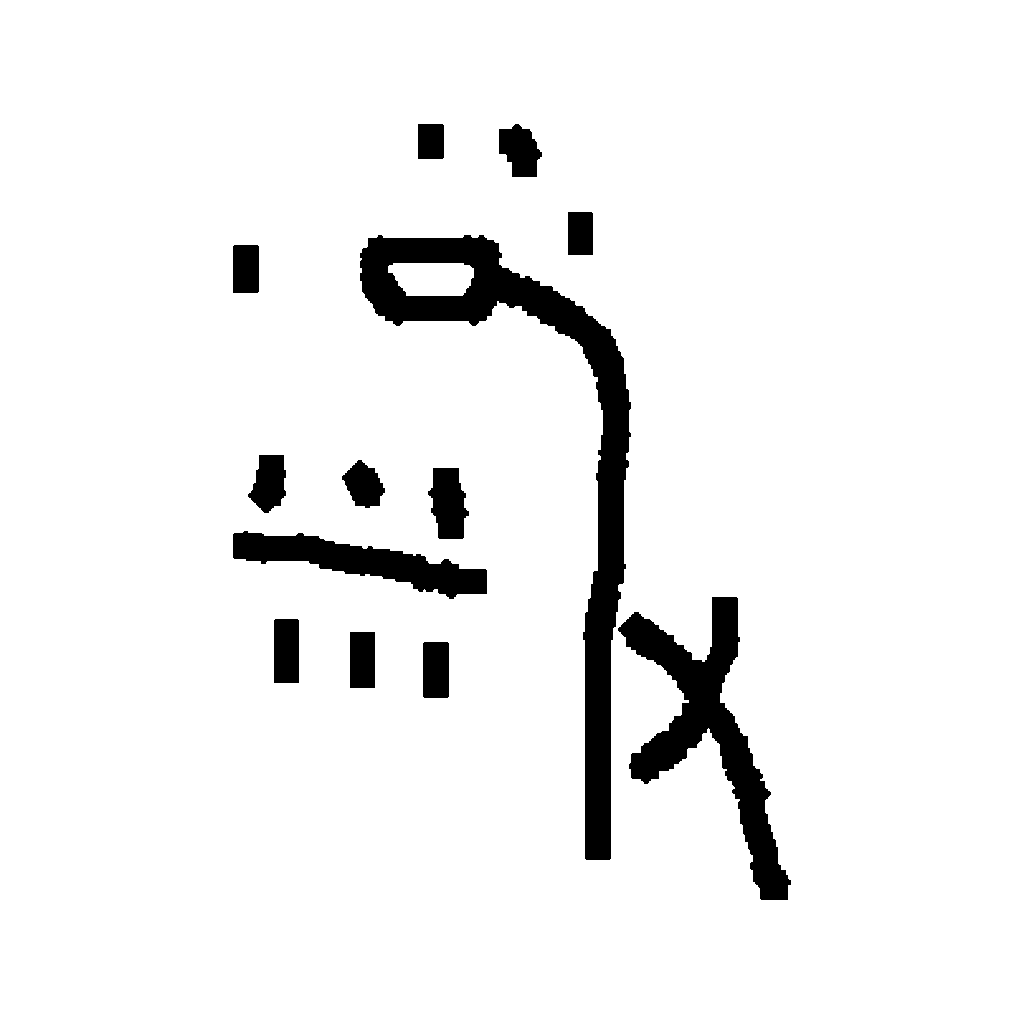} &
\imgwithbox[width=0.95\linewidth]{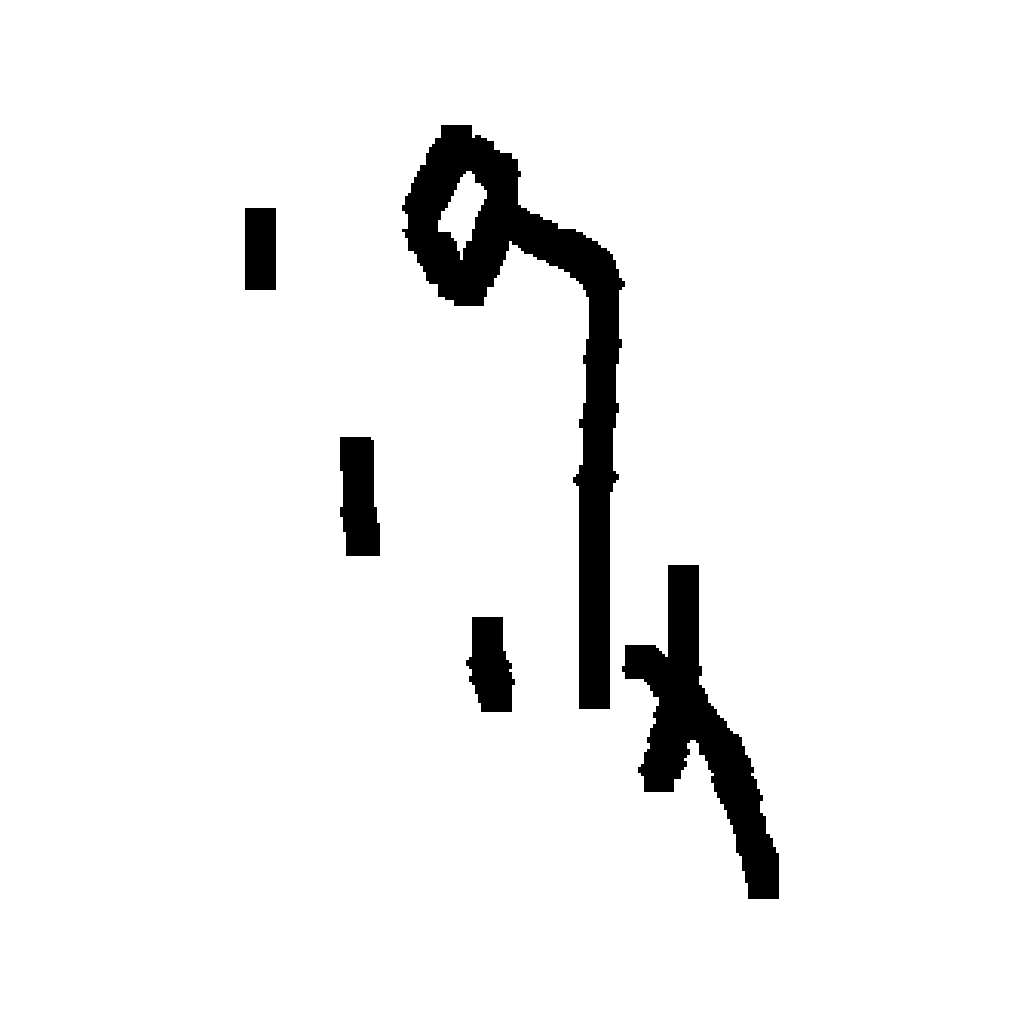} &
\imgfade[width=0.95\linewidth]{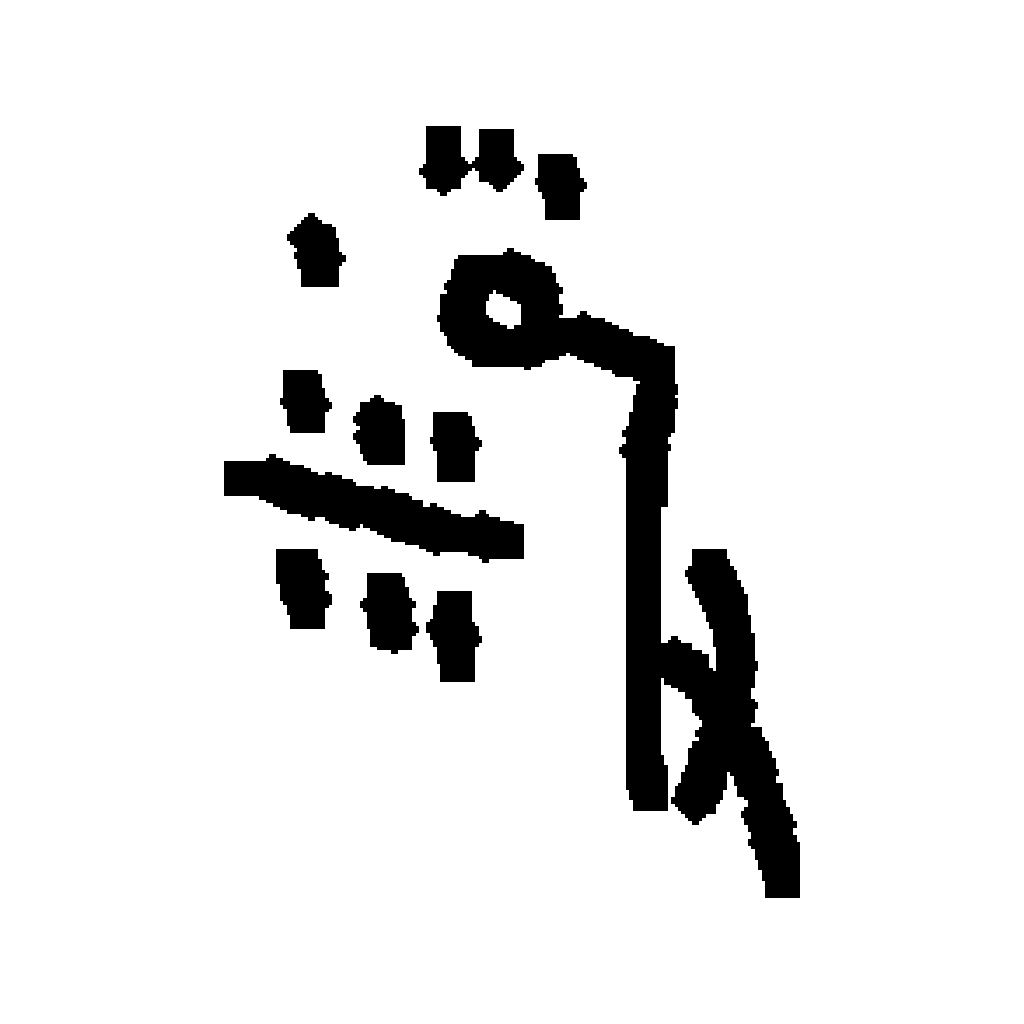} \\
3 &
\includegraphics[width=0.95\linewidth]{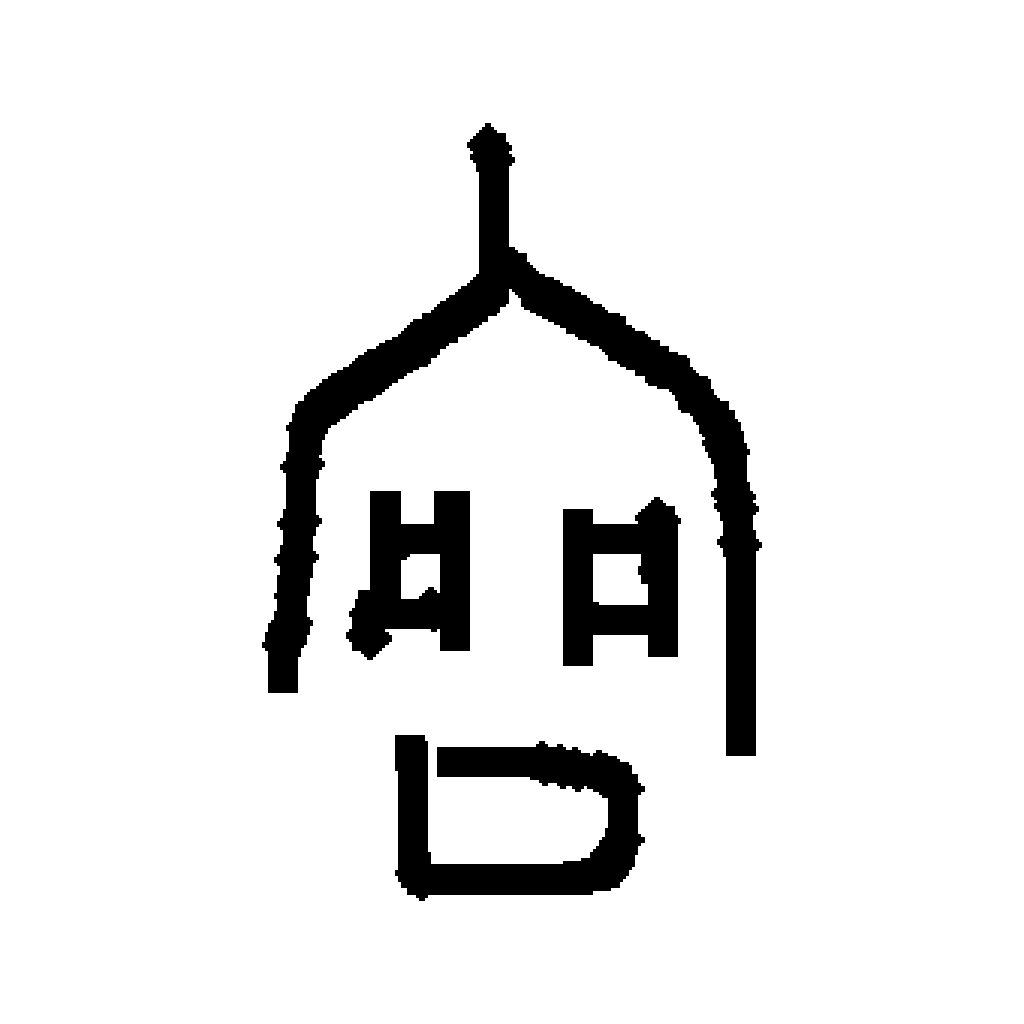} &
\imgfade[width=0.95\linewidth]{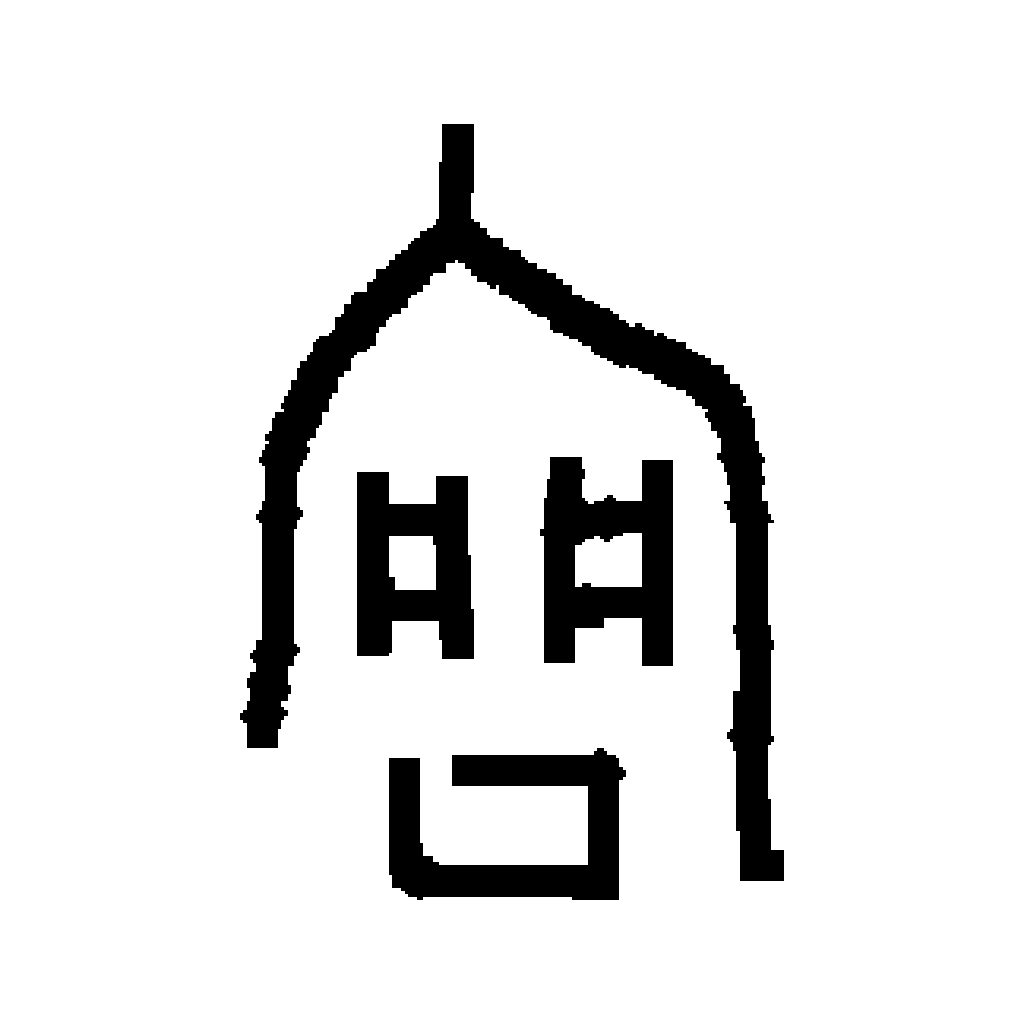} &
\imgfade[width=0.95\linewidth]{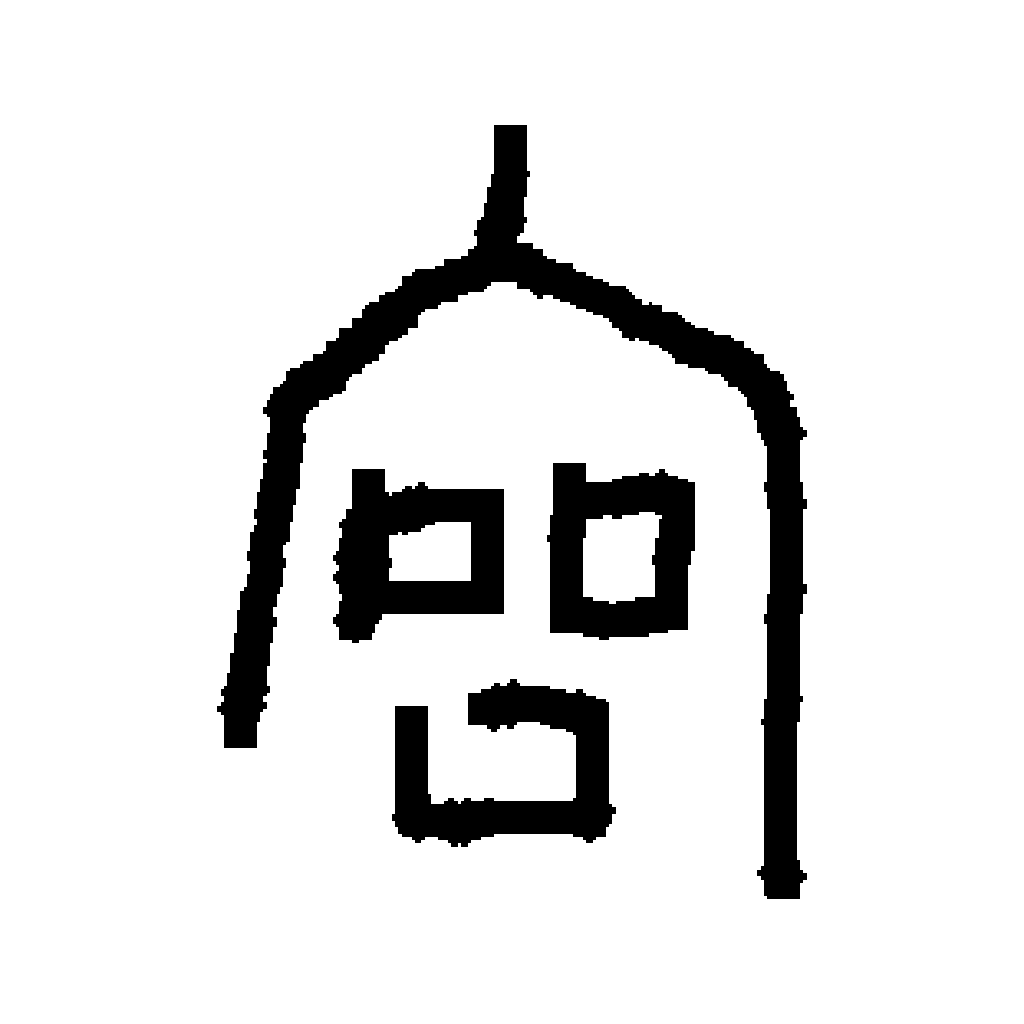} &
\imgfade[width=0.95\linewidth]{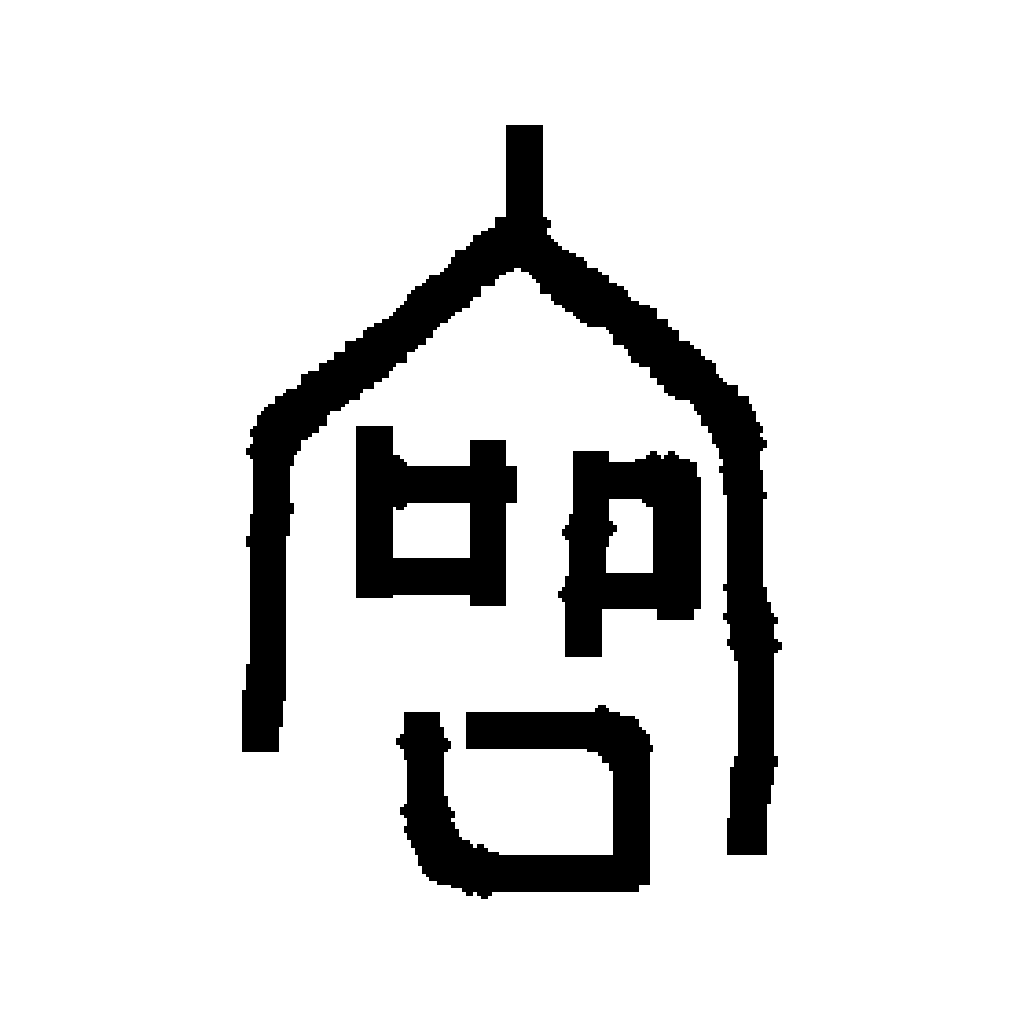} &
\imgfade[width=0.95\linewidth]{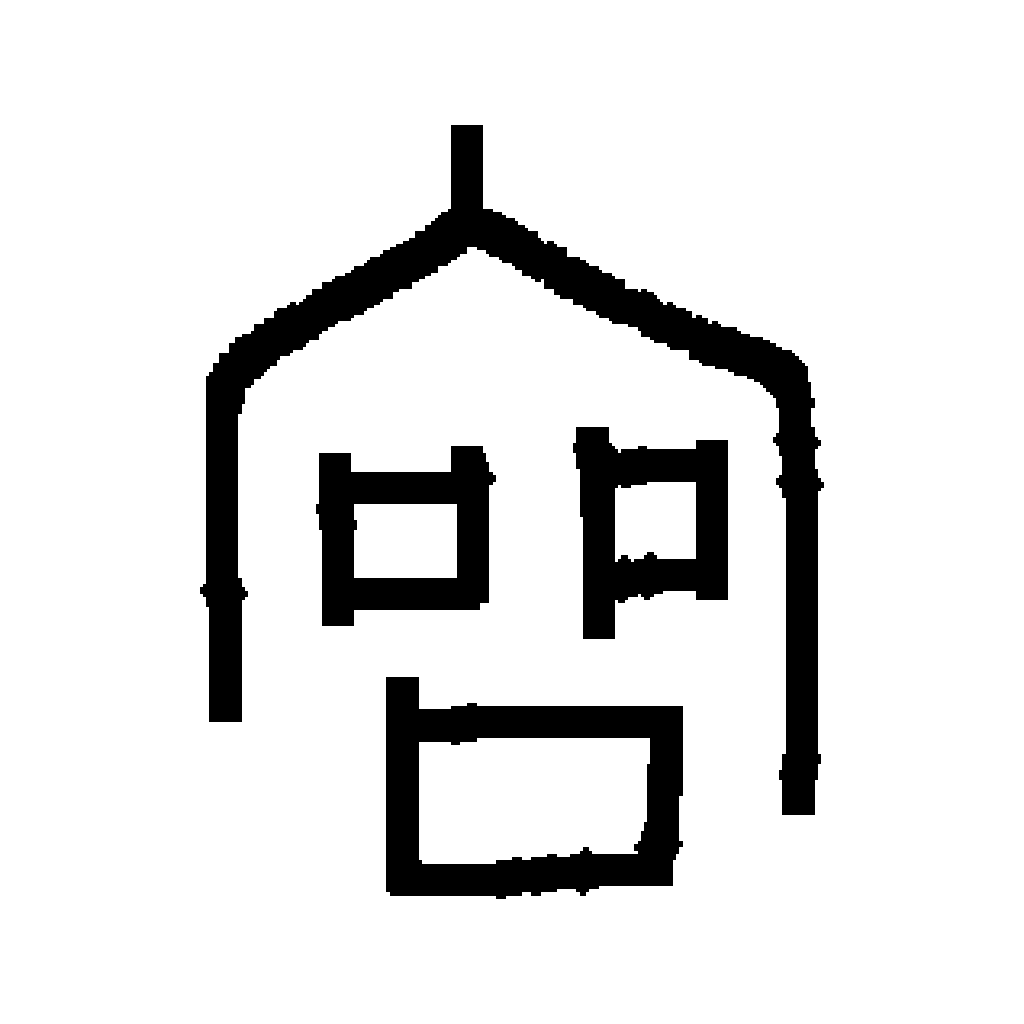} &
\imgfade[width=0.95\linewidth]{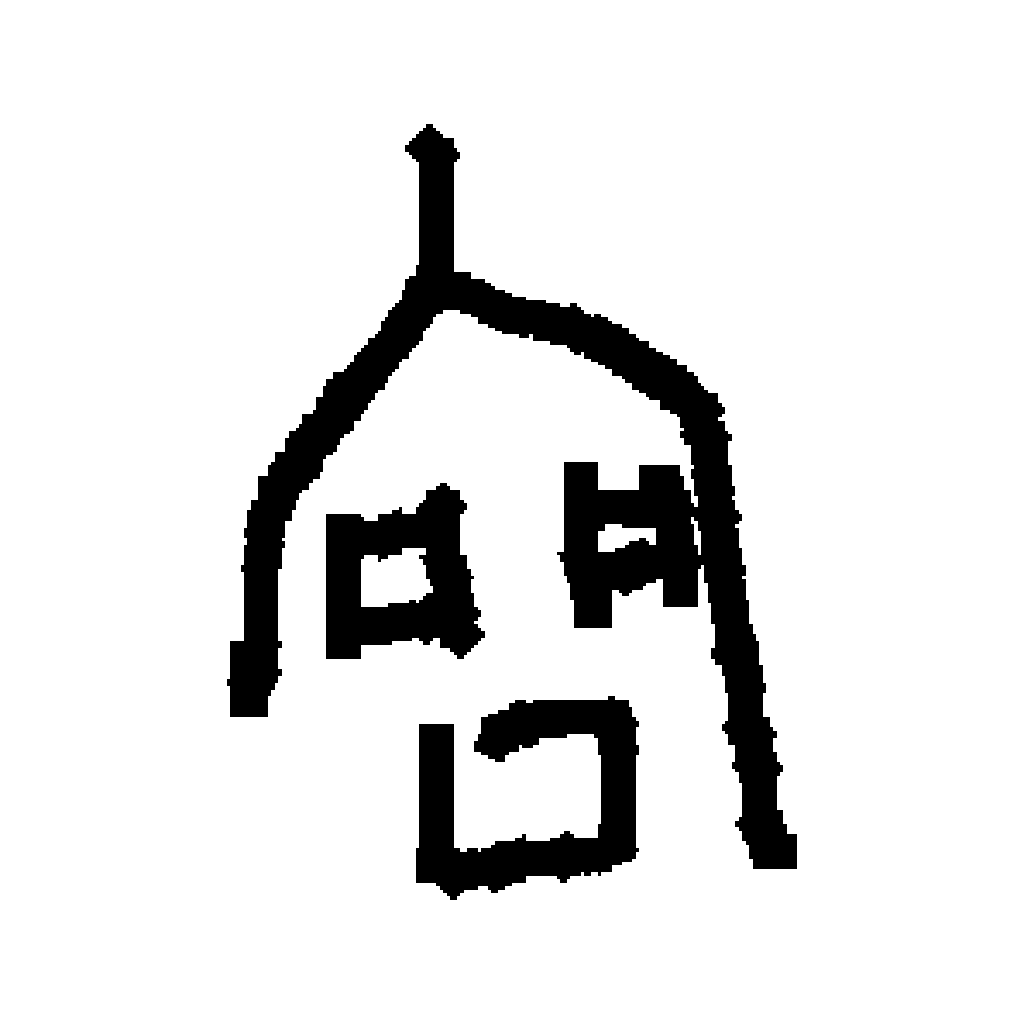} &
\imgfade[width=0.95\linewidth]{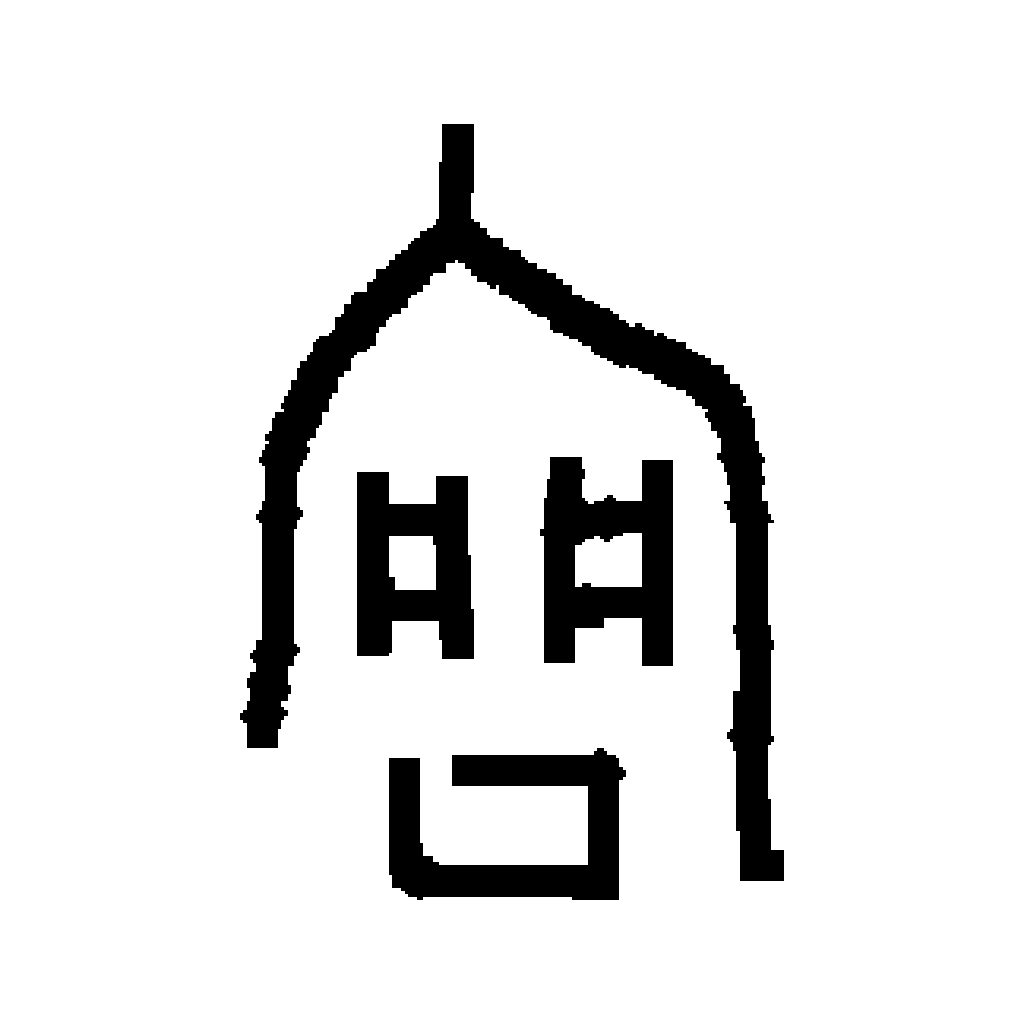} &
\imgfade[width=0.95\linewidth]{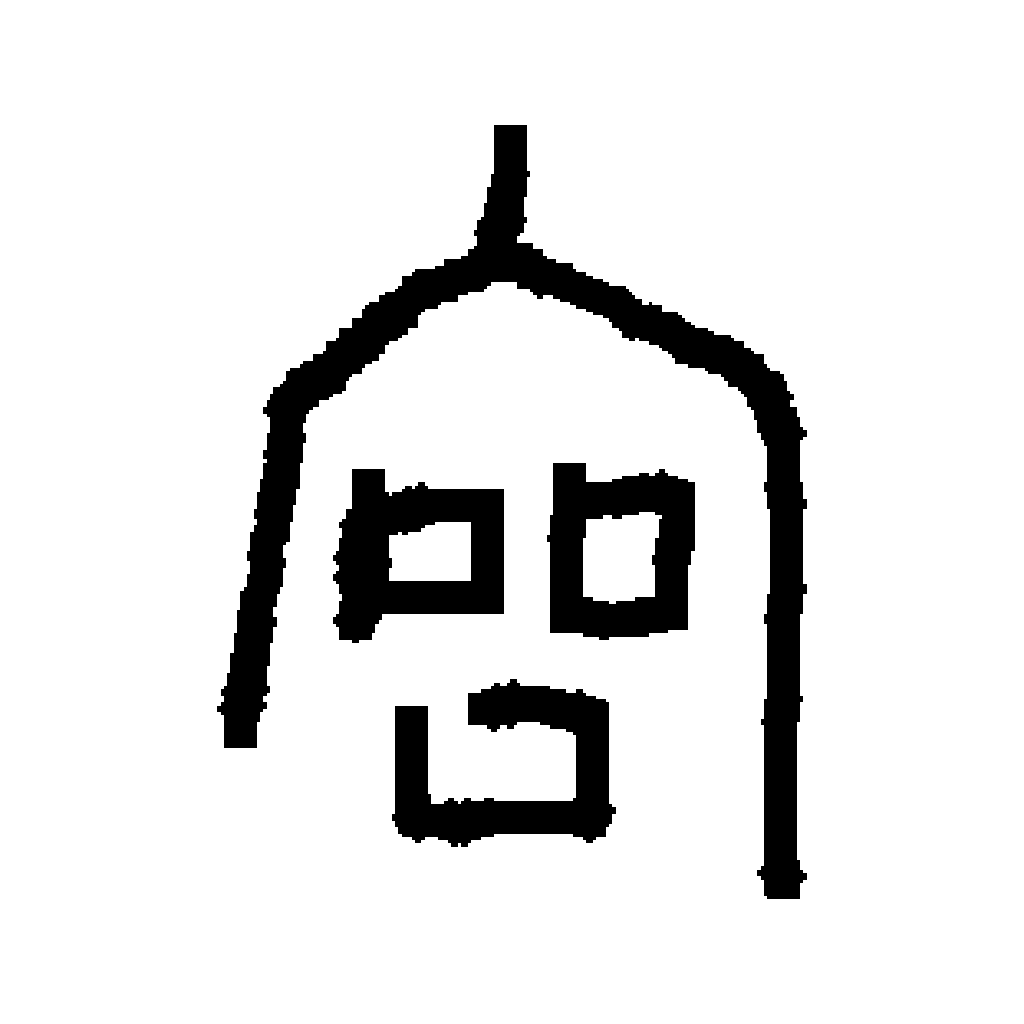} &
\imgwithbox[width=0.95\linewidth]{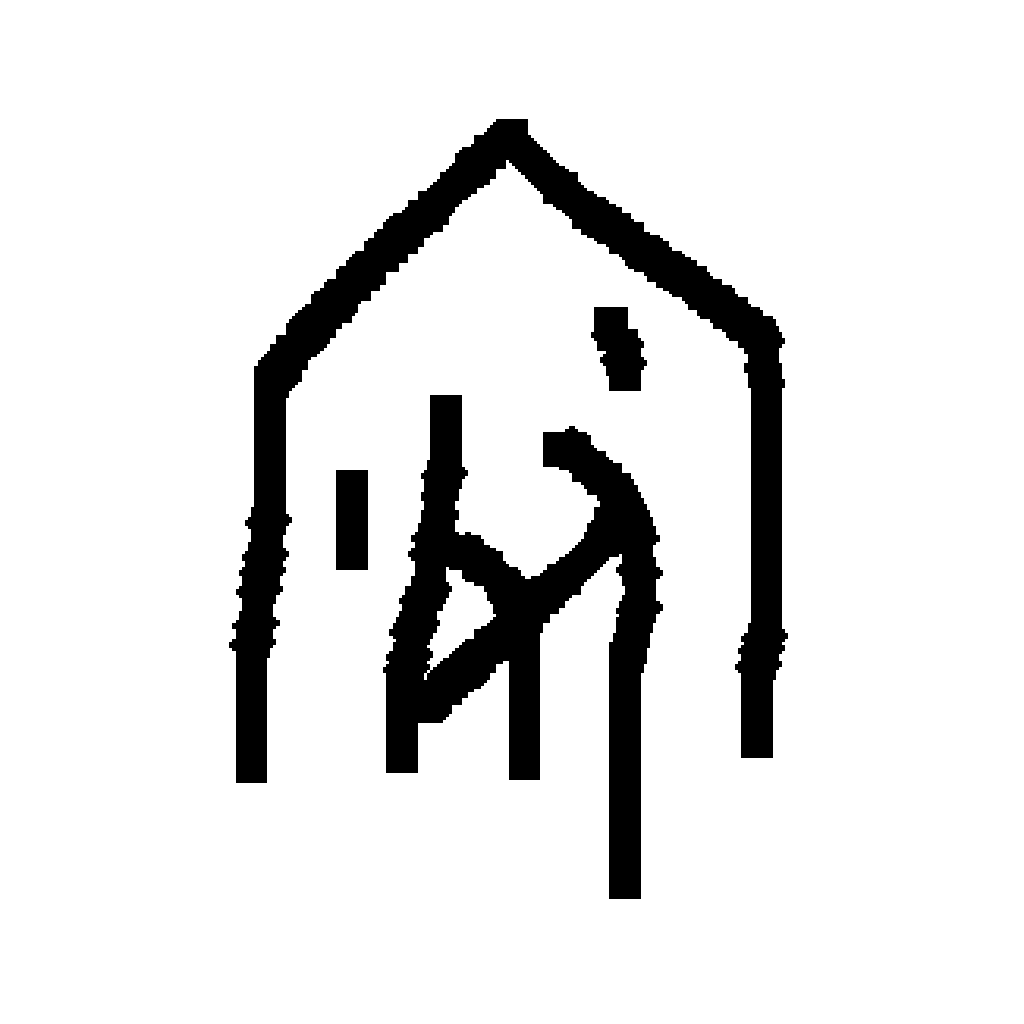} &
\imgwithbox[width=0.95\linewidth]{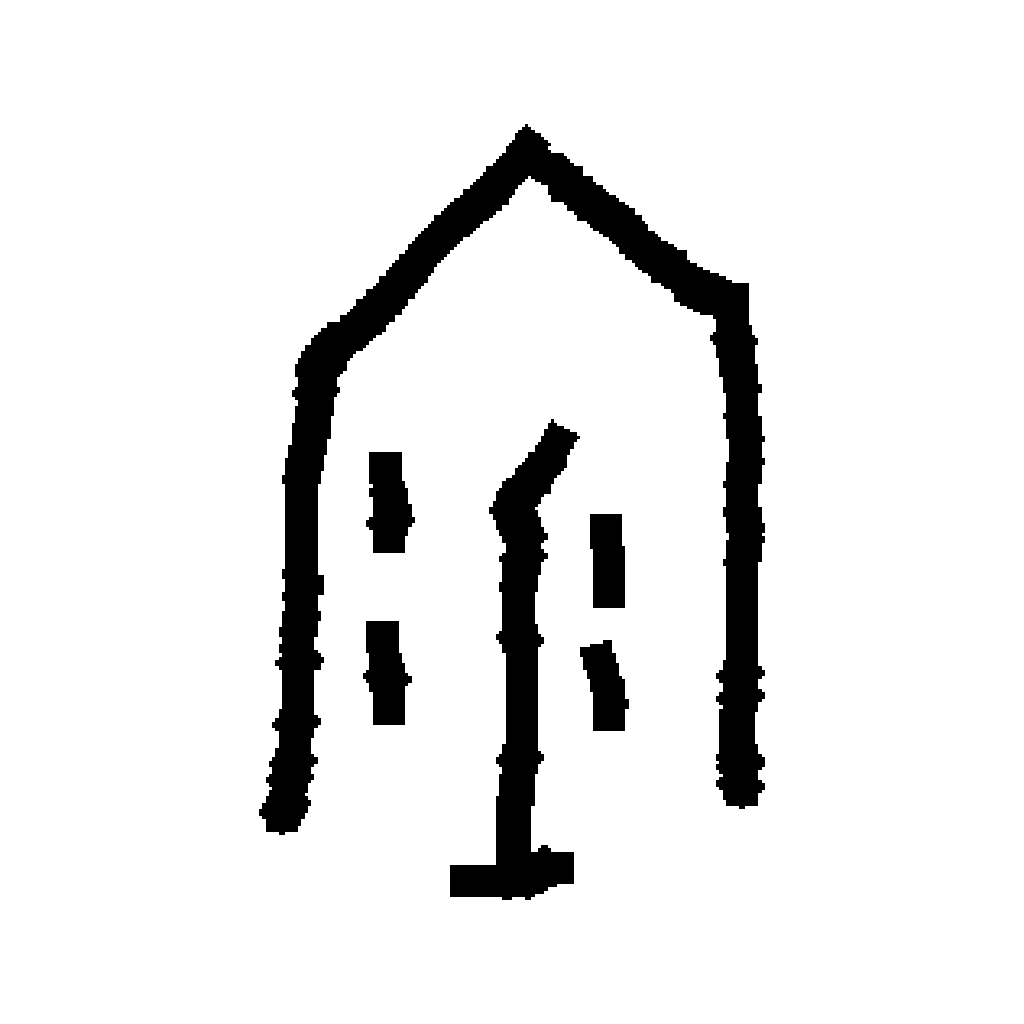} &
\imgwithbox[width=0.95\linewidth]{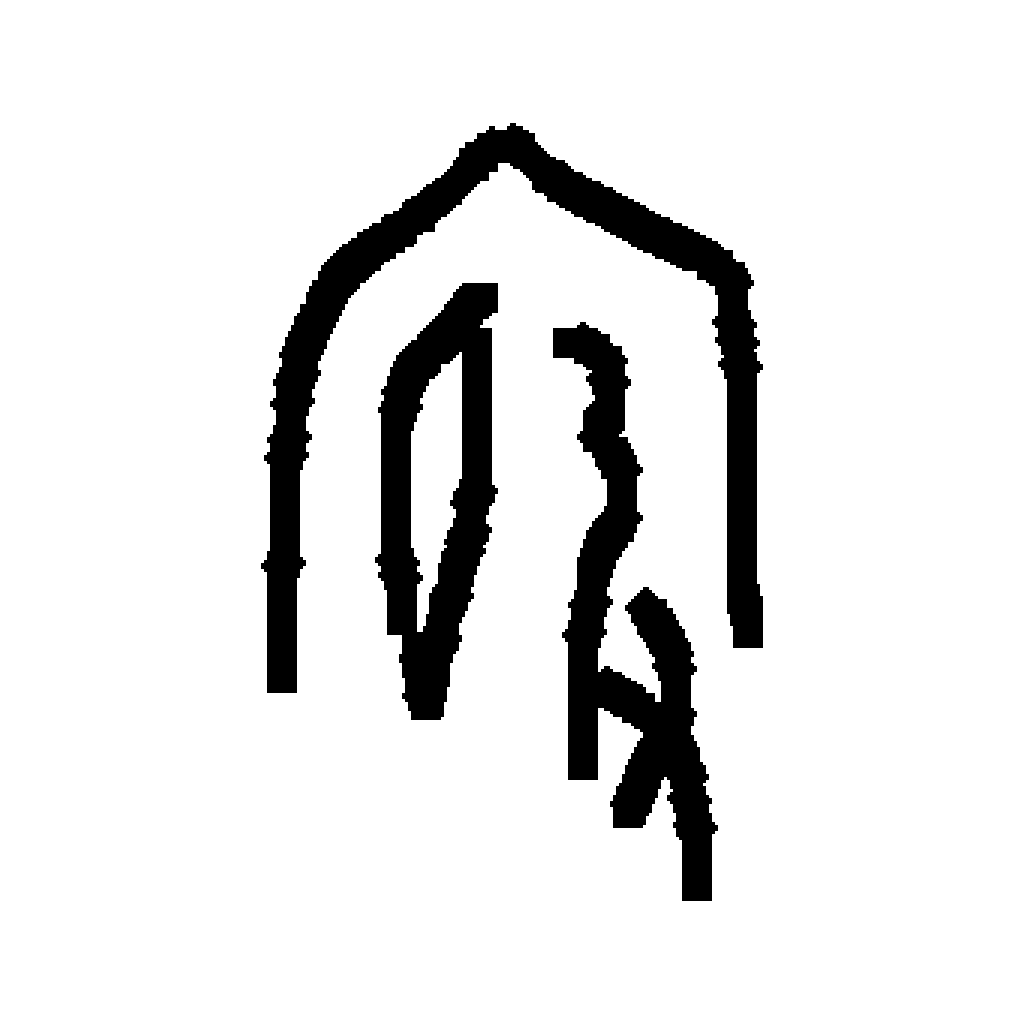} \\
\midrule\noalign{\vskip -3pt}\rowcolor{gray!20}\multicolumn{12}{c}{\textbf{Dongba Pictographs}} \\\noalign{\vskip -2pt}\midrule
4 &
\includegraphics[width=0.95\linewidth]{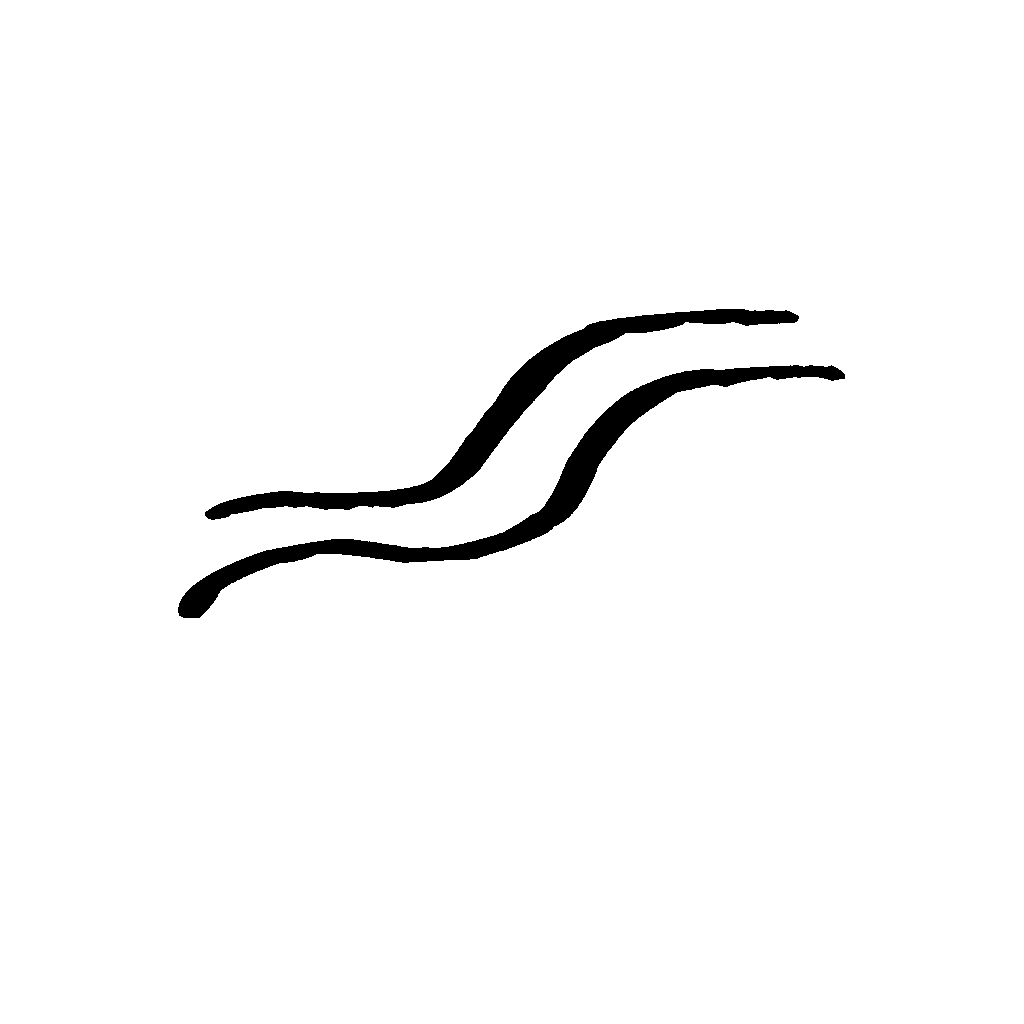} &
\imgwithbox[width=0.95\linewidth]{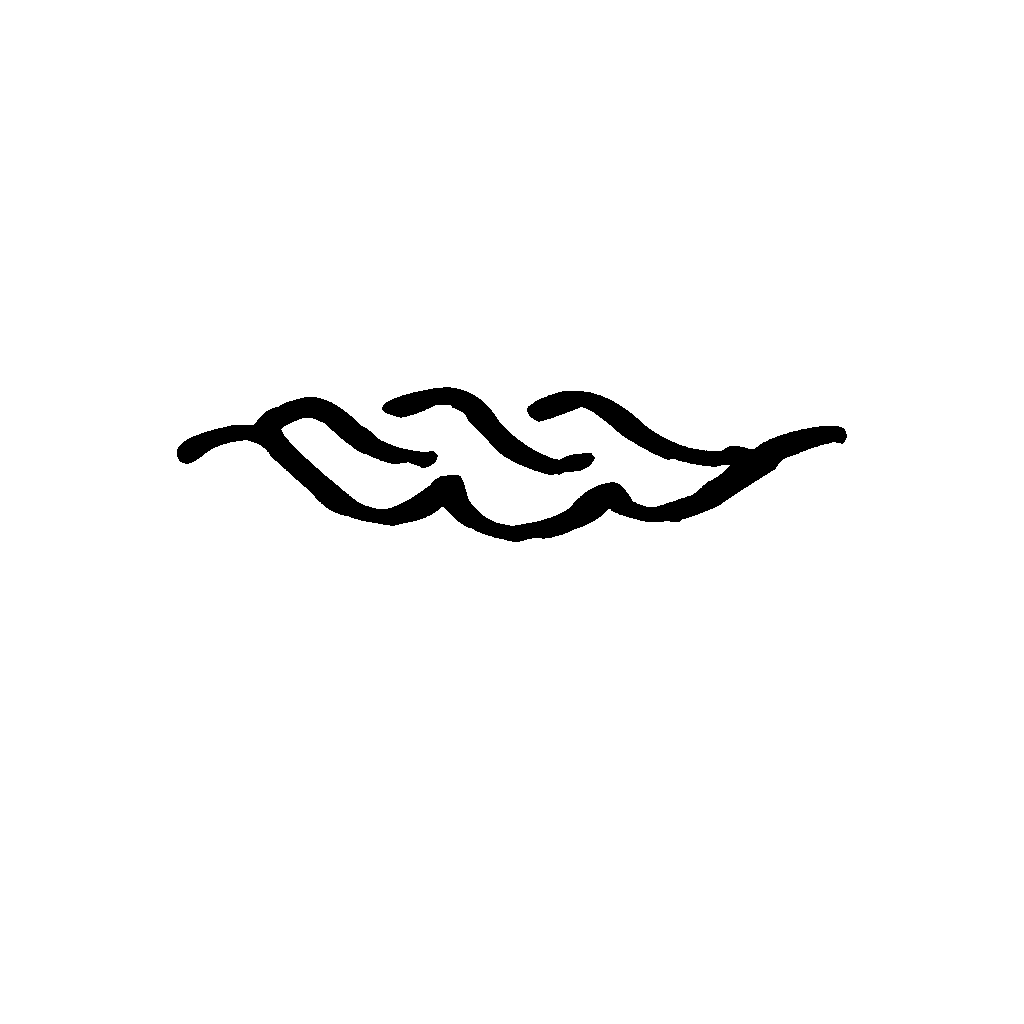} &
\includegraphics[width=0.95\linewidth]{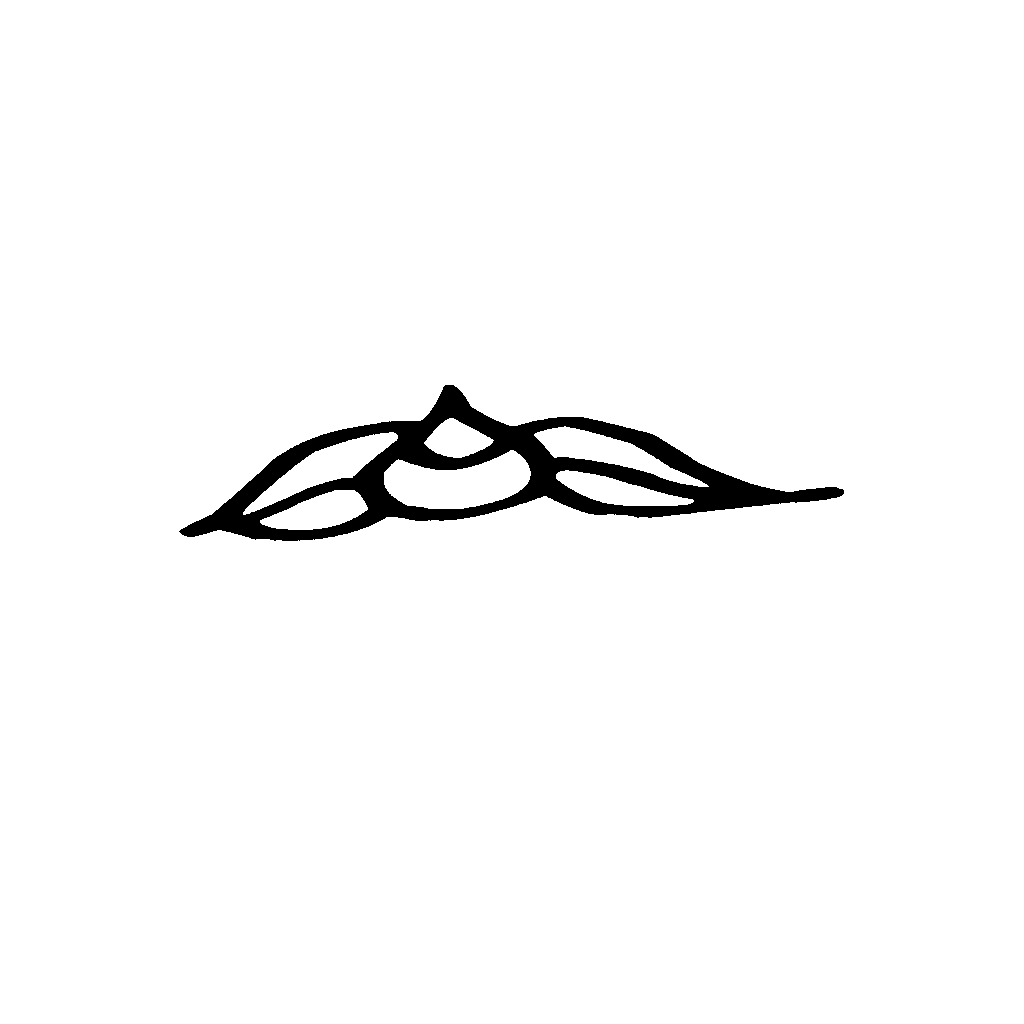} &
\includegraphics[width=0.95\linewidth]{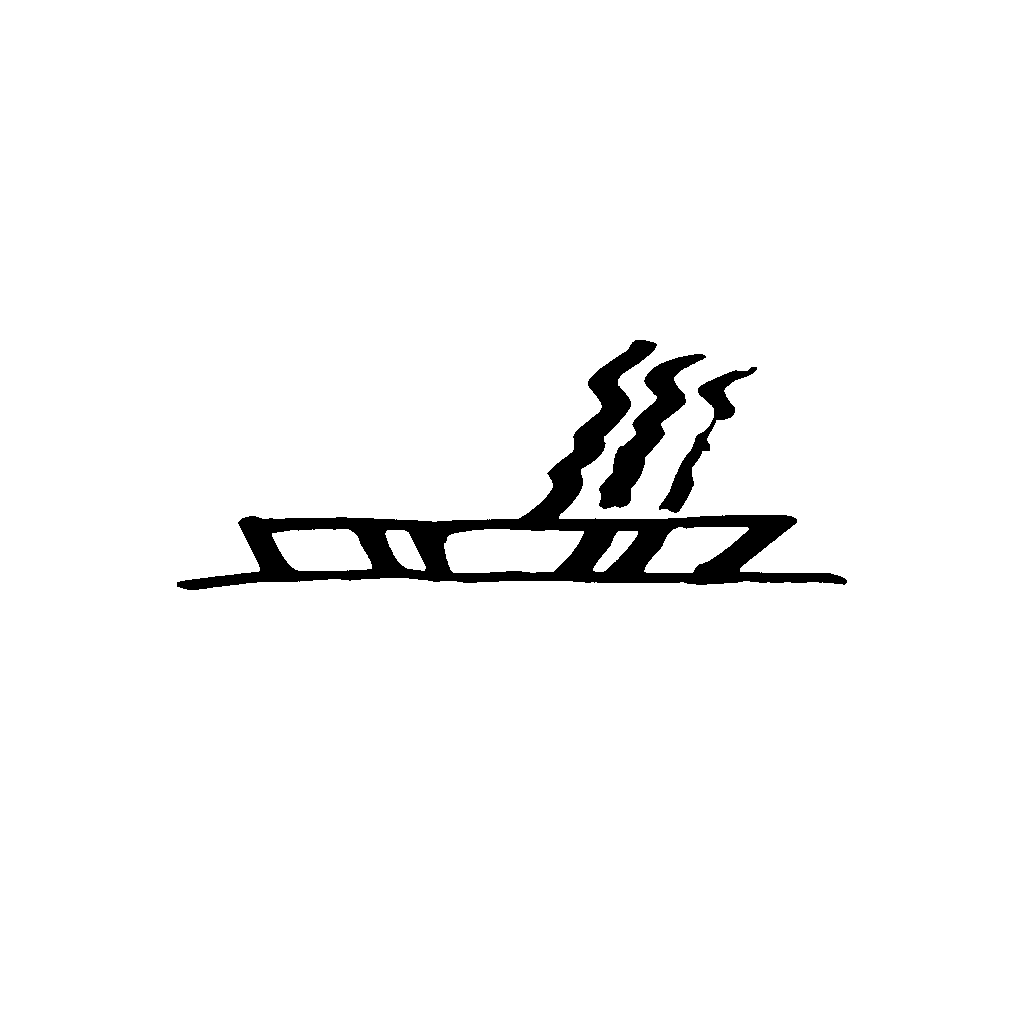} &
\includegraphics[width=0.95\linewidth]{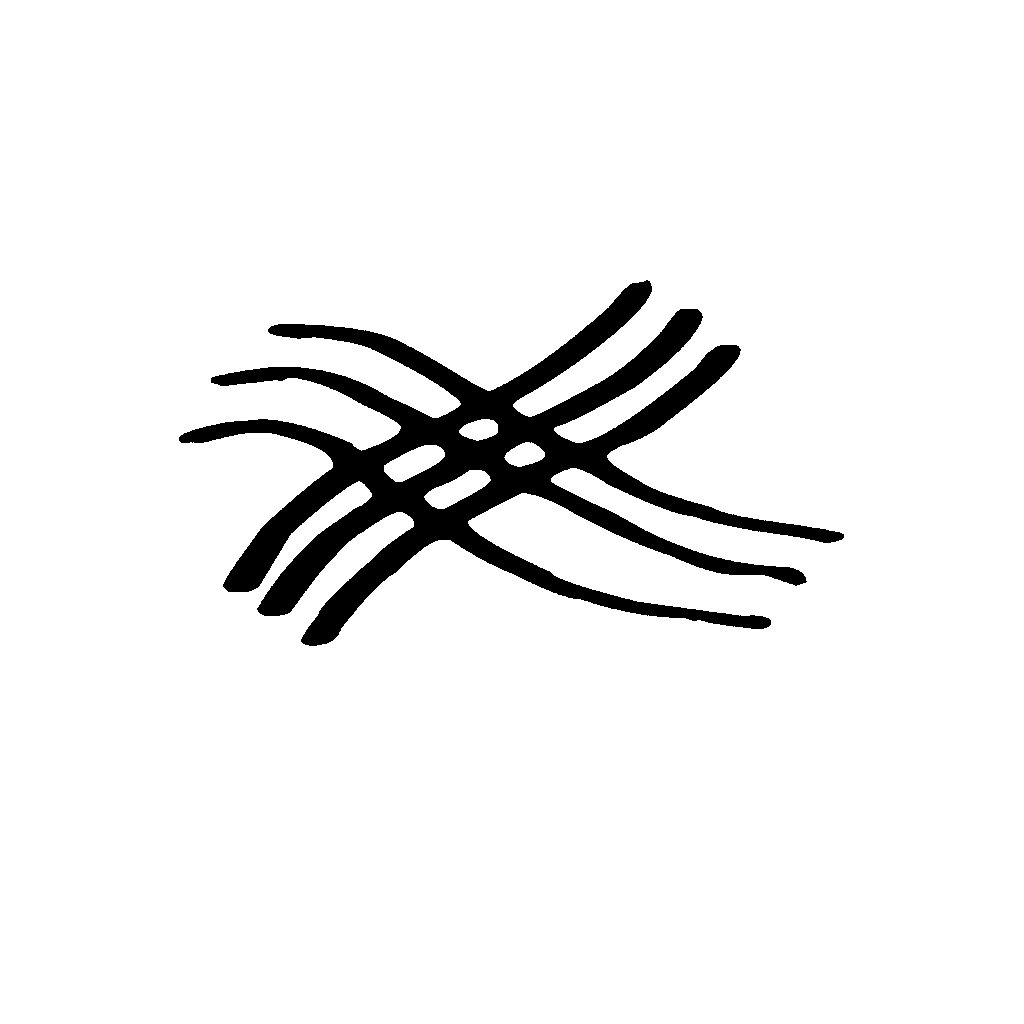} &
\includegraphics[width=0.95\linewidth]{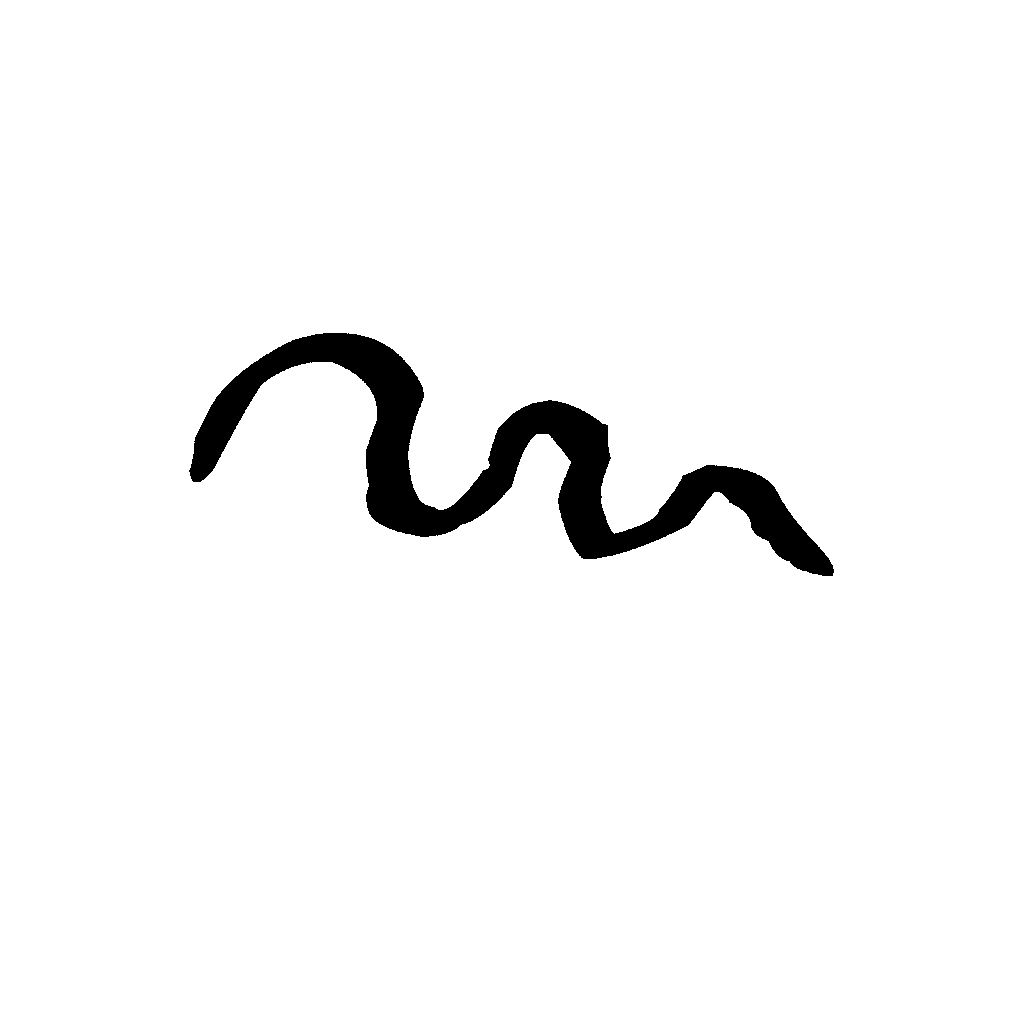} &
\imgwithbox[width=0.95\linewidth]{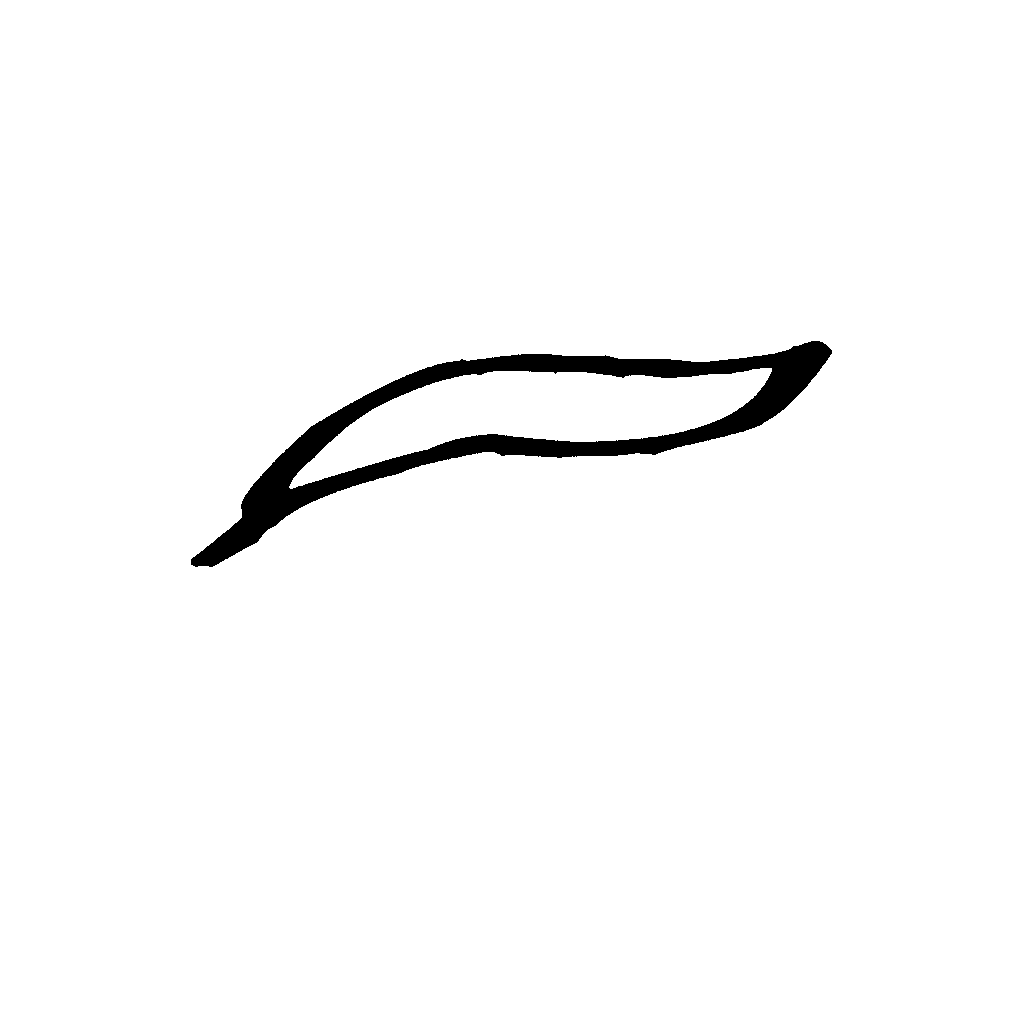} &
\includegraphics[width=0.95\linewidth]{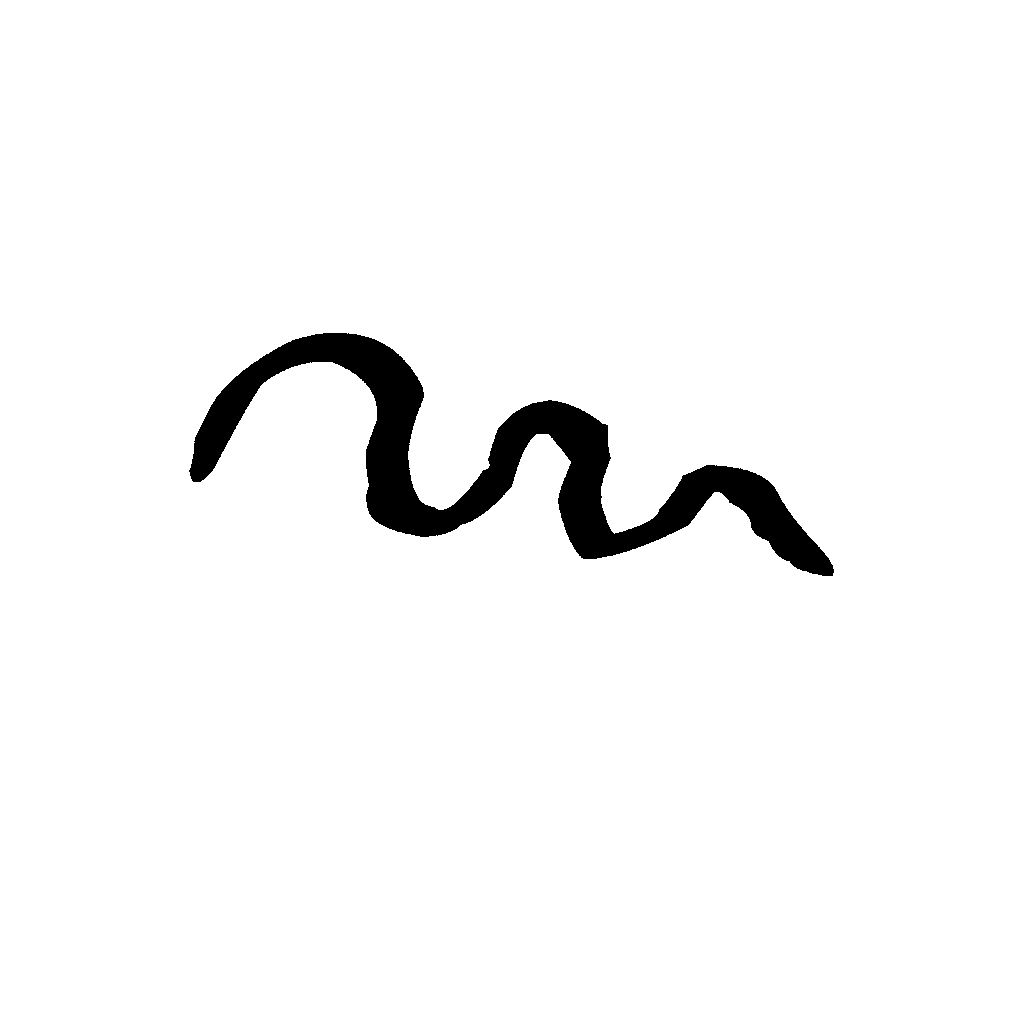} &
\imgwithbox[width=0.95\linewidth]{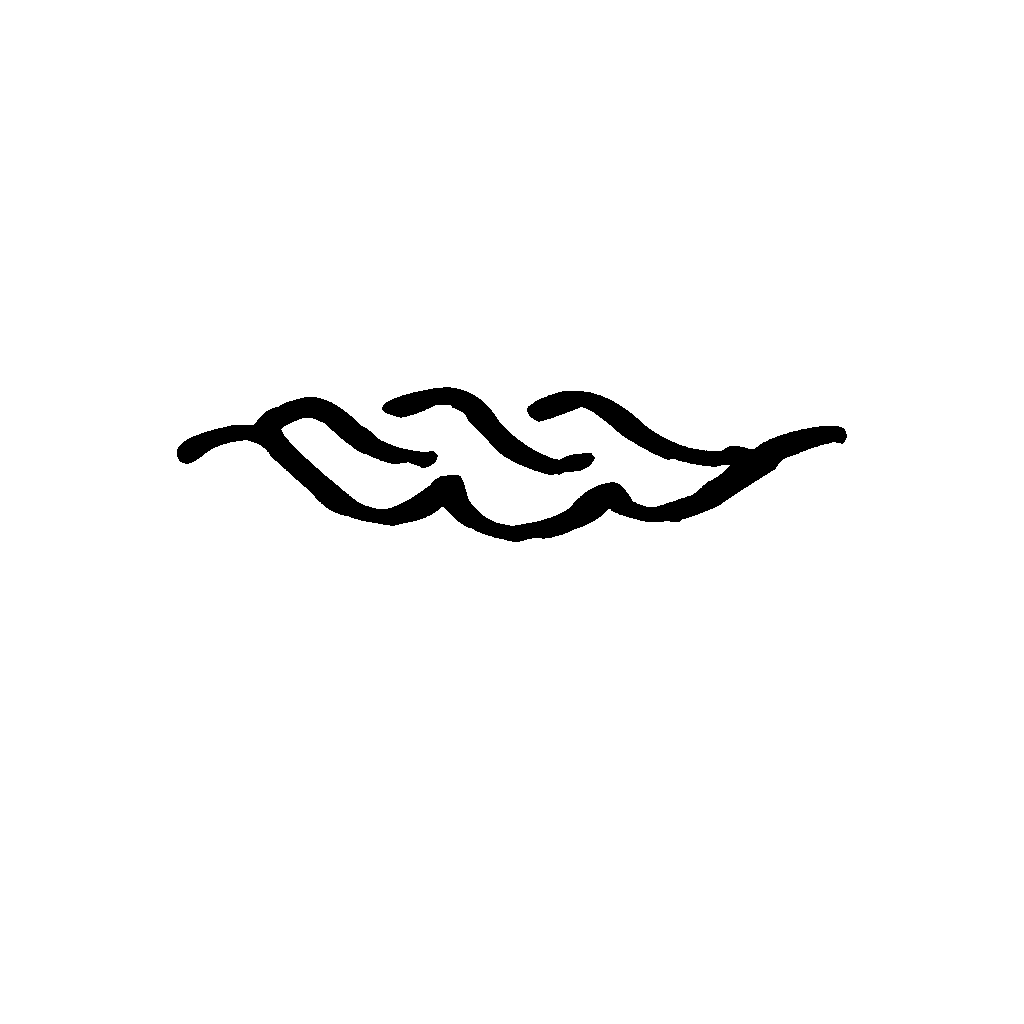} &
\includegraphics[width=0.95\linewidth]{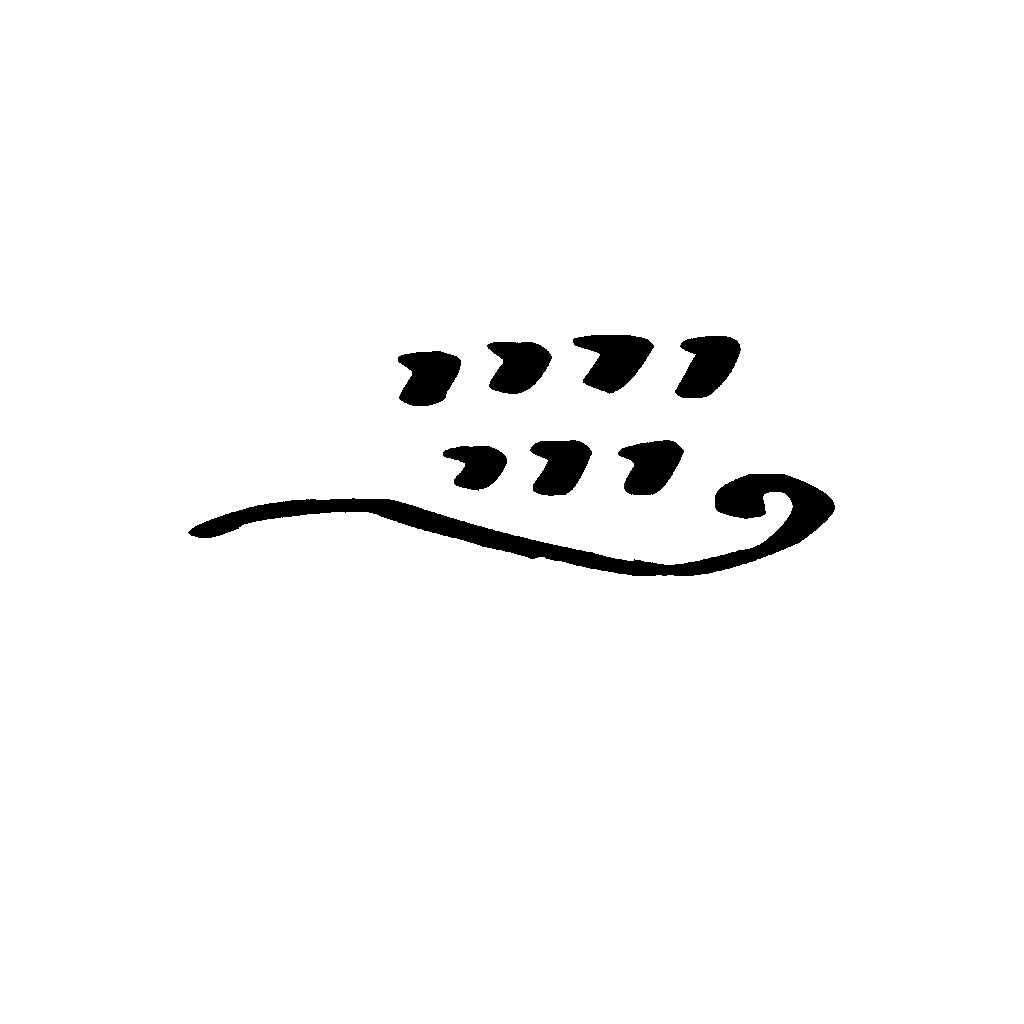} &
\includegraphics[width=0.95\linewidth]{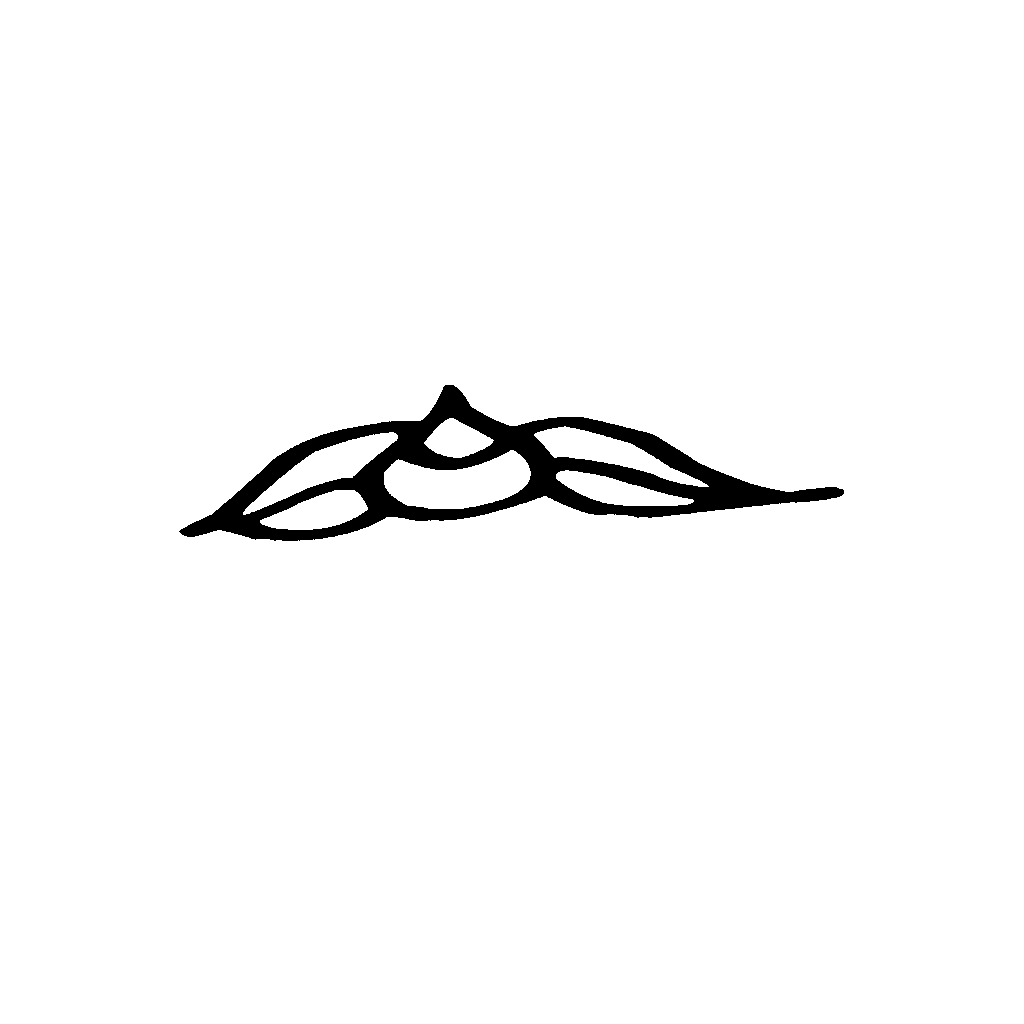} \\
5 &
\includegraphics[width=0.95\linewidth]{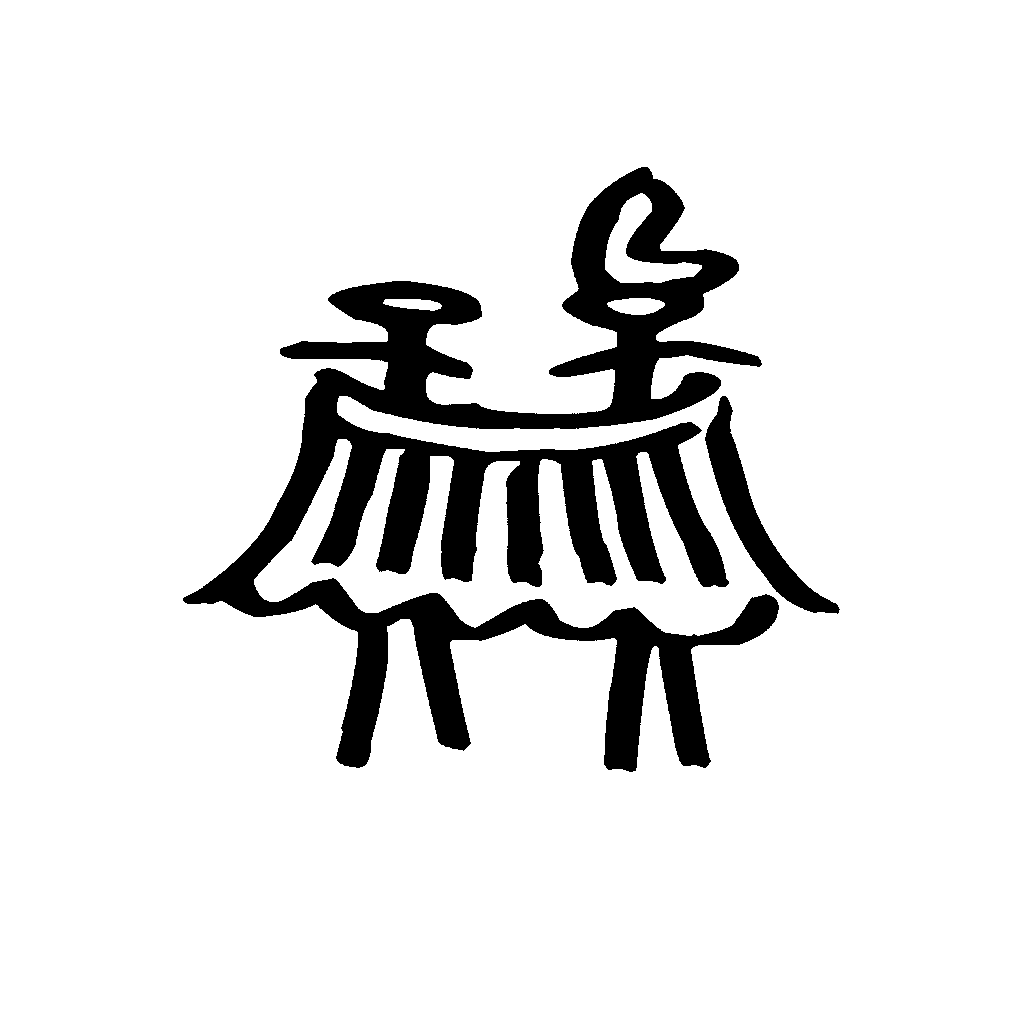} &
\includegraphics[width=0.95\linewidth]{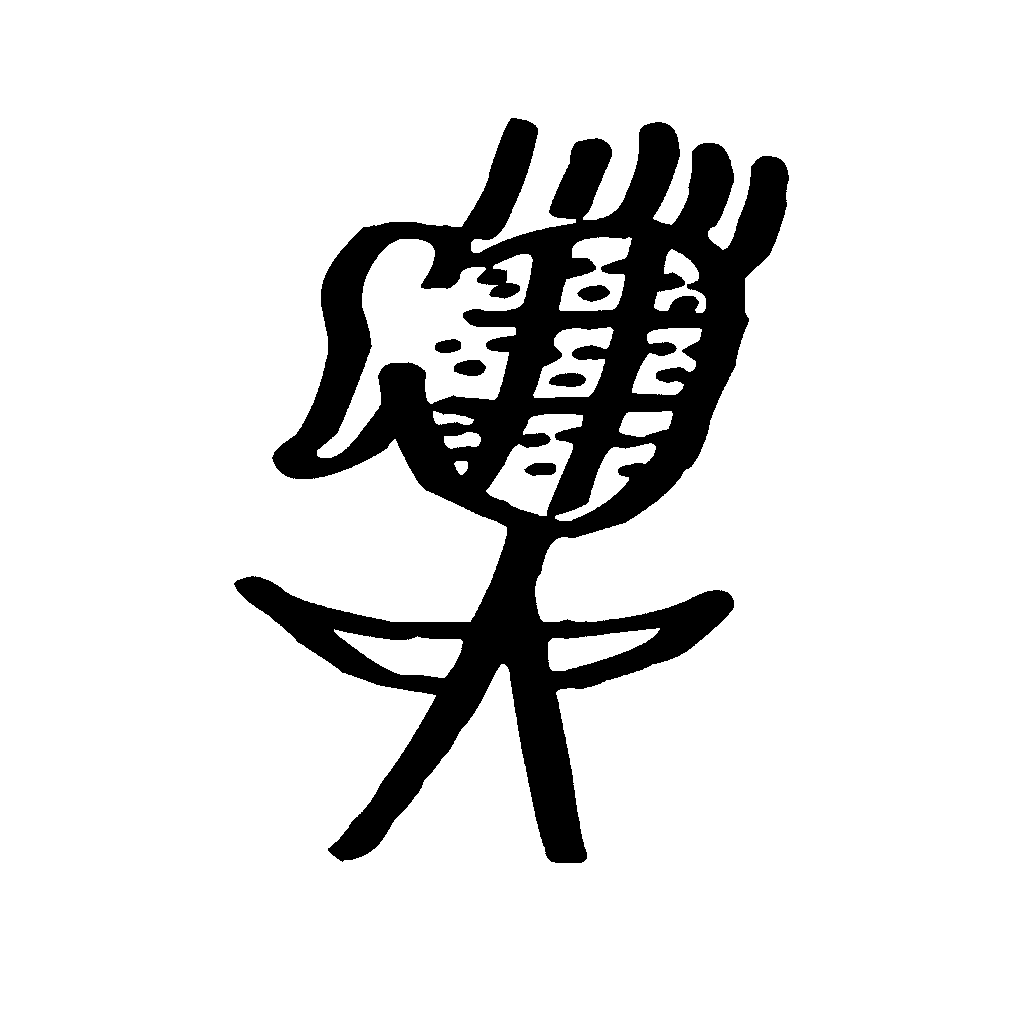} &
\includegraphics[width=0.95\linewidth]{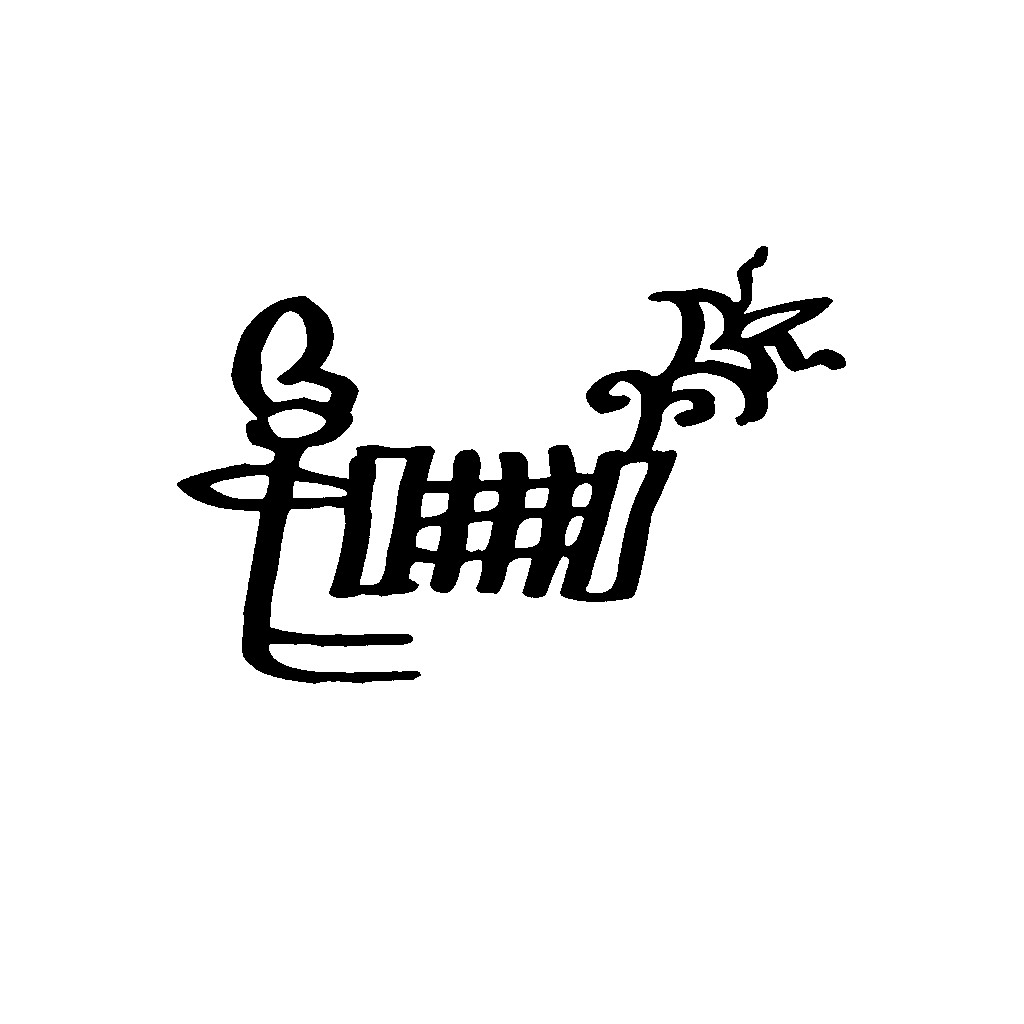} &
\includegraphics[width=0.95\linewidth]{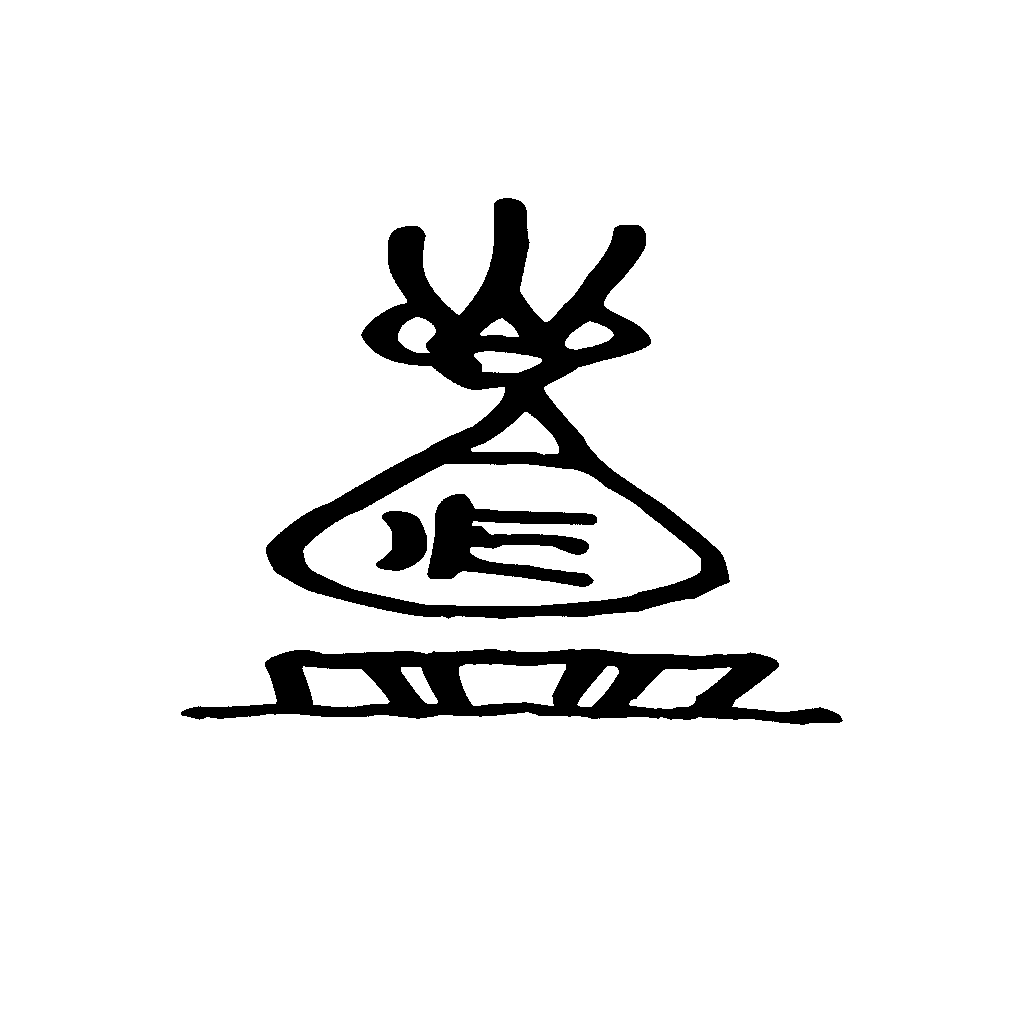} &
\includegraphics[width=0.95\linewidth]{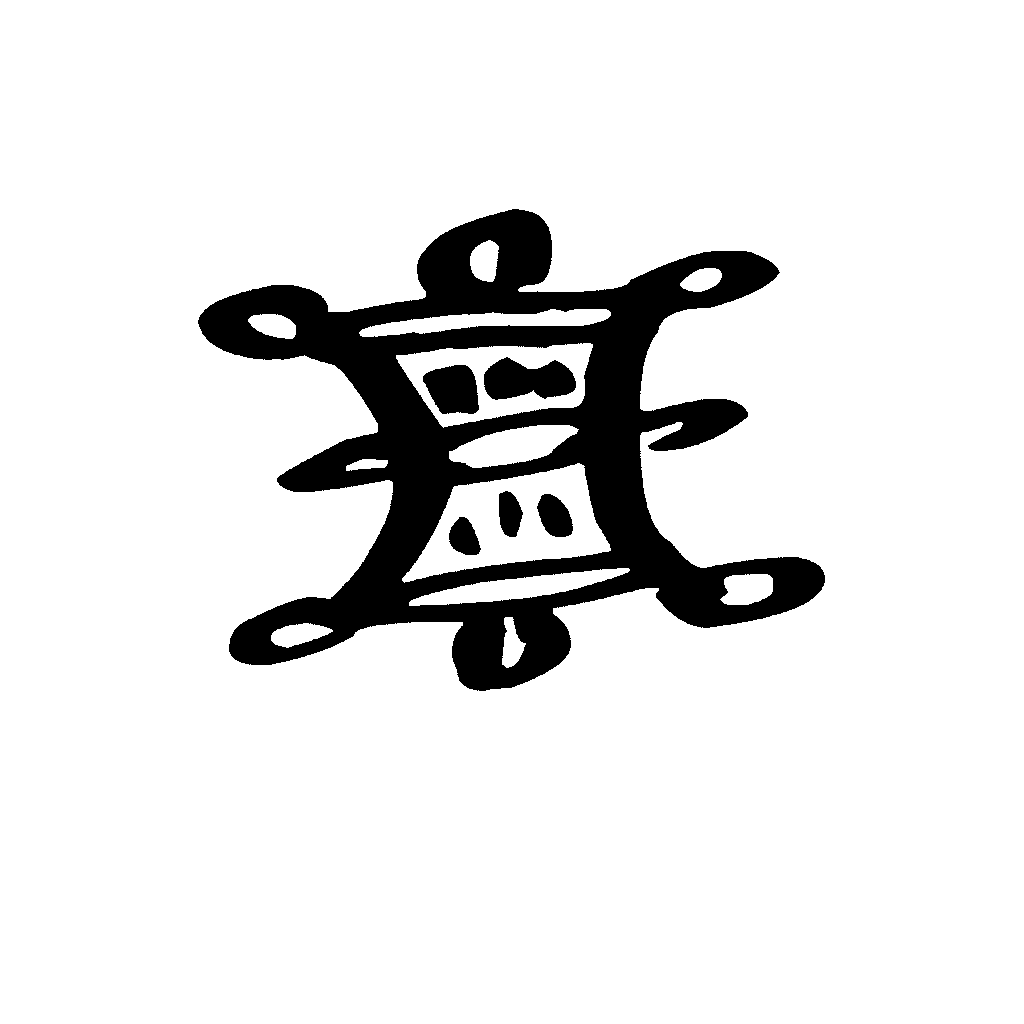} &
\includegraphics[width=0.95\linewidth]{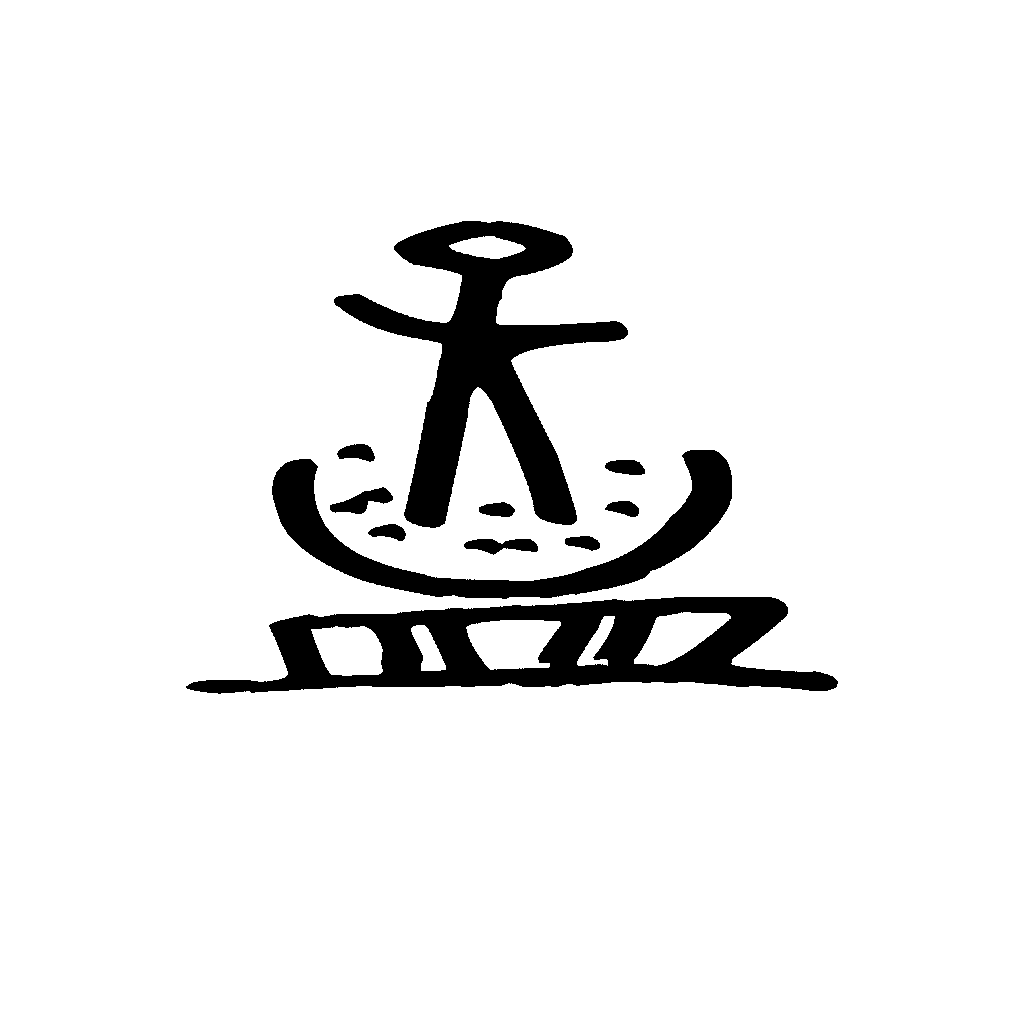} &
\imgwithbox[width=0.95\linewidth]{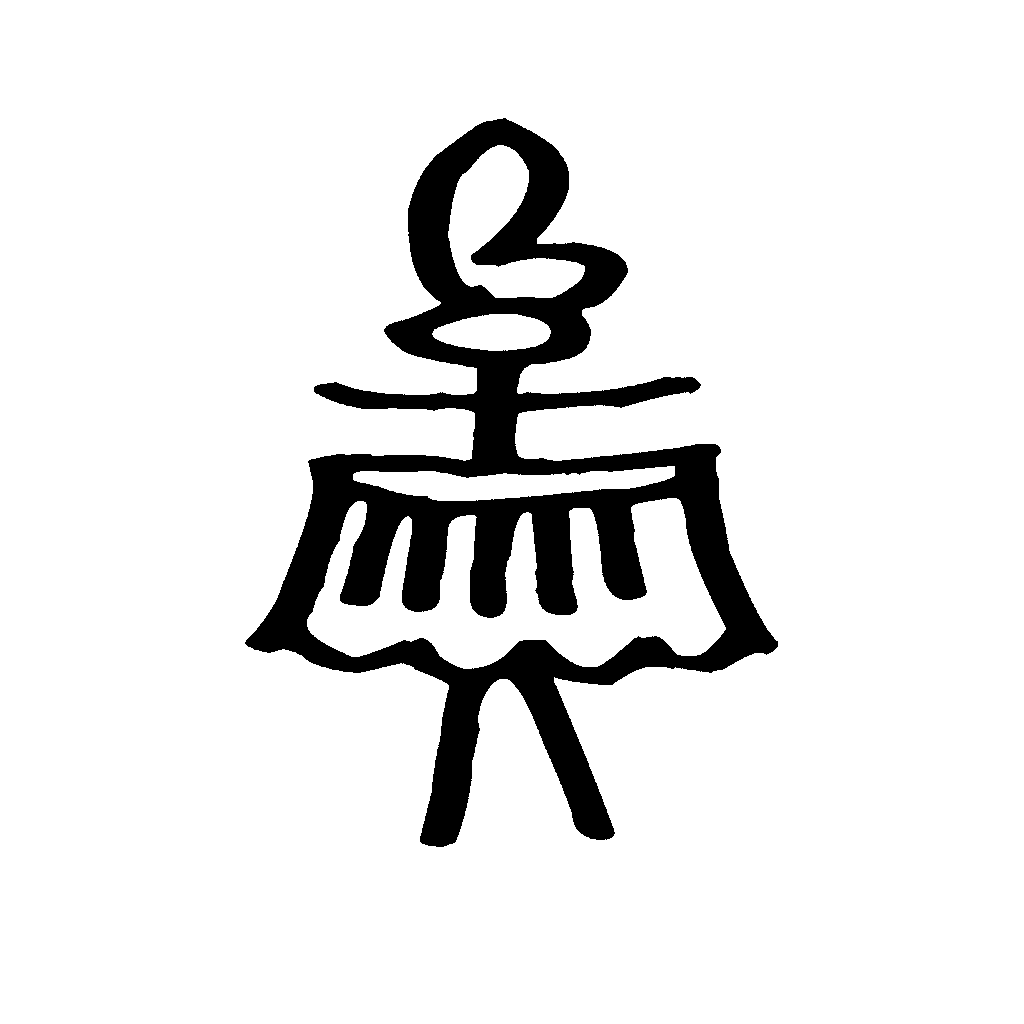} &
\includegraphics[width=0.95\linewidth]{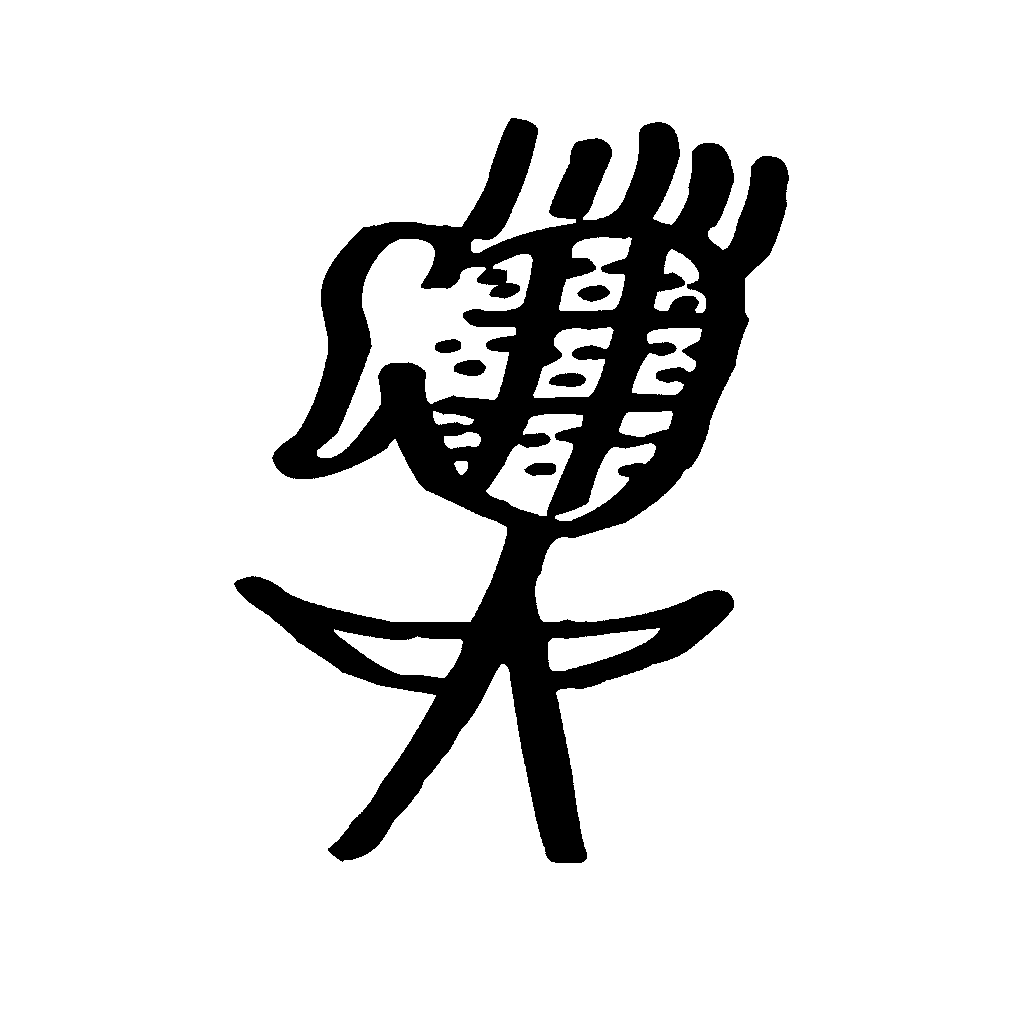} &
\includegraphics[width=0.95\linewidth]{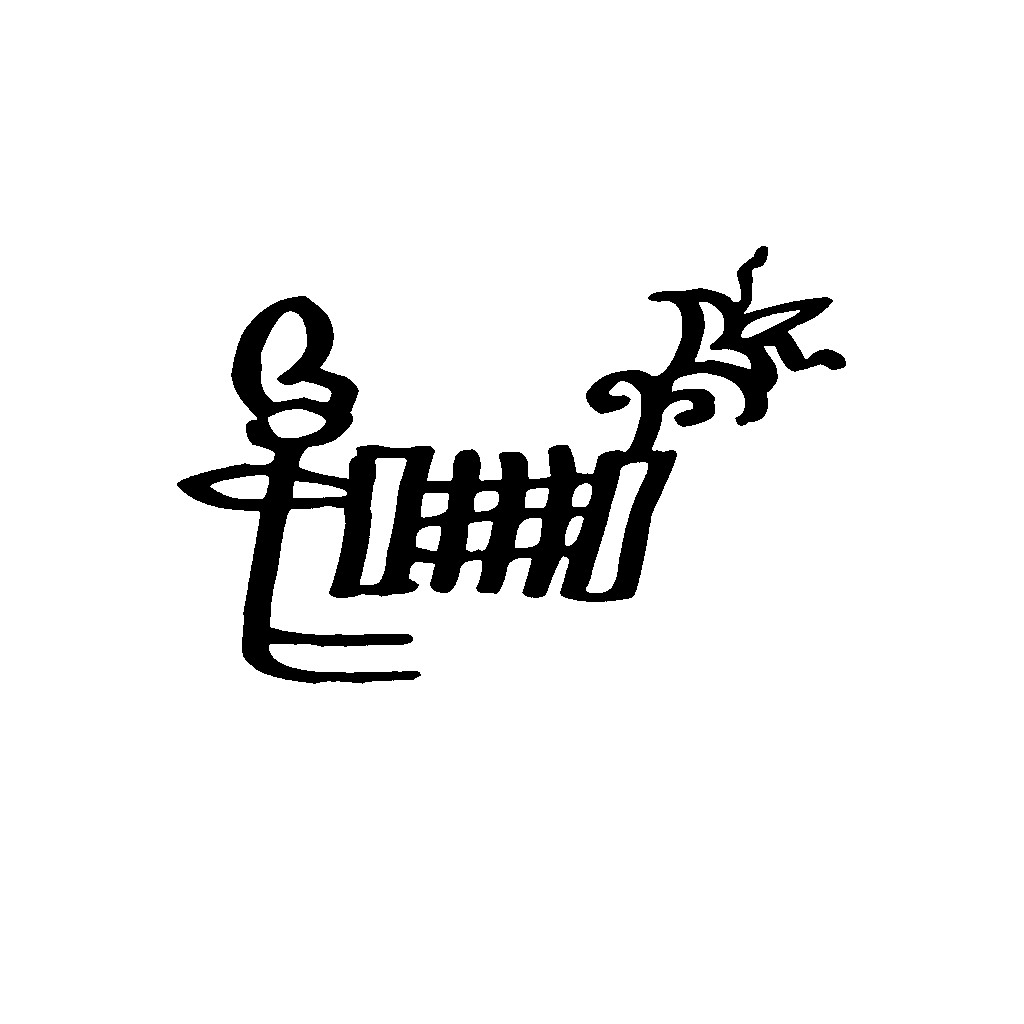} &
\includegraphics[width=0.95\linewidth]{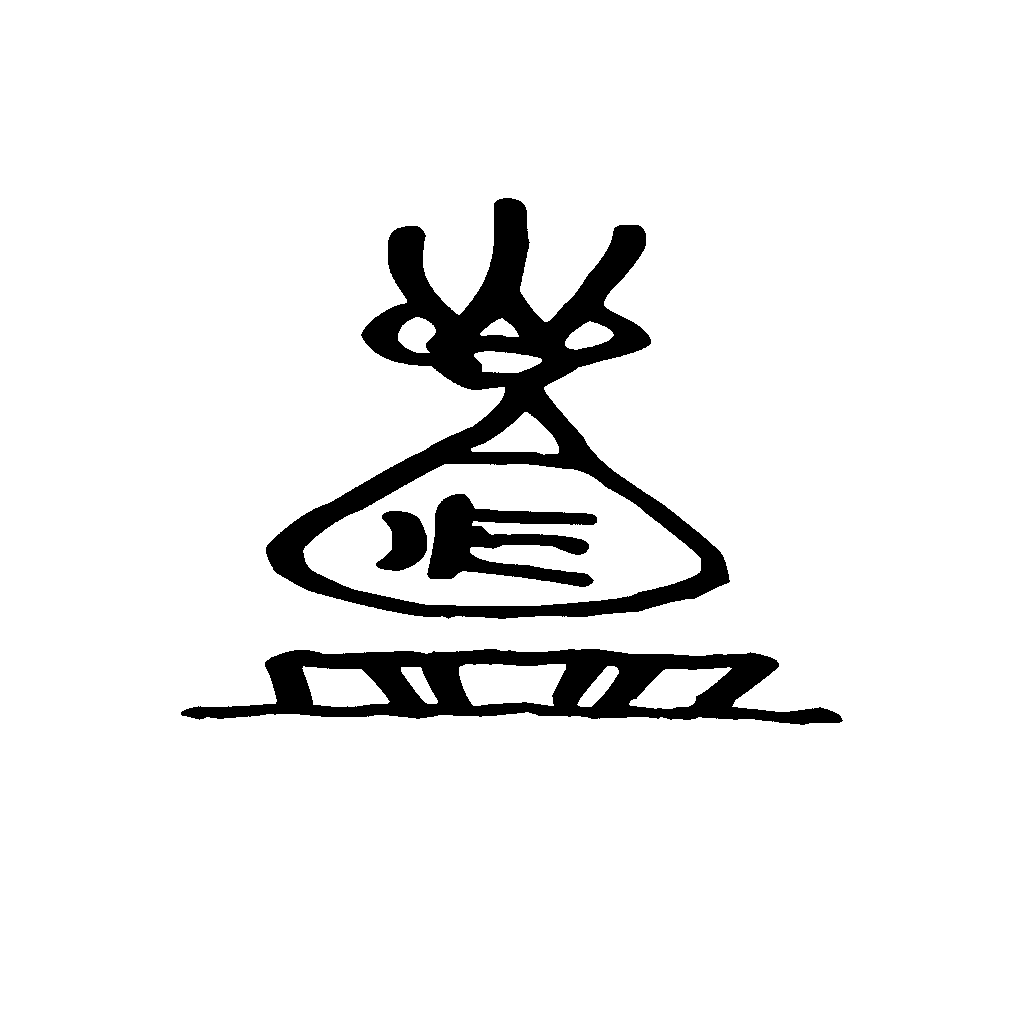} &
\includegraphics[width=0.95\linewidth]{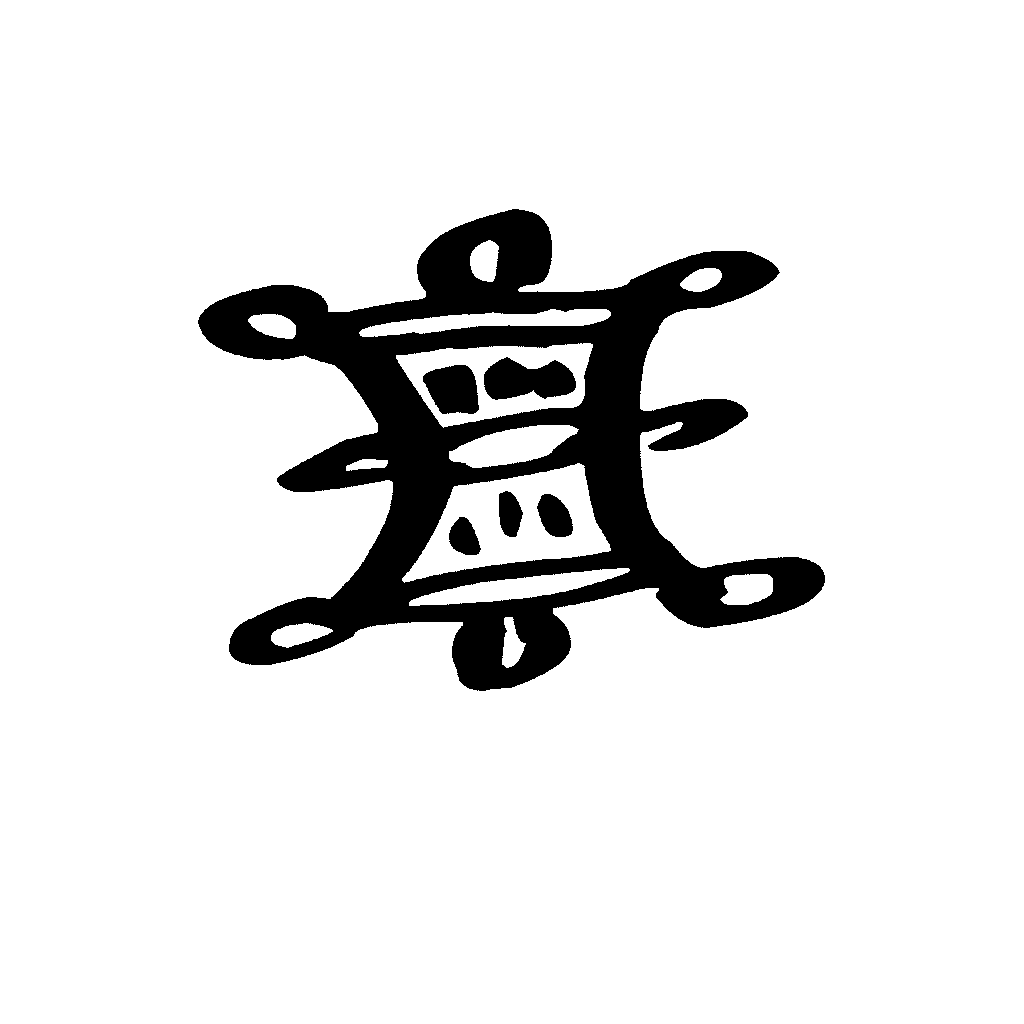} \\
6 &
\includegraphics[width=0.95\linewidth]{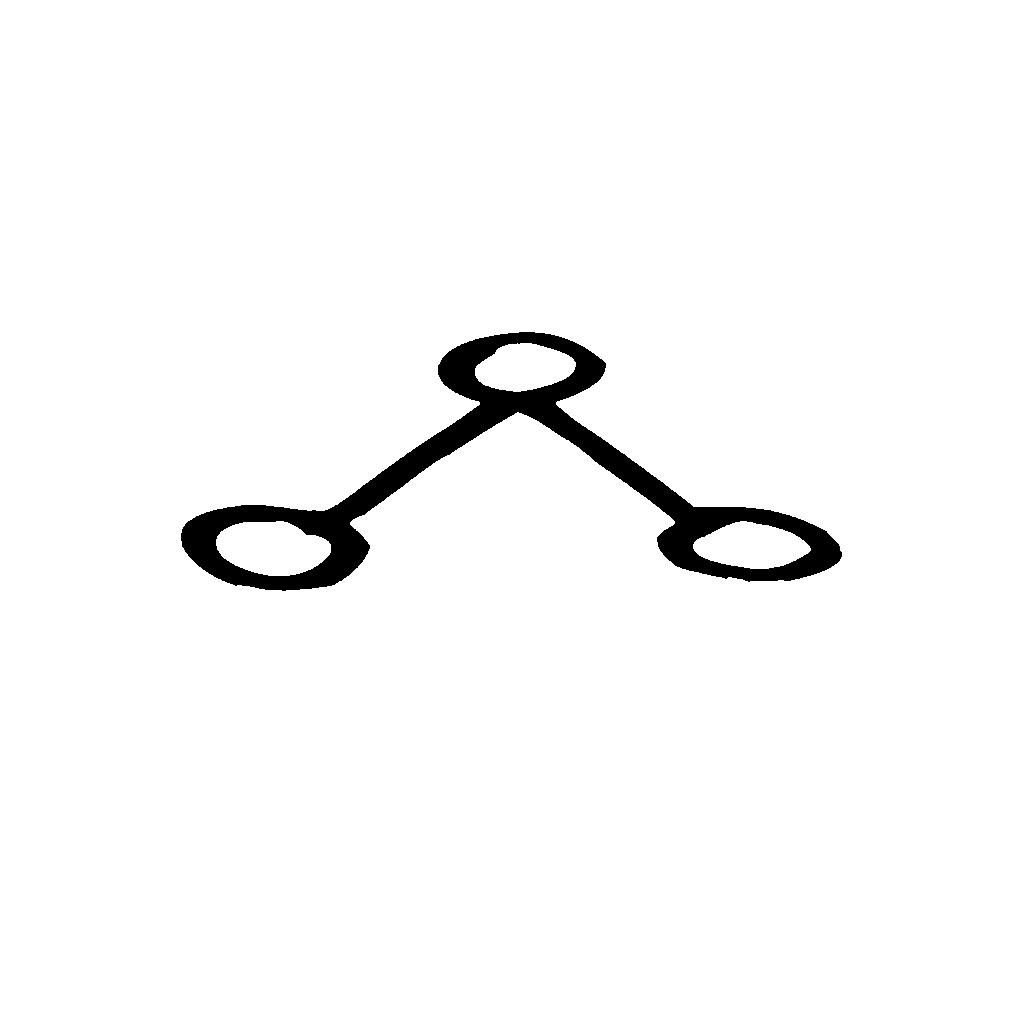} &
\includegraphics[width=0.95\linewidth]{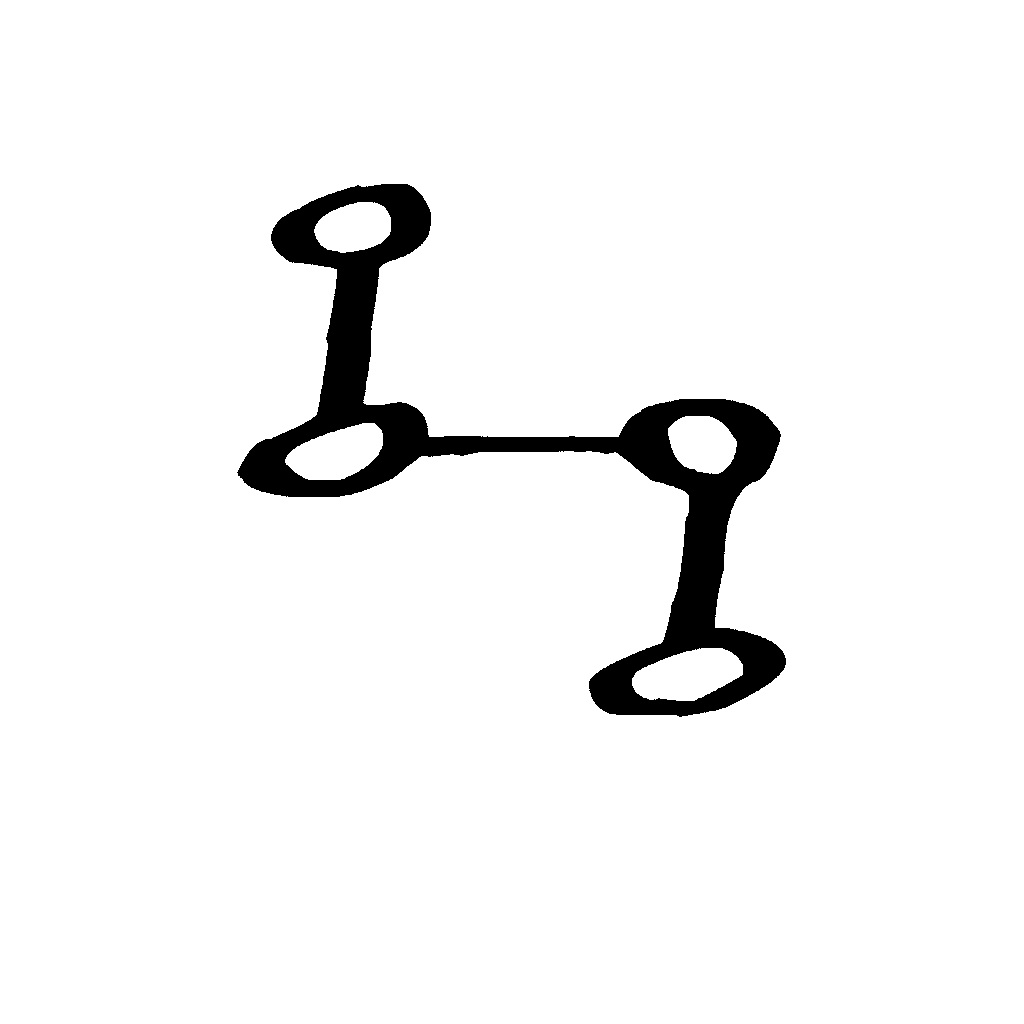} &
\imgwithbox[width=0.95\linewidth]{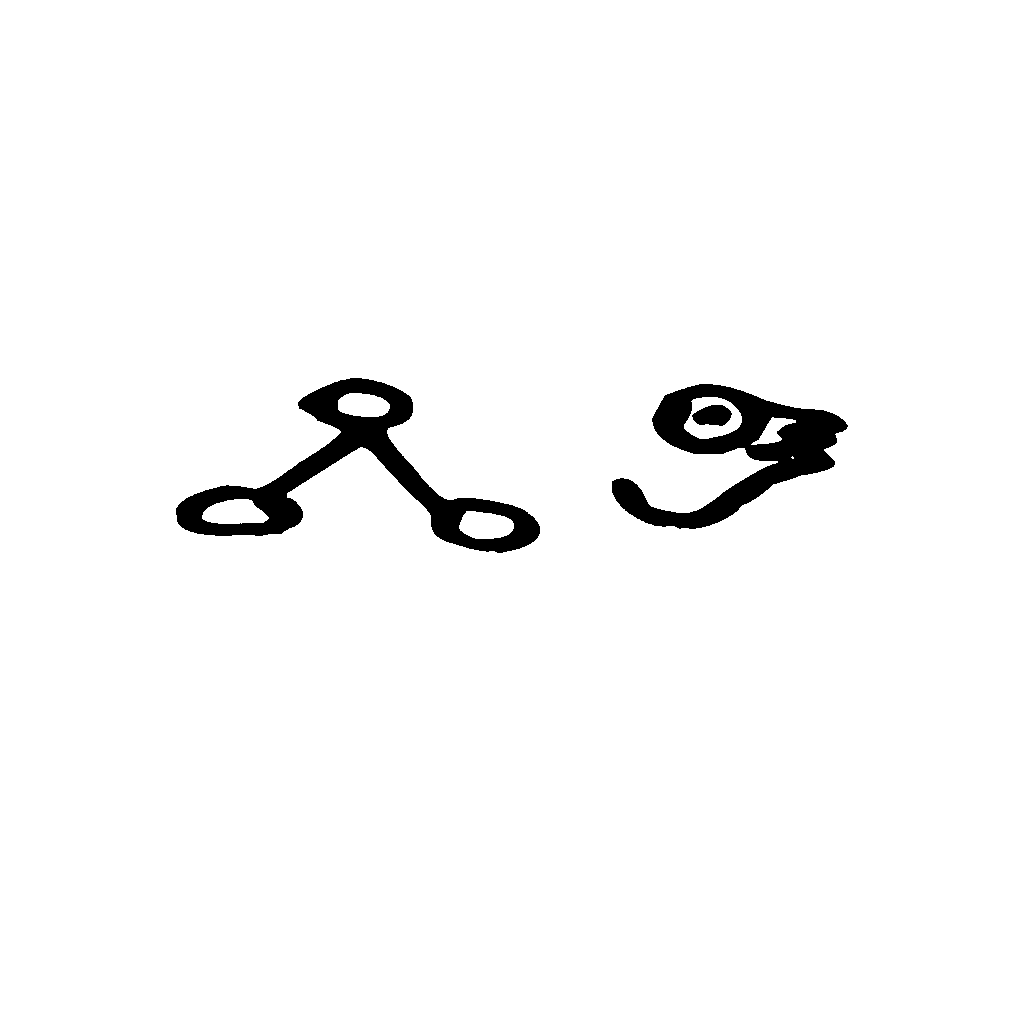} &
\imgwithbox[width=0.95\linewidth]{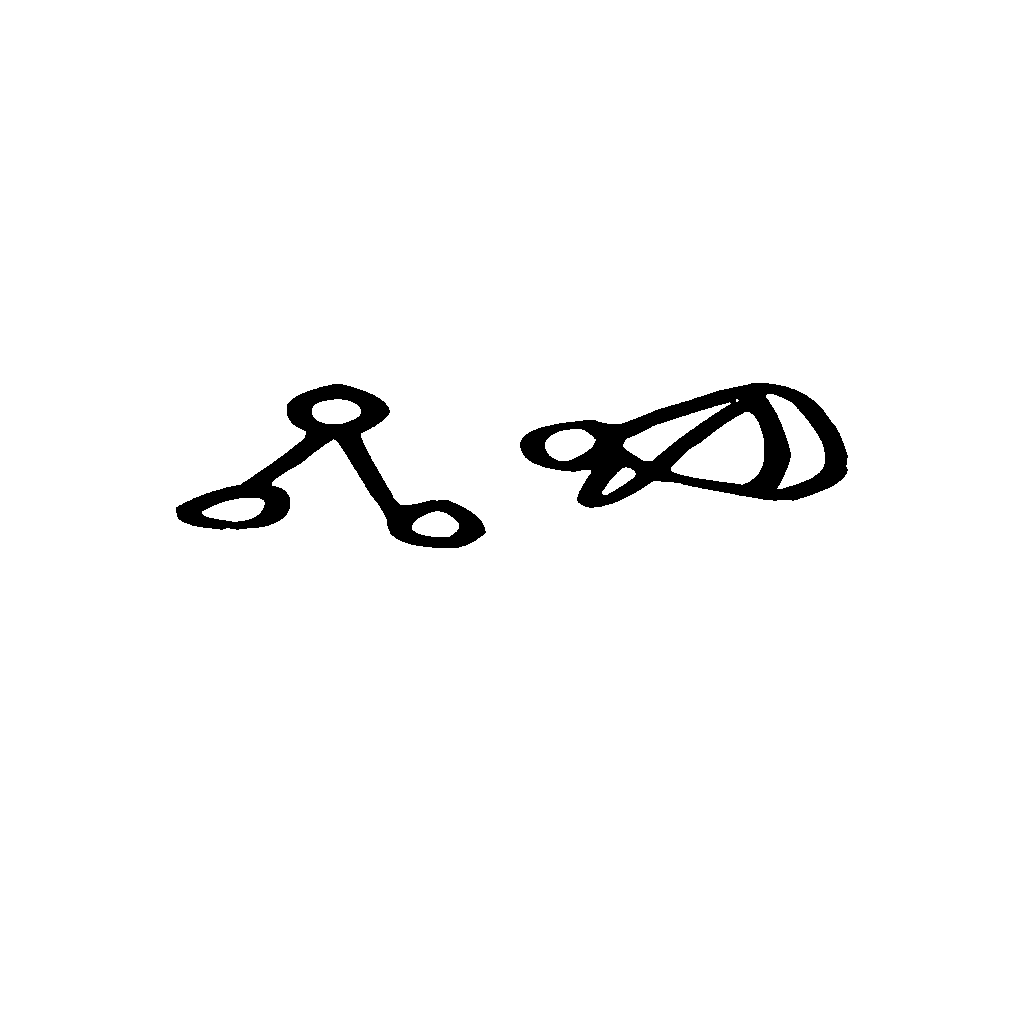} &
\includegraphics[width=0.95\linewidth]{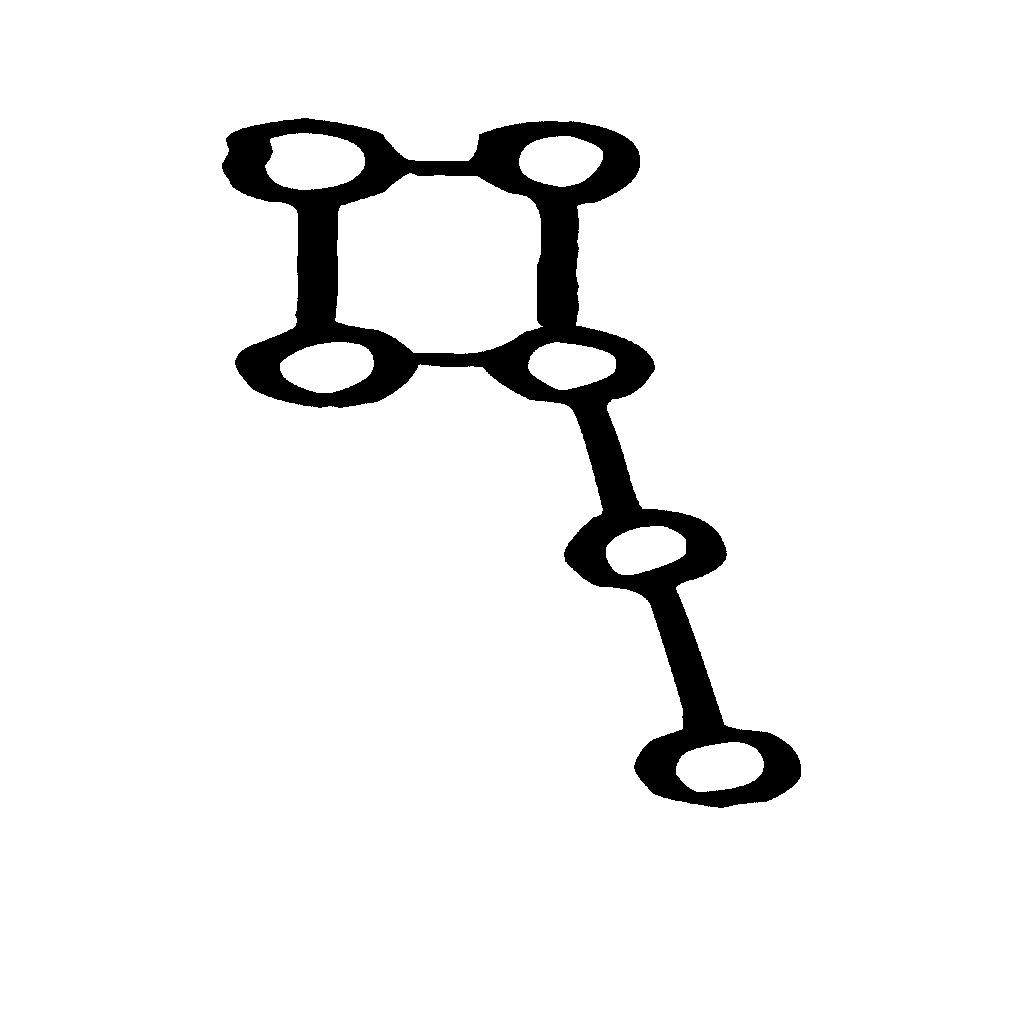} &
\includegraphics[width=0.95\linewidth]{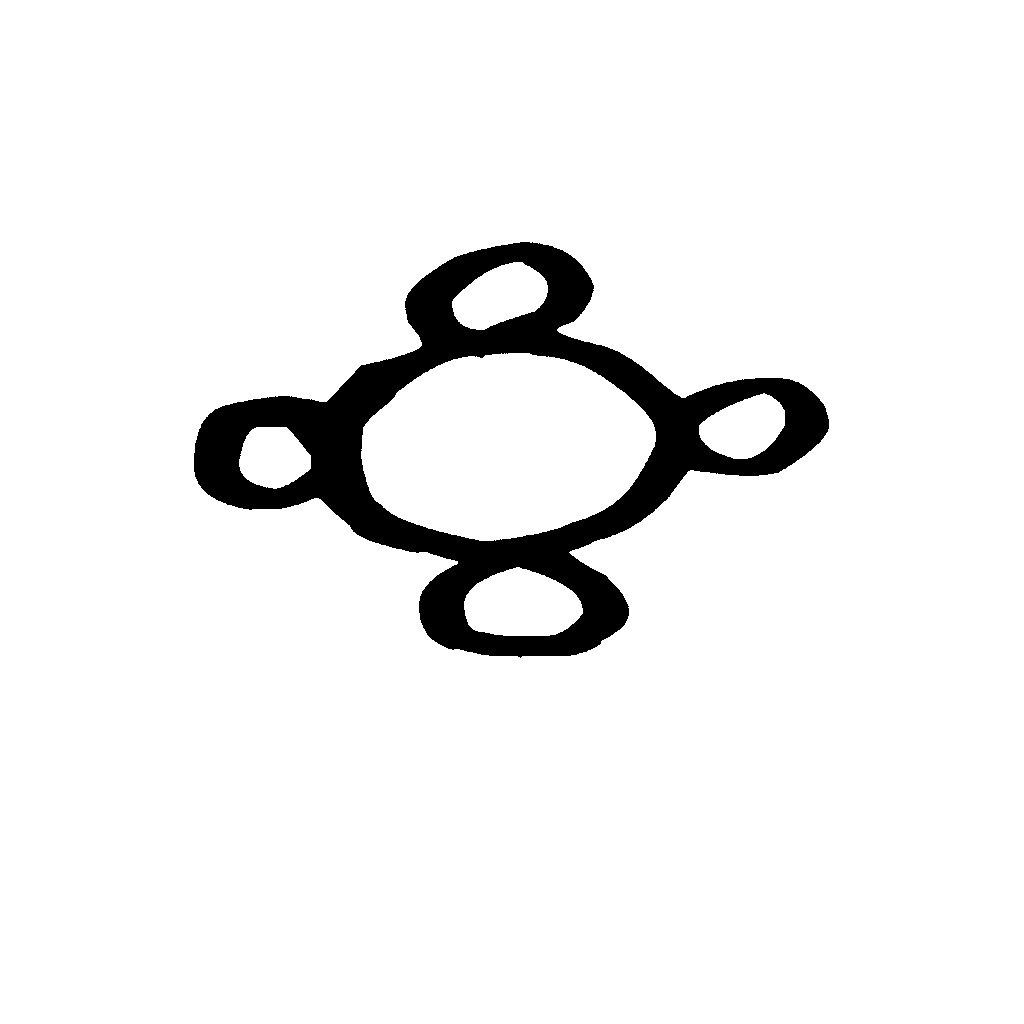} &
\imgwithbox[width=0.95\linewidth]{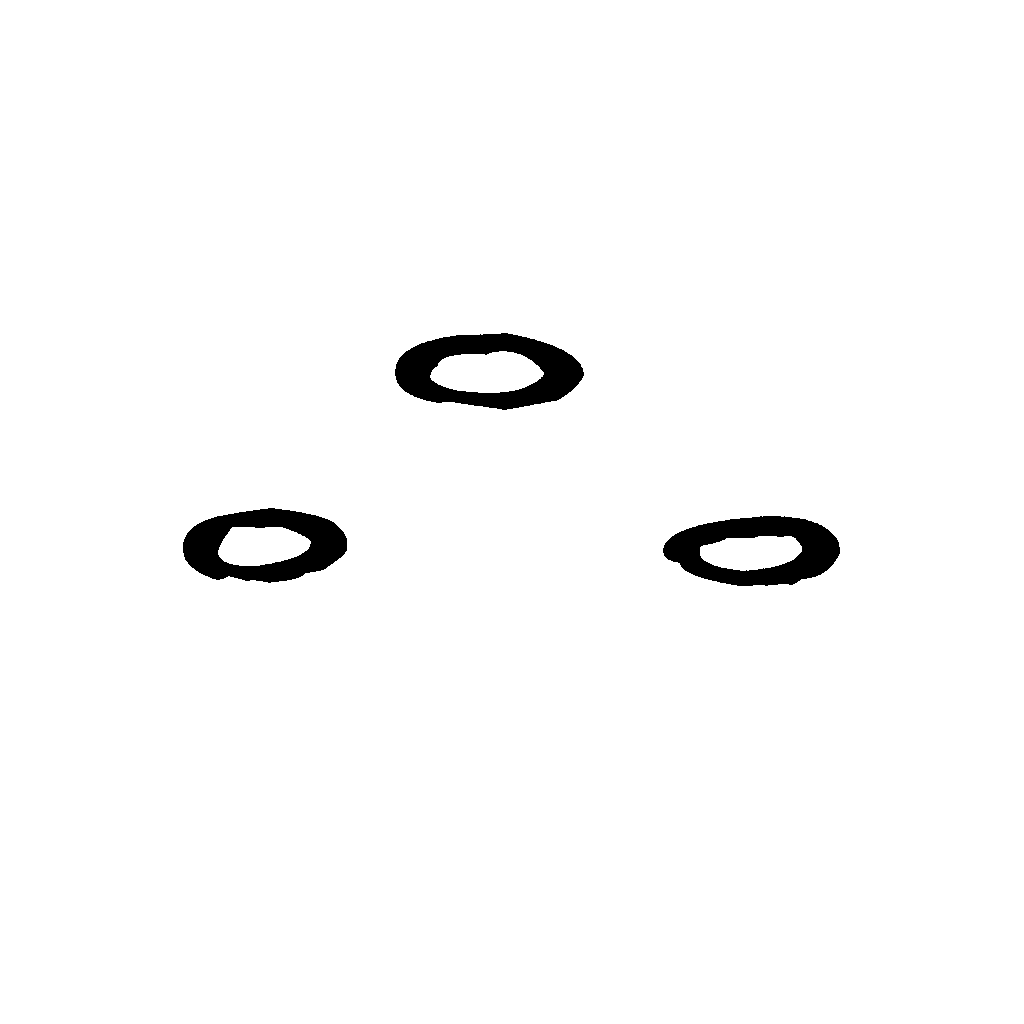} &
\imgwithbox[width=0.95\linewidth]{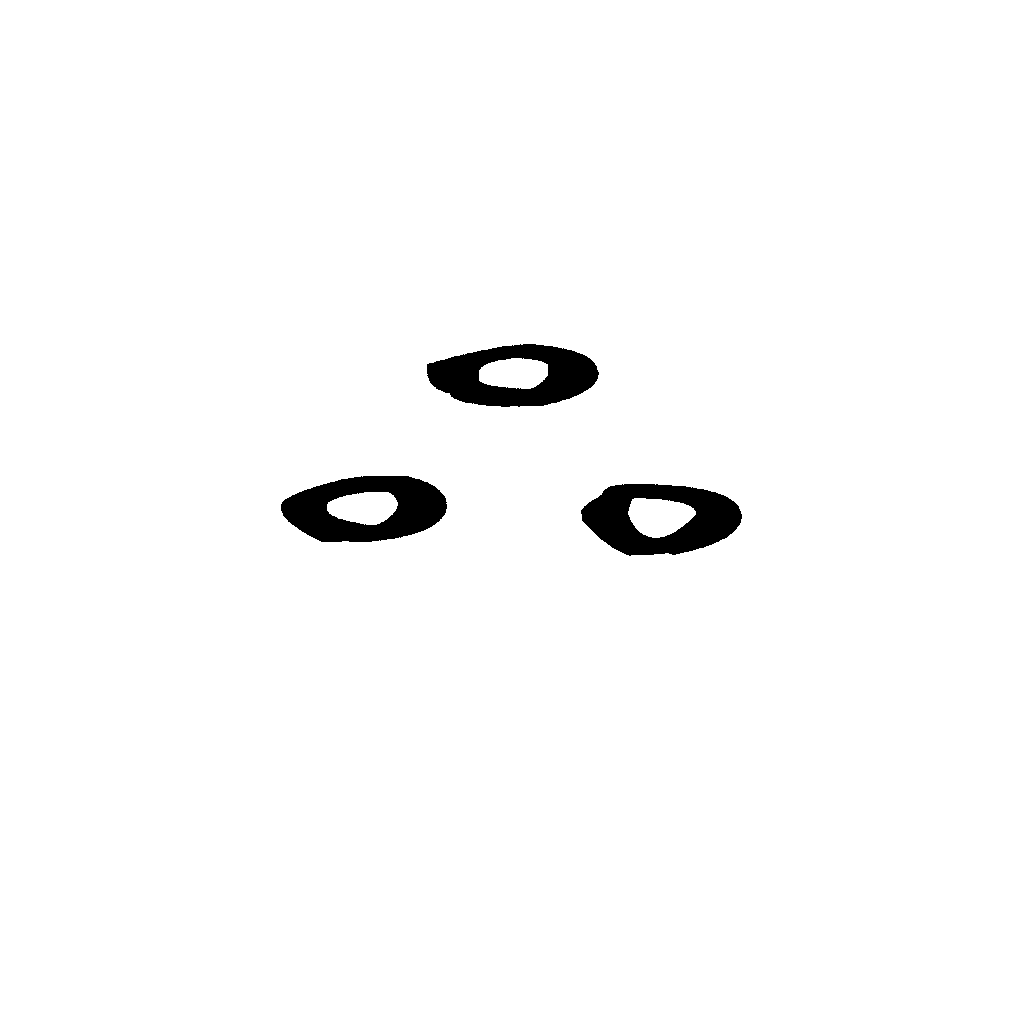} &
\includegraphics[width=0.95\linewidth]{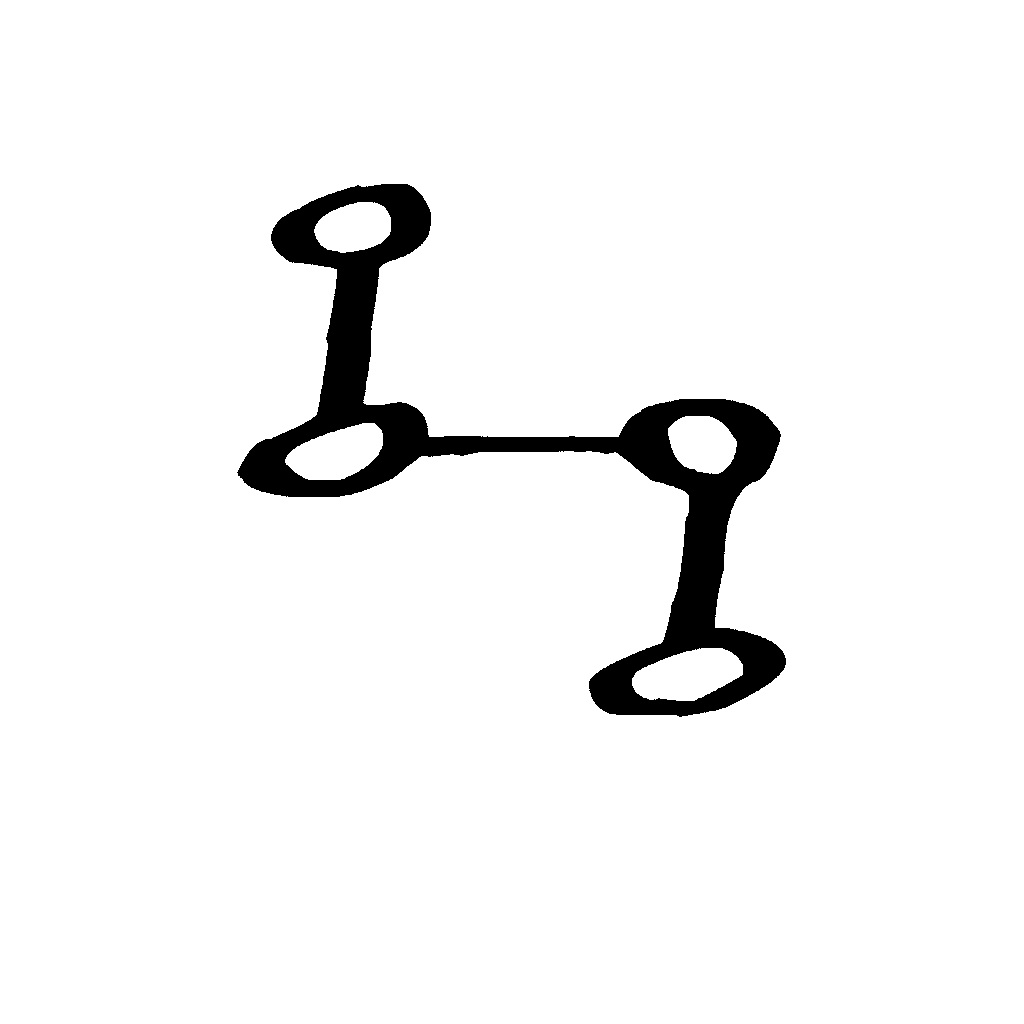} &
\imgwithbox[width=0.95\linewidth]{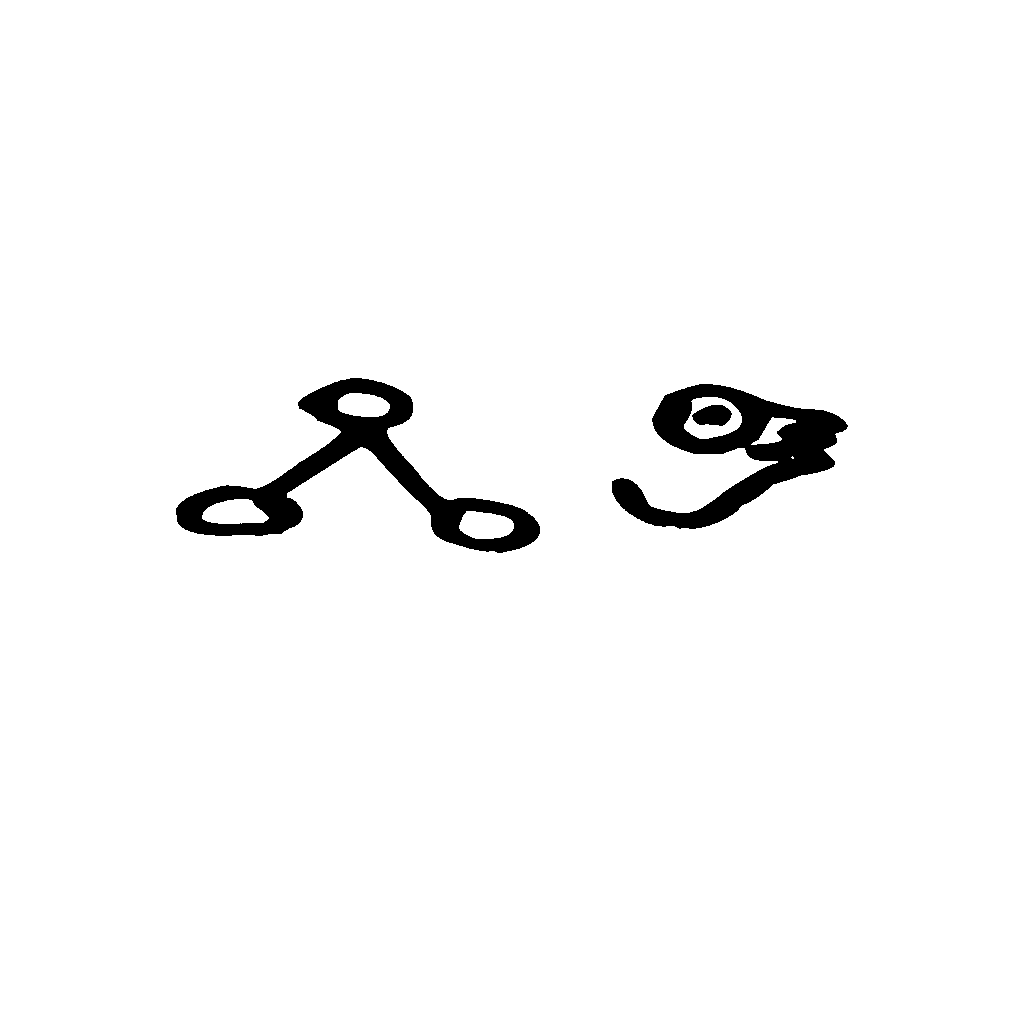} &
\imgwithbox[width=0.95\linewidth]{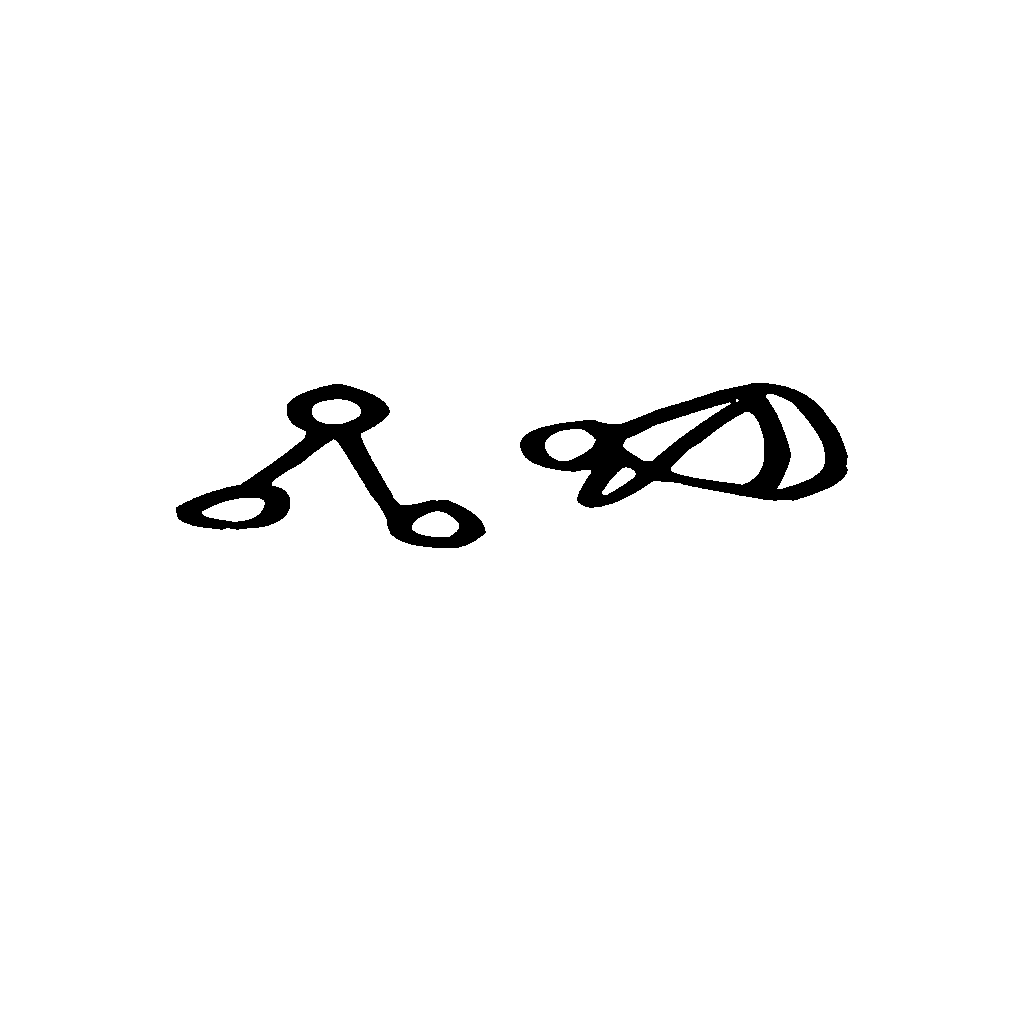} \\
\midrule\noalign{\vskip -3pt}\rowcolor{gray!20}\multicolumn{12}{c}{\textbf{Egyptian Hieroglyphs}} \\\noalign{\vskip -2pt}\midrule
7 &
\includegraphics[width=0.95\linewidth]{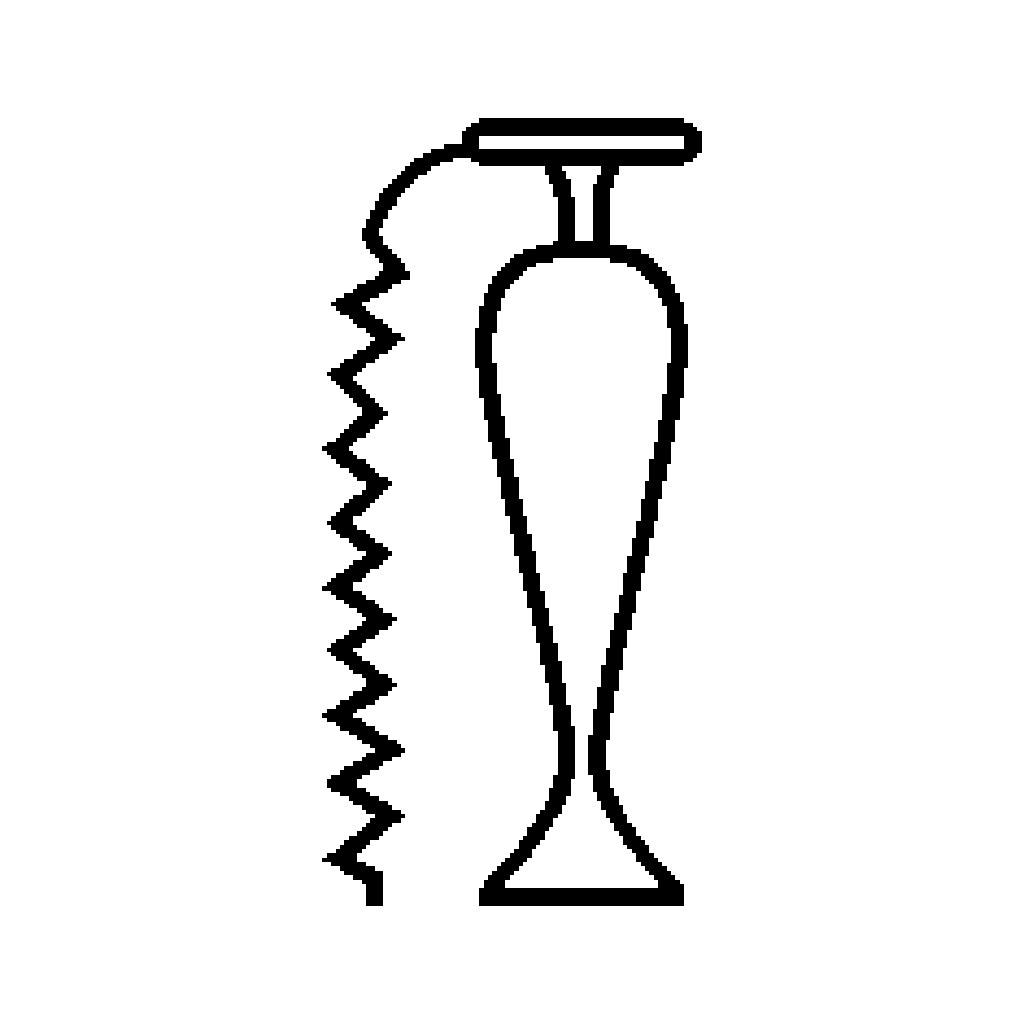} &
\imgwithbox[width=0.95\linewidth]{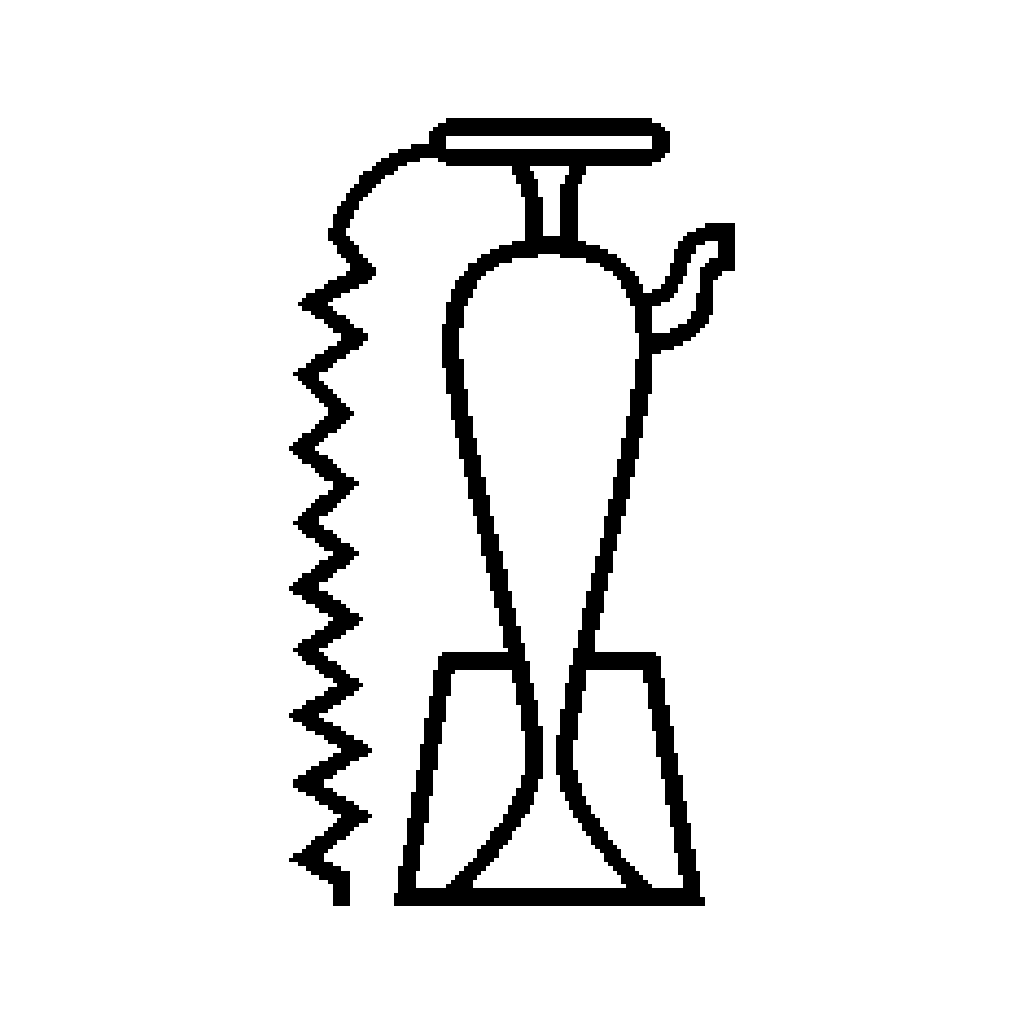} &
\imgwithbox[width=0.95\linewidth]{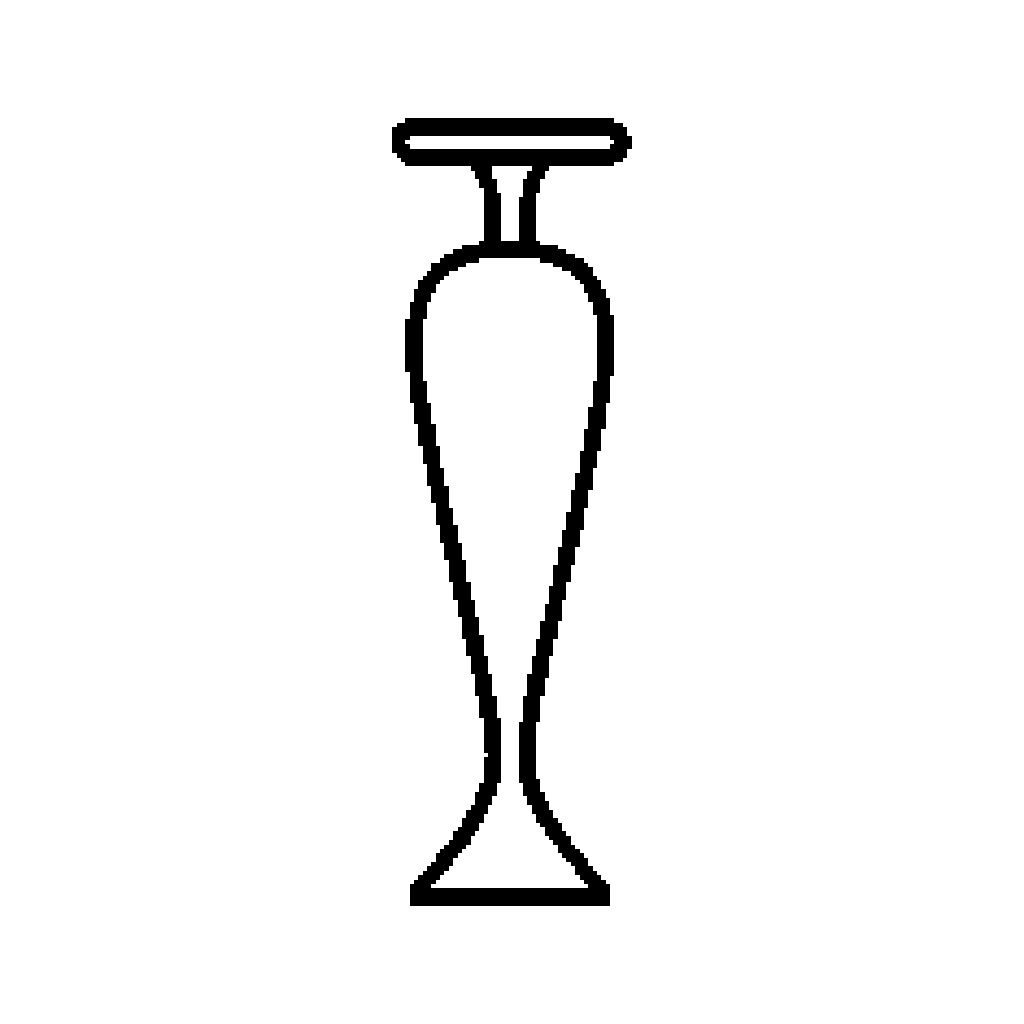} &
\imgwithbox[width=0.95\linewidth]{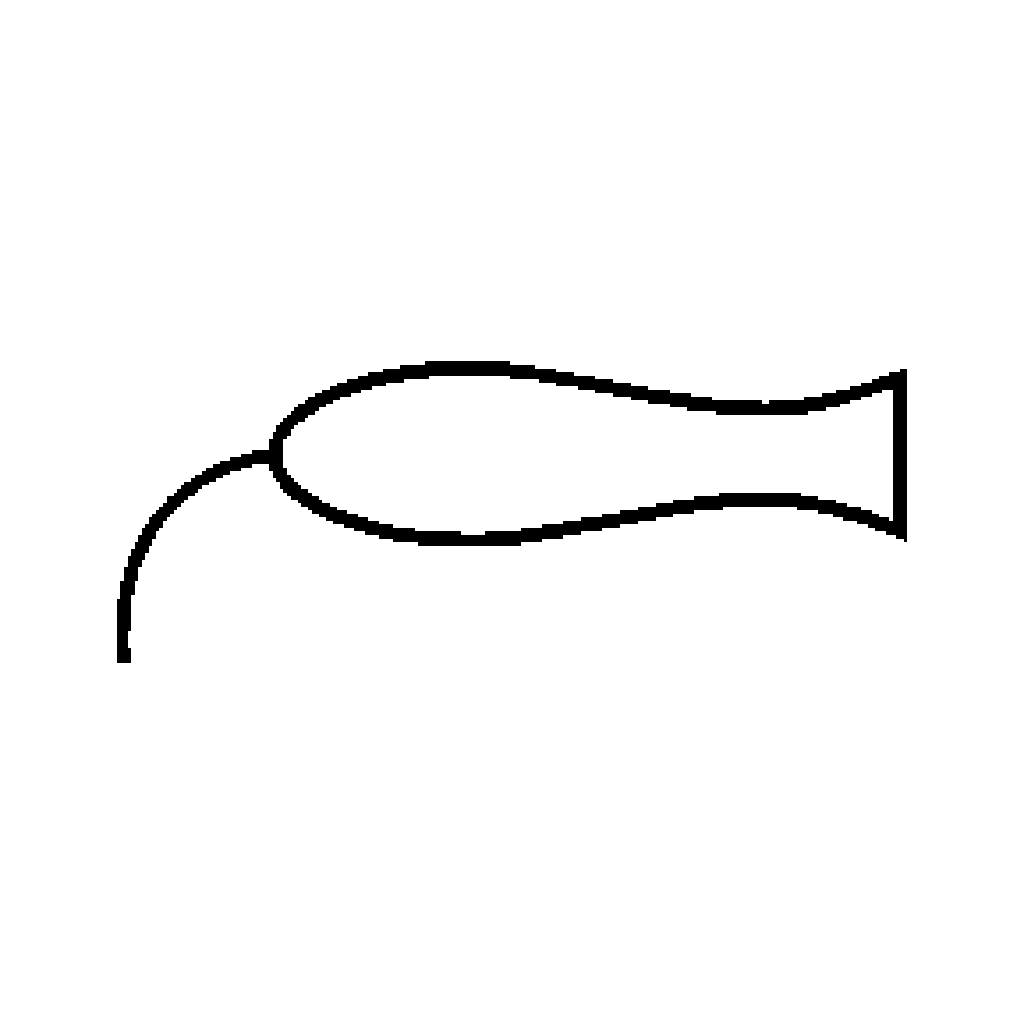} &
\includegraphics[width=0.95\linewidth]{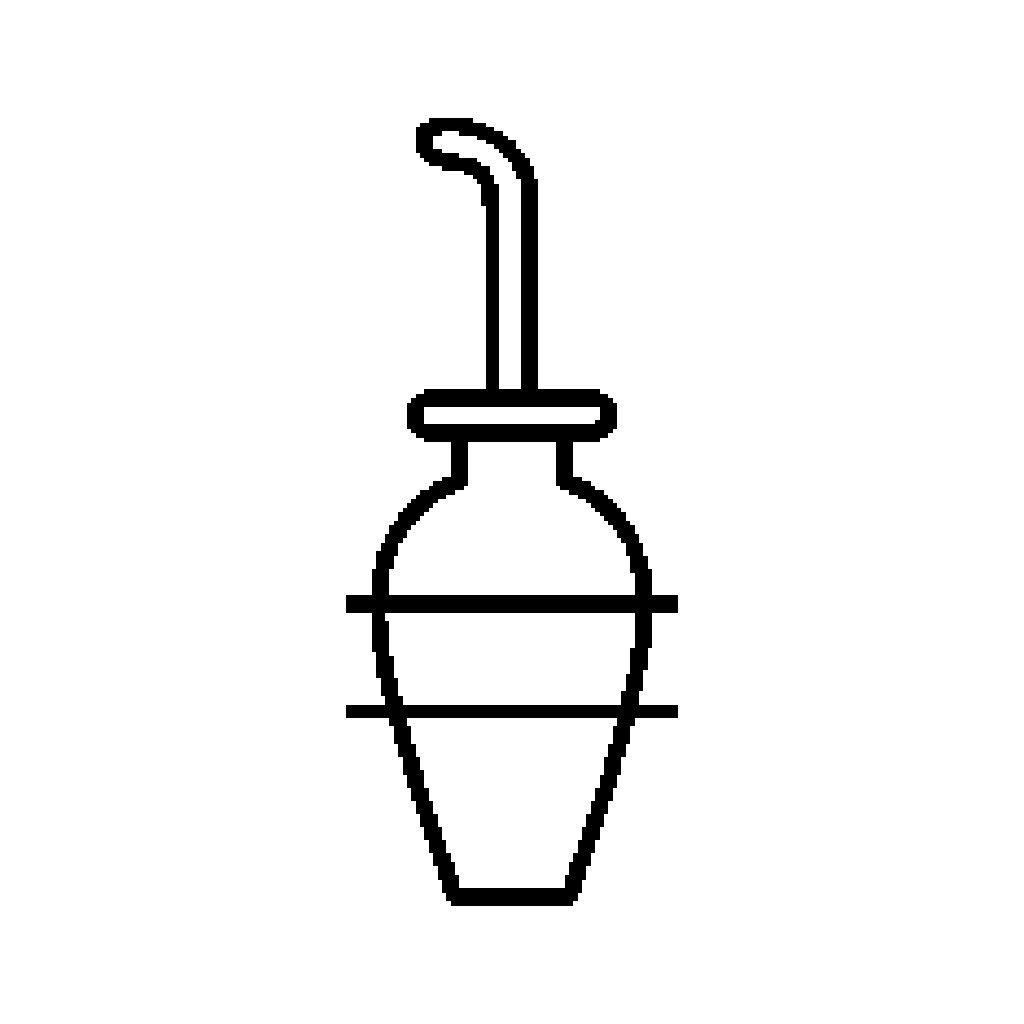} &
\includegraphics[width=0.95\linewidth]{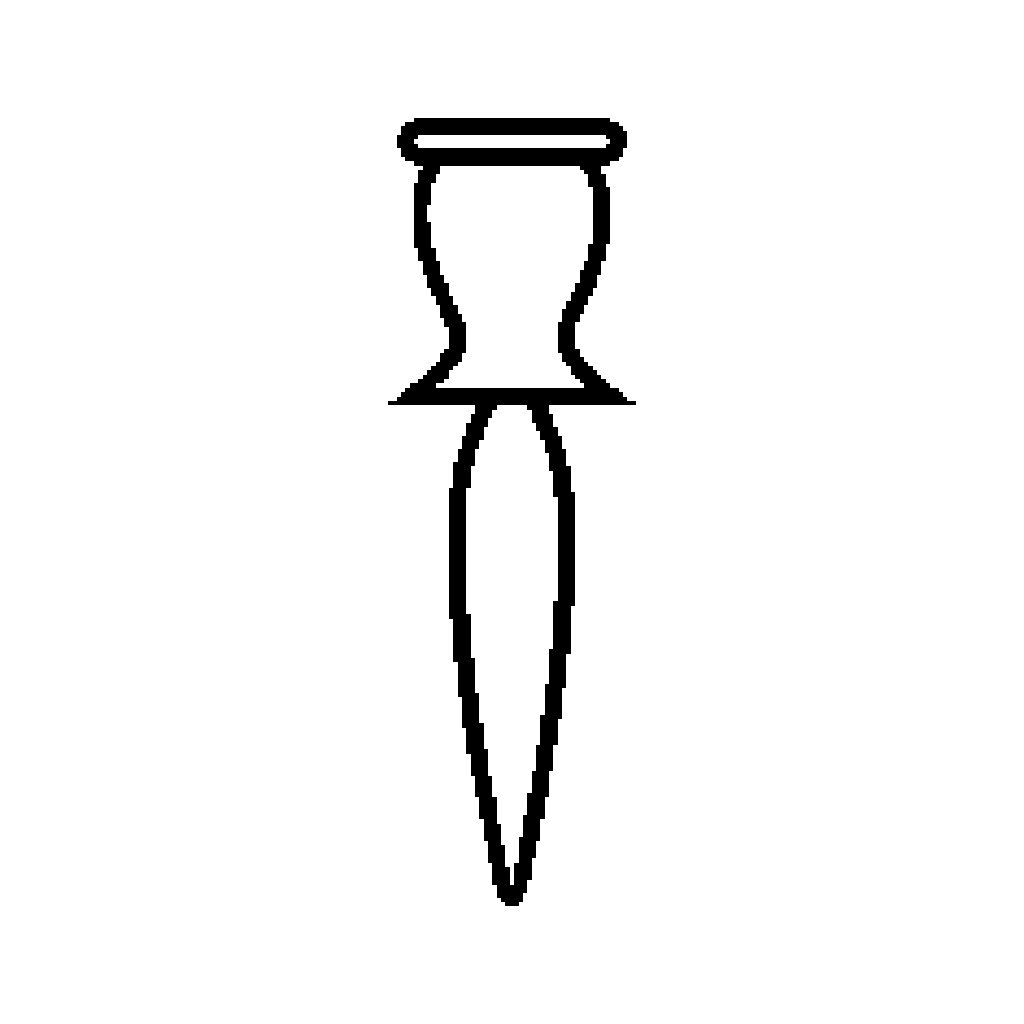} &
\imgwithbox[width=0.95\linewidth]{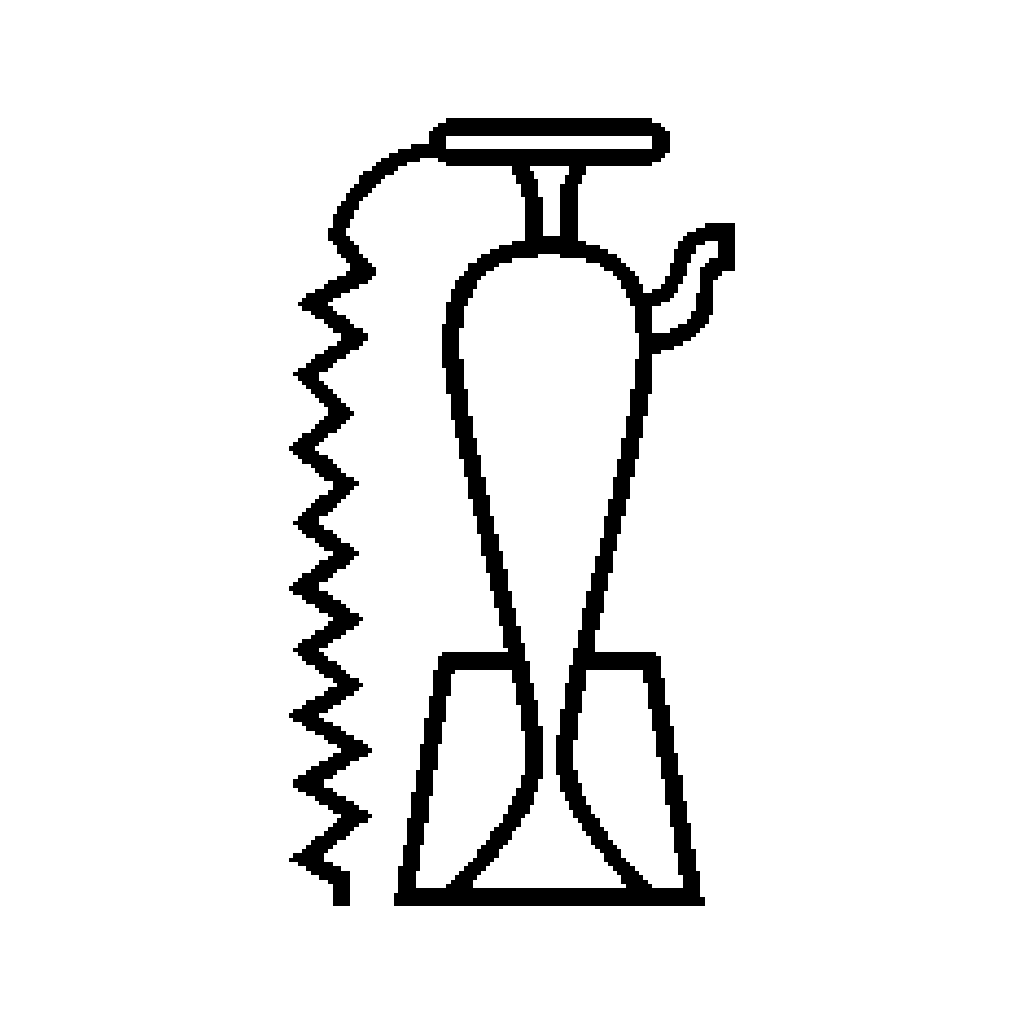} &
\imgwithbox[width=0.95\linewidth]{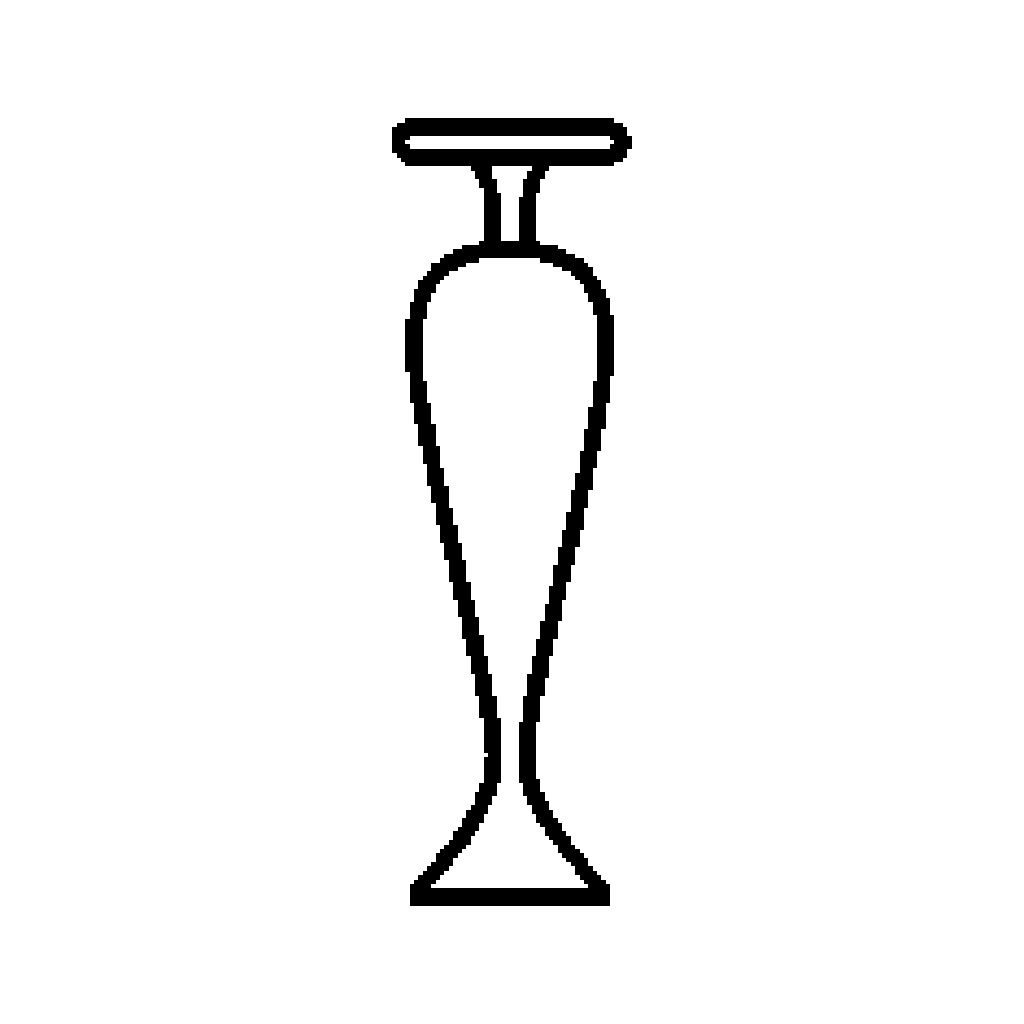} &
\imgwithbox[width=0.95\linewidth]{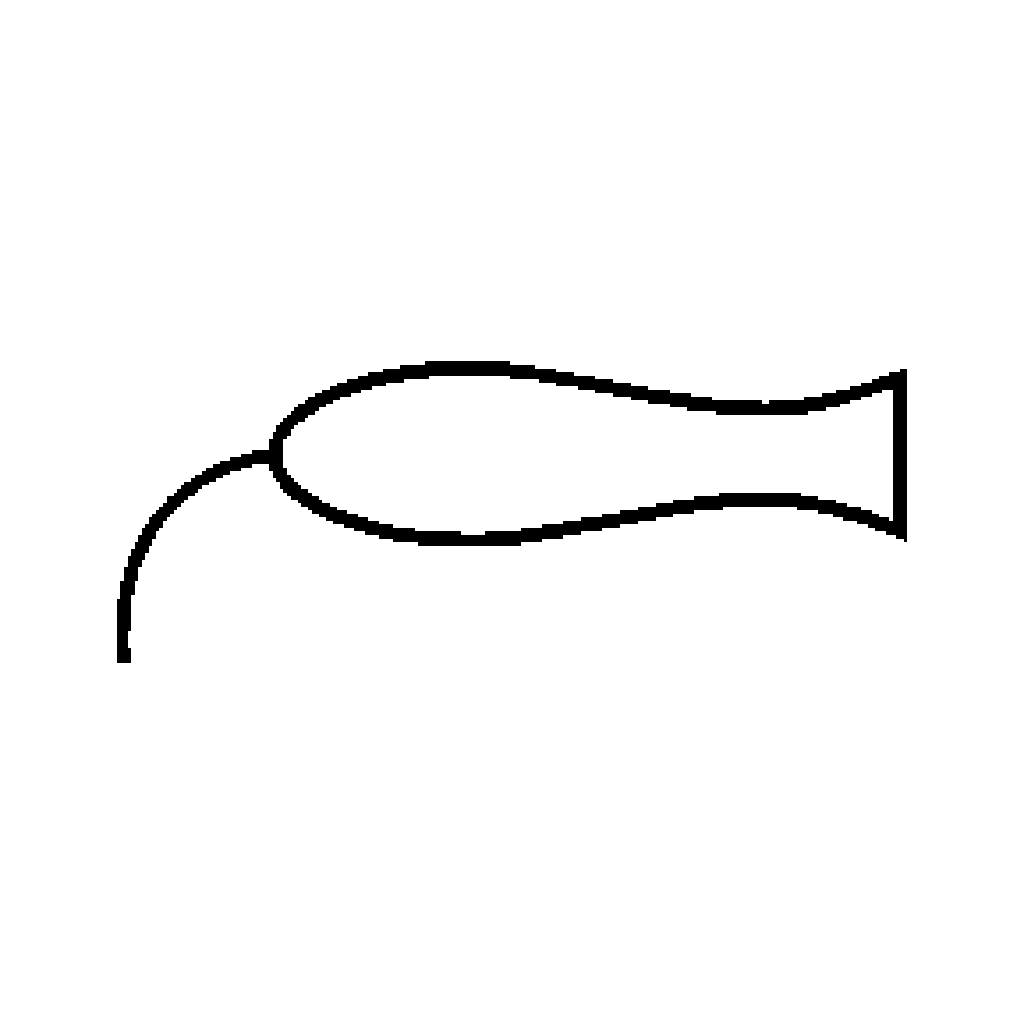} &
\imgwithbox[width=0.95\linewidth]{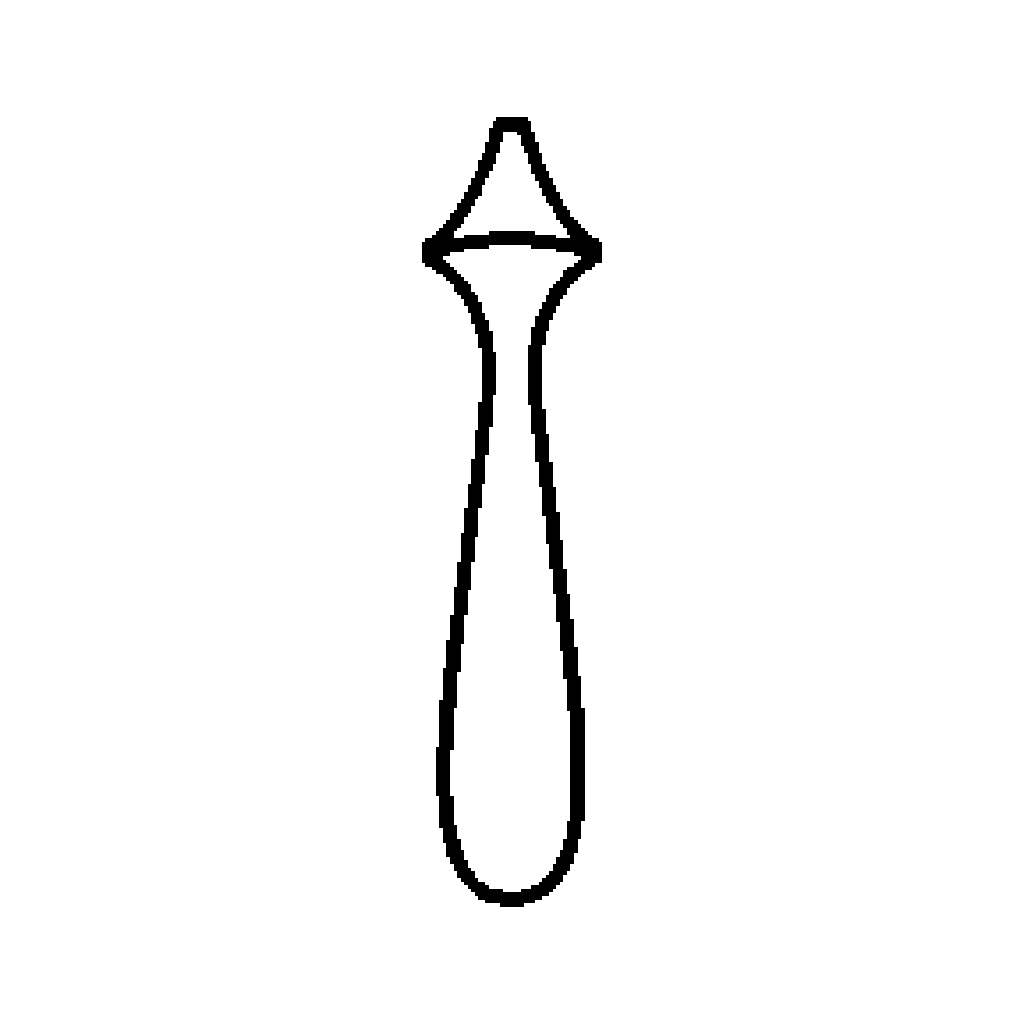} &
\imgwithbox[width=0.95\linewidth]{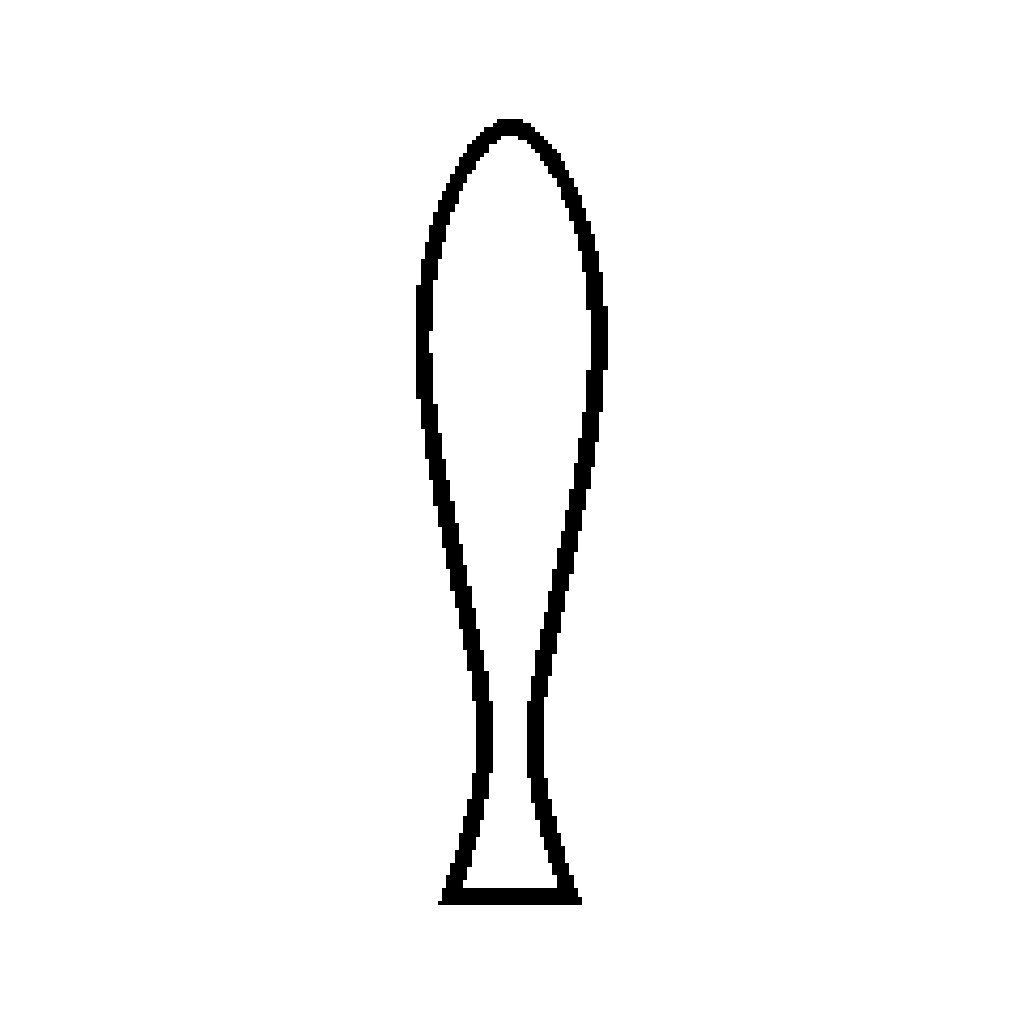} \\
8 &
\includegraphics[width=0.95\linewidth]{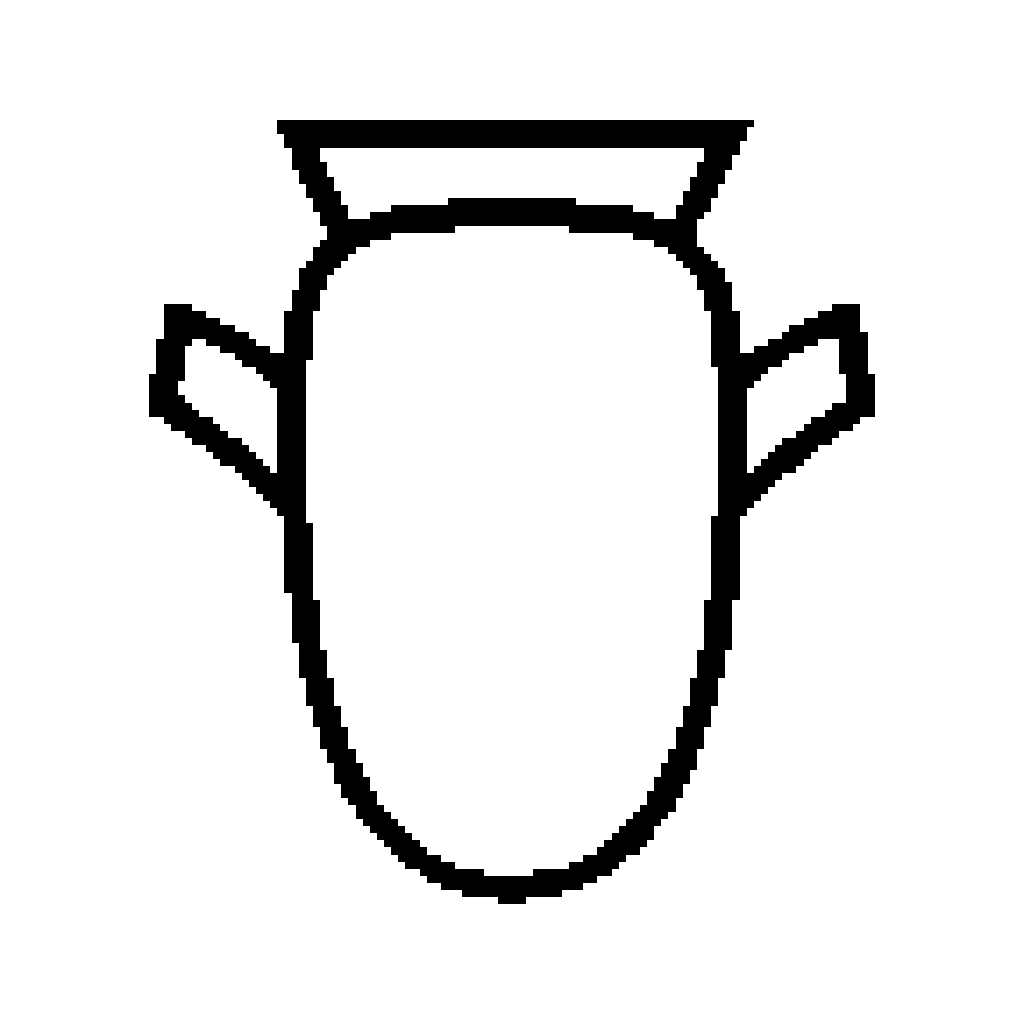} &
\includegraphics[width=0.95\linewidth]{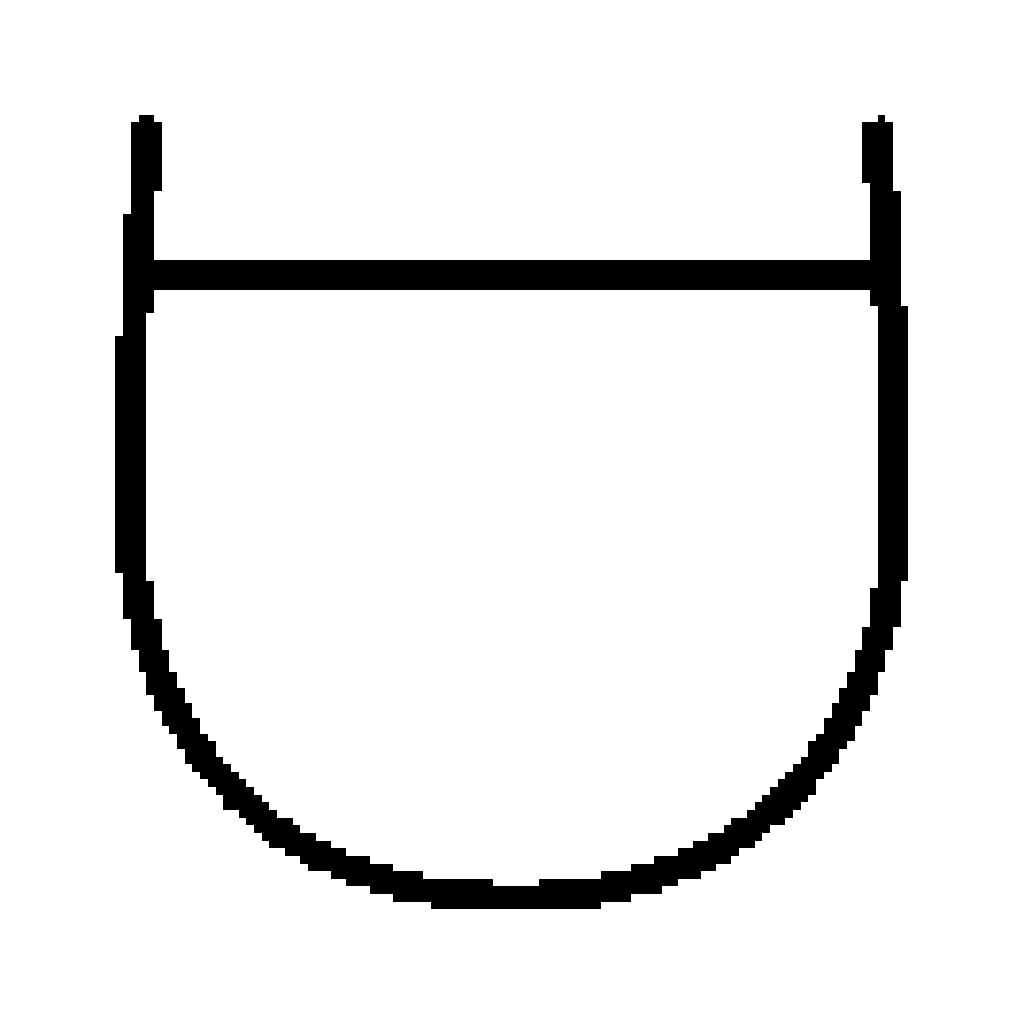} &
\includegraphics[width=0.95\linewidth]{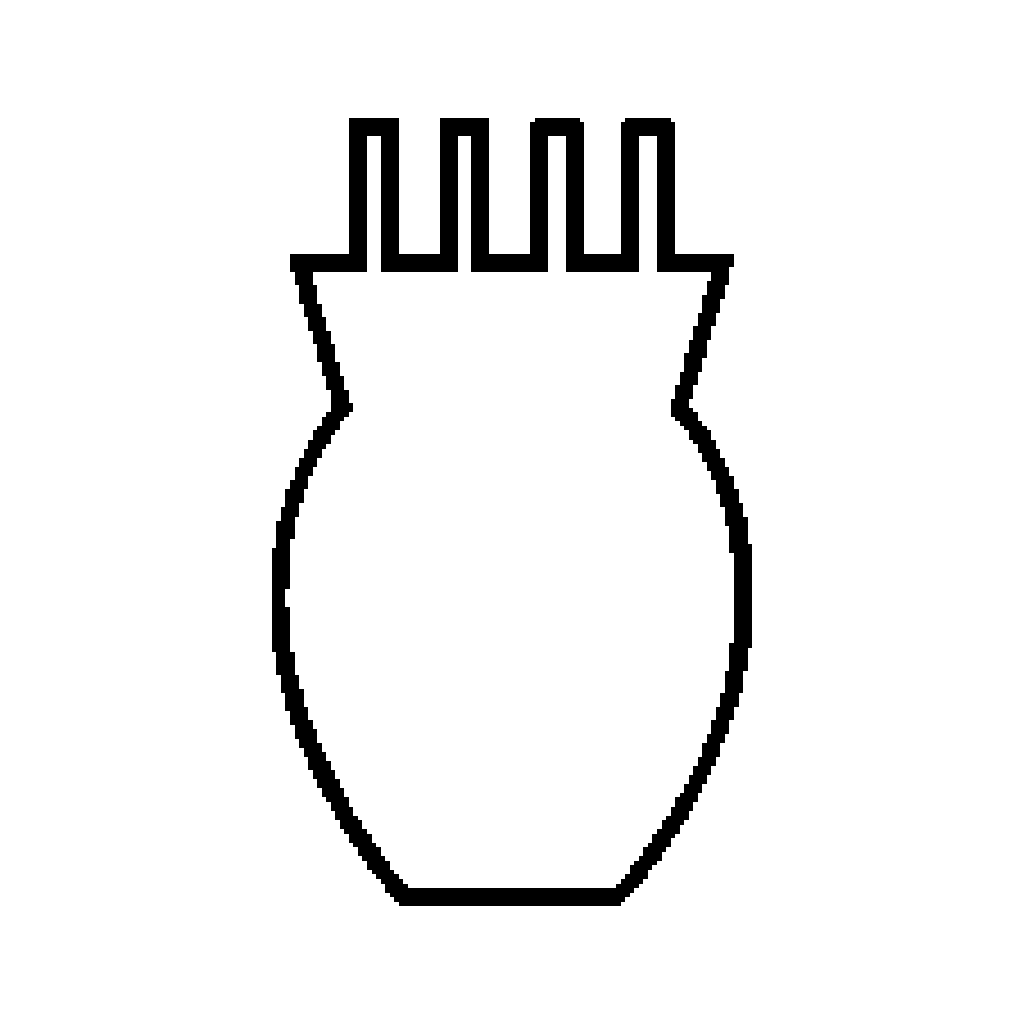} &
\includegraphics[width=0.95\linewidth]{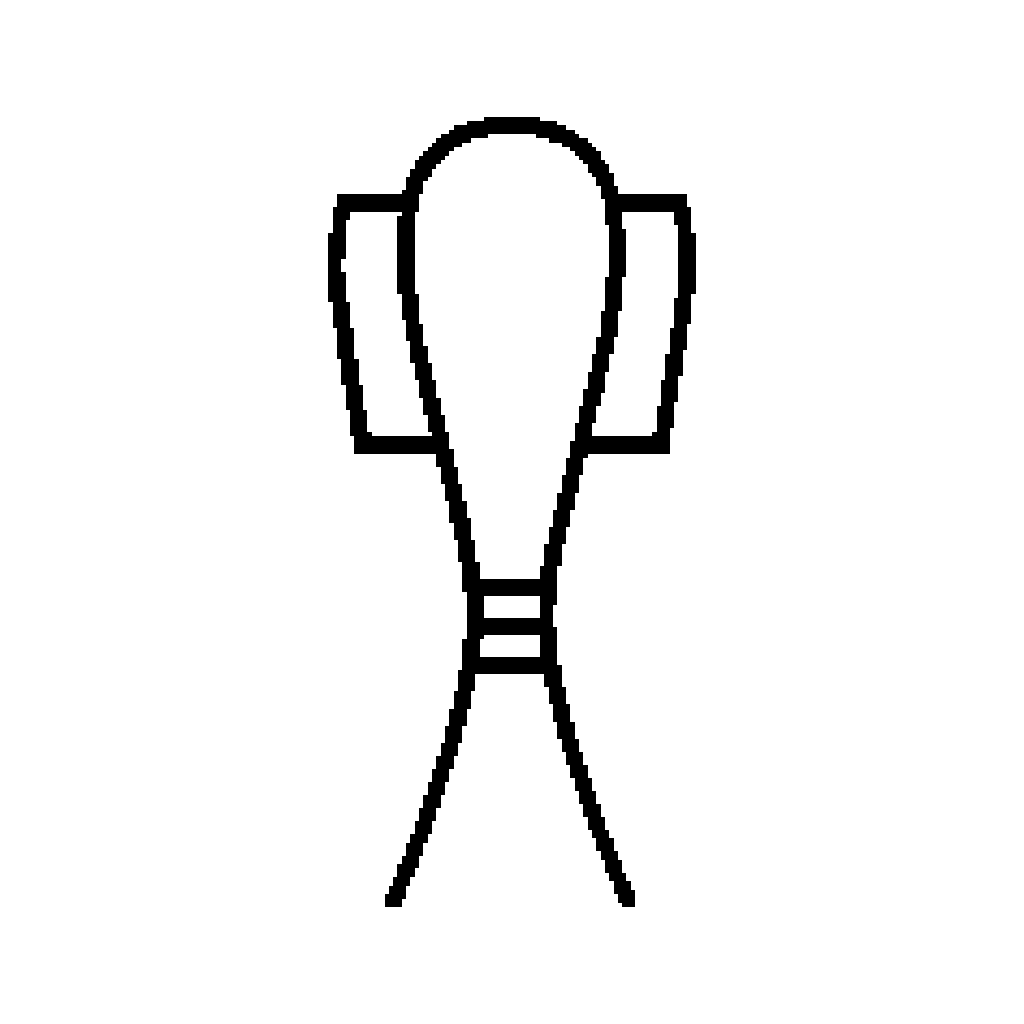} &
\includegraphics[width=0.95\linewidth]{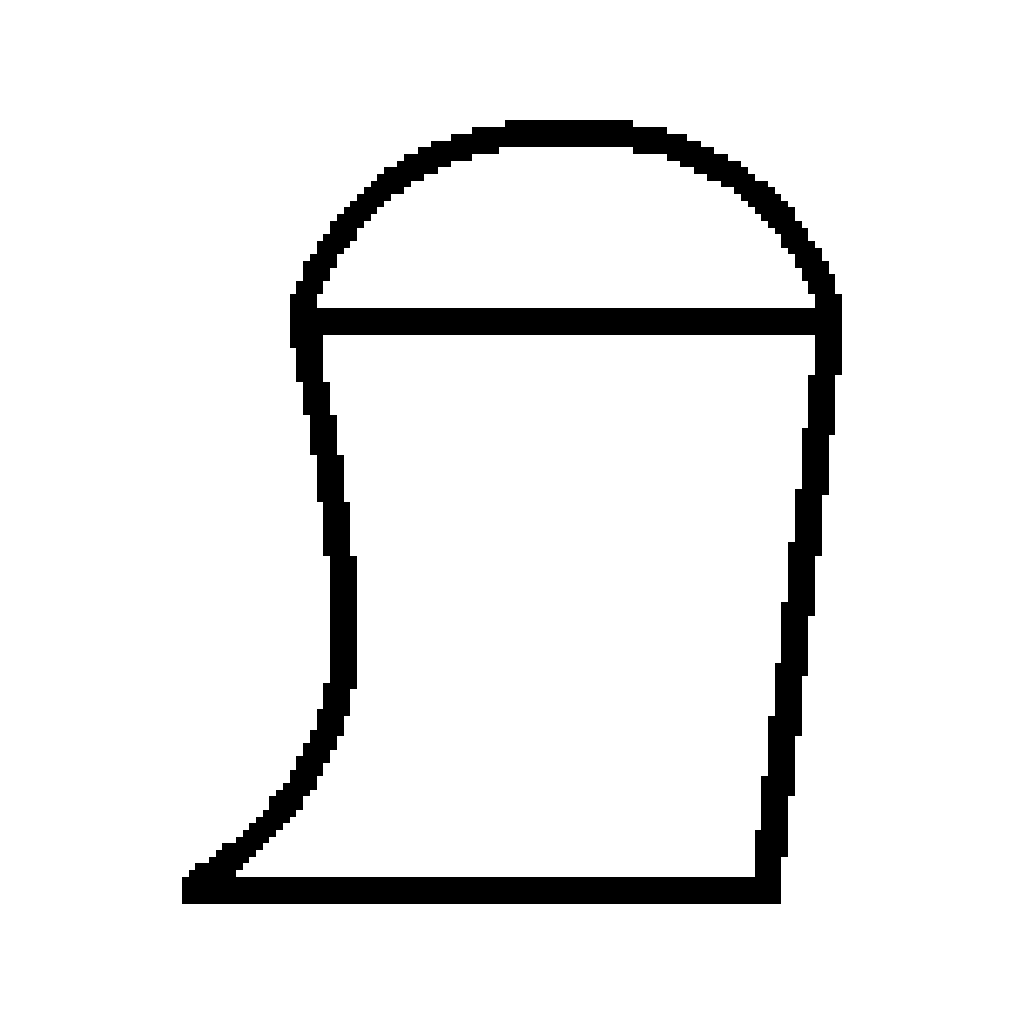} &
\includegraphics[width=0.95\linewidth]{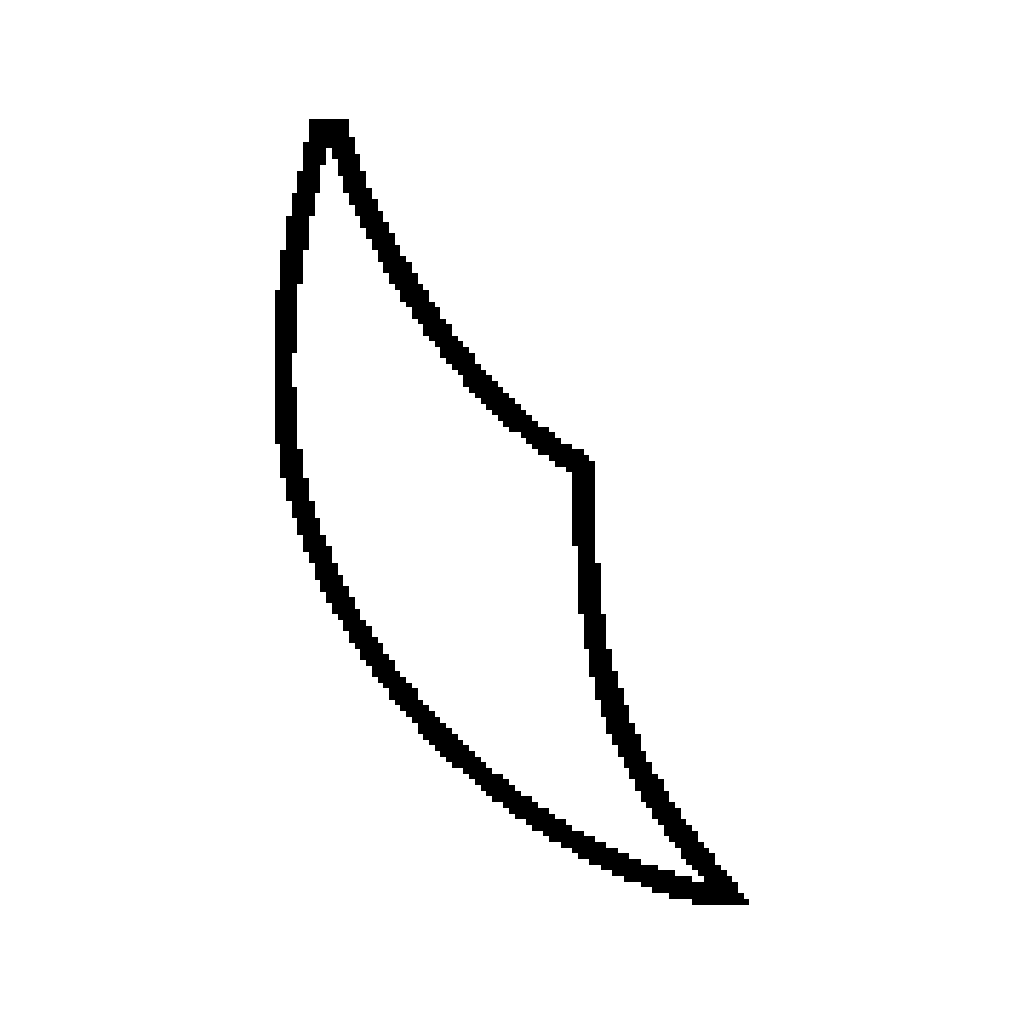} &
\imgwithbox[width=0.95\linewidth]{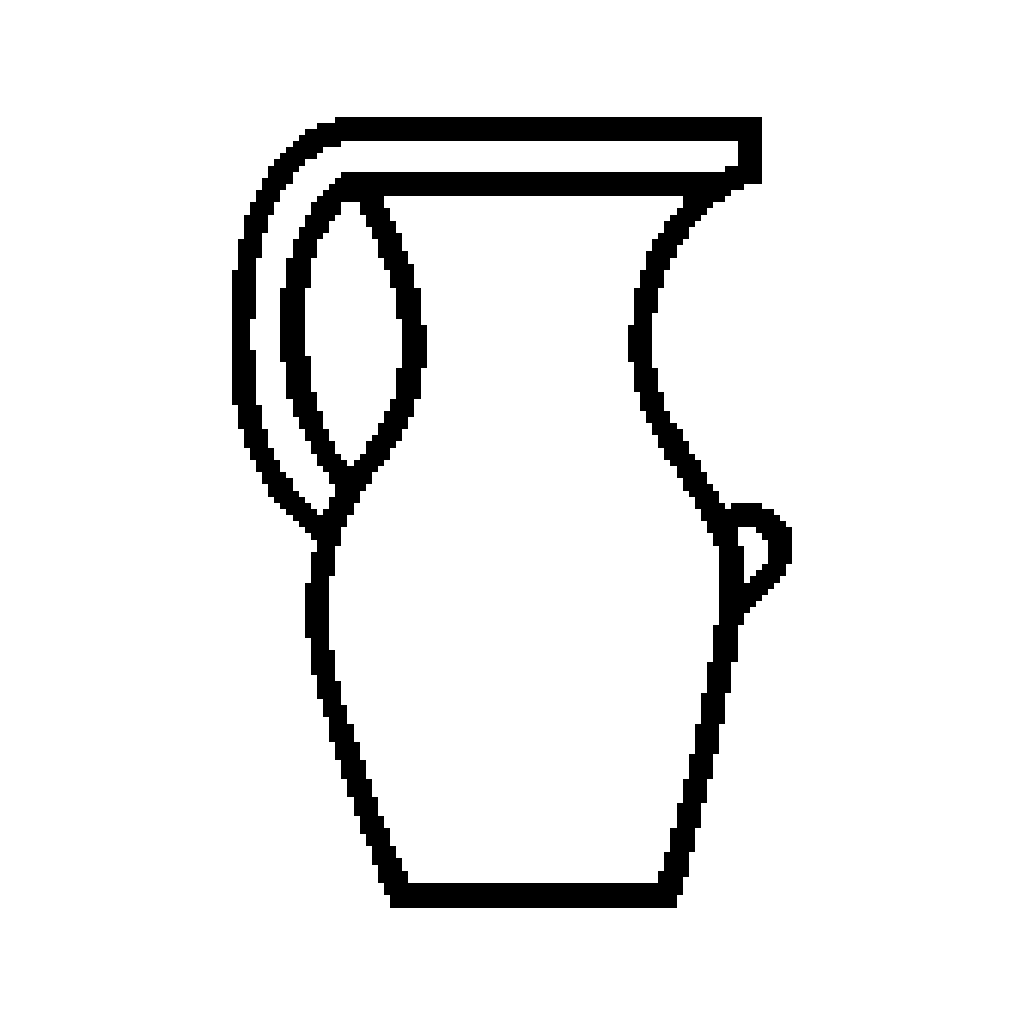} &
\imgwithbox[width=0.95\linewidth]{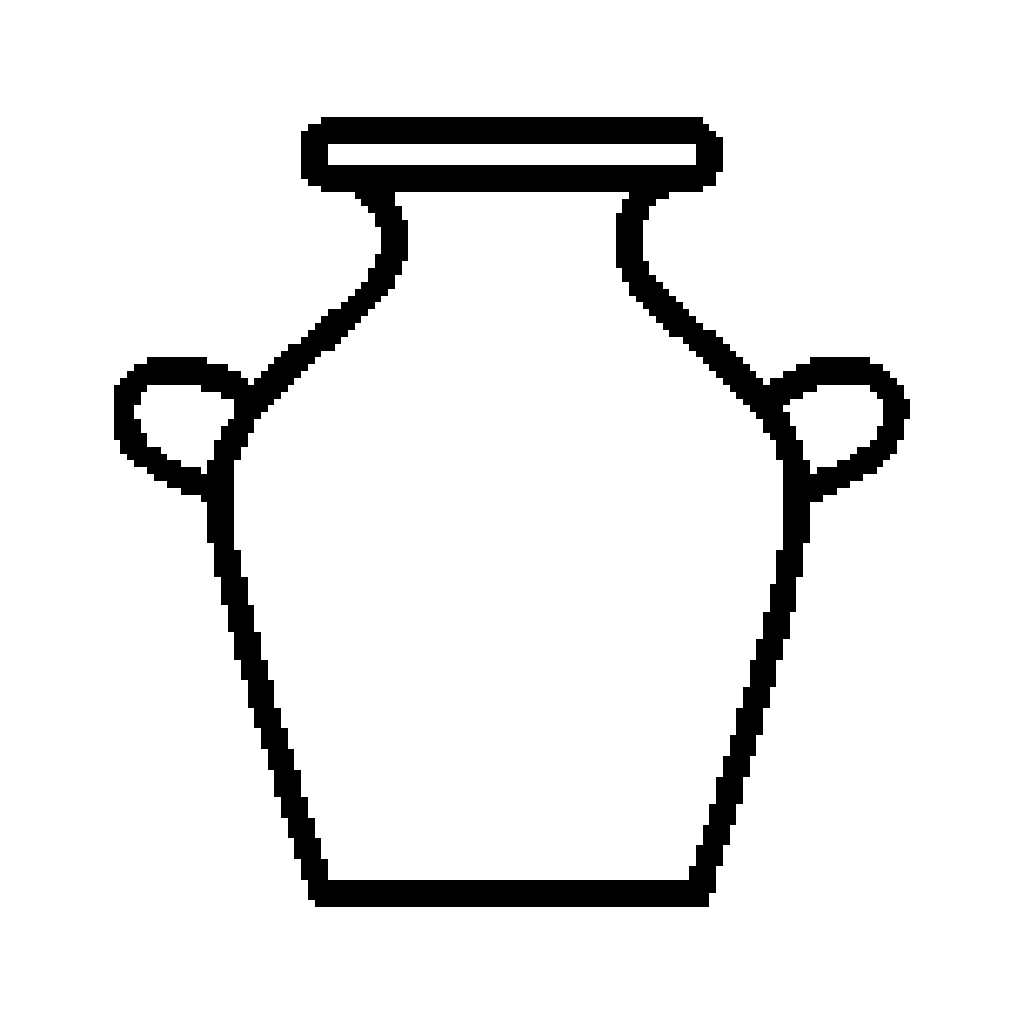} &
\imgwithbox[width=0.95\linewidth]{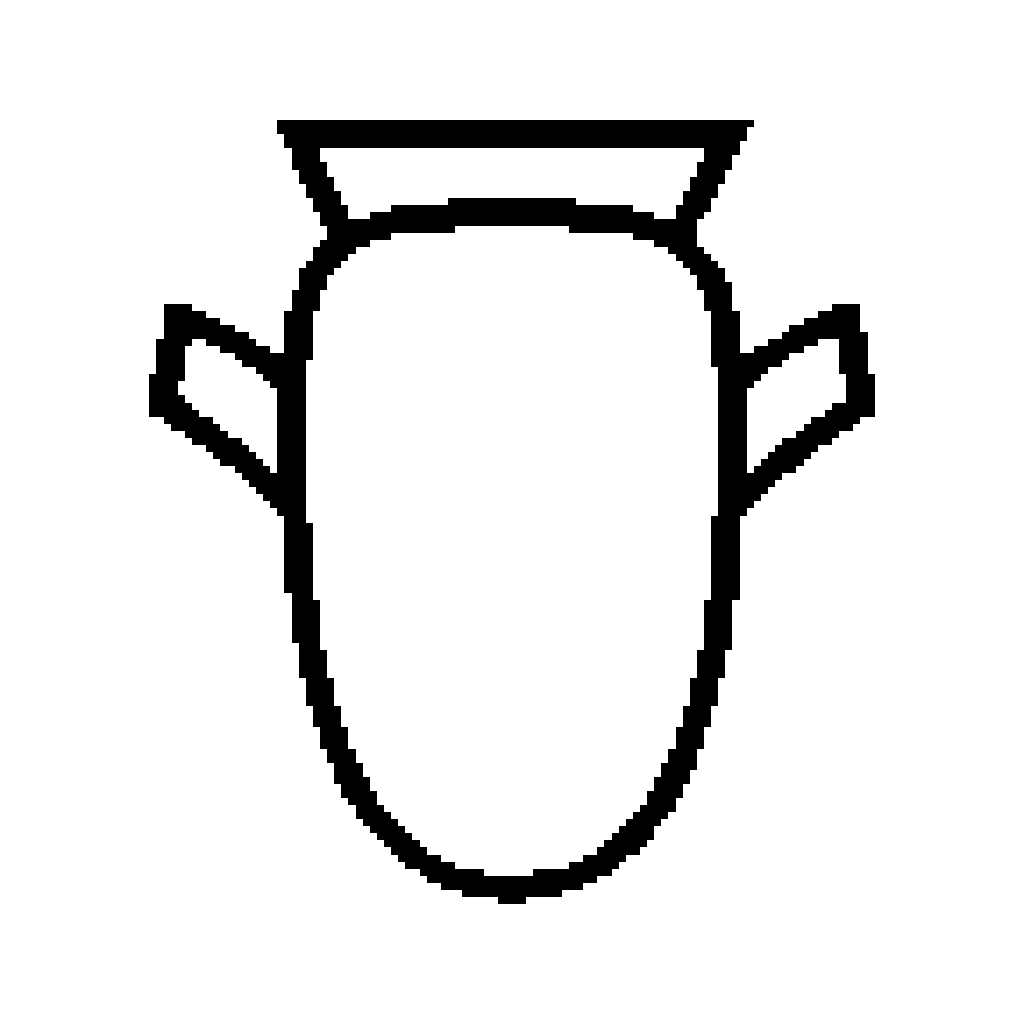} &
\imgwithbox[width=0.95\linewidth]{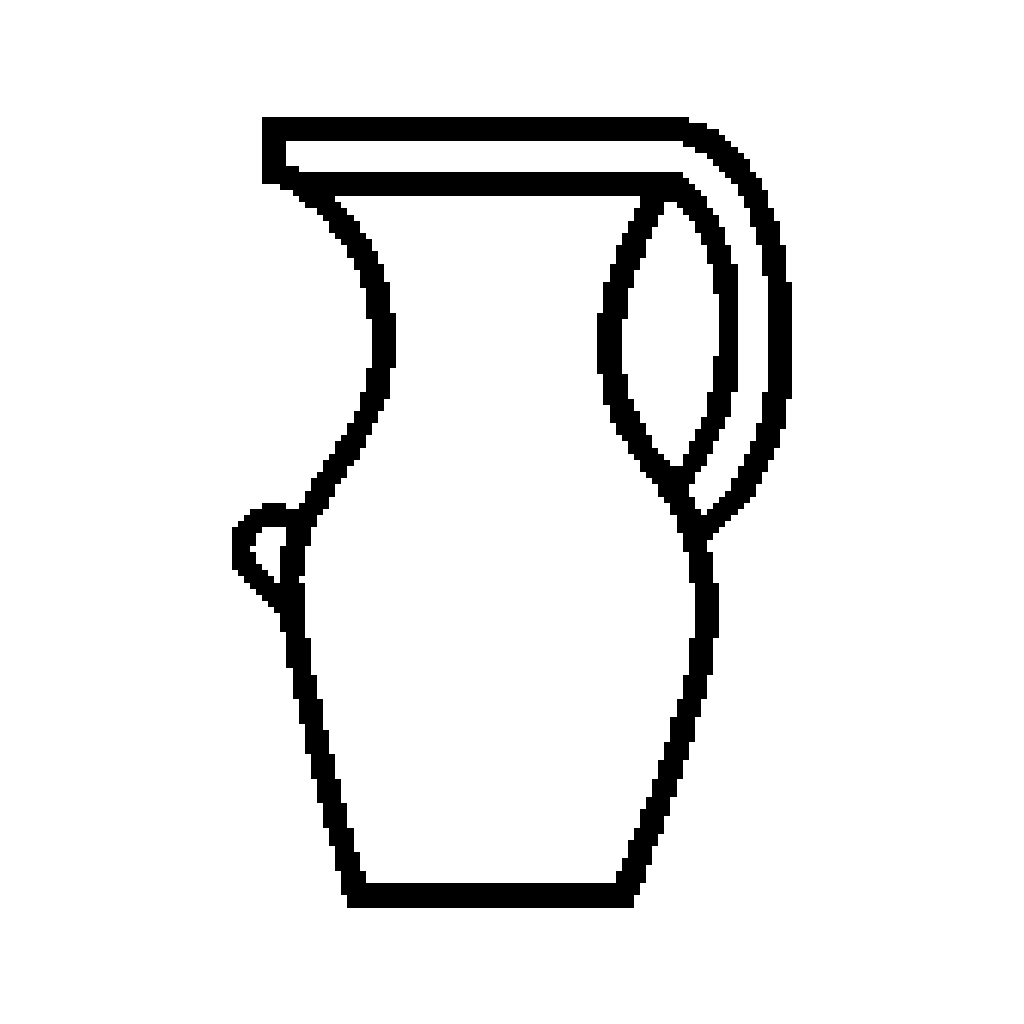} &
\includegraphics[width=0.95\linewidth]{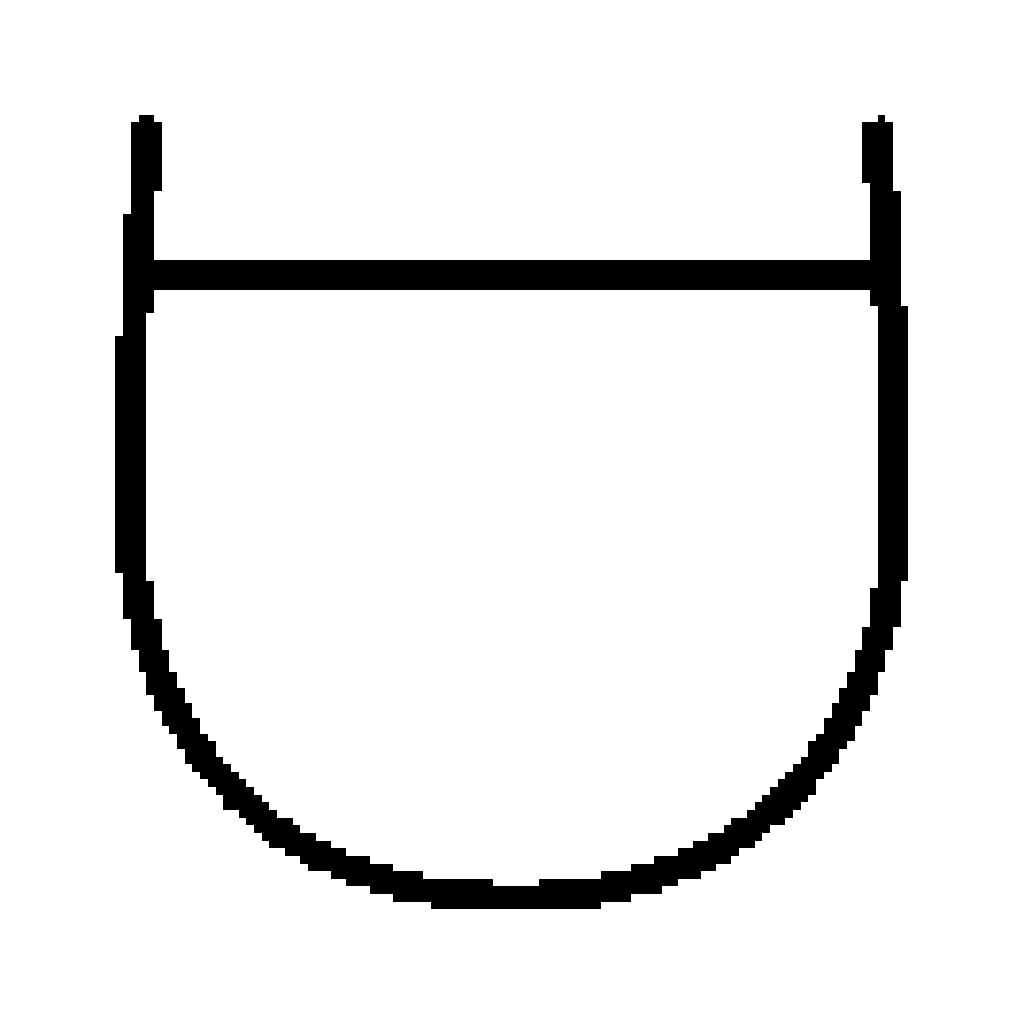} \\
9 &
\includegraphics[width=0.95\linewidth]{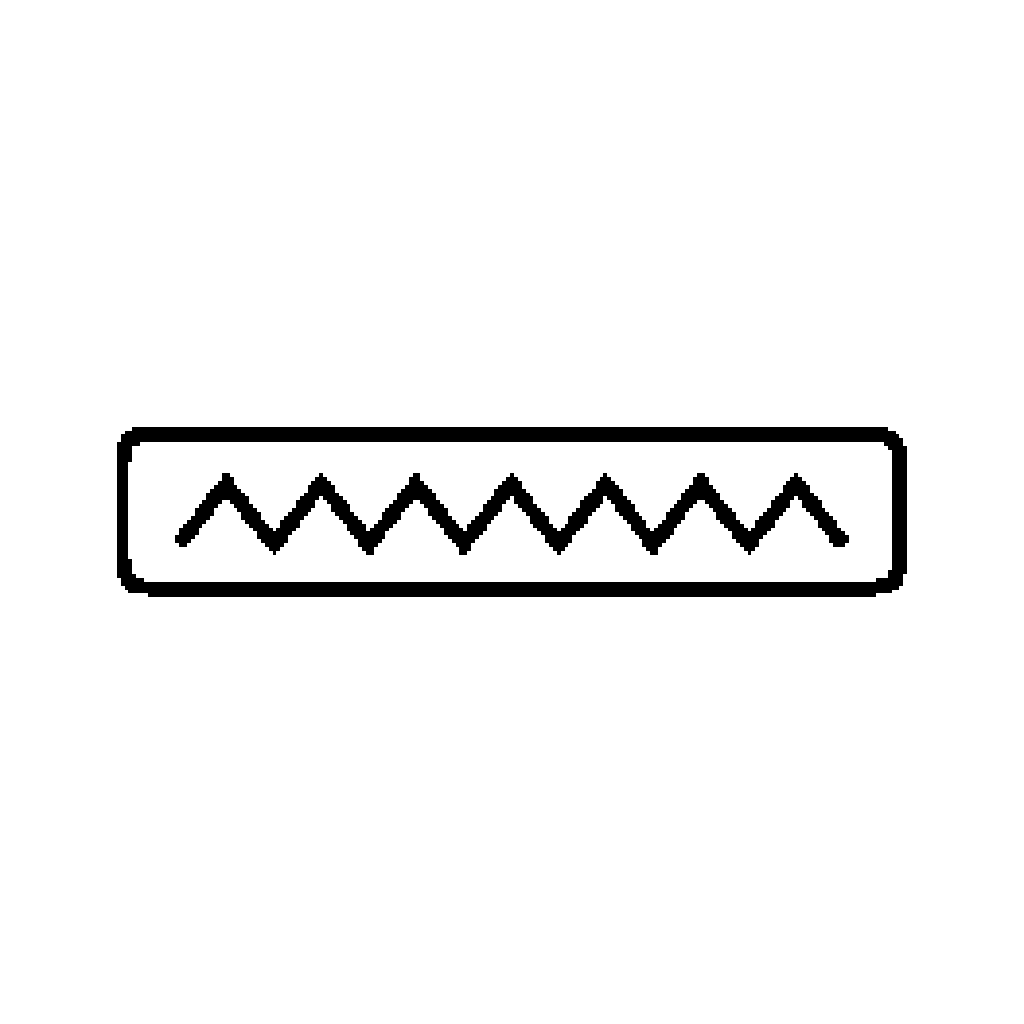} &
\includegraphics[width=0.95\linewidth]{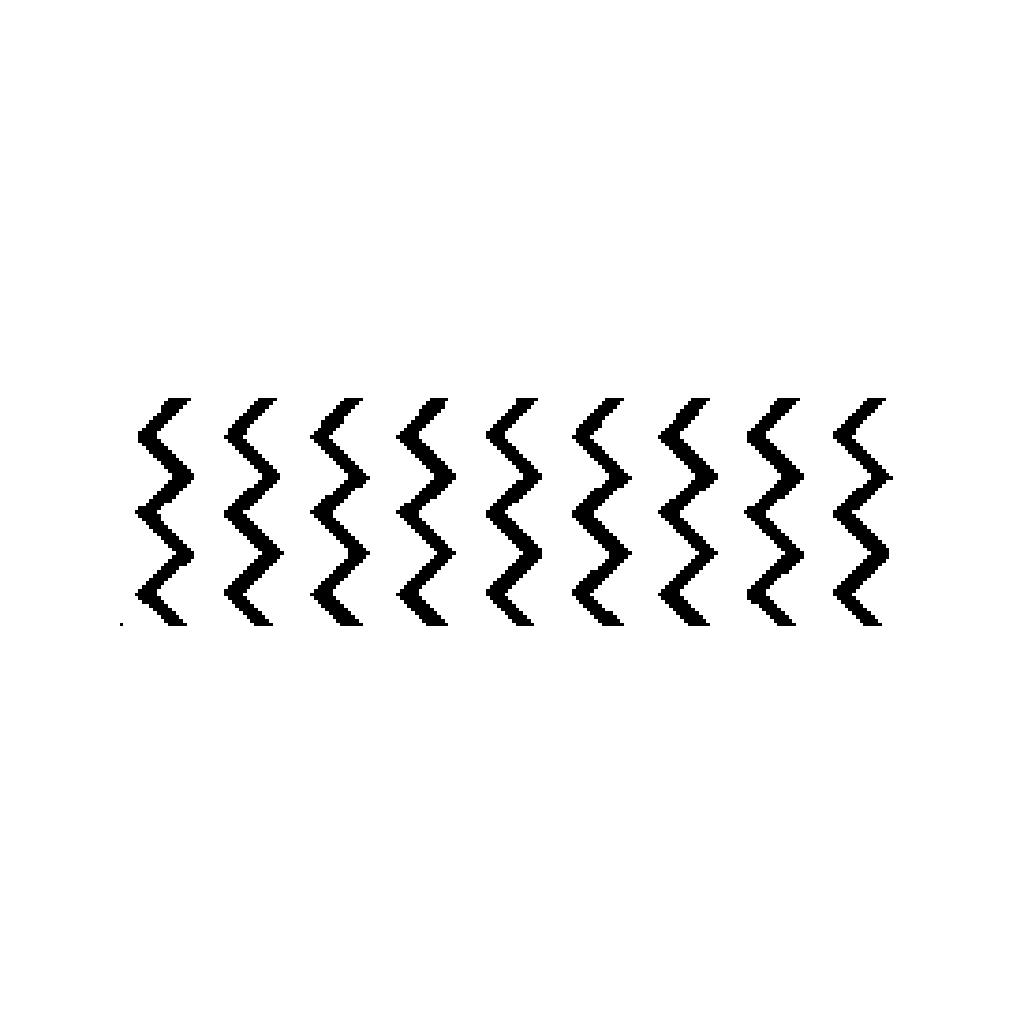} &
\includegraphics[width=0.95\linewidth]{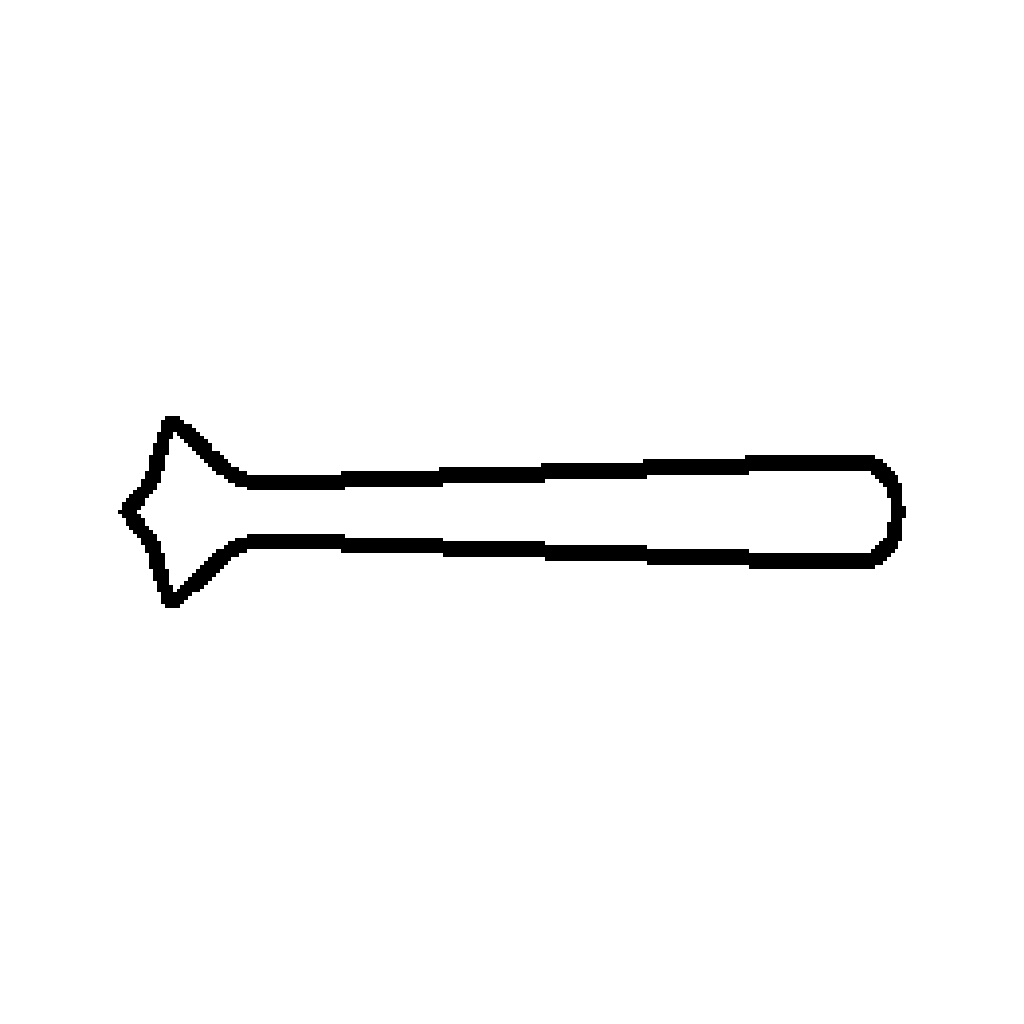} &
\includegraphics[width=0.95\linewidth]{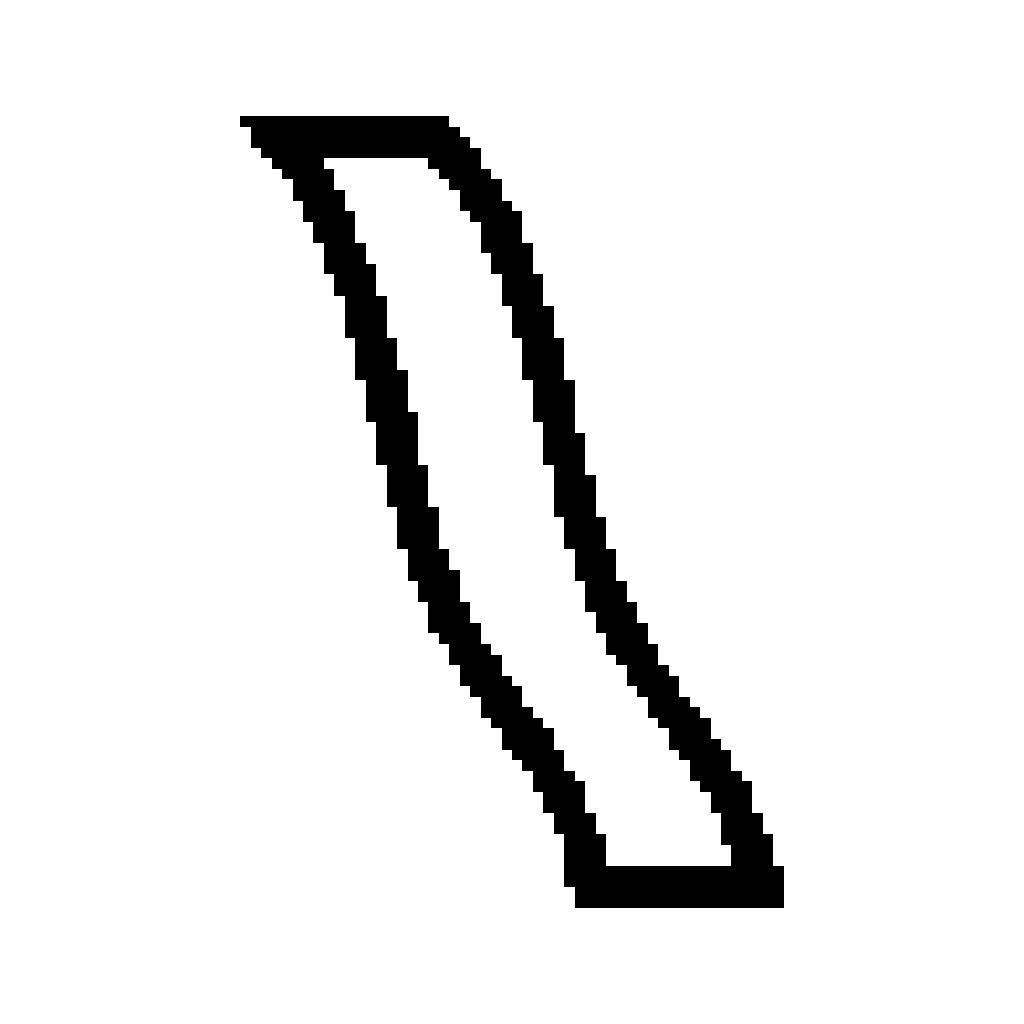} &
\includegraphics[width=0.95\linewidth]{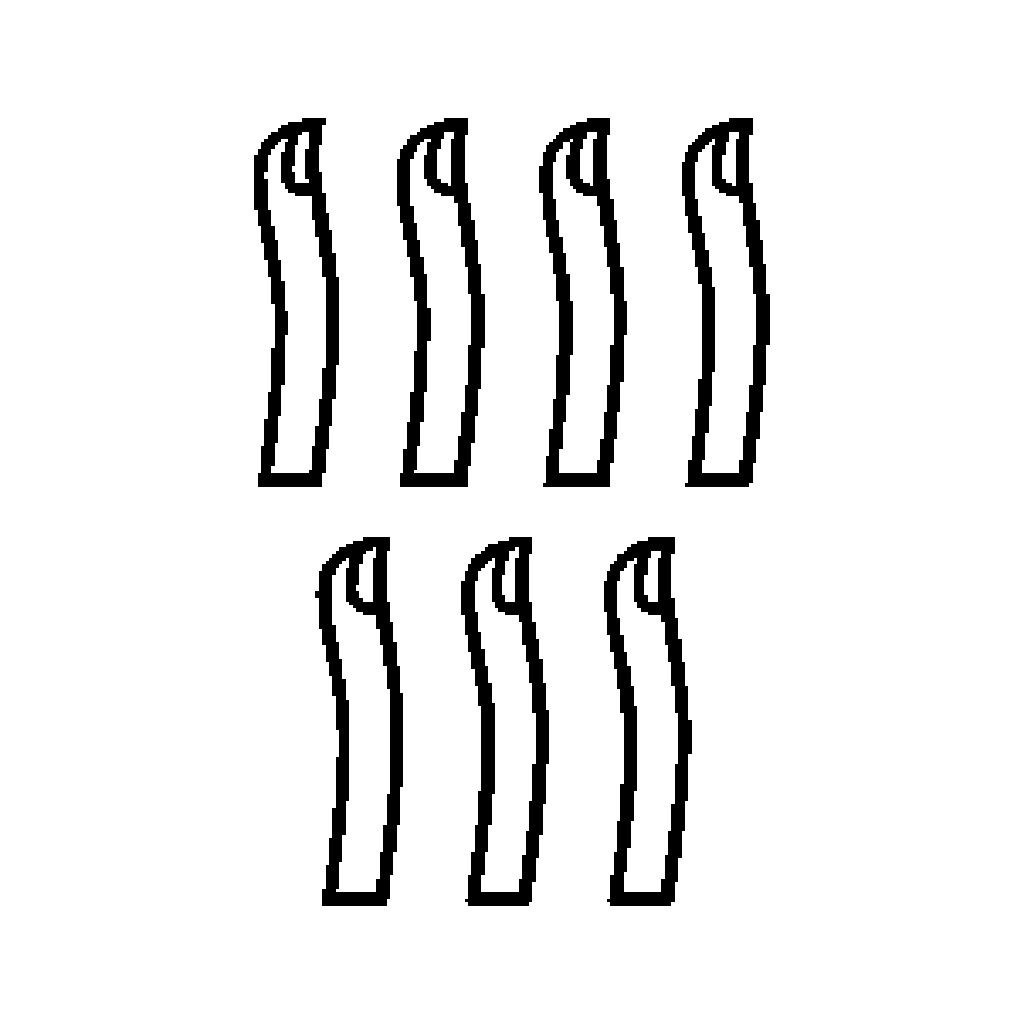} &
\includegraphics[width=0.95\linewidth]{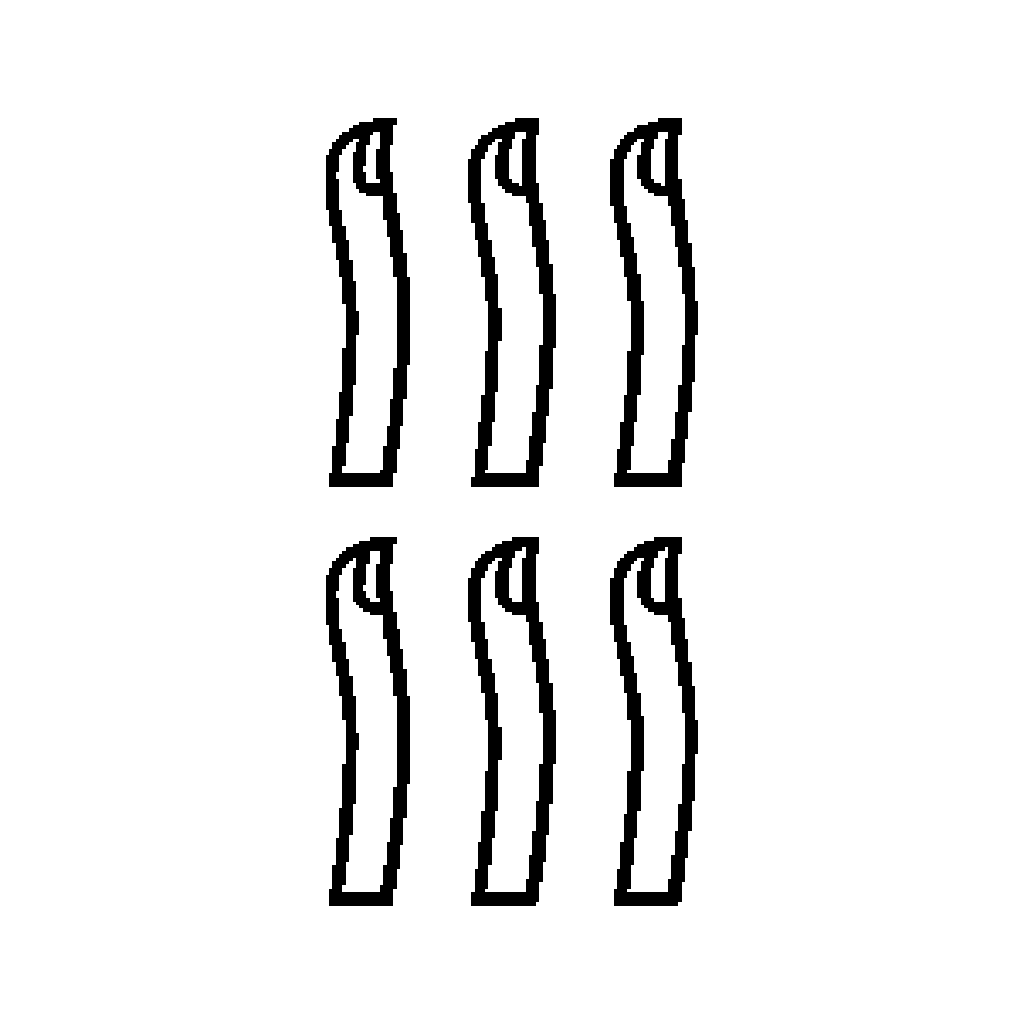} &
\includegraphics[width=0.95\linewidth]{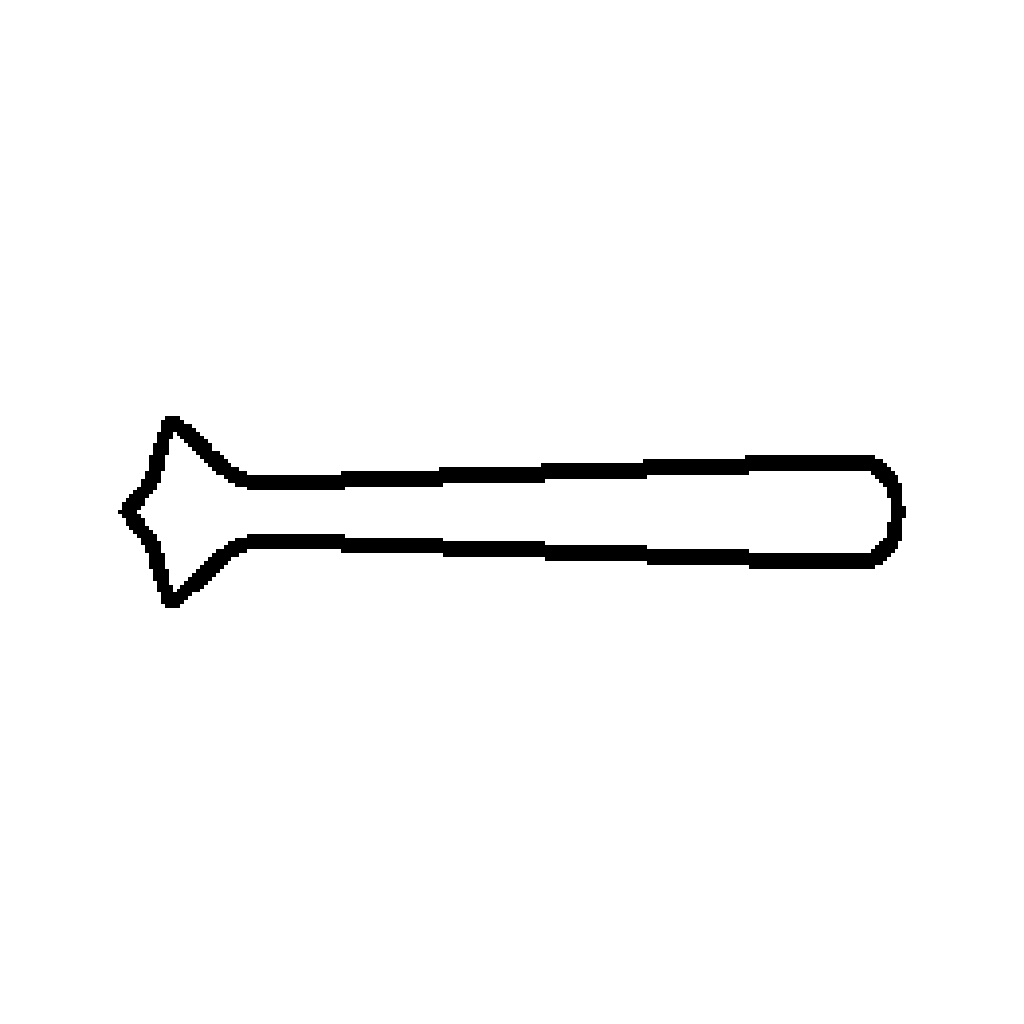} &
\includegraphics[width=0.95\linewidth]{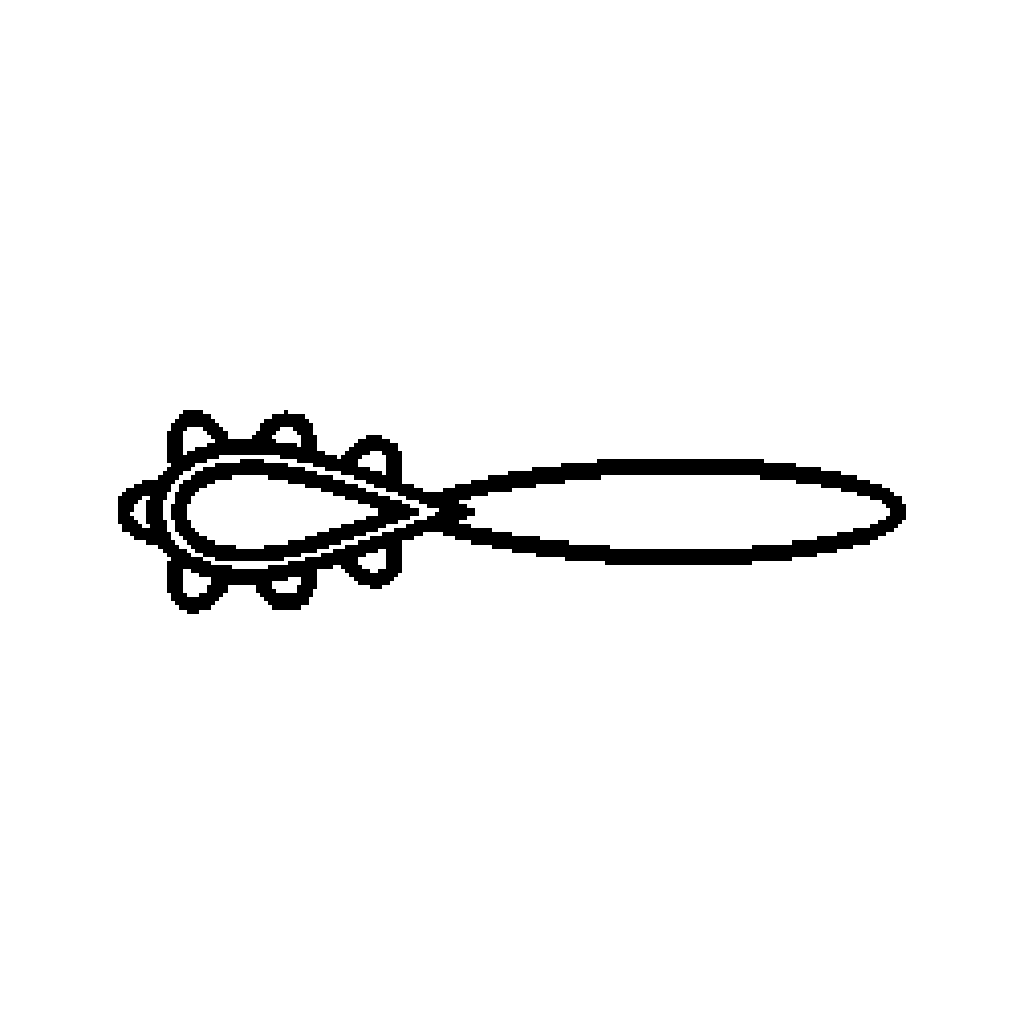} &
\includegraphics[width=0.95\linewidth]{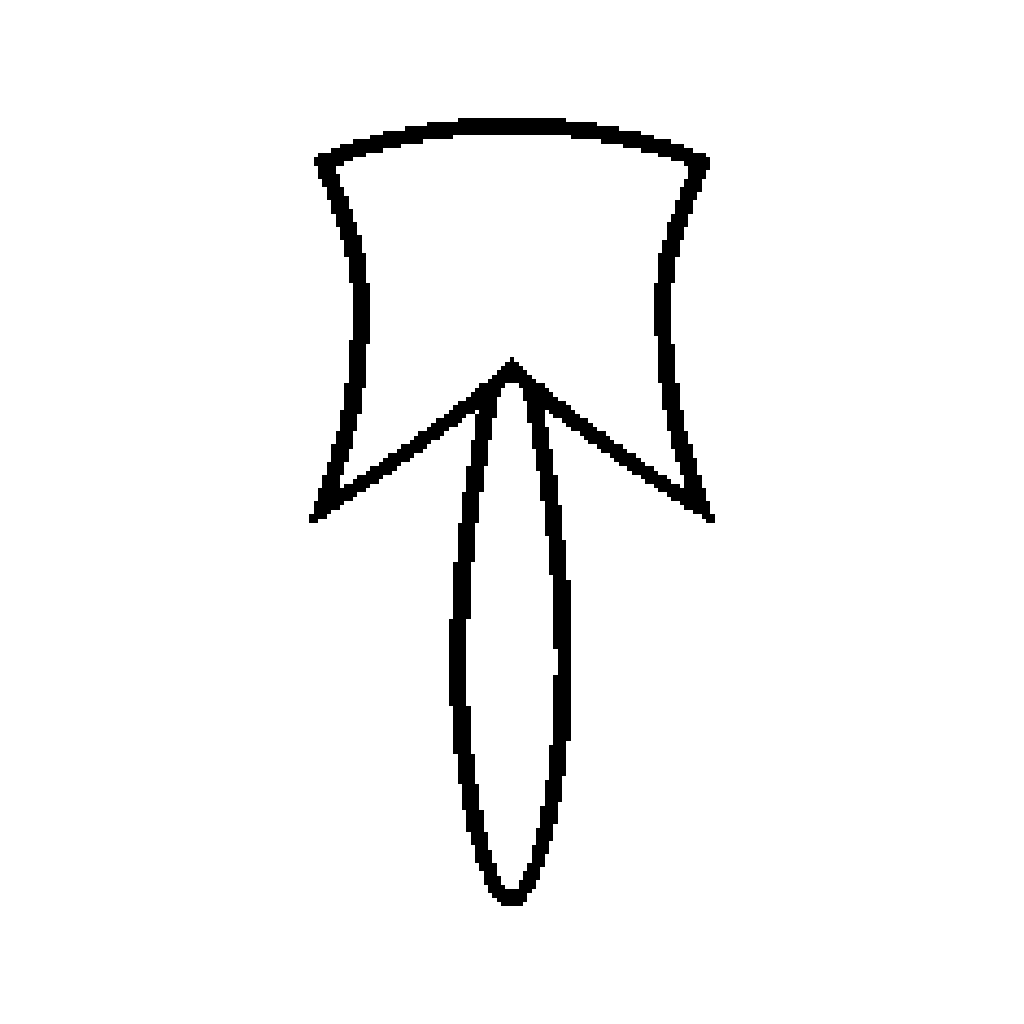} &
\imgwithbox[width=0.95\linewidth]{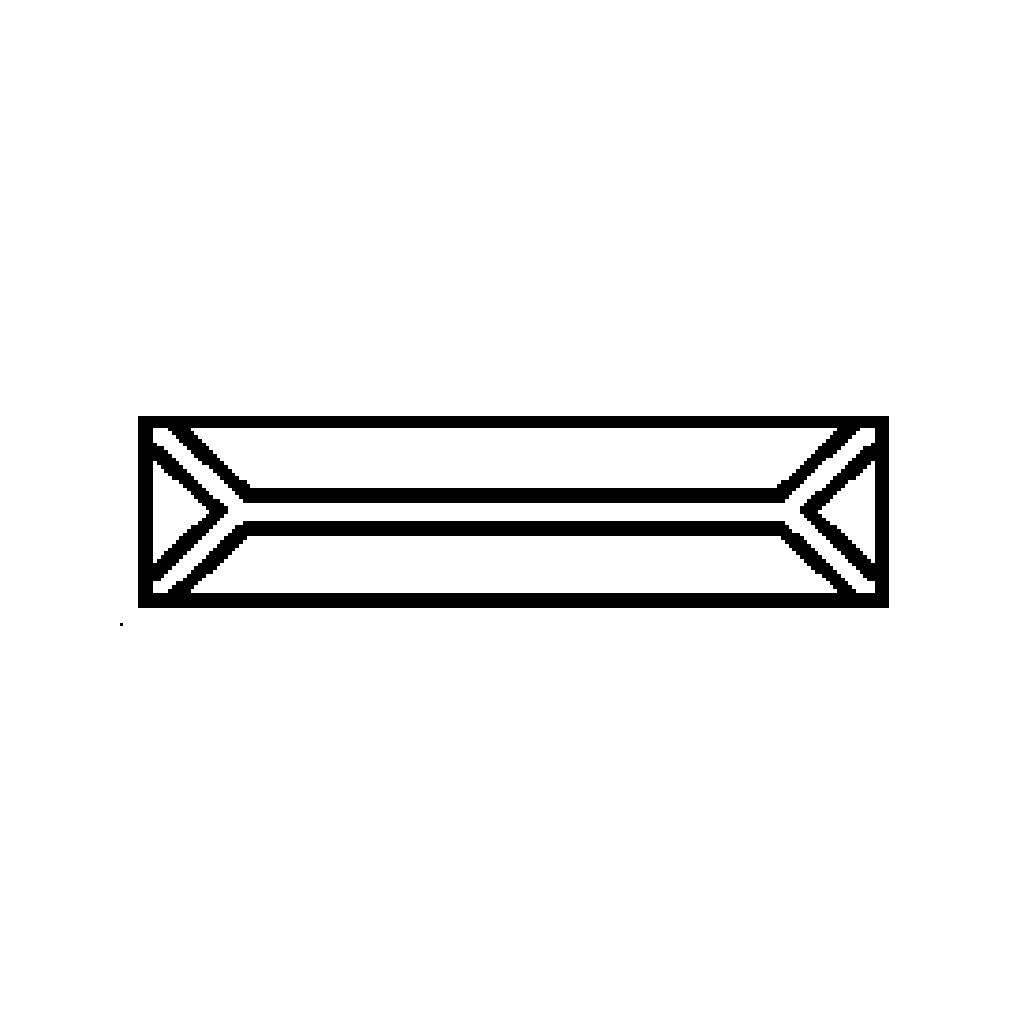} &
\includegraphics[width=0.95\linewidth]{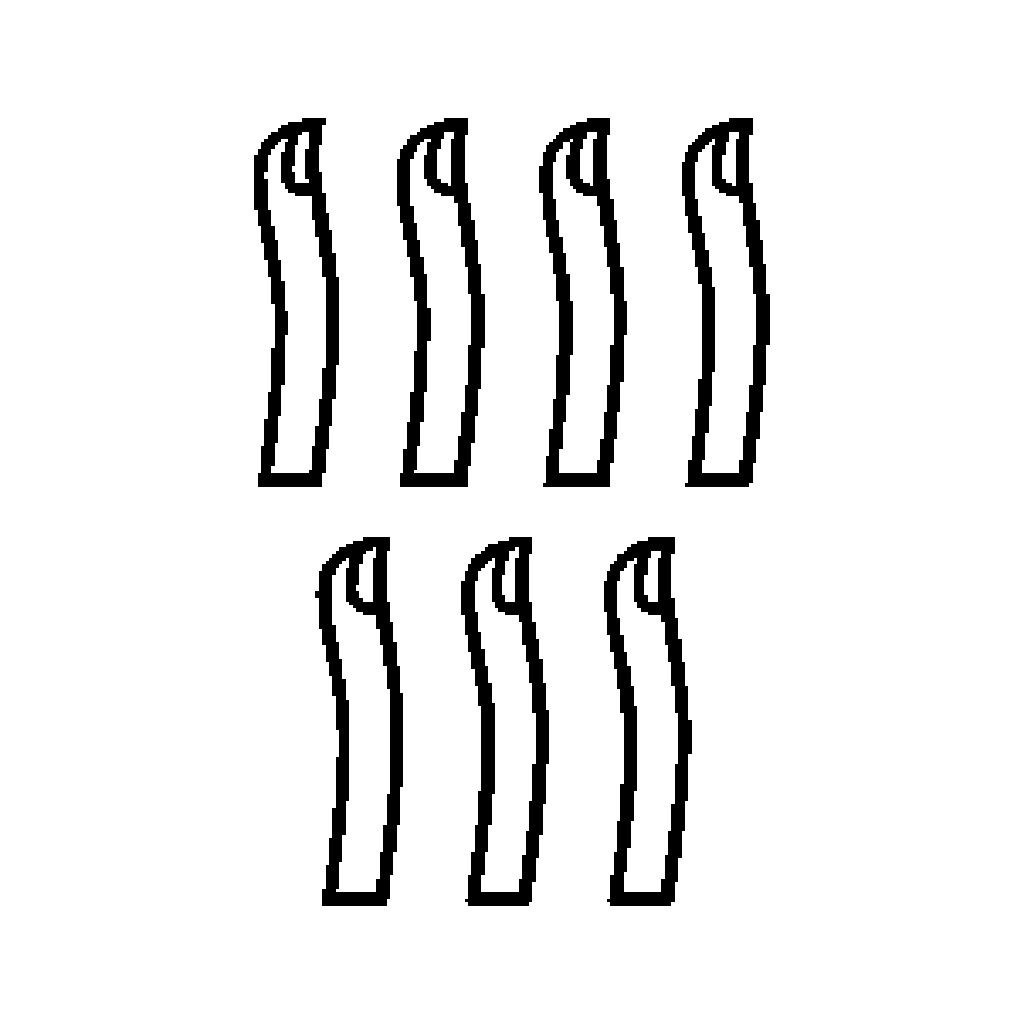} \\
\midrule\noalign{\vskip -3pt}\rowcolor{gray!20}\multicolumn{12}{c}{\textbf{Cuneiforms}} \\\noalign{\vskip -2pt}\midrule
10 &
\includegraphics[width=0.95\linewidth]{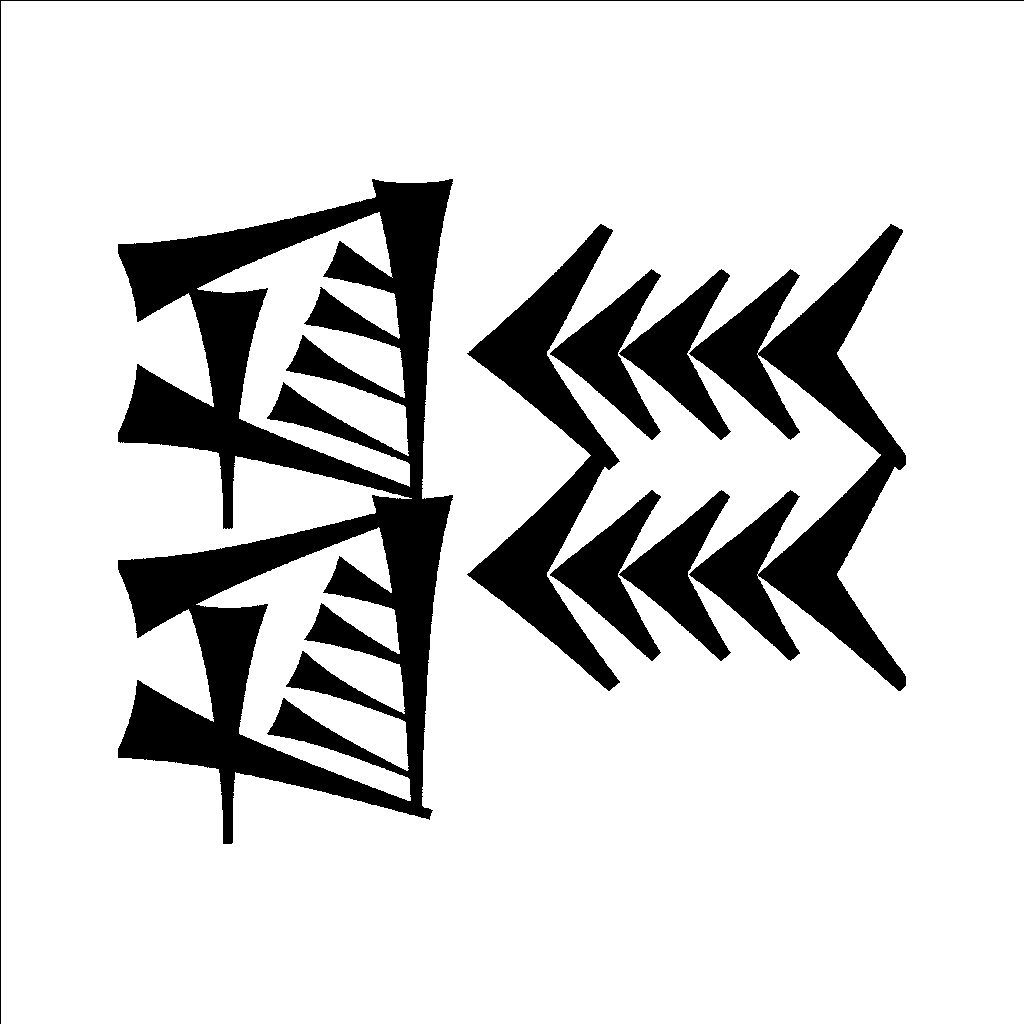} &
\imgwithbox[width=0.95\linewidth]{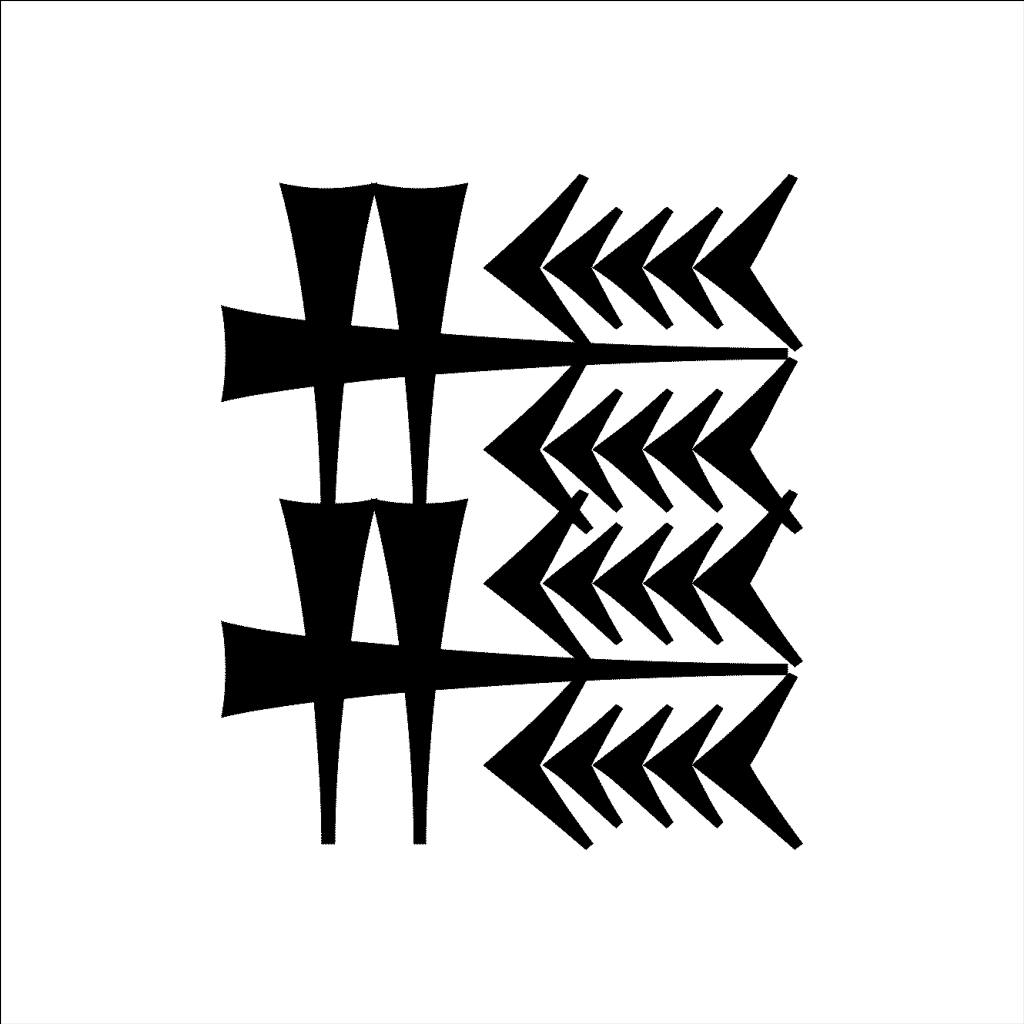} &
\includegraphics[width=0.95\linewidth]{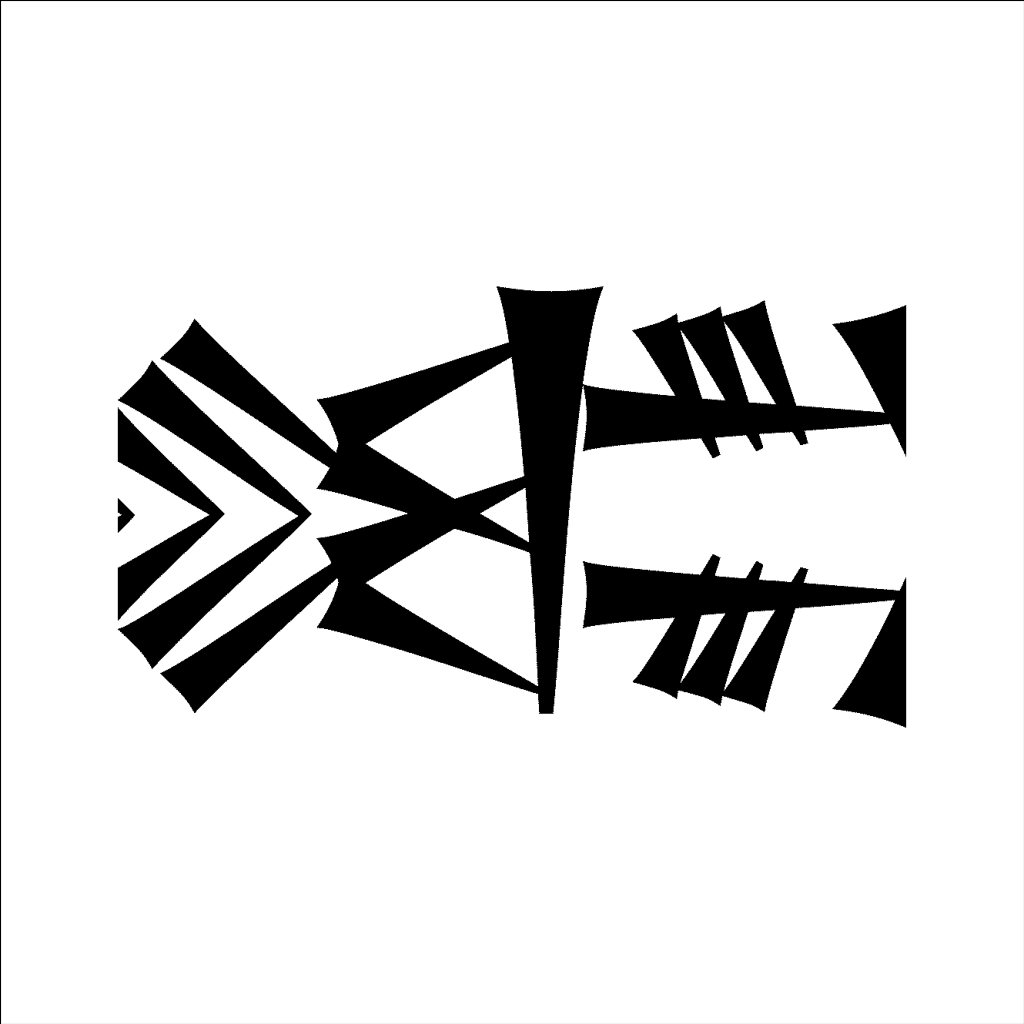} &
\includegraphics[width=0.95\linewidth]{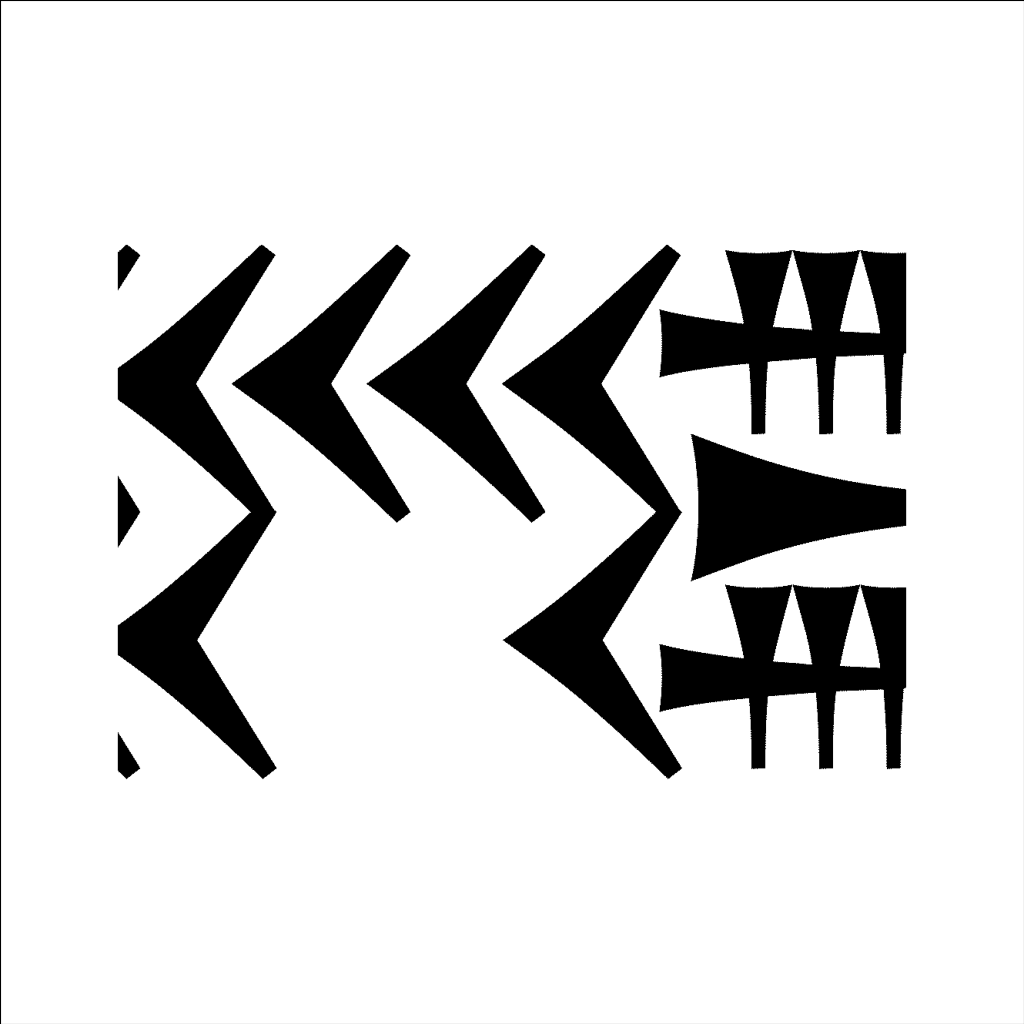} &
\includegraphics[width=0.95\linewidth]{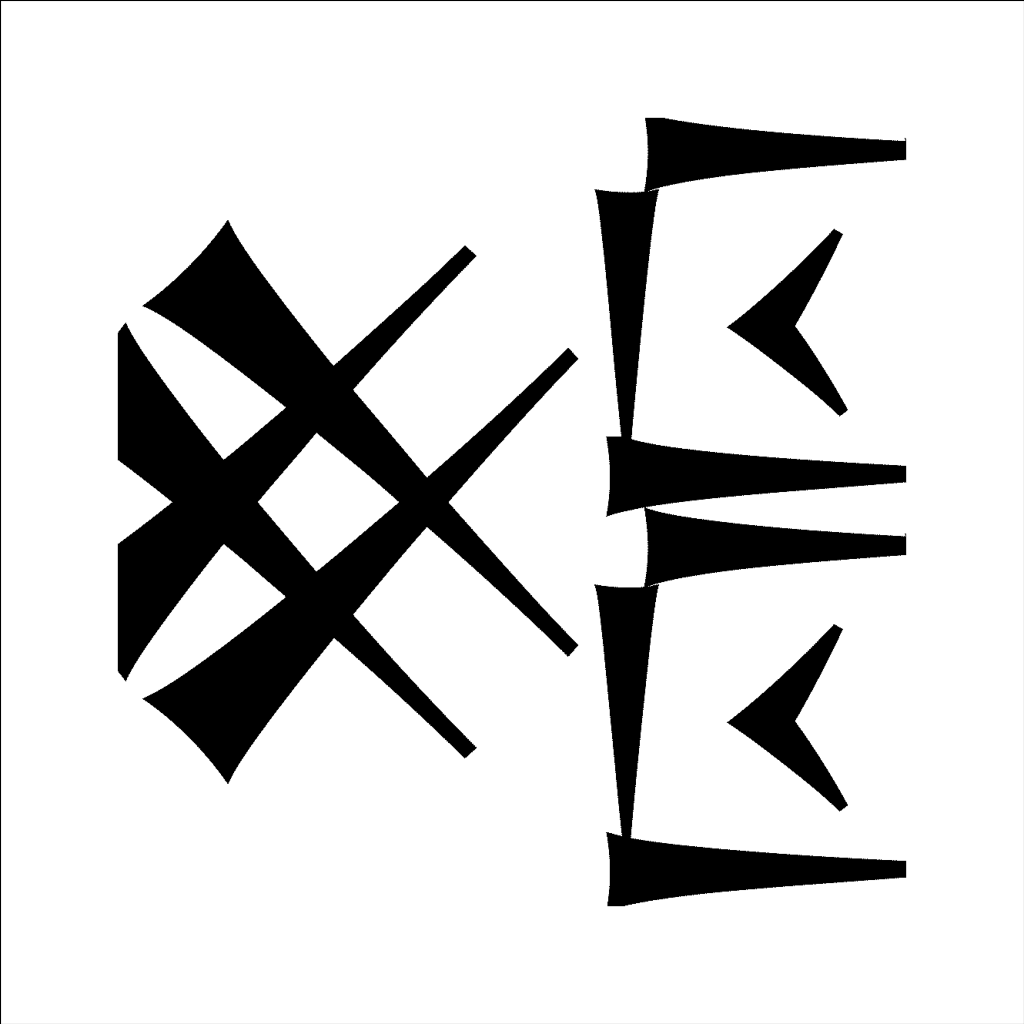} &
\includegraphics[width=0.95\linewidth]{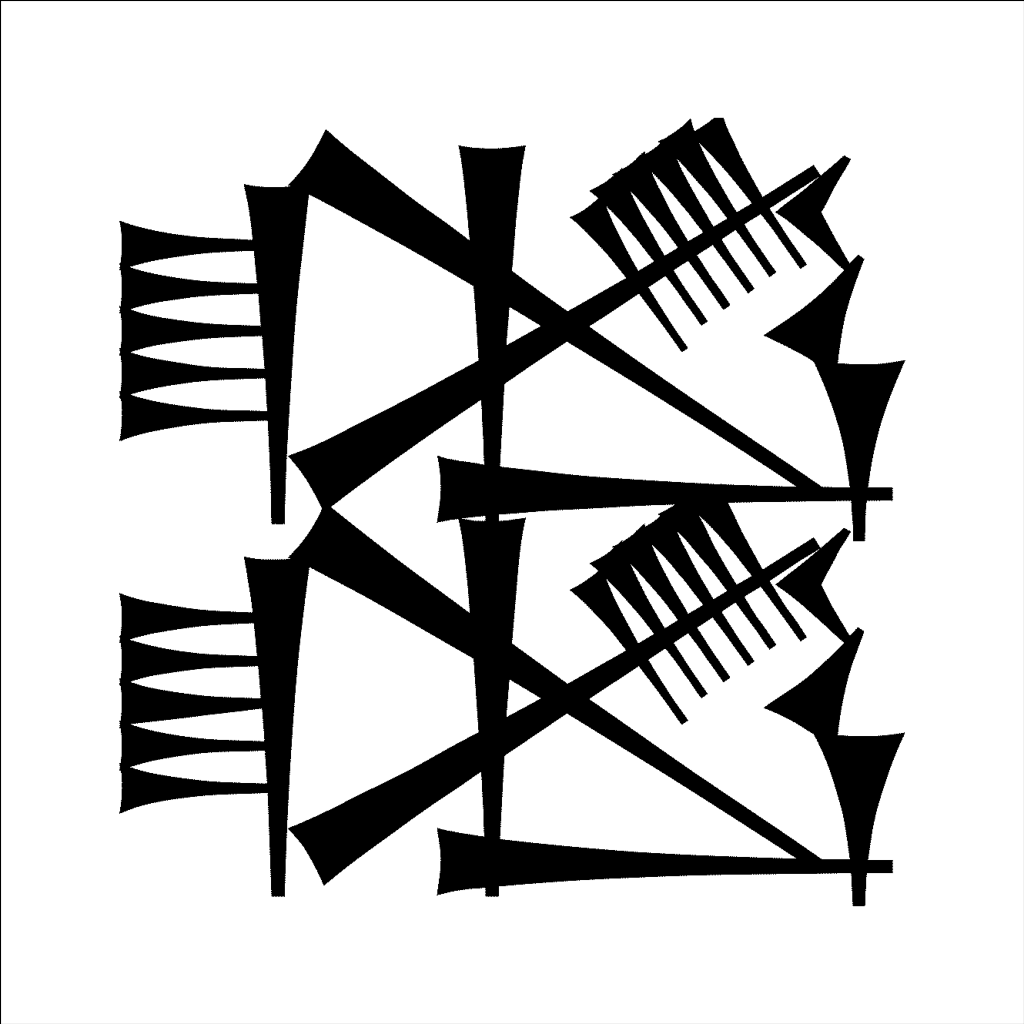} &
\imgwithbox[width=0.95\linewidth]{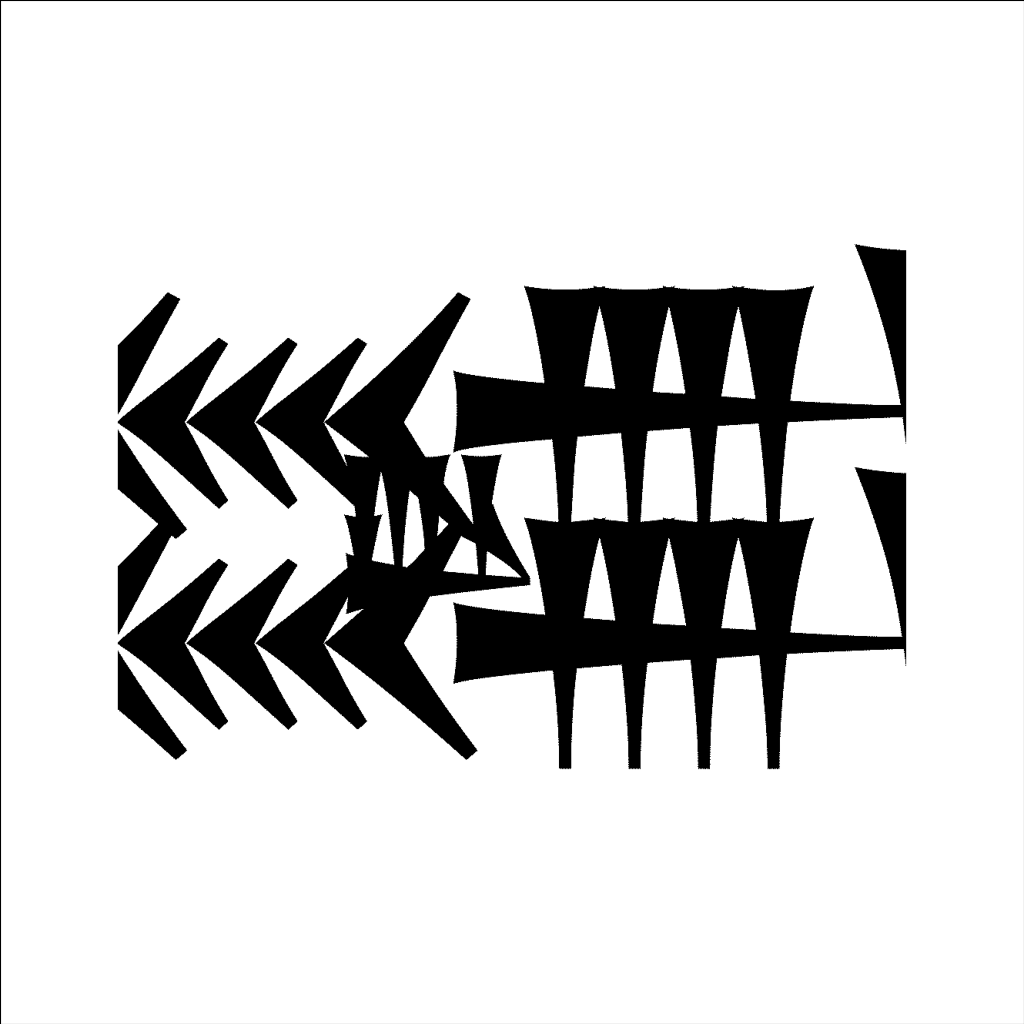} &
\imgwithbox[width=0.95\linewidth]{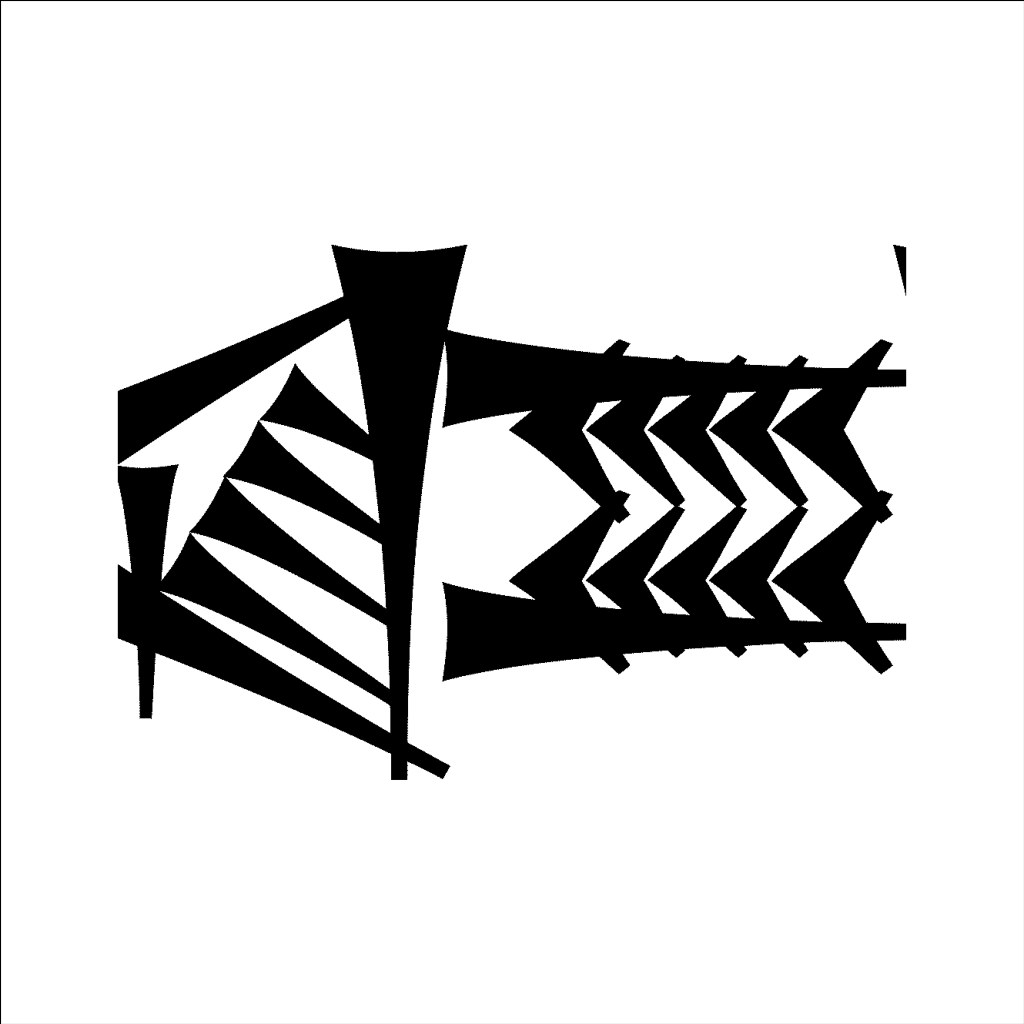} &
\imgwithbox[width=0.95\linewidth]{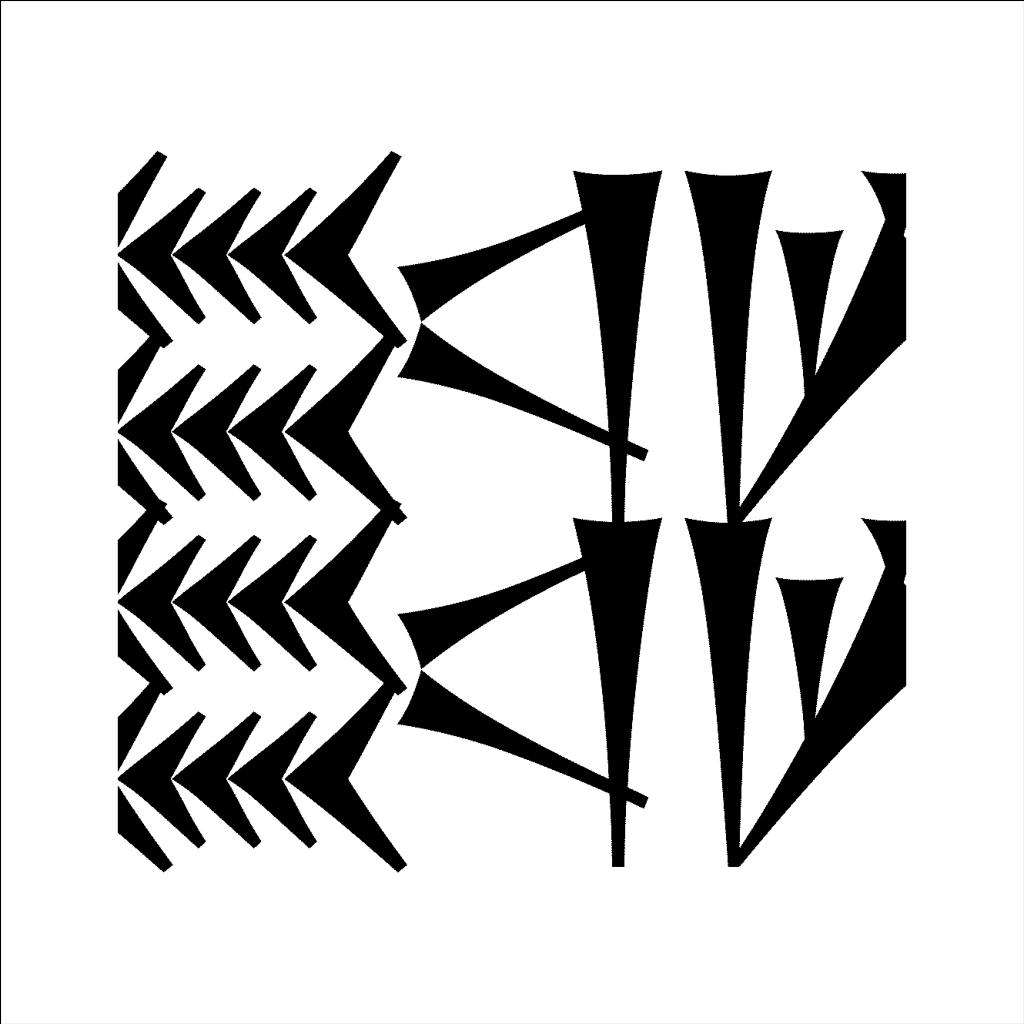} &
\imgwithbox[width=0.95\linewidth]{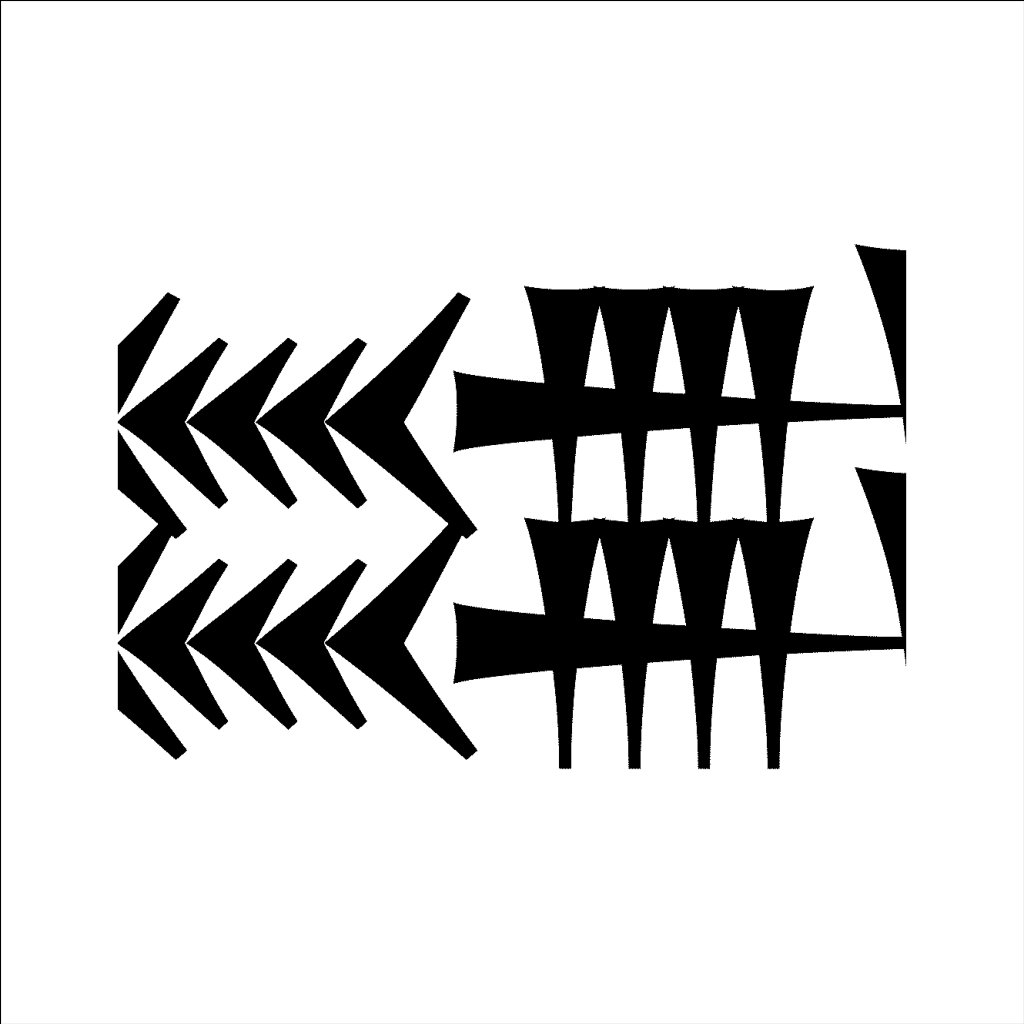} &
\includegraphics[width=0.95\linewidth]{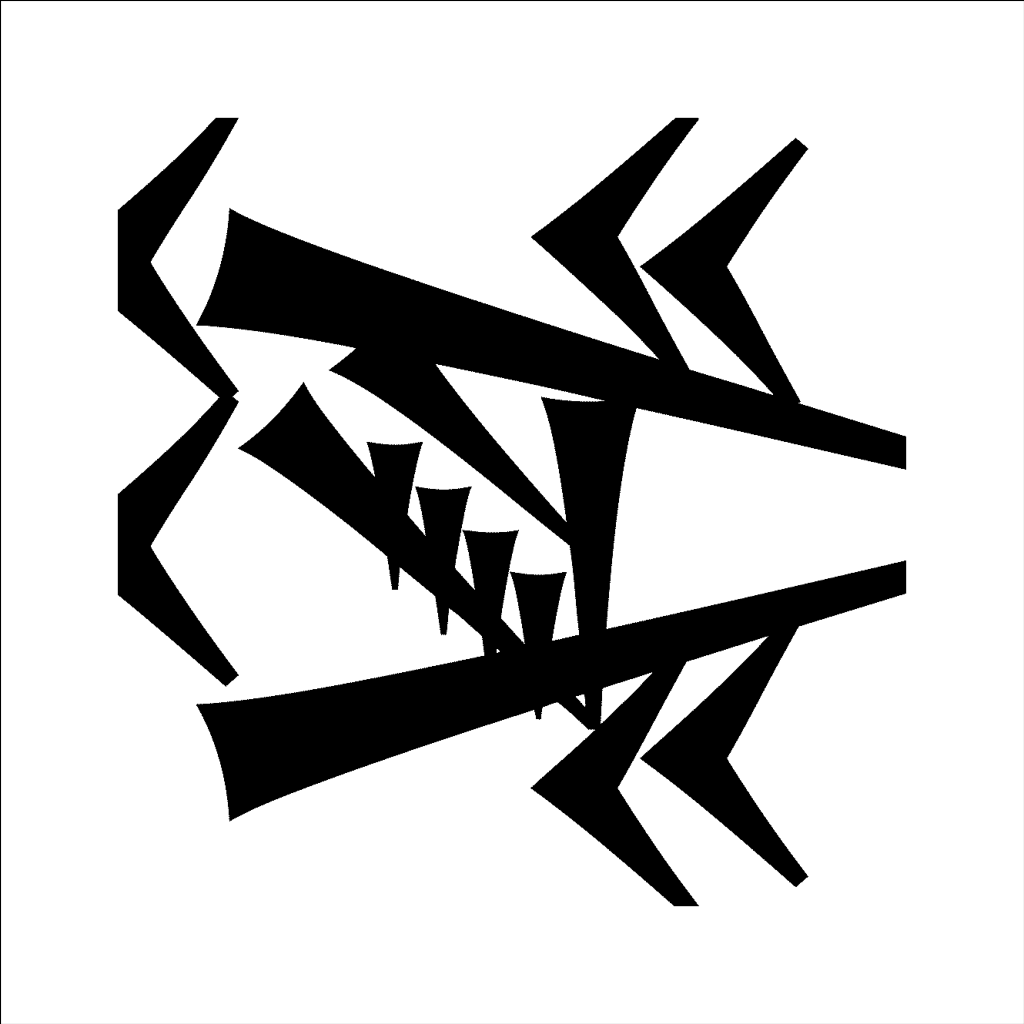} \\
11 &
\includegraphics[width=0.95\linewidth]{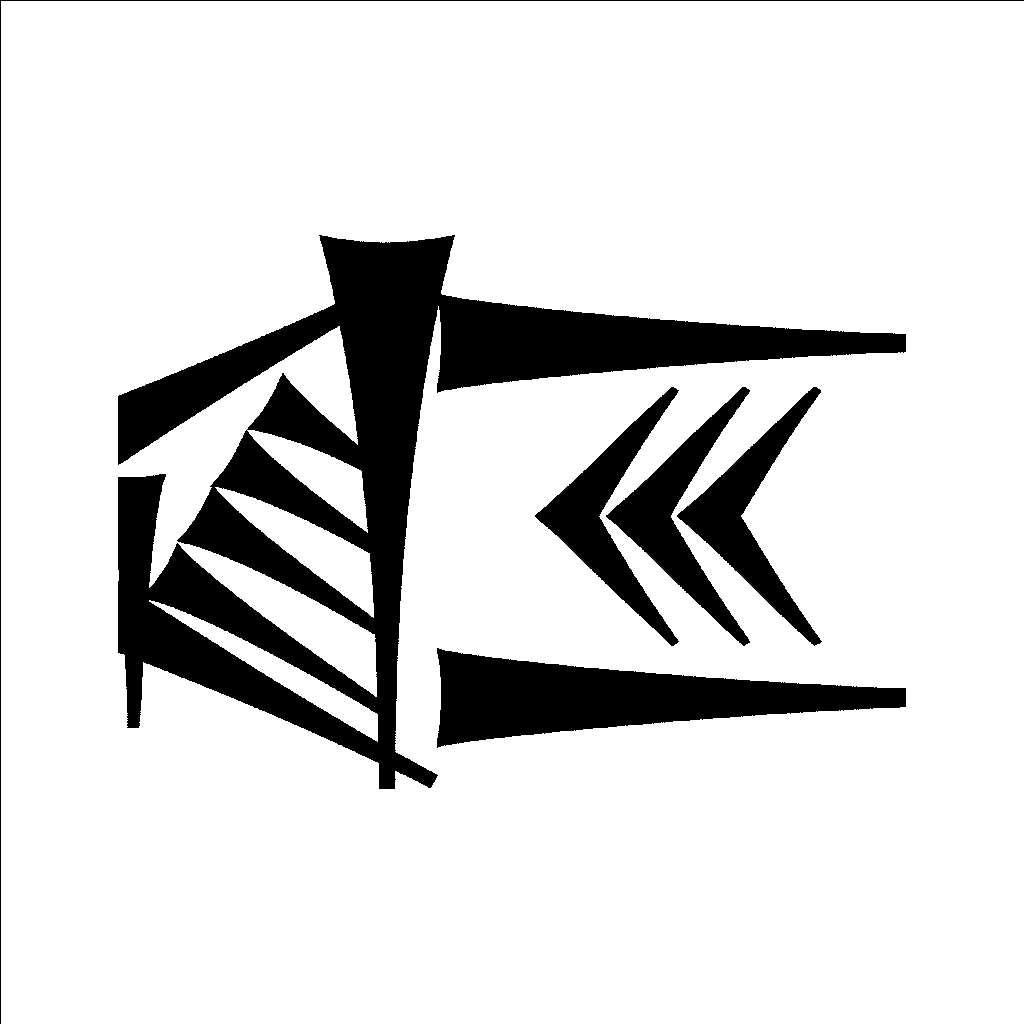} &
\includegraphics[width=0.95\linewidth]{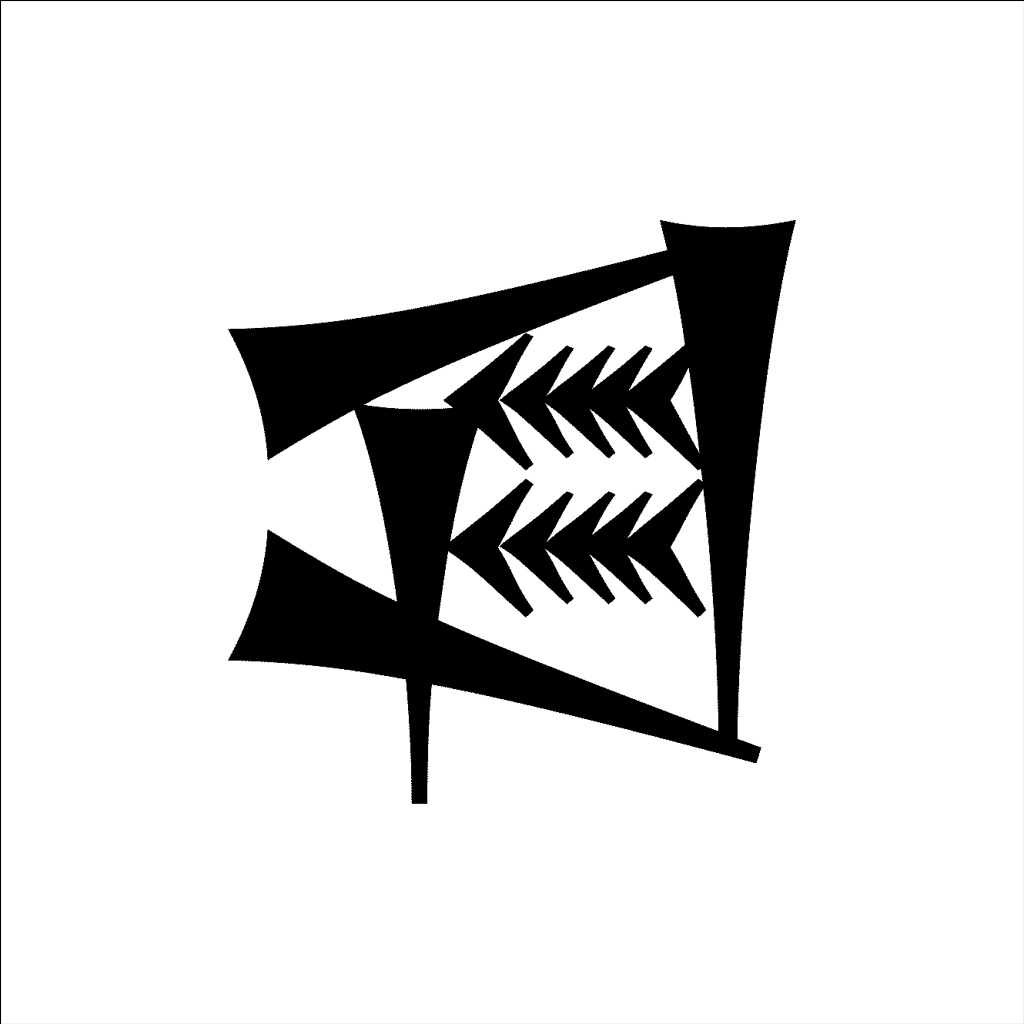} &
\imgwithbox[width=0.95\linewidth]{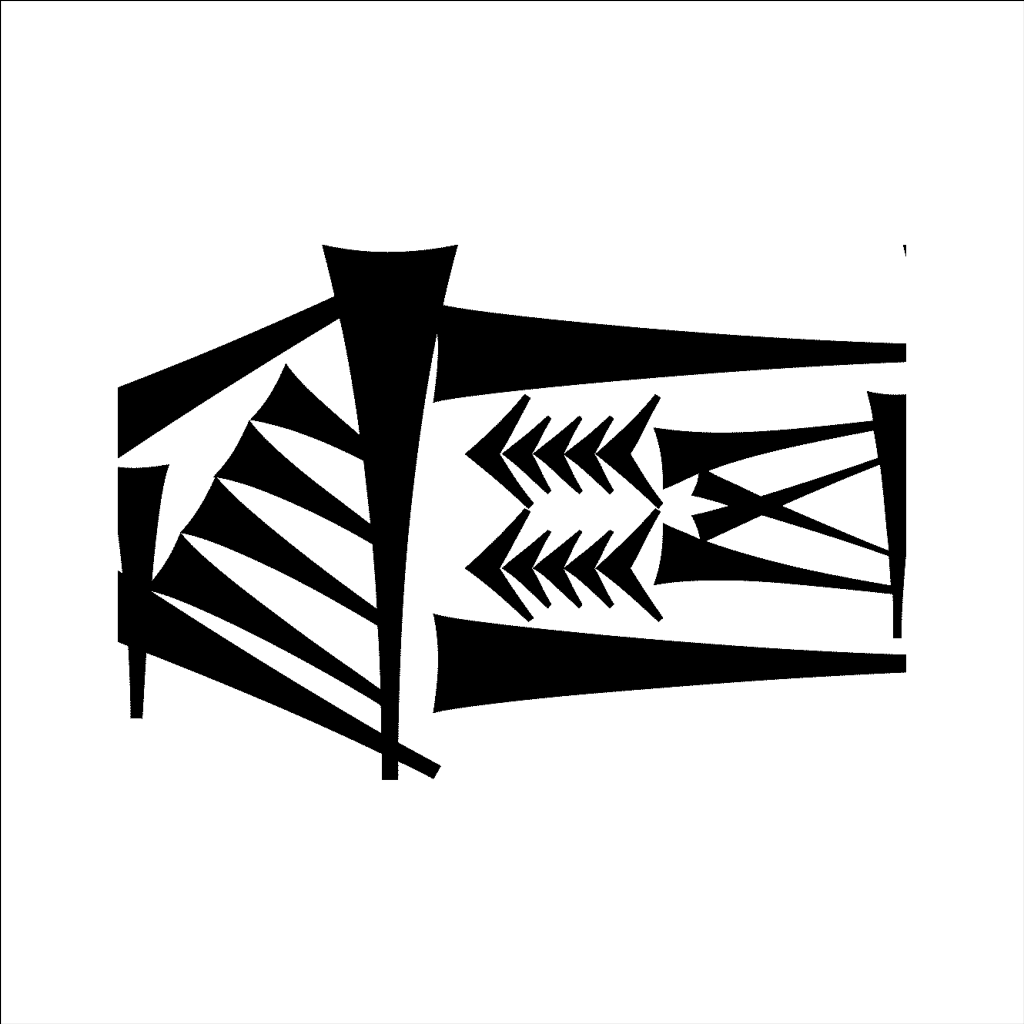} &
\includegraphics[width=0.95\linewidth]{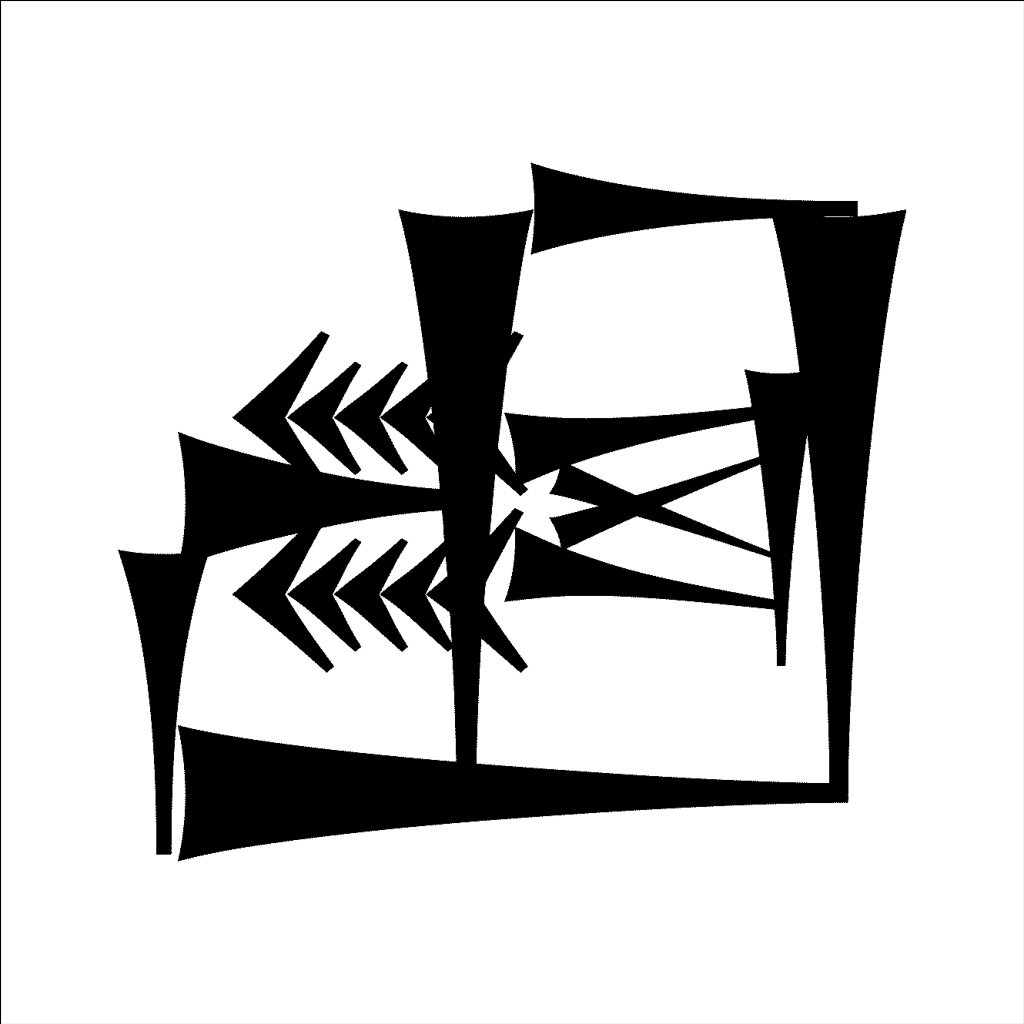} &
\includegraphics[width=0.95\linewidth]{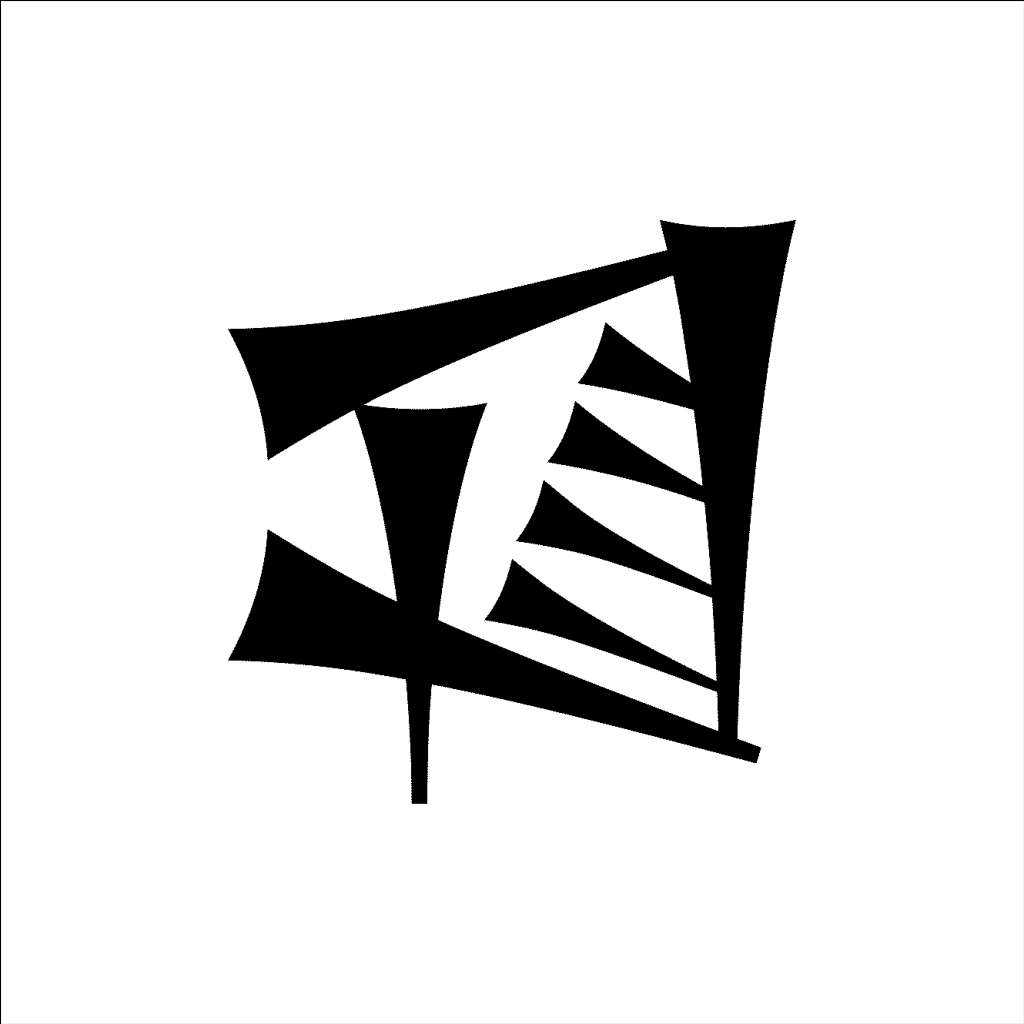} &
\includegraphics[width=0.95\linewidth]{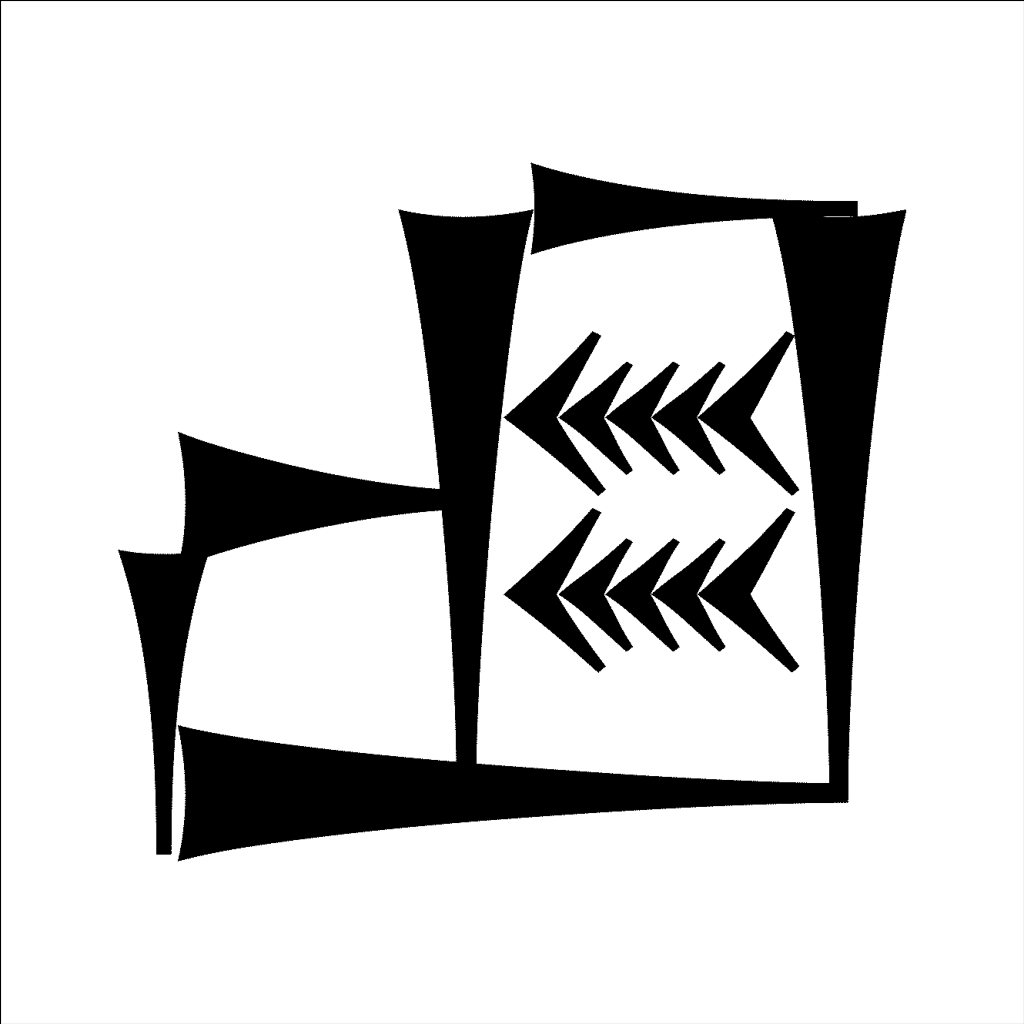} &
\imgwithbox[width=0.95\linewidth]{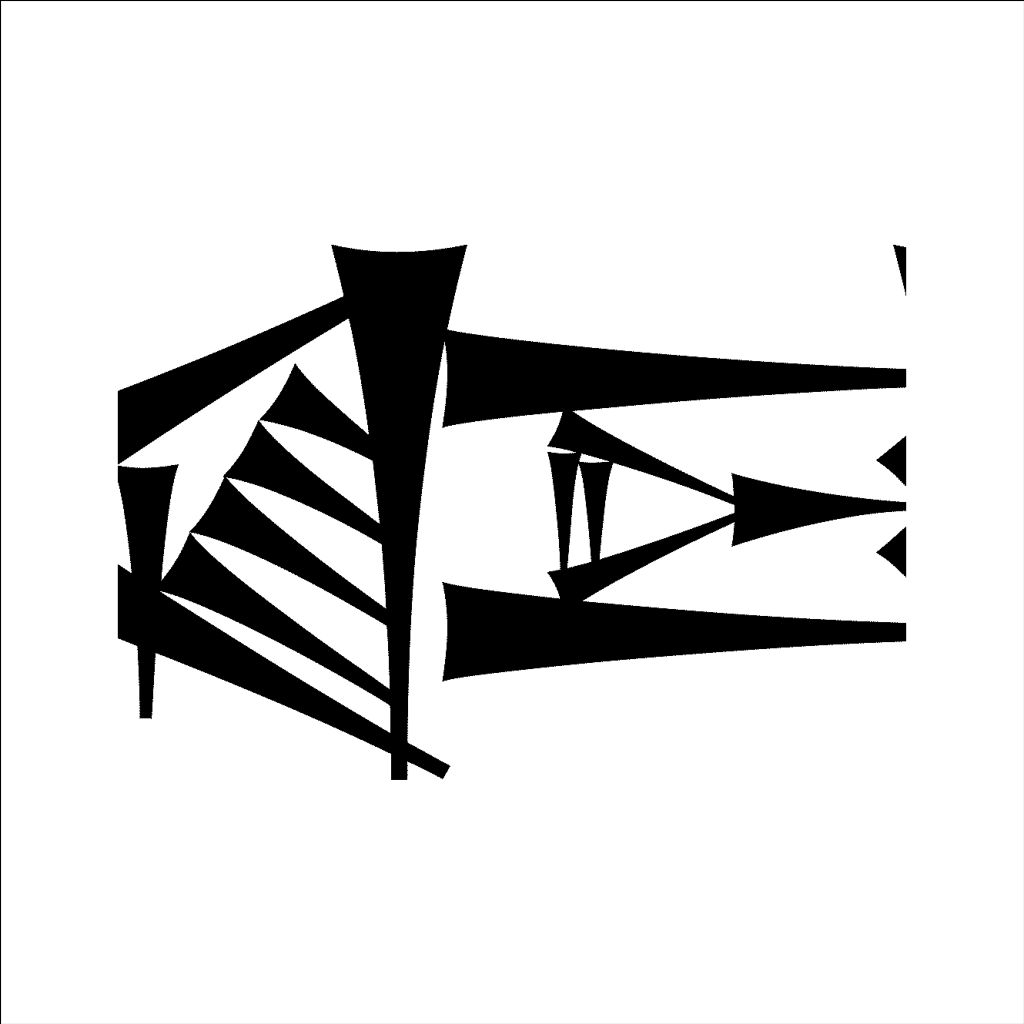} &
\imgwithbox[width=0.95\linewidth]{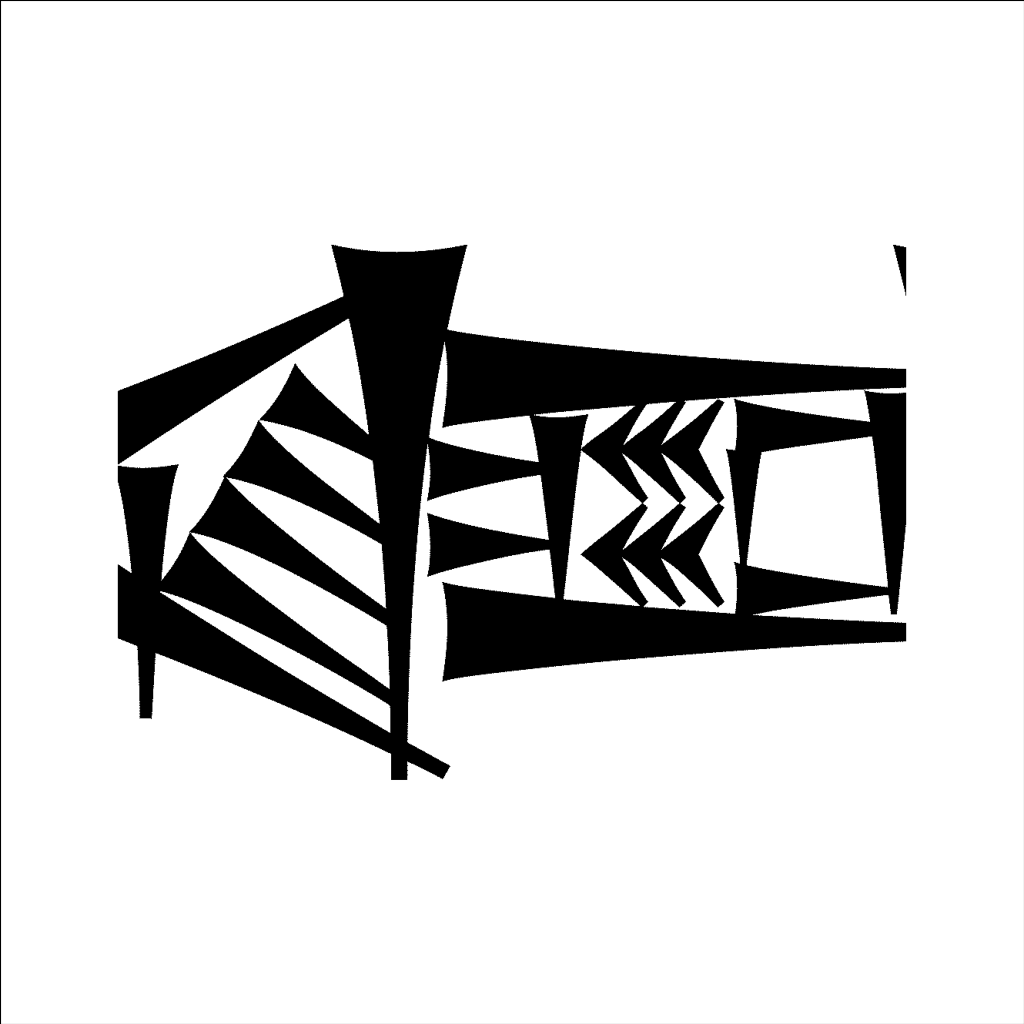} &
\includegraphics[width=0.95\linewidth]{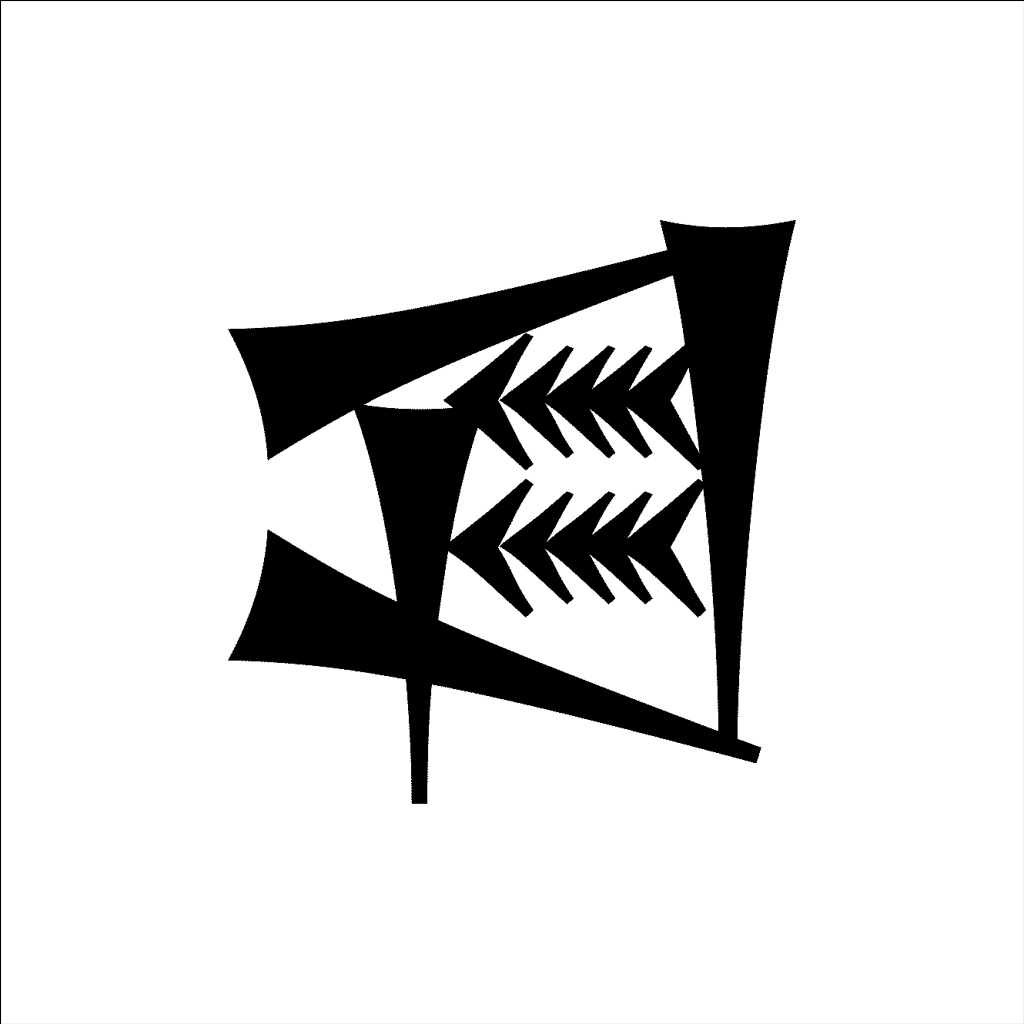} &
\imgwithbox[width=0.95\linewidth]{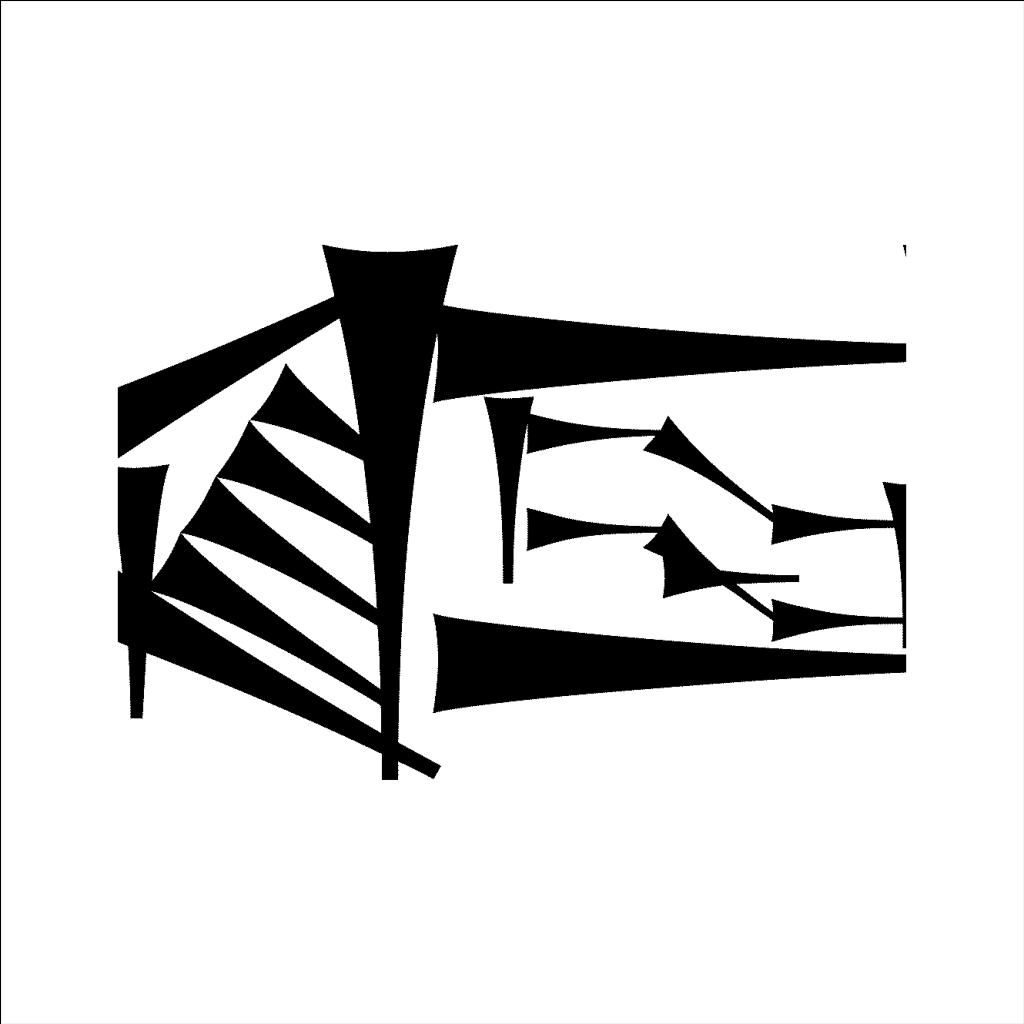} &
\imgwithbox[width=0.95\linewidth]{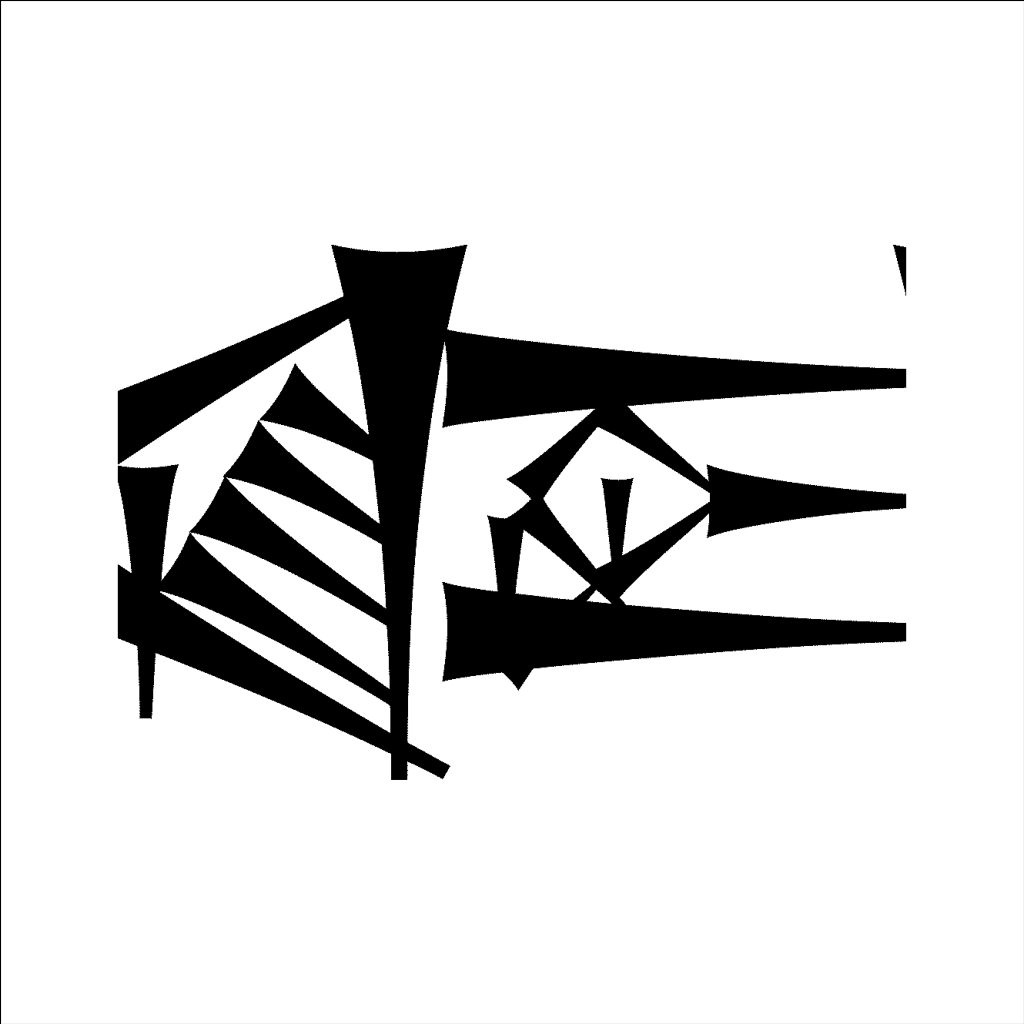} \\
12 &
\includegraphics[width=0.95\linewidth]{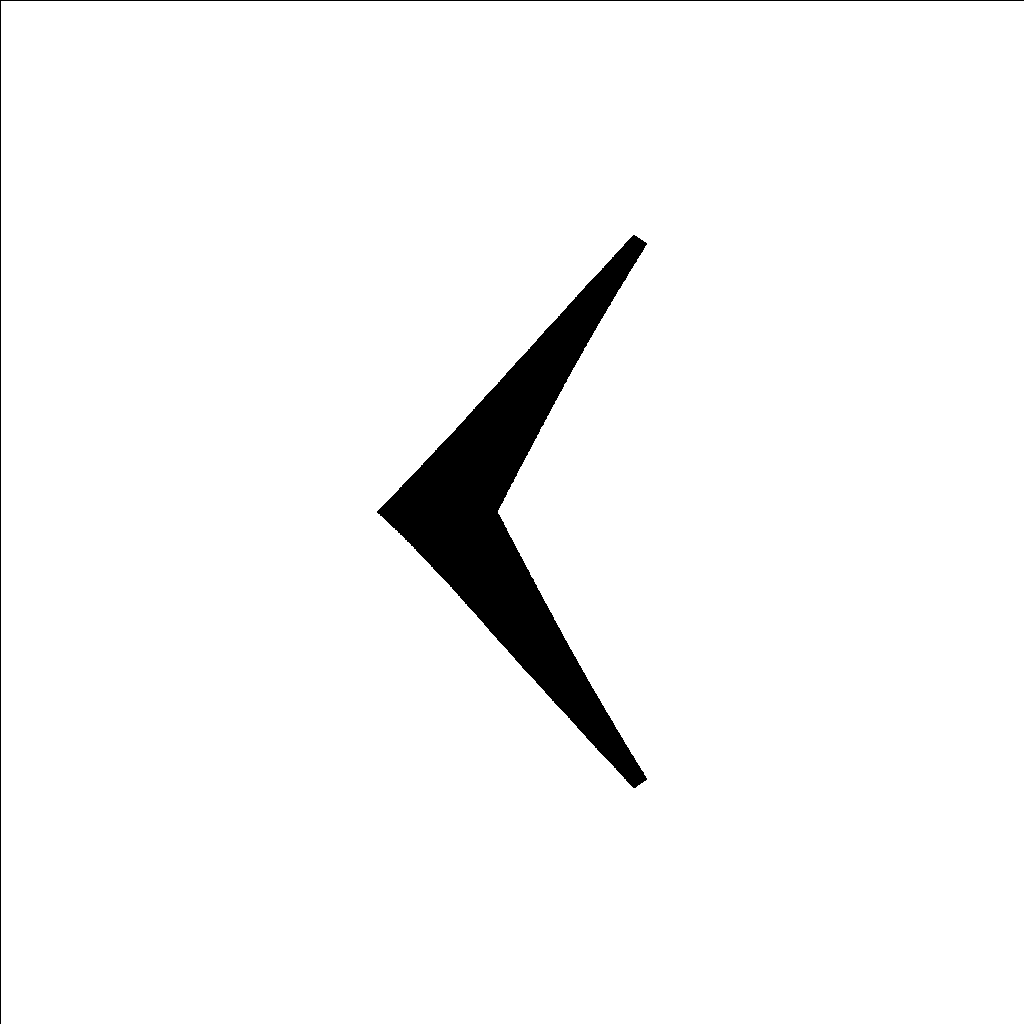} &
\imgwithbox[width=0.95\linewidth]{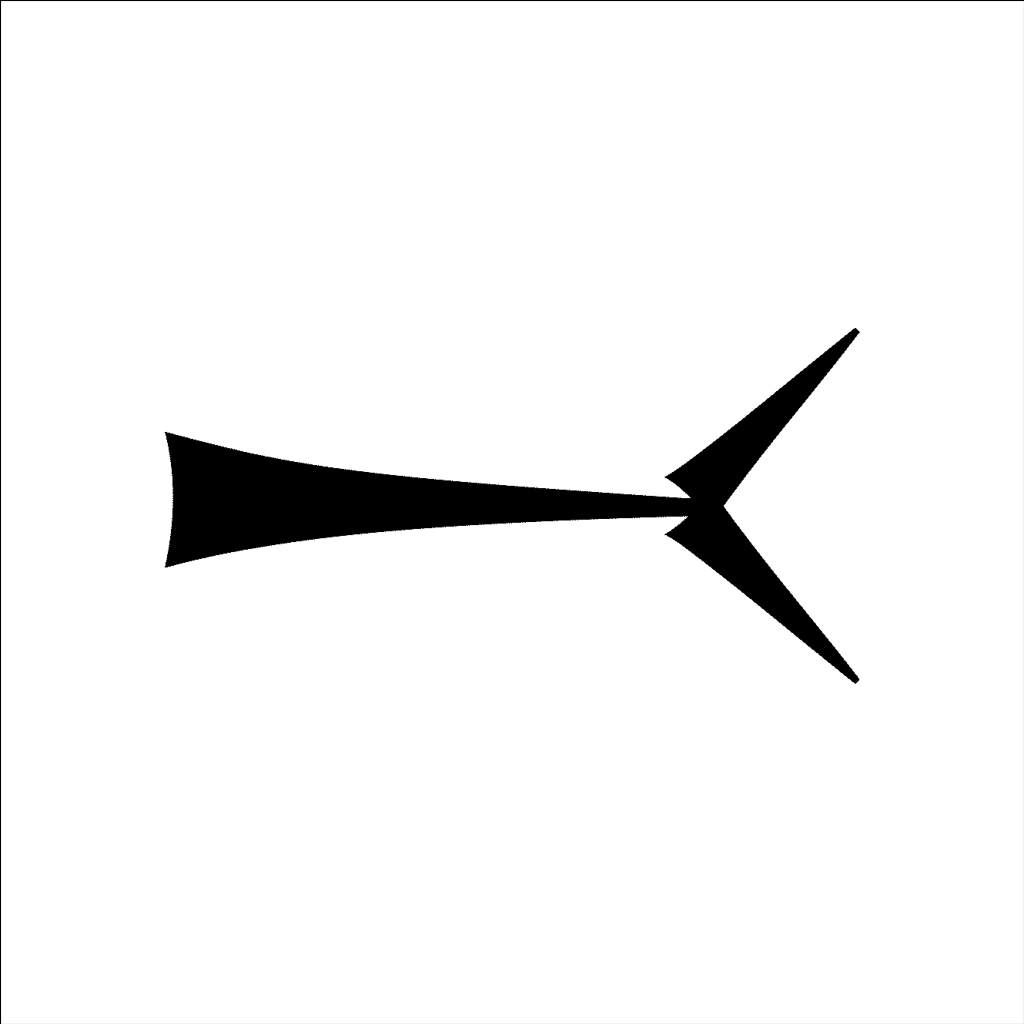} &
\includegraphics[width=0.95\linewidth]{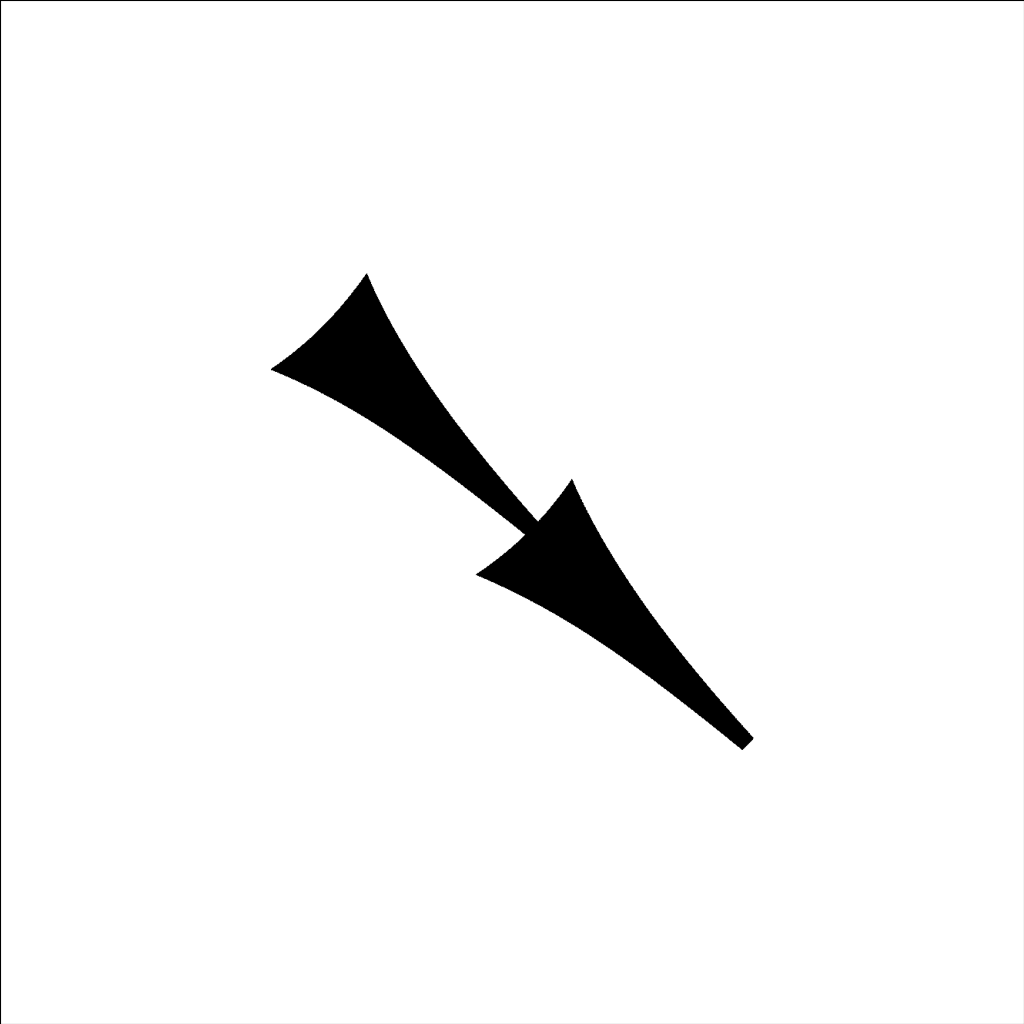} &
\includegraphics[width=0.95\linewidth]{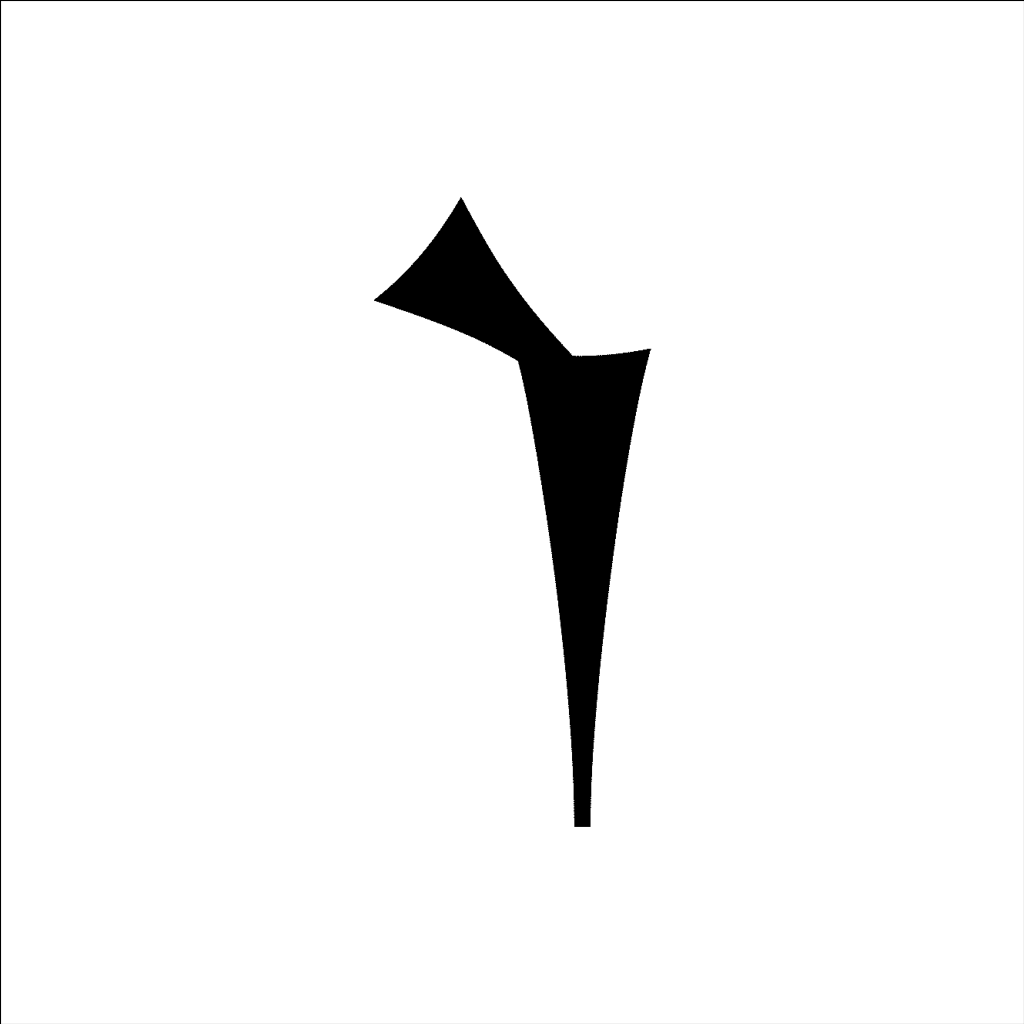} &
\includegraphics[width=0.95\linewidth]{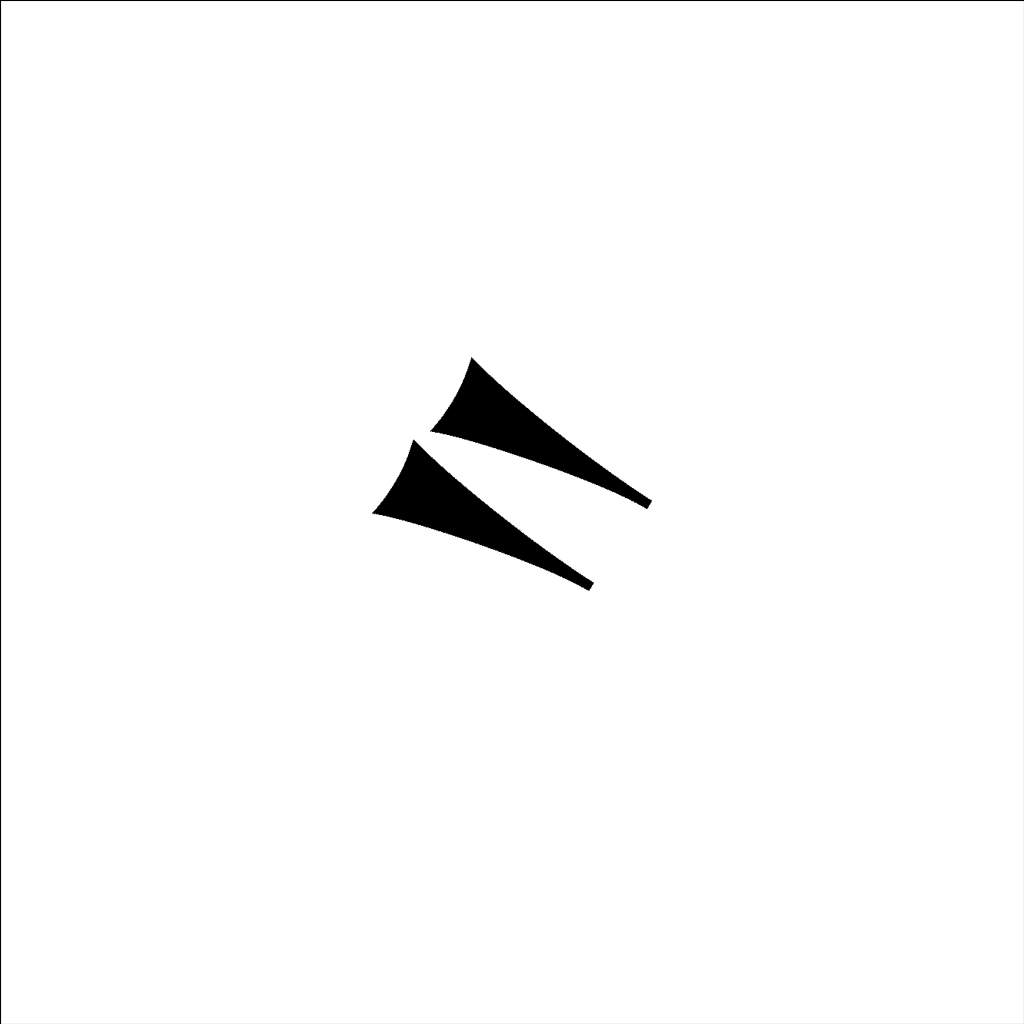} &
\includegraphics[width=0.95\linewidth]{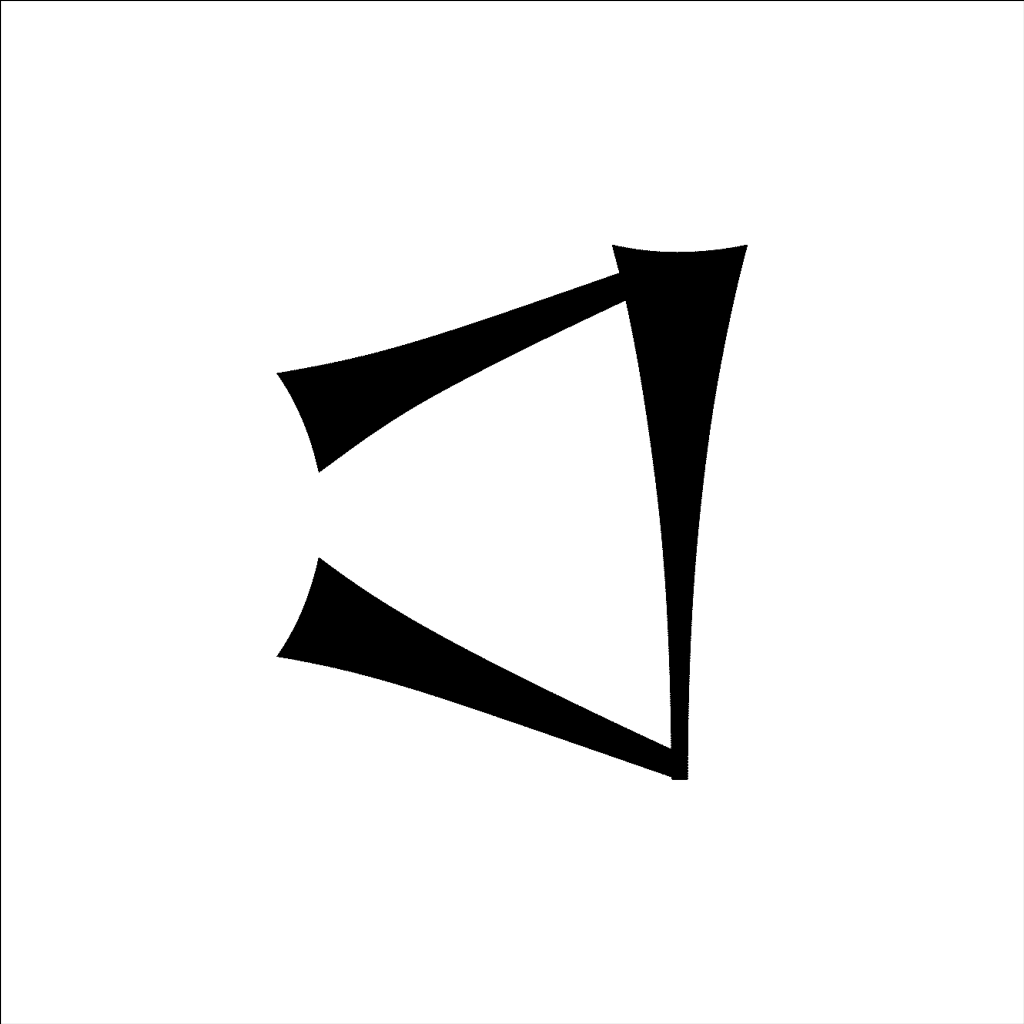} &
\imgwithbox[width=0.95\linewidth]{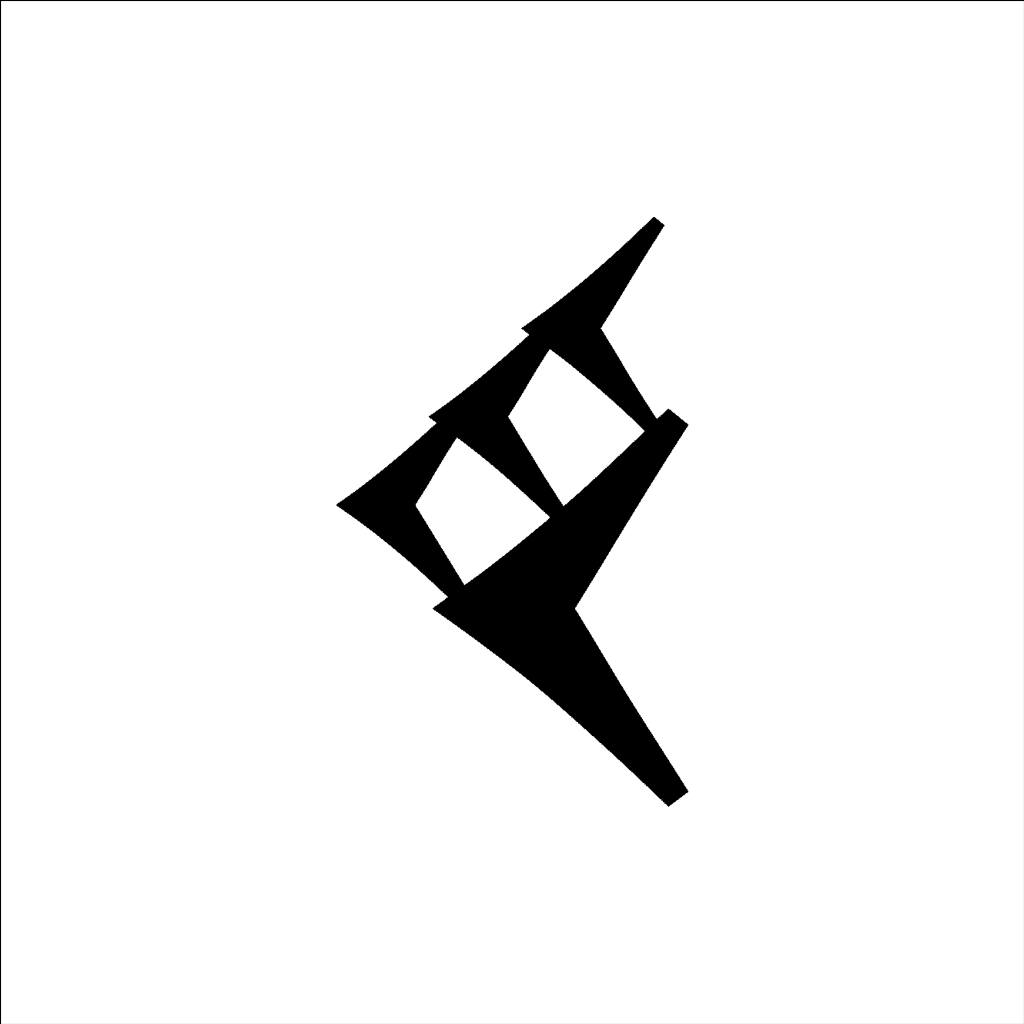} &
\imgwithbox[width=0.95\linewidth]{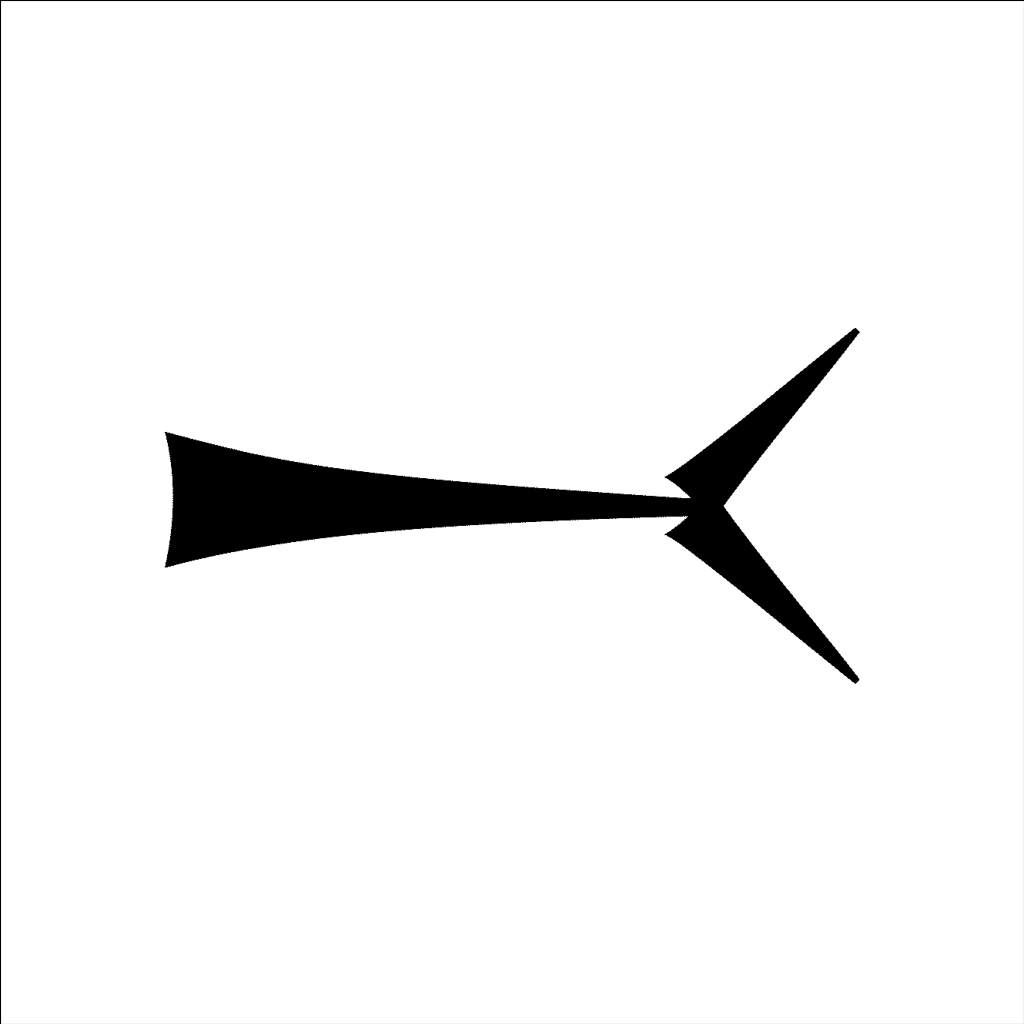} &
\includegraphics[width=0.95\linewidth]{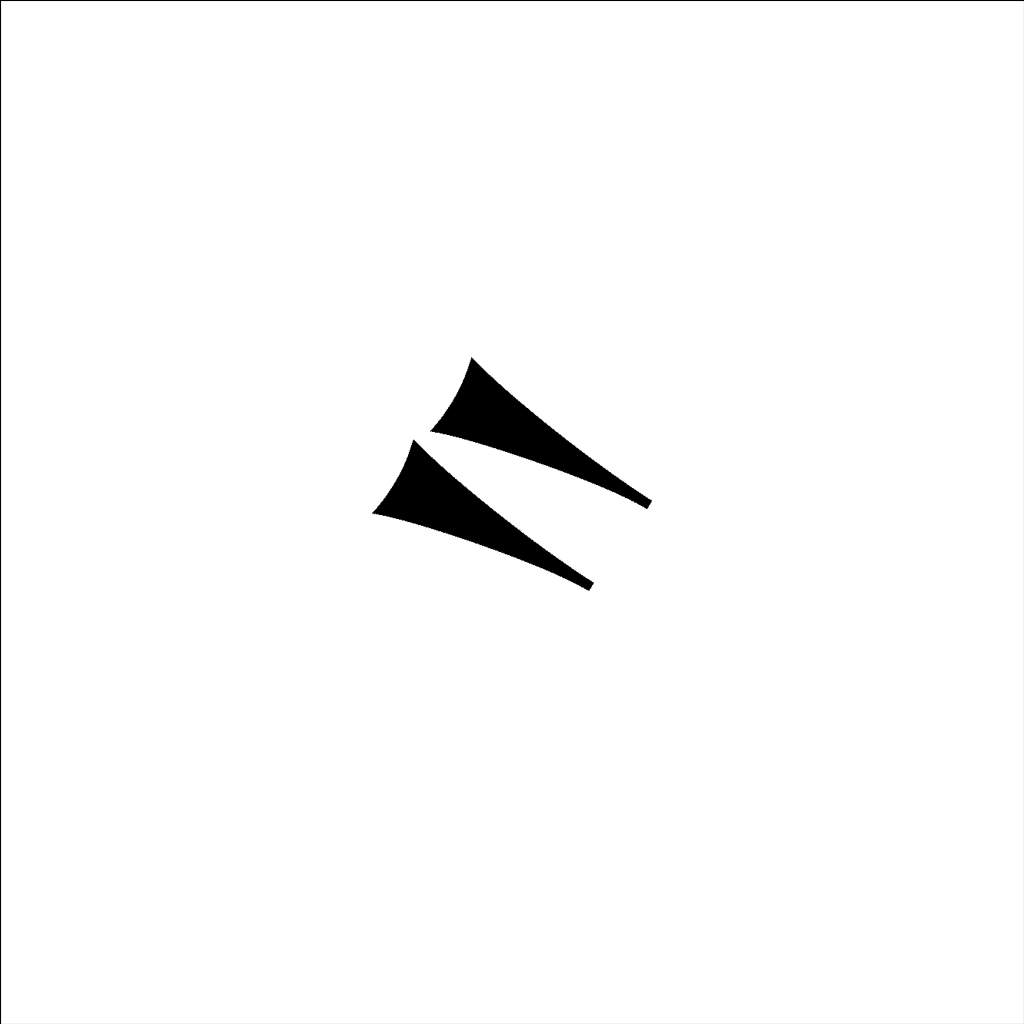} &
\includegraphics[width=0.95\linewidth]{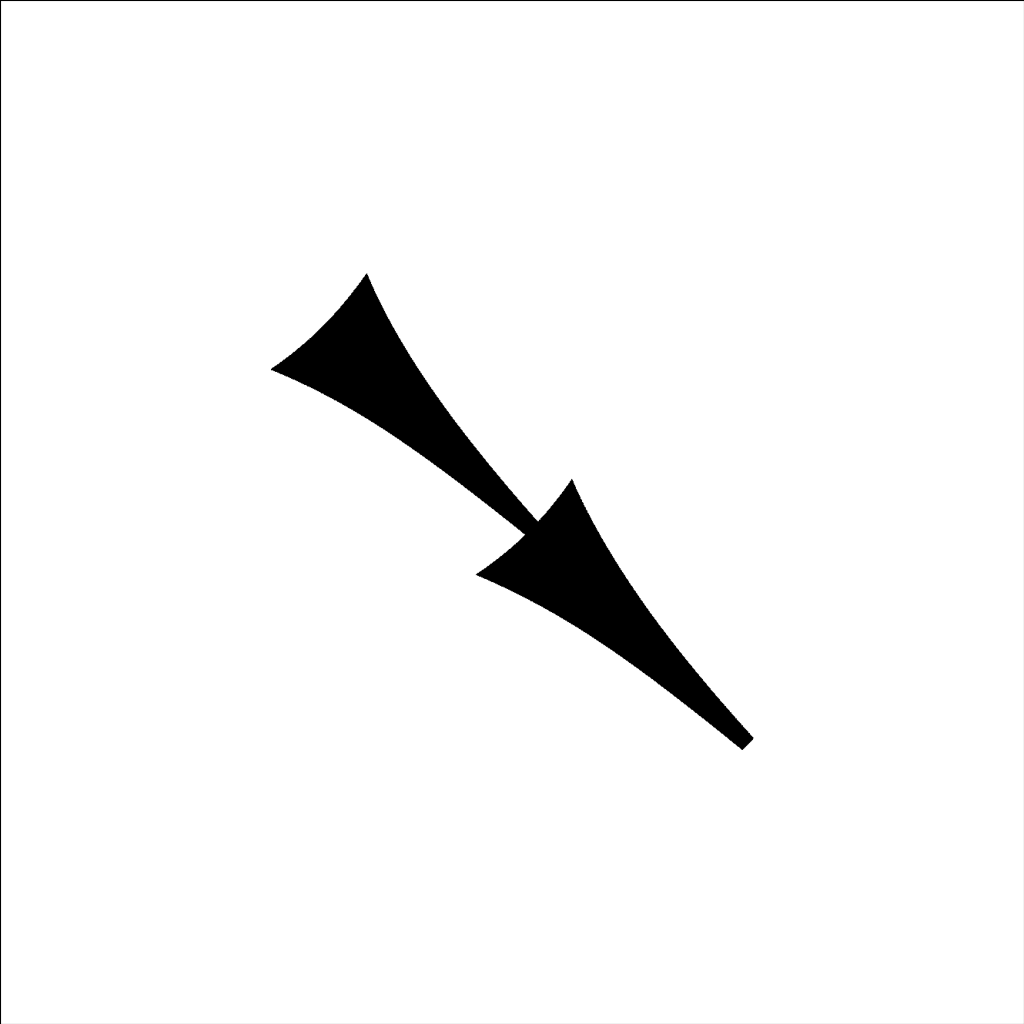} &
\includegraphics[width=0.95\linewidth]{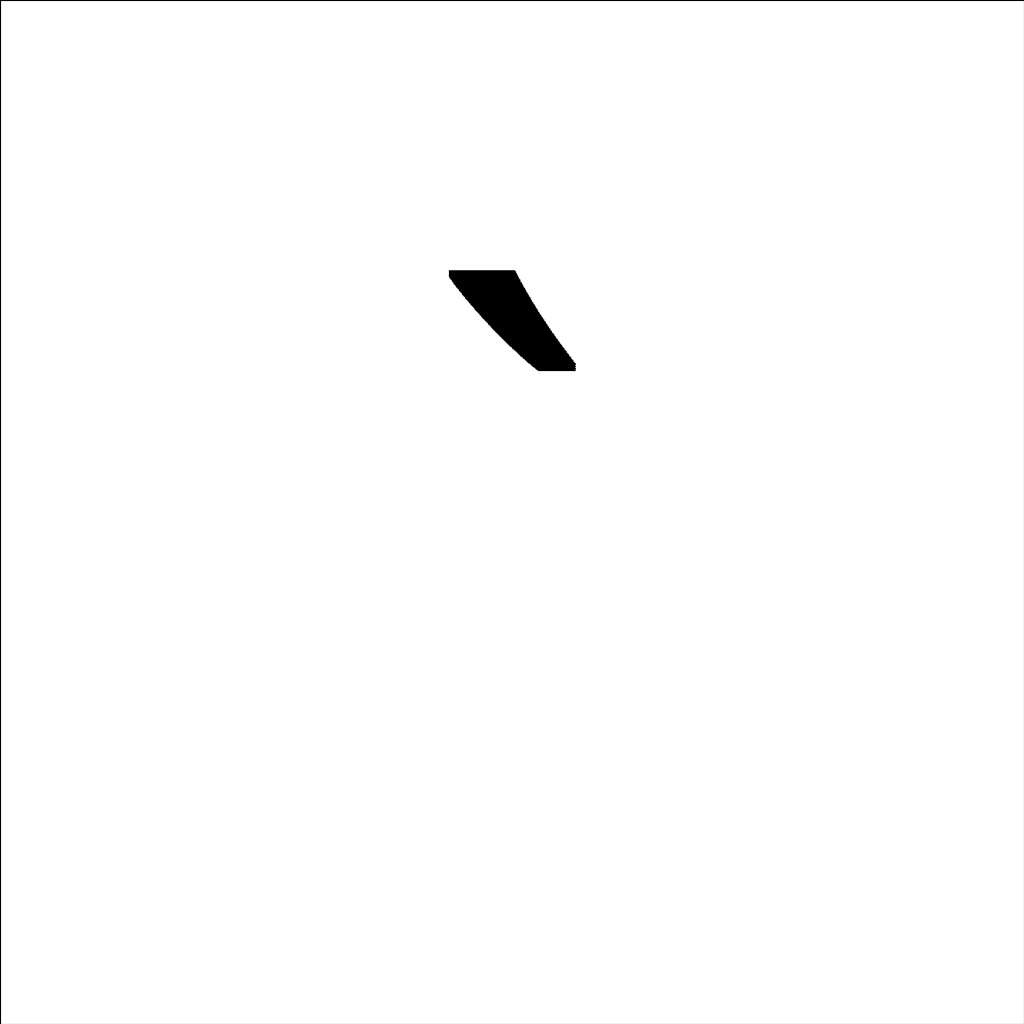} \\
\bottomrule
\end{tabular}}}
\vspace{-6pt}
\caption{More cases of structure-guided character exploration. More cases of structure-guided character exploration. Here, we additionally include examples from Cuneiform, one of the earliest known writing systems, developed in ancient Mesopotamia and characterized by wedge-shaped impressions pressed into clay tablets.}
\label{tab:character_exploration_appendix}
\vspace{-9pt}
\end{table*}

\section{More Cases of Structure-Guided Character Exploration}\label{app:more_cases}

Additional examples of structure-guided character exploration are presented in Table~\ref{tab:character_exploration_appendix}.

\section{LLM Usage Statement}\label{app:llm_usage_statement}

Throughout the preparation of this manuscript, LLMs are used exclusively for spelling and grammatical error checking. They are not employed for any other purposes, including the generation of research ideas or the validation of the proposed methods or experimental results.

\end{document}